\documentclass[10pt,twoside]{amundi_article}

\usepackage[table]{xcolor}
\usepackage{amssymb}
\usepackage{amsfonts}
\usepackage{amsmath}
\usepackage{amsthm}
\usepackage{fancybox}
\usepackage{fancyhdr}
\usepackage{graphicx}
\usepackage{subcaption}
\usepackage[hyphens]{url}
\usepackage{arydshln}
\usepackage{multirow}
\usepackage{pdflscape}
\usepackage{tikz}
\usepackage{comment}
\usepackage{nicefrac}
\usepackage{subcaption}
\usepackage{dsfont}
\usepackage{algorithmic}
\usepackage{algorithm}
\usepackage{smartdiagram}
\usepackage{natbib}
\usepackage[colorlinks = true, linkcolor = blue, urlcolor = blue,
            citecolor = blue, anchorcolor = blue]{hyperref}

\newlength{\figurewidth}
\newlength{\figureheight}
\def\figureskip{\vskip 10pt plus 2pt minus 2pt\relax}

\newtheorem{remark}{Remark}

\def\limfunc#1{\mathop{\rm #1}}
\def\func#1{\mathop{\rm #1}}

\newcommand{\TsIII}{\hspace{3pt}}
\newcommand{\TsVIII}{\hspace{8pt}}

\DeclareMathAlphabet\mathbfcal{OMS}{cmsy}{b}{n}

\newcommand{\Generator}{\mathbfcal{G}}
\newcommand{\Discriminator}{\mathbfcal{D}}

\usetikzlibrary{3d, decorations.text, shapes.arrows, positioning, fit, backgrounds}
\usetikzlibrary{shapes, calc, arrows, decorations.markings}

\newif\ifResearchVersion
\ResearchVersiontrue

\begin{document}

\ifResearchVersion

\title{\textbf{\color{amundi_blue}Improving the Robustness of Trading Strategy
Backtesting with Boltzmann Machines and Generative Adversarial
Networks}}
\author{
{\color{amundi_dark_blue} Edmond Lezmi} \\
Quantitative Research \\
Amundi Asset Management, Paris \\
\texttt{edmond.lezmi@amundi.com} \and
{\color{amundi_dark_blue} Jules Roche} \\
Department of Mathematics \\
Ecole des Ponts ParisTech, Paris \\
\texttt{jules.roche@eleves.enpc.fr} \and
{\color{amundi_dark_blue} Thierry Roncalli} \\
Quantitative Research \\
Amundi Asset Management, Paris \\
\texttt{thierry.roncalli@amundi.com} \and
{\color{amundi_dark_blue} Jiali Xu} \\
Quantitative Research \\
Amundi Asset Management, Paris \\
\texttt{jiali.xu@amundi.com}}

\date{\color{amundi_dark_blue}June 2020}

\maketitle

\begin{abstract}
In this article, we explore generative models in order to build a market
generator. The underlying idea is to simulate artificial multi-dimensional
financial time series, whose statistical properties are the same as those
observed in the financial markets. In particular, these synthetic data must
preserve the first four statistical moments (mean, standard deviation,
skewness and kurtosis), the stochastic dependence between the different
dimensions (copula structure) and across time (autocorrelation
function). The first part of the article reviews the more relevant generative
models, which are restricted Boltzmann machines, generative adversarial
networks, and convolutional Wasserstein models. The second part of the
article is dedicated to financial applications by considering the simulation
of multi-dimensional times series and estimating the probability distribution
of backtest statistics. The final objective is to develop a framework for
improving the risk management of quantitative investment strategies.
\end{abstract}

\medskip


\section{Introduction}

In machine learning, we generally distinguish two types of statistical modeling
\citep{Jebara-2004}:
\begin{itemize}
\item the generative approach models the unconditional probability
    distribution $\mathbb{P}\left( X\right) $ given a set of observable
    variables $X$;

\item the discriminative approach models the conditional probability
    distribution $\mathbb{P}\left( Y\mid X\right) $ given a set of observable
    variables $X$ and a target variable $Y$.
\end{itemize}
For instance, examples of generative models are the principal component
analysis (PCA) or the method of maximum likelihood (ML) applied to a parametric
probability distribution. Examples of discriminative models are the linear
regression, the linear discriminant analysis or support vector machines. In the
first case, generative models can be used to simulate samples that capture and
reproduce the statistical properties of a training dataset. In the second case,
discriminative models can be used to predict the target variable $Y$ for new
examples of $X$. Said differently, the distinction between generative and
discriminative models can be seen as a reformulation of the distinction between
unsupervised and supervised machine learning. More specifically, generative
models can be used to learn the underlying probability distributions over data
manifolds. The objective of these models is to estimate the statistical properties and
correlation structure of real data and simulate synthetic data with a
probability distribution, which is close to the real one.\smallskip

\fi

In finance, we generally observe only one sample path of market prices. For
instance, if we would like to build a trading strategy on the S\&P 500 index,
we can backtest the strategy using the historical values of the S\&P 500 index.
We can then measure the performance and the risk of this investment strategy by
computing annualized return, volatility, Sharpe ratio, maximum
drawdown, etc. In this case, it is extremely difficult to assess the robustness
of the strategy, since we can consider that the historical sample of the S\&P
500 index is one realization of the unknown stochastic process. Therefore,
portfolio managers generally split the study period into two subperiods: the
\textquoteleft \textit{in-sample}\textquoteright\ period and the \textquoteleft
\textit{out-of-sample}\textquoteright\ period. The objective is to calibrate
the parameters of the trading strategy with one subperiod and measure the
financial performance with the other period in order to reduce the overfitting
bias. However, if the out-of-sample approach is appealing, it is limited for
two main reasons. First, by splitting the study period into two subperiods, the
calibration procedure is performed with fewer observations, and does not
generally take into account the most recent period. Second, the validation step
is done using only one sample path. Again, we observe only one realization of
the risk/return statistics. Of course, we could use different splitting
methods, but we know that these bootstrap techniques are not well-adapted to
times series and must be reserved for modeling random variables. For stochastic
processes, statisticians prefer to consider Monte Carlo methods. Nevertheless,
financial times series are difficult to model, because they exhibit non-linear
autocorrelations, fat tails, heteroscedasticity, regime switching and
non-stationary properties \citep{Cont-2001}.\smallskip

In this article, we are interested in generative models in order to obtain
several training/validation sets. The underlying idea is then to generate
artificial but realistic financial market prices. The choice of generative
model and market data leads naturally to the concept of market generator
introduced by \citet{Kondratyev-2019}. If the market generator is able to
replicate the probability distribution of the original market data, we can then
backtest quantitative investment strategies on several financial markets. The
backtesting procedure is then largely improved, since we obtain a probability
distribution of performance and risk statistics, and not only one value.
Therefore, we can reduce the in-sample property of the backtest procedure and
the overfitting bias of the parameters that define the trading strategy.
However, the challenge is not simple, since the market generator must be
sufficiently robust and flexible in order to preserve the uni-dimensional
statistical properties of the original financial time series, but also the
multi-dimensional dependence structure.\smallskip

This paper is organized as follows. Section Two reviews the more promising
generative models that may be useful in finance. In particular, we focus on
restricted Boltzmann machines, generative adversarial networks and Wasserstein
distance models. In Section Three, we apply these models in the context of
trading strategies. We first consider the joint simulation of S\&P 500 and VIX
indices. We then build an equity/bond risk parity strategy with an investment
universe of six futures contracts. Finally, Section Four offers some concluding
remarks.

\section{Generative models}
\label{section:generative-models}

In this section, we consider two main approaches. The first one is based on a
restricted Boltzmann machine (RBM), while the second one uses a conditional
generative adversarial network (GAN). Both are stochastic artificial neural
networks that learn the probability distribution of a real data sample.
However, the objective function strongly differs. Indeed, RBMs consider a
log-likelihood maximization problem, while the framework of GANs corresponds to
a minimax two-player game. In the first case, the difficulty lies in the
gradient approximation of the log-likelihood function. In the second case, the
hard task is to find a learning process algorithm that solves the two-player
game. This section also presents an extension of generative adversarial
networks, which is called a convolutional Wasserstein model.

\subsection{Restricted Boltzmann machines}
\label{section:rbm}

Restricted Boltzmann machines were initially invented under the name
\textquoteleft \textit{Harmonium}\textquoteright\ by \citet{Smolensky-1986}.
Under the framework of undirected graph models\footnote{Fundamental concepts
of undirected graph model are explained in Appendix
\ref{appendix:graph-model} on page \pageref{appendix:graph-model}.}, an RBM
is a Markov random field (MRF) associated with a bipartite undirected
graph\footnote{A bipartite graph is a graph whose nodes can be divided into
two disjoint and independent sets $\mathcal{U}$ and $\mathcal{V}$ such that
every edge connects a node in $\mathcal{U}$ to one in $\mathcal{V}$.}. RBMs
are made of only two layers as shown in Figure \ref{fig:RBM}. We distinguish
visible units belonging to the visible layer from hidden units belonging to
the hidden layer. Each unit of visible and hidden layers is respectively
associated with visible and hidden random variables. The term \textquoteleft
\textit{restricted}\textquoteright\ comes from the facts that the connections
are only between the visible units and the hidden units and that there is no
connection between two different units in the same layer.

\tikzstyle{neuron}=[draw,circle,minimum size=20pt, inner sep=0pt, fill=white]
\tikzstyle{stateTransition}=[thick] \tikzstyle{learned}=[text=red]
\tikzstyle{place}=[circle, draw=black, minimum size = 8mm]

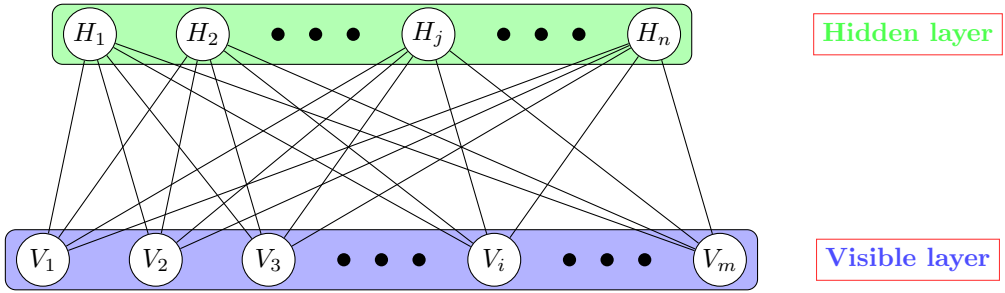
\begin{figure}[tbh]
\centering
\caption{The schema of a RBM with $m$ visible units and $n$ hidden units}
\label{fig:RBM}
\begin{tikzpicture}[scale=2]
	\draw[fill=green!30, rounded corners] (0.8125, 1.3) rectangle (5.0625, 1.7) {};
    \draw node at (1.0625, 1.5) [neuron] (H1) {$H_1$};
    \draw node at (1.8125, 1.5) [neuron] (H2) {$H_2$};
    \foreach \x in {1,...,3}
    	\fill (2.0625 +\x*0.25, 1.5) circle (1.25pt);
    \draw node at (3.3125, 1.5) [neuron] (Hj) {$H_j$};
    \foreach \x in {1,...,3}
    	\fill (3.5625 +\x*0.25, 1.5) circle (1.25pt);
    \draw node at (4.8125, 1.5) [neuron] (Hn) {$H_n$};
    \node[draw=red!80,text=green!70] at (6.5,1.5) (H_name) {\textbf{Hidden layer}};
	
	\draw[fill=blue!30, rounded corners] (0.5, -0.2) rectangle (5.5, 0.2) {};
    \foreach \x in {1,...,3}
    	\draw node at (\x*0.75, 0) [neuron] (V\x) {$V_\x$};
    \foreach \x in {1,...,3}
    	\fill (2.5 +\x*0.25, 0) circle (1.25pt);    	
    \draw node at (3.75, 0) [neuron] (Vi) {$V_i$};
    \foreach \x in {1,...,3}
    	\fill (4 +\x*0.25, 0) circle (1.25pt);
    \draw node at (5.25, 0) [neuron] (Vm) {$V_m$};
    \node[draw=red!80,text=blue!70] at (6.5,0.0) (V_name) {\textbf{Visible layer}};

    \foreach \visible in {1, 2, 3, i, m}
    	\foreach \hidden in {1, 2, j, n}
    		\path (V\visible) edge (H\hidden);
    \end{tikzpicture}
\end{figure}

We would like to model the distribution of $m$ visible variables $V=\left(
V_{1},V_{2},\ldots ,V_{m}\right) $ representing the observable data whose
elements $V_{i}$ are highly dependent. A first way to directly model these
dependencies is to introduce a Markov chain or a Bayesian network. In this
case, networks are no longer restricted and those methods are computationally
expensive particularly when $V$ is a high-dimensional vector. The RBM
approach consists in introducing hidden variables $H=\left(
H_{1},H_{2},\ldots ,H_{n}\right) $ as latent variables which will indirectly
capture dependencies. Therefore, the hidden layer can be considered as an
alternative representation of the visible layer.

\subsubsection{Bernoulli RBMs}
\label{section:bernoulli-rbm}

\paragraph{Definition}

A Bernoulli RBM is the standard type of RBMs and has binary-valued visible
and hidden variables. Let us denote by $v=\left( v_{1},v_{2},\ldots
,v_{m}\right) $ and $h=\left( h_{1},h_{2},\ldots ,h_{n}\right) $ the
configurations of visible variables $V=\left( V_{1},V_{2},\ldots
,V_{m}\right) $ and hidden variables $H=\left( H_{1},H_{2},\ldots
,H_{n}\right) $, where $v_{i}$ and $h_{j}$ are the binary states of the
$i^{\mathrm{th}}$ visible variable $V_{i}$ and the $j^{\mathrm{th}}$ hidden
variable $H_{j}$ such that $\left( v,h\right) \in \left\{ 0,1\right\}
^{m+n}$. The joint probability distribution of a Bernoulli RBM is given by
the Boltzmann distribution:
\begin{equation*}
\mathbb{P}\left( v,h\right) =\frac{1}{Z}e^{-E\left( v,h\right) }
\end{equation*}%
where the energy function $E\left( v,h\right) $ is defined as:
\begin{eqnarray}
E\left( v,h\right)
&=&-\sum_{i=1}^{m}a_{i}v_{i}-\sum_{j=1}^{n}b_{j}h_{j}-\sum_{i=1}^{m}%
\sum_{j=1}^{n}w_{i,j}v_{i}h_{j}  \notag \\
&=&-a^{\top }v-b^{\top }h-v^{\top }Wh  \label{eq:Energyfunction}
\end{eqnarray}%
where $a=\left( a_{i}\right) $ and $b=\left( b_{j}\right) $ are the two
vectors of bias terms associated with $V$ and $H$ and $W=\left(
w_{i,j}\right) $ is the matrix of weights associated with the edges between
$V$ and $H$. The normalizing constant $Z$ is the partition function and
ensures the overall distribution sums to one\footnote{We have
$Z=\sum_{v,h}e^{-E\left( v,h\right) }$.}. It follows that the marginal
probability distributions of the visible and hidden unit states are:
\begin{equation*}
\mathbb{P}\left( v\right) =\frac{1}{Z}\sum_{h}e^{-E\left( v,h\right) }
\end{equation*}%
and:%
\begin{equation*}
\mathbb{P}\left( h\right) =\frac{1}{Z}\sum_{v}e^{-E\left( v,h\right) }
\end{equation*}
The underlying idea of a Bernoulli RBM is to learn the unconditional
probability distributions $\mathbb{P}\left( v\right) $ of the observable
data.

\paragraph{Conditional Distributions}

According to \citet{Long-2010}, the partition function $Z$ is intractable in
the case of Bernoulli RBMs, since its calculus requires summing up $2^{m+n}$
elements. Therefore, the probability distribution $\mathbb{P}\left( v\right)$
is also intractable when $m+n$ increases. However, we can take advantage of
the property of the bipartite graph structure of the RBM. Indeed, there are
no connections between two different units in the same layer. The
probabilities $\mathbb{P}\left( v_{i}\mid h\right) $ and $\mathbb{P}\left(
h_{j}\mid v\right) $ are then independent for all $i$ and $j$. It follows
that:%
\begin{equation*}
\mathbb{P}\left( v\mid h\right) =\mathbb{P}\left( v_{1},v_{2},\ldots
,v_{m}\mid h\right) =\prod_{i=1}^{m}\mathbb{P}\left( v_{i}\mid h\right)
\end{equation*}%
and:%
\begin{equation*}
\mathbb{P}\left( h\mid v\right) =\mathbb{P}\left( h_{1},h_{2},\ldots
,h_{n}\mid v\right) =\prod_{j=1}^{n}\mathbb{P}\left( h_{j}\mid v\right)
\end{equation*}
With these properties, we can find some useful results that help when
computing the gradient of the log-likelihood function on page
\pageref{section:bernoulli-rbm-training}. For instance, we can show
that\footnote{See Appendix \ref{appendix:rbm-cp} on page
\pageref{appendix:rbm-cp}.}:
\begin{equation*}
\sum_{h}\mathbb{P}\left( h\mid v\right) h_{j}=\mathbb{P}\left( h_{j}=1\mid
v\right)
\end{equation*}

\paragraph{A neural network perspective of RBMs}

In Appendix \ref{appendix:rbm-nn} on page \pageref{appendix:rbm-nn}, we show
that:
\begin{equation}
\mathbb{P}\left( v_{i}=1\mid h\right) =\sigma \left(
a_{i}+\sum_{j=1}^{n}w_{i,j}h_{j}\right) \label{eq:rbm-pr-visible}
\end{equation}%
and:%
\begin{equation}
\mathbb{P}\left( h_{j}=1\mid v\right) =\sigma \left(
b_{j}+\sum_{i=1}^{m}w_{i,j}v_{i}\right) \label{eq:rbm-pr-hidden}
\end{equation}%
where $\sigma \left( x\right) $ is the sigmoid function:%
\begin{equation*}
\sigma \left( x\right) =\frac{1}{1+e^{-x}}
\end{equation*}%
Thus, a Bernoulli RBM can be considered as a stochastic artificial neural
network, meaning that the nodes and edges correspond to neurons and synaptic
connections. For a given vector $v$, $h$ is obtained as follows:
\begin{equation*}
h_{j}=f\left( \sum_{i=1}^{m}w_{i,j}v_{i}+b_{j}\right)
\end{equation*}%
where $f=\varphi \circ \sigma $, $\varphi :x\in \left[ 0:1\right] \mapsto
X\sim \mathcal{B}\left( x\right) $ is a binarizer function and $\sigma $ is
the sigmoid activation function. In addition, we may also go backward in the
neural network as follows:%
\begin{equation*}
v_{i}=f\left( \sum_{j=1}^{n}w_{i,j}h_{j}+a_{i}\right)
\end{equation*}%
According to \citet{Fischer-2014}, \textquotedblleft \textsl{an RBM can
[then] be reinterpreted as a standard feed-forward neural network with one
layer of nonlinear processing units}\textquotedblright .

\paragraph{Training process}
\label{section:bernoulli-rbm-training}

Let $\theta =\left( a,b,W\right) $ be the set of parameters to estimate. The
objective is to find a value of $\theta $ such that $\mathbb{P}_{\theta
}\left( v\right) \approx \mathbb{P}_{\mathrm{data}}\left( v\right) $. Since
the log-likelihood function $\ell \left( \theta \mid v\right) $ of the input
vector $v$ is defined as $\ell \left( \theta \mid v\right) =\log
\mathbb{P}_{\theta }\left( v\right) $, the Bernoulli RBM model is trained in
order to maximize the log-likelihood function of a training set of $N$
samples $\left\{ v_{\left( 1\right) },\ldots ,v_{\left( N\right) }\right\} $:
\begin{equation}
\theta ^{\star }=\arg \max_{\theta }\sum_{s=1}^{N}\ell \left( \theta \mid
v_{\left( s\right) }\right)   \label{eq:rbm-ml1}
\end{equation}%
where:%
\begin{eqnarray}
\ell \left( \theta \mid v\right)  &=&\log \left( \sum_{h}\frac{e^{-E\left(
v,h\right) }}{Z}\right)  \notag \\
&=&\log \left( \sum_{h}e^{-E\left( v,h\right) }\right) -\log \left(
\sum_{v^{\prime },h}e^{-E\left( v^{\prime },h\right) }\right) \label{eq:rbm-ml2}
\end{eqnarray}
\citet{Hinton-2002} proposed to use gradient ascent method with the following
update rule between iteration steps $t$ and $t+1$:
\begin{eqnarray*}
\theta ^{\left( t+1\right) } &=&\theta ^{\left( t\right) }+\eta ^{\left(
t\right) }\frac{\partial }{\partial \,\theta }\left(
\sum_{s=1}^{N}\ell \left( \theta ^{\left( t\right) }\mid v_{\left( s\right)
}\right) \right)  \\
&=&\theta ^{\left( t\right) }+\eta ^{\left( t\right) }\Delta \theta ^{\left(
t\right) }
\end{eqnarray*}%
where $\eta ^{\left( t\right) }$ is the learning rate parameter, $\Delta
\theta ^{\left( t\right) }=\sum_{s=1}^{N}\nabla _{\theta ^{\left( t\right)
}}\left( v_{\left( s\right) }\right) $ and $\nabla _{\theta }\left( v\right)
$ is the gradient vector given in Appendix \ref{appendix:rbm-logl-gradient}
on page \pageref{appendix:rbm-logl-gradient}.

\paragraph{Gibbs sampling}

\citet{Ackley-1985} and \citet{Hinton-1986} showed that the expectation over
$\mathbb{P}\left( v\right) $ can be approximated by Gibbs sampling, which
belongs to the family of MCMC algorithms. The goal of Gibbs sampling is to
simulate correlated random variables by using a Markov chain. Usually, we
initialize the Gibbs sampling with a random vector and the algorithm updates one variable iteratively, based on its conditional distribution given the
state of the remaining variables. After a sufficiently large number of
sampling steps, we get the unbiased samples from the joint probability
distribution. Formally, Gibbs sampling of the joint probability distribution
of $n$ random variables $X=\left( X_{1},X_{2},\ldots ,X_{n}\right) $ consists
in sampling $x_{i}\sim \mathbb{P}\left( X_{i}\mid
X_{-i}=x_{-i}\right) $ iteratively.

\begin{algorithm}[tbh]
\caption{Gibbs sampling}
\label{alg:gibbs}
\begin{algorithmic}
    \STATE \textbf{initialization}: $x^{\left( 0\right) }=\left( x_{1}^{\left( 0\right)},x_{2}^{\left( 0\right) },\ldots ,x_{n}^{\left( 0\right) }\right) $
    \FOR{step $k= 1, 2, 3, \ldots $}
    \STATE Sample $x^{\left( k\right) }$ as follows:
        \begin{eqnarray*}
        x_{1}^{(k)} &\sim &\mathbb{P}\left( X_{1}\mid x_{2}^{\left( k-1\right)
        },x_{3}^{\left( k-1\right) },\ldots ,x_{n}^{\left( k-1\right) }\right)  \\
        x_{2}^{(k)} &\sim &\mathbb{P}\left( X_{2}\mid x_{1}^{\left( k\right)
        },x_{3}^{\left( k-1\right) },\ldots ,x_{n}^{\left( k-1\right) }\right)  \\
        &&\vdots  \\
        x_{i}^{(k)} &\sim &\mathbb{P}\left( X_{i}\mid x_{1}^{\left( k\right)
        },\ldots ,x_{i-1}^{\left( k\right) },x_{i+1}^{\left( k-1\right) },\ldots
        ,x_{n}^{\left( k-1\right) }\right)  \\
        &&\vdots  \\
        x_{n}^{(k)} &\sim &\mathbb{P}\left( X_{n}\mid x_{1}^{\left( k\right)
        },x_{2}^{\left( k\right) },\ldots ,x_{n-1}^{\left( k\right) }\right)
        \end{eqnarray*}
    \ENDFOR
\end{algorithmic}
\end{algorithm}

Let us consider a Bernoulli RBM as a Markov random field defined by the
random variables $V=\left( V_{1},V_{2},\ldots ,V_{m}\right) $ and $H=\left(
H_{1},H_{2},\ldots ,H_{n}\right) $. Since an RBM is a bipartite undirected
graph, we can take advantage of conditional independence properties between
the variables in the same layer. At each step, we jointly sample the states
of all variables in one layer, as shown in Figure \ref{fig:gibbs}. Thus, Gibbs
sampling for an RBM consists in alternating\footnote{This method is also
known as block Gibbs sampling.} between sampling a new state $h$ for all
hidden units based on $\mathbb{P}\left( h\mid v\right) $ and sampling a new
state $v$ for all visible units based on $\mathbb{P}\left( v\mid h\right) $.
Let $v^{\left( k\right) }=\left( v_{1}^{\left( k\right) },v_{2}^{\left(
k\right) },\ldots ,v_{m}^{\left( k\right) }\right) $ and $h^{\left( k\right)
}=\left( h_{1}^{\left( k\right) },h_{2}^{\left( k\right) },\ldots
,h_{n}^{\left( k\right) }\right) $ denote the states of the visible layer and
the hidden layer at time step $k$. For each unit, we rely on the fact that
the conditional probabilities $\mathbb{P}\left( v_{i}^{\left( k\right) }\mid
h^{\left( k\right) }\right) $ and $\mathbb{P}\left( h_{j}^{\left( k\right)
}\mid v^{\left( k-1\right) }\right) $ are easily tractable according to
Equations (\ref{eq:rbm-pr-visible}) and (\ref{eq:rbm-pr-hidden}). We start by
initializing the state of the visible units and we can choose a binary random
vector for the first time step. At time step $k$, here are the steps of the
algorithm:

\begin{enumerate}
\item We go forward in the network by computing simultaneously for each
hidden unit $j$ the following probability $p\left( h_{j}^{\left( k\right)
}\right) =\mathbb{P}\left( h_{j}^{\left( k\right) }\mid v^{\left( k-1\right)
}\right) $. The state of the $j^{\mathrm{th}}$ is then simulated according
to the Bernoulli distribution $\mathcal{B}\left( p\left( h_{j}^{\left(
k\right) }\right) \right) $.

\item We go backward in the network by computing simultaneously for each
visible unit $i$ the following probability $p\left( v_{i}^{\left( k\right)
}\right) =\mathbb{P}\left( v_{i}^{\left( k\right) }\mid h^{\left( k\right)
}\right) $. The state of the $i^{\mathrm{th}}$ unit is then simulated
according to the Bernoulli distribution $\mathcal{B}\left( p\left(
v_{i}^{\left( k\right) }\right) \right) $.
\end{enumerate}

\tikzstyle{neuron}=[draw,circle,minimum size=20pt, inner sep=0pt, fill=white]
\tikzstyle{stateTransition}=[thick]
\tikzstyle{learned}=[text=red]
\tikzstyle{place}=[circle, draw=black, minimum size = 8mm]

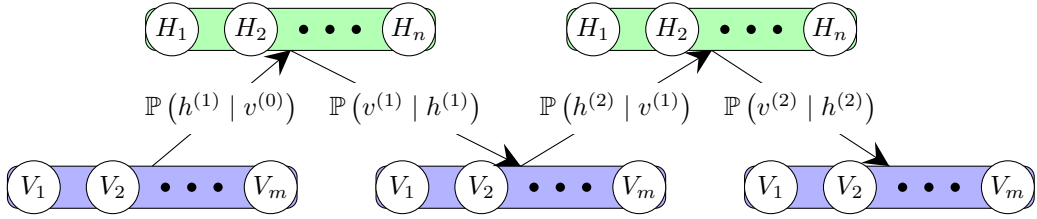
\begin{figure}[!h]
\centering
\caption{The schema of blocks Gibbs sampling for an RBM}
\label{fig:gibbs}
\begin{tikzpicture}[scale=1.4]
	\draw[fill=green!30, rounded corners] (0.8125, 1.3) rectangle (3.5625, 1.7) {};
    \draw node at (1.0625, 1.5) [neuron] (H11) {$H_1$};
    \draw node at (1.8125, 1.5) [neuron] (H12) {$H_2$};
    \foreach \x in {1,...,3}
    	\fill (2.0625 +\x*0.25, 1.5) circle (1.25pt);
    \draw node at (3.3125, 1.5) [neuron] (H1n) {$H_n$};
	
	\draw[fill=green!30, rounded corners] (0.8125 + 4, 1.3) rectangle (3.5625 + 4, 1.7) {};
    \draw node at (1.0625 + 4, 1.5) [neuron] (H21) {$H_1$};
    \draw node at (1.8125 + 4, 1.5) [neuron] (H22) {$H_2$};
    \foreach \x in {1,...,3}
    	\fill (2.0625 + 4 +\x*0.25, 1.5) circle (1.25pt);
    \draw node at (3.3125 + 4, 1.5) [neuron] (H2n) {$H_n$};
	
	\draw[fill=blue!30, rounded corners] (0.5-1, -0.2) rectangle (3.25-1, 0.2) {};
    \foreach \x in {1,...,2}
    	\draw node at (\x*0.75-1, 0) [neuron] (V1\x) {$V_\x$};
    \foreach \x in {1,...,3}
    	\fill (1.75 +\x*0.25 - 1, 0) circle (1.25pt);
   	\draw node at (3-1, 0) [neuron] (V1m) {$V_m$};  	

	\draw[fill=blue!30, rounded corners] (0.5+2.5, -0.2) rectangle (3.25+2.5, 0.2) {};
    \foreach \x in {1,...,2}
    	\draw node at (\x*0.75+2.5, 0) [neuron] (V2\x) {$V_\x$};
    \foreach \x in {1,...,3}
    	\fill (1.75 +\x*0.25 +2.5, 0) circle (1.25pt);
   	\draw node at (3+2.5, 0) [neuron] (V2m) {$V_m$};  	

	\draw[fill=blue!30, rounded corners] (0.5+6, -0.2) rectangle (3.25+6, 0.2) {};
    \foreach \x in {1,...,2}
    	\draw node at (\x*0.75+6, 0) [neuron] (V3\x) {$V_\x$};
    \foreach \x in {1,...,3}
    	\fill (1.75 +\x*0.25 +6, 0) circle (1.25pt);
   	\draw node at (3+6, 0) [neuron] (V3m) {$V_m$};

    \draw [decoration={markings,mark=at position 1 with {\arrow[scale=3,>=stealth]{>}}},postaction={decorate}] (0.875, 0.2) -- (2.1875, 1.3) node [midway, fill = white] {$\mathbb{P}\left( h^{(1)} \mid v^{(0)} \right)$} ;
    \draw [decoration={markings,mark=at position 1 with {\arrow[scale=3,>=stealth]{>}}},postaction={decorate}] (2.1875, 1.3) -- (0.875+3.5, 0.2)node [midway, fill = white] {$\mathbb{P}\left( v^{(1)} \mid h^{(1)}  \right)$} ;
    \draw [decoration={markings,mark=at position 1 with {\arrow[scale=3,>=stealth]{>}}},postaction={decorate}] (0.875+3.5, 0.2) -- (2.1875+4, 1.3)node [midway, fill = white] {$\mathbb{P}\left( h^{(2)} \mid v^{(1)}  \right)$} ;
    \draw [decoration={markings,mark=at position 1 with {\arrow[scale=3,>=stealth]{>}}},postaction={decorate}] (2.1875+4, 1.3) -- (0.875+7, 0.2)node [midway, fill = white] {$\mathbb{P}\left( v^{(2)} \mid h^{(2)}  \right)$} ;
    \end{tikzpicture}
\end{figure}

\paragraph{Contrastive divergence algorithm}

As shown in the previous paragraph, it is possible to use Gibbs sampling to
get samples from the joint distribution $\mathbb{P}\left( v, h \right)$.
However, the computational effort is still too large since the sampling chain
needs many sampling steps to obtain unbiased samples. To address this issue,
\citet{Hinton-2002} initialized the Gibbs sampling with a sample from the
real data, instead of using a random vector and suggested the use of only a few
sampling steps to get a sample that could produce a sufficiently good
approximation of log-likelihood gradient. This faster method is called
contrastive divergence algorithm.\smallskip

Given a training set of $N$ samples $\left\{ v_{\left( 1\right) },\ldots
,v_{\left( N\right) }\right\} $, we obtain:
\begin{eqnarray}
\frac{1}{N}\sum_{s=1}^{N}\ell \left( \theta \mid v_{\left( s\right) }\right)
&=&\frac{1}{N}\sum_{s=1}^{N}\log \mathbb{P}_{\theta }\left( v_{\left(
s\right) }\right)   \notag \\
&=&\mathbb{E}_{\mathbb{P}_{\mathrm{data}}}\left[ \log \mathbb{P}_{\theta
}\left( v\right) \right]   \label{eq:rbm-logl2}
\end{eqnarray}%
We note $\mathbb{P}_{\mathrm{model}}\left( v\right) =\mathbb{P}_{\theta
}\left( v\right) $. Since $\mathbb{P}_{\mathrm{data}}$ is independent of
$\theta $, maximizing the log-likelihood function (\ref{eq:rbm-ml2}) is
equivalent to minimizing the Kullback-Leibler divergence between
$\mathbb{P}_{\mathrm{data}}$ and $\mathbb{P}_{\mathrm{model}}$:
\begin{eqnarray}
\theta ^{\star } &=&\arg \max_{\theta }\mathbb{E}_{\mathbb{P}_{\mathrm{data}%
}}\left[ \log \mathbb{P}_{\theta }\left( v\right) \right] -\mathbb{E}_{%
\mathbb{P}_{\mathrm{data}}}\left[ \log \mathbb{P}_{\mathrm{data}}\left(
v\right) \right]   \notag \\
&=&\arg \max_{\theta }\sum_{v}\mathbb{P}_{\mathrm{data}}\left( v\right) \log
\mathbb{P}_{\mathrm{model}}\left( v\right) -\sum_{v}\mathbb{P}_{\mathrm{data}%
}\left( v\right) \log \mathbb{P}_{\mathrm{data}}\left( v\right)   \notag \\
&=&\arg \max_{\theta }\sum_{v}\mathbb{P}_{\mathrm{data}}\left( v\right) \log
\left( \frac{\mathbb{P}_{\mathrm{model}}\left( v\right) }{\mathbb{P}_{%
\mathrm{data}}\left( v\right) }\right)   \notag \\
&=&\arg \max_{\theta }-\func{KL}\left( \mathbb{P}_{\mathrm{data}}\parallel
\mathbb{P}_{\mathrm{model}}\right)   \notag \\
&=&\arg \min_{\theta }\func{KL}\left( \mathbb{P}_{\mathrm{data}}\parallel
\mathbb{P}_{\mathrm{model}}\right)   \label{eq:rbm-kl1}
\end{eqnarray}%
\smallskip

In contrastive divergence algorithms, we note the distribution of
starting values as $\mathbb{P}^{\left( 0\right) }$, which is also
the distribution of real data in training set: $\mathbb{P}^{\left(
0\right) }=\mathbb{P}_{\mathrm{data}}$. Let $\mathbb{P}^{\left(
k\right) }$ and $\mathbb{P}^{\left( \infty \right) }$ be the
distribution after running $k$ steps of Gibbs sampling and the
equilibrium distribution. Compared to $\mathbb{P}^{\left( 0\right)
}$, $\mathbb{P}^{\left( k\right) }$ is $k$ steps closer to the
equilibrium distribution $\mathbb{P}^{\left( \infty \right) }$, so
the divergence measure $\func{KL}\left( \mathbb{P}^{\left( 0\right)
}\parallel \mathbb{P}^{\left( \infty \right) }\right) $ should be
greater than or equal to $\func{KL}\left( \mathbb{P}^{\left(
k\right) }\parallel \mathbb{P}^{\left( \infty \right) }\right) $. In
particular, when the model is well-trained, $\mathbb{P}^{\left(
0\right) }$, $\mathbb{P}^{\left( k\right) }$, and
$\mathbb{P}^{\left( \infty \right) }$ should have the same
distribution as $\mathbb{P}_{\mathrm{data}}$. In this case,
$\func{KL}\left( \mathbb{P}^{\left( 0\right) }\parallel
\mathbb{P}^{\left( \infty \right) }\right) $ is equal to 0 and the
difference $\func{KL}\left( \mathbb{P}^{\left( 0\right) }\parallel
\mathbb{P}^{\left( \infty \right) }\right) -\func{KL}\left(
\mathbb{P}^{\left( k\right) }\parallel \mathbb{P}^{\left( \infty
\right) }\right) $ is also equal to 0. Thus, according to
\citet{Hinton-2002}, we can find the optimal parameter $\theta
^{\star }$ by minimizing the difference $\func{KL}\left(
\mathbb{P}^{\left( 0\right) }\parallel \mathbb{P}^{\left( \infty
\right) }\right) -\func{KL}\left( \mathbb{P}^{\left( k\right)
}\parallel \mathbb{P}^{\left( \infty \right) }\right) $ instead of
minimizing directly $\func{KL}\left( \mathbb{P}^{\left( 0\right)
}\parallel \mathbb{P}^{\left( \infty \right) }\right) $ in Equation
(\ref{eq:rbm-kl1}). This quantity is called a contrastive divergence
since we compare two KL divergence measures. Therefore, we can
rewrite the objective function in Equation (\ref{eq:rbm-kl1}) as
follows:
\begin{eqnarray}
\theta ^{\star } &=&\arg \min_{\theta }\limfunc{CD}\nolimits^{\left(
k\right) } \notag \\
&=&\arg \min_{\theta }\func{KL}\left( \mathbb{P}^{\left( 0\right) }\parallel
\mathbb{P}^{\left( \infty \right) }\right) -\func{KL}\left( \mathbb{P}%
^{\left( k\right) }\parallel \mathbb{P}^{\left( \infty \right) }\right) \label{eq:kl2}
\end{eqnarray}%
Again, we can use the gradient descent method to find the minimum of the
objective function. In this case, the update rule between iteration steps $t$
and $t+1$ is:
\begin{equation*}
\theta ^{\left( t+1\right) }=\theta ^{\left( t\right) }-\eta ^{\left(
t\right) }\frac{\partial \,\limfunc{CD}\nolimits^{\left( k\right) }\left(
\theta ^{\left( t\right) }\right) }{\partial \,\theta }
\end{equation*}%
where the gradient vector $\partial _{\theta }\limfunc{CD}\nolimits^{\left(
k\right) }\left( \theta ^{\left( t\right) }\right) $ is given in Appendix
\ref{appendix:rbm-cd-gradient} on page \pageref{appendix:rbm-cd-gradient}.
Thus, Algorithm (\ref{alg:cd}) summarizes the $k$-step contrastive divergence
algorithm for calibrating $\theta ^{\star }$.

\begin{algorithm}[tbh]
    \caption{$k$-step contrastive divergence algorithm}
    \label{alg:cd}
    \begin{algorithmic}

    \STATE \textbf{input}: $\left\{ v_{\left( 1\right) },\ldots ,v_{\left( N\right) }\right\} $
    \STATE \textbf{initialization}: We note $\Delta a_{i}=\partial _{a_{i}}\limfunc{CD}\nolimits^{\left(k\right) }$,
        $\Delta b_{j}=\partial _{b_{i}}\limfunc{CD}\nolimits^{\left(k\right) }$ and
        $\Delta w_{i,j}=\partial _{w_{i,j}}\limfunc{CD}\nolimits^{\left( k\right) }$,
        and we set $\Delta a_{i}=\Delta b_{j}=\Delta w_{i,j}=0$

    \FOR{$v$ from $v_{\left( 1 \right)}$ to $v_{\left( N \right)}$}
        \STATE $v^{\left( 0 \right)}\leftarrow v$

        \STATE \COMMENT{Gibbs sampling}
        \FOR{$t=1$ to $k$}
            \FOR{$j=1$ to $n$}
                \STATE compute $p\left( h_j^{\left(t\right)} \right) = \mathbb{P}\left(h_{j}^{\left(t\right)} \mid v^{\left( t-1 \right)}\right)$
                \STATE sample the $j^{\mathrm{th}}$ hidden unit state: $h_j^{\left( t \right)} \sim \mathcal{B} \left( p\left( h_j^{\left(t\right)} \right) \right)$
            \ENDFOR
            \FOR{$i=1$ to $m$}
                \STATE compute $p\left(v_i^{\left(t\right)}\right) = \mathbb{P}\left(v_{i}^{\left(t\right)} \mid h^{\left( t \right)}\right)$
                \STATE sample the $i^{\mathrm{th}}$ visible unit state: $v_i^{\left( t \right)} \sim \mathcal{B} \left( p\left(v_i^{\left(t\right)}\right) \right)$
            \ENDFOR
        \ENDFOR

        \STATE \COMMENT{Gradient approximation}

        \FOR{$i=1$ to $m$}
            \STATE $\Delta a_{i}\leftarrow \Delta a_{i}+\dfrac{1}{N}\left( v_{i}^{\left(k\right) }-v_{i}^{\left( 0\right) }\right) $
            \FOR{$j=1$ to $n$}
                \STATE $\Delta b_{j}\leftarrow \Delta b_{j}+\dfrac{1}{N}\left( \mathbb{P}\left(
                        h_{j}=1\mid v^{\left( k\right) }\right) -\mathbb{P}\left( h_{j}=1\mid
                        v^{\left( 0\right) }\right) \right) $
                \STATE  $\Delta w_{i,j}\leftarrow \Delta w_{i,j}+\dfrac{1}{N}\left( \mathbb{P}\left(
                        h_{j}=1\mid v^{\left( k\right) }\right) \cdot v_{i}^{\left( k\right) }-%
                        \mathbb{P}\left( h_{j}=1\mid v^{\left( 0\right) }\right) \cdot v_{i}^{\left(0\right) }\right) $
            \ENDFOR
         \ENDFOR
    \ENDFOR
    \STATE \textbf{Output}: gradient estimation $\Delta a_i$, $\Delta b_j$ and $\Delta w_{i,j}$
\end{algorithmic}
\end{algorithm}

\subsubsection{Gaussian-Bernoulli RBMs}
\label{section:gaussian-rbm}

A Bernoulli RBM is limited to modelling the probability distribution
$\mathbb{P}\left( v\right) $ where $v$ is a binary vector. To address this
issue, given an RBM with $m$ visible units and $n$ hidden units, we can
associate a normally distributed variable to each visible unit and a binary
variable to each hidden unit. This type of RBMs is called Gaussian-Bernoulli
RBM. We are free to choose the expression of the energy function as long as
it satisfies the Hammersley-Clifford theorem\footnote{See Appendix
\ref{appendix:Hammersley-Clifford} on page
\pageref{appendix:Hammersley-Clifford}.} and its partition function is well
defined. For instance, \citet{Cho-2011} defined the energy function of this
RBM as follows:
\begin{equation*}
E_{g}\left( v,h\right) =\sum_{i=1}^{m}\frac{\left( v_{i}-a_{i}\right) ^{2}}{%
2\sigma _{i}^{2}}-\sum_{j=1}^{n}b_{j}h_{j}-\sum_{i=1}^{m}%
\sum_{j=1}^{n}w_{i,j}\frac{v_{i}h_{j}}{\sigma _{i}^{2}}
\end{equation*}%
where $a_{i}$ and $b_{j}$ are bias terms associated with visible variables
$V_{i}$ and hidden variables $H_{j}$, $w_{i,j}$ is the weight associated with
the edge between $V_{i}$ and $H_{j}$, and $\sigma _{i}$ is a new parameter
associated with $V_{i}$. Following the same calculus in Equation
\ref{eq:rbm-pr-hidden}, we can show that the conditional probability is equal
to:
\begin{equation}
\mathbb{P}\left( h_{j}=1\mid v\right) =\limfunc{sigmoid}\left(
b_{j}+\sum_{i=1}^{m}w_{i,j}\frac{v_{i}}{\sigma _{i}^{2}}\right)
\label{eq:grbm1}
\end{equation}%
Moreover, according to \citet{Krizhevsky-2009}, the conditional distribution
of $V_{i}$ given $h$ is Gaussian and we obtain:
\begin{equation}
V_{i}\mid h\sim \mathcal{N}\left( a_{i}+\sum_{j=1}^{n}w_{i,j}h_{j},\sigma
_{i}^{2}\right)   \label{eq:grbm2}
\end{equation}%
\smallskip

As we have previously seen, we can maximize the log-likelihood function in
order to train a Gaussian-Bernoulli RBM with $m$ visible units and $n$ hidden
units. As all visible units are associated with continuous probability
distribution, the log-likelihood function $\ell \left( \theta \mid v\right) $
of an input vector $v=\left( v_{1},v_{2},\ldots
,v_{m}\right) $ is equal to $\log p_{\theta }\left( v\right) $ where $%
p_{\theta }\left( v\right) $ is the probability density function of $%
V=\left( V_{1},V_{2},\ldots ,V_{m}\right) $. In Appendix \ref{appendix:gaussian-rbm-grad} on page \pageref{appendix:gaussian-rbm-grad}, we
give the expression of the gradient vector. Therefore, there is no difficulty
to use the gradient ascent method or the $k$-step contrastive divergence
algorithm to train a Gaussian-Bernoulli RBM.

\begin{remark}
If we normalize the data of the training set using a $z$-score function, we
can set the standard deviation $\sigma _{i}$ to 1 during the training
process. This reduces the number of parameter and accelerates the convergence
of the training process.
\end{remark}

\subsubsection{Conditional RBM structure}

Traditional Bernoulli and Gaussian-Bernoulli RBMs can only model the static
dependence of variables. However, in the practice of financial data modeling,
we also want to capture the temporal dependencies between variables. To address
this issue, \citet{Taylor-2011} introduced a conditional RBM structure by
adding a new layer to the Gaussian-Bernoulli RBM. Thus, this conditional RBM is
made of three parts as shown in Figure \ref{fig:crbm}:

\begin{enumerate}
\item a hidden layer with $n$ binary units;

\item a visible layer with $m$ units;

\item a conditional layer with $d$ conditional meta-units $\left\{
    C_{1},\ldots ,C_{d}\right\} $. Since we model the temporal structure,
    we note the observation at time $t$ as $v_{t}=\left(
    v_{t,1},v_{t,2},\ldots ,v_{t,m}\right) $. The conditional layer is fed
    with $d$ past values $\left(v_{t-1},\ldots ,v_{t-d}\right) $ that are
    concatenated into a $d\times m$-dimensional vector\footnote{A
    conditional meta-unit is then composed of $m$ values: $C_{k}=\left(
    V_{t-k,1}\ldots ,V_{t-k,m}\right) $.} and we note it as $c_{t}$. The
    conditional layer is fully linked to visible and hidden layers with
    directed connections. Let us denote by $Q=\left( q_{d\times m,m}\right)
    $ the weight matrix connecting the conditional layer to the visible
    layer and $P=\left( p_{d\times m,n}\right) $ the weight matrix
    connecting the conditional layer to the hidden layer.
\end{enumerate}
\bigskip

\tikzstyle{neuron}=[draw,circle,minimum size=30pt, inner sep=0pt, fill=white]
\tikzstyle{Cneuron}=[draw,ellipse,minimum size=60pt, inner sep=20pt, fill=white]
\tikzstyle{stateTransition}=[thick]
\tikzstyle{learned}=[text=red]
\tikzstyle{place}=[circle, draw=black, minimum size = 8mm]

\vspace*{-25pt}
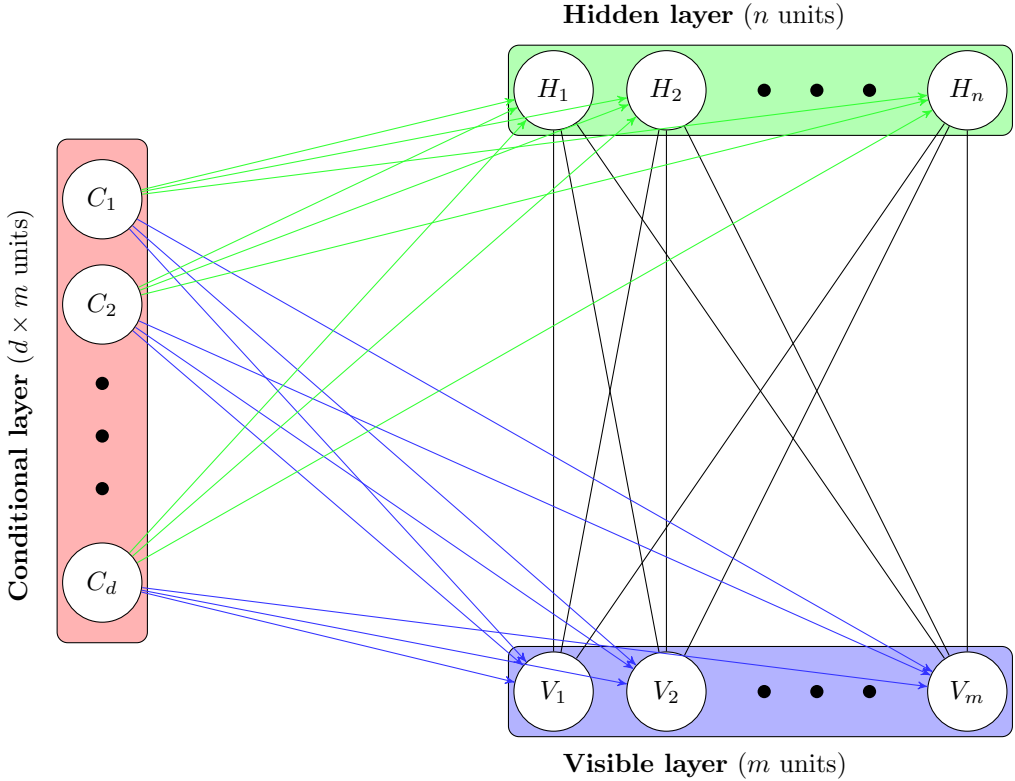
\begin{figure}[tbh]
\centering
\caption{The schema of a conditional RBM with $m$ visible layer units, $n$ hidden layer units and $d \times m$ conditional layer units}
\label{fig:crbm}

\begin{tikzpicture}[scale=2.0]
	\draw[fill=green!30, rounded corners] (0.7, 1.7) rectangle (4.05, 2.3) {};
    \draw node at (1, 2) [neuron] (H1) {$H_1$};
    \draw node at (1.75, 2) [neuron] (H2) {$H_2$};
    \foreach \x in {1,...,3}
    	\fill (2.05 +\x*0.35, 2) circle (1.25pt);
    \draw node at (3.75, 2) [neuron] (Hn) {$H_n$};
    \node[above=0.45cm of H1,right] (H_name) {\textbf{Hidden layer} ($n$ units)};

	\draw[fill=red!30, rounded corners] (-2.3, -1.675) rectangle (-1.7, 1.675) {};
    \draw node at (-2.0, 1.275) [neuron] (C1) {$C_1$};
    \draw node at (-2.0, 0.575) [neuron] (C2) {$C_2$};
    \draw node at (-2.0,-1.275) [neuron] (Cd) {$C_d$};
   	\foreach \x in {1,...,3}
    	\fill (-2.0,-1.00+\x*0.35) circle (1.25pt);
	\node[left=0.55cm of C1, rotate=90] (C_name) {\textbf{Conditional layer} ($d \times m$ units)};
    	
	\draw[fill=blue!30, rounded corners] (0.7, -1.7) rectangle (4.05, -2.3) {};
    \draw node at (1, -2) [neuron] (V1) {$V_1$};
    \draw node at (1.75, -2) [neuron] (V2) {$V_2$};
    \foreach \x in {1,...,3}
    	\fill (2.05 +\x*0.35, -2) circle (1.25pt);
    \draw node at (3.75, -2) [neuron] (Vm) {$V_m$};
    \node[below=0.45cm of V1,right] (V_name) {\textbf{Visible layer} ($m$ units)};

    \foreach \visible in {1, 2, m}
    	\foreach \hidden in {1, 2, n}
    		\path (V\visible) edge (H\hidden);
    		
    \foreach \conditional in {1, 2, d}
		\foreach \visible in {1, 2, m}
 			\path[->,-stealth',blue!80] (C\conditional) edge (V\visible);	
			
    \foreach \conditional in {1, 2, d}
		\foreach \hidden in {1, 2, n}
			\path[->,-stealth',green!80] (C\conditional) edge (H\hidden);	
			
\end{tikzpicture}

\end{figure}

A conditional RBM contains both undirected and directed connections in the
graph. Thus, it can't be defined as an MRF or Bayesian network. However,
conditionally on $c_{t}$, we can consider both visible and hidden layers as
an undirected graph and we can compute $\mathbb{P}\left( v_{t},h_{t}\mid
c_{t}\right) $ instead of computing $\mathbb{P}\left(
v_{t},h_{t},c_{t}\right) $ in order to take still advantage of undirected
graph properties. Thus, according to the Hammersley-Clifford theorem, the
conditional probability distribution has the following form:
\begin{eqnarray*}
\mathbb{P}\left( v_{t},h_{t}\mid c_{t}\right)  &=&\frac{1}{Z}e^{-\tilde{E}%
_{g}\left( v_{t},h_{t},c_{t}\right) } \\
&=&\frac{1}{Z}e^{-\sum_{i=1}^{m}\tilde{E}_{i}\left( v_{t,i},c_{t}\right)
-\sum_{j=1}^{n}\tilde{E}_{j}\left( h_{t,j},c_{t}\right)
-\sum_{i=1}^{m}\sum_{j=1}^{n}\tilde{E}_{i,j}\left(
v_{t,i},h_{t,j},c_{t}\right) }
\end{eqnarray*}
\newpage
\noindent where $Z$ is the partition function. \citet{Taylor-2011} proposed a
form of energy function that is an extension of the Gaussian-Bernoulli RBM
energy:
\begin{eqnarray*}
\tilde{E}_{i}\left( v_{t,i},c_{t}\right)  &=&\frac{\left( v_{t,i}-\tilde{a}%
_{t,i}\right) ^{2}}{2\sigma _{i}^{2}} \\
\tilde{E}_{j}\left( h_{t,j},c_{t}\right)  &=&-\tilde{b}_{t,j}h_{t,j} \\
\tilde{E}_{i,j}\left( v_{t,i},h_{t,j},c_{t}\right)  &=&-w_{i,j}\frac{%
v_{t,i}h_{t,j}}{\sigma _{i}^{2}}
\end{eqnarray*}%
where $\tilde{a}_{t}=a+c_{t}Q^{\top }$\ and $\tilde{b}_{t}=b+c_{t}P^{\top }$
are dynamic bias terms with respect to $a$ and $b$. Thus, the energy function
$\tilde{E}_{g}$ corresponds to a Gaussian-Bernoulli RBM energy function
$E_{g}$ by replacing constant biases $a$ and $b$ by dynamic bias
$\tilde{a}_{t}$ and $\tilde{b}_{t}$. By updating these terms in Equations
(\ref{eq:grbm1}) and (\ref{eq:grbm2}), we obtain:
\begin{equation}
\mathbb{P}\left( h_{j}=1\mid v_{t},c_{t}\right) =\limfunc{sigmoid}\left(
\tilde{b}_{t,j}+\sum_{i=1}^{m}w_{i,j}\frac{v_{t,i}}{\sigma _{i}^{2}}\right)
\label{eq:crbm1}
\end{equation}%
and:
\begin{equation}
V_{t,i}\mid h_{t}, c_{t} \sim \mathcal{N}\left( \tilde{a}_{t,i}+%
\sum_{j=1}^{n}w_{i,j}h_{t,j},\sigma _{i}^{2}\right)   \label{eq:crbm2}
\end{equation}%
Again, the partition function $Z$ is still intractable:
\begin{equation*}
Z=\sum_{h_{t}}\int_{v_{t}}e^{-\tilde{E}_{g}(v_{t},h_{t},c_{t}})\,\mathrm{d}v_{t}
\end{equation*}

\begin{remark}
As for Gaussian-Bernoulli RBMs, we can use the $k$-step contrastive
divergence algorithm for the training process with some modifications of the
gradient vector\footnote{The calculations are given in Appendix
\ref{appendix:cond-rbm-grad} on page \pageref{appendix:cond-rbm-grad}.}.
\end{remark}

\subsection{Generative adversarial networks}

Generative models are an important part of machine learning algorithms that
learn the underlying probability distributions of the real data sample. In
other words, given a finite sample data with a distribution
$\mathbb{P}_{\mathrm{data}}\left( x\right) $, can we build a model such that
$\mathbb{P}_{\mathrm{model}}\left( x;\theta \right) \approx
\mathbb{P}_{\mathrm{data}}\left( x\right) $? The goal is to learn to sample a
complex distribution given a real sample. Generative adversarial networks
(GANs) belong to the class of generative models and move away from the
classical likelihood maximization approach, whose objective is to estimate the
parameter $\theta $. GAN models enable the estimation of the high dimensional
underlying statistical structure of real data and simulate synthetic data,
whose probability distribution is close to the real one. To assess the
difference between real and simulated data, GANs are trained using a
discrepancy measure. Two widely classes of discrepancy measures are
information-theoretic divergences and integral probability metrics. The choice
of the GAN objective function associated with the selected discrepancy measure
explains the multitude of GAN models appearing in the machine learning
literature. However, the different GAN models share a common framework. Indeed,
Appendix \ref{appendix:unified} page \pageref{appendix:unified} shows how the
original formulation of GAN models is a particular case of the theory of
$\phi$-divergence and probability functional descent. Moreover, the link
between influence function used in robust statistics and the formulation of the
discriminator and generator is derived.\smallskip

According to \citet{Goodfellow-2014}, \textquotedblleft \textsl{a generative
adversarial process trains two models: a generative model $\Generator$ that
captures the data distribution, and a discriminative model $\Discriminator$
that estimates the probability that a sample came from the training data rather
than $\Generator$. The training procedure for $\Generator$ is to maximize the
probability of $\Discriminator$ making a mistake. This framework corresponds to
a minimax two-player game. In the space of arbitrary functions $\Generator$ and
$\Discriminator$, a unique solution exists, with $\Generator$ recovering the
training data distribution and $\Discriminator$ equal to $\nicefrac{1}{2}$
everywhere}\textquotedblright. Therefore, a generative adversarial model
consists of two neural networks. The first neural network simulates a new data
sample, and is interpreted as a data generation process. The second neural
network is a classifier. The input data are both the real and simulated
samples. If the generative model has done a good job, the discriminative model
is unable to know whether an observation comes from the real dataset or the
simulated dataset.

\subsubsection{Adversarial training problem}

We use the framework of \citet{Goodfellow-2014} and \citet{Wiese-2019}. Let
$\mathcal{Z}$ be a random noise space. $z\in \mathcal{Z}$ is sampled from the
prior distribution $\mathbb{P}_{\mathrm{noize}}\left( z\right)$. The generative
model $\Generator$ is then specified as follows:
\begin{equation}
\Generator:\left\{
\begin{array}{l}
\left( \mathcal{Z},\Theta _{\Generator}\right) \longrightarrow \mathcal{X} \\
\left( z,\theta _{g}\right) \longmapsto x_{0}=\Generator\left( z;\theta_{g}\right)
\end{array}%
\right.  \label{eq:gan-generator}
\end{equation}%
where $\mathcal{X}$ denotes the data space and $\Theta _{\Generator}$ defines
the parameter space including generator weights and bias that will be
optimized. The generator $\Generator$ helps to simulate data $x_0$. The
discriminative model is defined as follows:
\begin{equation}
\Discriminator:\left\{
\begin{array}{l}
\left( \mathcal{X},\Theta _{\Discriminator}\right) \longrightarrow \left[ 0,1\right]  \\
\left( x,\theta _{d}\right) \longmapsto p=\Discriminator\left( x;\theta_{d}\right)
\end{array}%
\right. \label{eq:gan-discriminator}
\end{equation}%
where $x$ corresponds to the set of simulated data $x_{0}$ and training (or
real) data $x_{1}$. In this approach, the statistical model of $X$
corresponds to:%
\begin{equation}
\mathbb{P}_{\textrm{model}}\left( x;\theta \right) \sim \Generator\left( \mathbb{P}_{\mathrm{%
noise}}\left( z\right) ;\theta _{g}\right)   \label{eq:gan-model}
\end{equation}%
The probability that the observation comes from the real (or true) data is
equal to $p_{1}=\Discriminator\left( x_{1};\theta _{d}\right) $, whereas the
probability that the observation is simulated is given by $p_{0}=%
\Discriminator\left( x_{0};\theta _{d}\right) $. If the model (\ref%
{eq:gan-model}) is wrong and does not reproduce the statistical properties of
the real data, the classifier has no difficulty in separating the simulated data
from the real data, and we obtain $p_{0}\approx 0$ and $p_{1}\approx 1$.
Otherwise, if the model is valid, we must verify that:%
\begin{equation*}
p_{0}\approx p_{1}\approx \frac{1}{2}
\end{equation*}

The main issue of GANs is the specification of the two functions
$\Generator$ and $\Discriminator$ and the estimation of the
parameters $\theta _{g}$ and $\theta _{d}$ associated to
$\Generator$ and $\Discriminator$. For the first step,
\citet{Goodfellow-2014} proposed to use two multi-layer neural
networks $\Generator$ and $\Discriminator$, whereas they consider
the following cost function for the second step:
\begin{eqnarray*}
\mathcal{C}\left( \theta _{g},\theta _{d}\right)  &=&\mathbb{E}\left[
\log \Discriminator\left( x_{1};\theta _{d}\right) \mid x_{1}\sim
\mathbb{P}_{\mathrm{data}}\left( x\right) \right] + \\
&&\mathbb{E}\left[ \log \left( 1-\Discriminator\left( x_{0};\theta
_{d}\right) \right) \mid x_{0} \sim \Generator\left( z;\theta _{g}\right)
,z\sim \mathbb{P}_{\mathrm{noise}}\left( z\right) \right]
\end{eqnarray*}%
The optimization problem becomes:
\begin{equation*}
\left\{ \hat{\theta}_{g},\hat{\theta}_{d}\right\} =\arg \, \underset{\theta
_{g}\in \Theta _{\Generator}}{\min }\, \underset{\theta _{d}\in \Theta _{%
\Discriminator}}{\max }\,\mathcal{C}\left( \theta _{g},\theta _{d}\right)
\end{equation*}%
In other words, the discriminator is trained in order to maximize the
probability to correctly classify historical samples from simulated samples.
The objective is to obtain a good classification model since the maximum value
$\mathcal{C}\left( \theta _{g},\theta _{d}\right) $ with respect to $\theta
_{d}$ is reached when:
\begin{equation*}
\left\{
\begin{array}{c}
\Discriminator\left( x_{1};\theta _{d}\right) =1 \\
\Discriminator\left( x_{0};\theta _{d}\right) =0%
\end{array}%
\right.
\end{equation*}%
In the meantime, the generator is trained in order to minimize the probability
that the discriminator is able to perform a correct classification or
equivalently to maximize the probability to fool the discriminator, since the
minimum value $\mathcal{C}\left( \theta _{g},\theta _{d}\right) $ with respect
to $\theta _{g}$ is reached when:
\begin{equation*}
\left\{
\begin{array}{l}
\Discriminator\left( x_{0};\theta _{d}\right) =1 \\
x_{0}\sim \Generator\left( z;\theta _{g}\right)
\end{array}%
\right.
\end{equation*}

\begin{remark}
In Appendix \ref{appendix:minimax} on page \pageref{appendix:minimax}, we show
that the cost function is related to the binary cross-entropy measure or the
opposite of the log-likelihood function of the logit model.
Moreover, the cost function can be interpreted as a $\phi$-divergence measure as
explained in Appendix \ref{appendix:phi-gan} on page \pageref{appendix:phi-gan}.
\end{remark}

\subsubsection{Solving the optimization problem}

\begin{figure}[tbh]
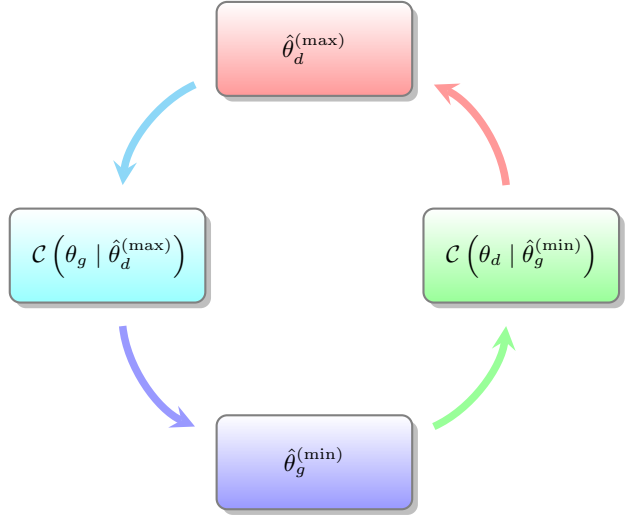

\centering
\caption{Two-stage minimax algorithm}
\label{fig:minimax1}
\figureskip
\smartdiagram[circular diagram]{
{$\hat{\theta}_{d}^{\left( \max \right) }$},
{$\mathcal{C}\left( \theta_{g} \mid \hat{\theta}_{d}^{\left( \max \right)}\right)$},
{$\hat{\theta}_{g}^{\left( \min \right) }$},
{$\mathcal{C}\left( \theta_{d} \mid \hat{\theta}_{g}^{\left( \min \right)}\right)$}}
\end{figure}

The minimax optimization problem is difficult to solve directly, because the
gradient vector $\nabla_{\theta}\,\mathcal{C}\left( \theta _{g},\theta
_{d}\right) $ is not well informative if the discriminative model is poor.
Therefore, the traditional way to solve this problem is to use a two-stage
approach:
\begin{enumerate}
\item In a first stage, the vector of parameters $\theta _{g}$ is considered
to be constant whereas the vector of parameters $\theta _{d}$ is unknown.
This implies that the minimax problem reduces to a maximization step:%
\begin{equation*}
\hat{\theta}_{d}^{\left( \max \right) }=\arg \underset{\theta _{d}\in \Theta
_{\Discriminator}}{\max }\mathcal{C}^{\left( \max \right) }\left( \theta _{d} \mid \theta _{g} \right)
\end{equation*}
where $\mathcal{C}^{\left( \max \right) }\left( \theta _{d} \mid \theta _{g}
\right)$ corresponds to the cost function by assuming that $\theta _{g}$ is
given.

\item In a second stage, the vector of parameters $\theta _{d}$ is
considered to be constant whereas the vector of parameters $\theta _{g}$ is
unknown. This implies that the minimax problem reduces to a minimization
step:
\begin{equation*}
\hat{\theta}_{g}^{\left( \min \right) }=\arg \underset{\theta _{g}\in \Theta
_{\Generator}}{\min }\mathcal{C}^{\left( \min \right) }\left( \theta _{g} \mid \theta _{d}\right)
\end{equation*}
where $\mathcal{C}^{\left( \min \right) }\left( \theta _{g} \mid \theta _{d}
\right)$ corresponds to the cost function by assuming that $\theta _{d}$ is
given.
\end{enumerate}
The two-stage approach is repeated until convergence by setting $\theta _{g}$
to the value $\hat{\theta}_{g}^{\left( \min \right) }$ calculated at the
minimization step and $\theta _{d}$ to the value $\hat{\theta}_{d}^{\left(
\max \right) }$ calculated at the maximization step. The cycle sequence is
given in Figure \ref{fig:minimax1}. A drawback of this approach is the
computational time. Indeed, this implies to solve two optimization problems
at each iteration.\smallskip

\smartdiagramset{module minimum width = 2.6cm, module minimum height = 1.25cm, text width = 2cm}

Therefore, \citet{Goodfellow-2014} proposed another two-stage
approach, which converges more rapidly. The underlying idea is not
to estimate the optimal generator at each iteration. The objective
is, rather, to improve the generative model at each iteration, such
that the new estimated model is better than the previous. The
convergence is only needed for the discriminator step
\citep[Proposition 2]{Goodfellow-2014}. Moreover, the authors
applied a mini-batch sampling in order to reduce the computational
time. It follows that the cost functions become:
\begin{equation*}
\mathcal{C}^{\left( \max \right) }\left( \theta _{d}\mid \theta _{g}\right)
\approx \frac{1}{m}\sum_{i=1}^{m}\log \Discriminator\left( x_{1}^{\left(
i\right) };\theta _{d}\right) +\log \left( 1-\Discriminator\left(
x_{0}^{\left( i\right) };\theta _{d}\right) \right)
\end{equation*}%
and:%
\begin{equation*}
\mathcal{C}^{\left( \min \right) }\left( \theta _{g}\mid \theta _{d}\right)
\approx c+\frac{1}{m}\sum_{i=1}^{m}\log \left( 1-\Discriminator\left( %
\Generator\left( z^{\left( i\right) };\theta _{g}\right) ;\theta _{d}\right)
\right)
\end{equation*}%
where $m$ is the size of the mini-batch sample and $c$ is a constant that does
not depend on the generator parameters $\theta _{g}$:
\begin{equation*}
c=\frac{1}{m}\sum_{i=1}^{m}\log \Discriminator\left( x_{1}^{\left( i\right)
};\theta _{d}\right)
\end{equation*}

\begin{algorithm}[tbph]
\caption{Stochastic gradient optimization for GAN training}
\label{alg:minimax1}
\begin{algorithmic}
\STATE The goal is to compute the optimal parameters $\hat{\theta}_g$ and $\hat{\theta}_d$
\STATE We note $n_{\max}$ the number of iterations to apply to the maximization problem (discriminator step) and
$n_{\min}$ the number of iterations to apply to the minimization problem (generator step)
\STATE $m$ is the size of the mini-batch sample
\item The starting values are denoted by $\theta _{g}^{\left( 0\right) }$
and $\theta _{d}^{\left( 0\right) }$
\FOR{$j=1$ to $n_{\min}$}
    \STATE $\theta _{d}^{\left( j,0\right) }\leftarrow \theta _{d}^{\left(j-1\right) }$
    \FOR{$k=1$ to $n_{\max}$}
        \STATE Sample $m$ random noise vector $\left(z^{\left(1\right)}, \ldots ,z^{\left(m\right)}\right)$
                from the prior distribution $\mathbb{P}_{\mathrm{noise}}\left(z\right)$
        \STATE Compute the simulated data $\left(x_0^{\left(1\right)}, \ldots
        ,x_0^{\left(m\right)}\right)$:
        \begin{equation*}
        x_0^{\left(i\right)}\sim \Generator\left( z^{\left( i\right) };\theta _{g}^{\left( j-1\right)}\right)
        \end{equation*}
        \STATE Sample $m$ examples $\left(x^{\left(1\right)}, \ldots ,x^{\left(m\right)}\right)$
                from the data distribution $\mathbb{P}_{\mathrm{data}}\left(x\right)$
        \STATE Compute the gradient vector of the maximization problem with respect to the parameter $\theta_{d}$:
            \begin{equation*}
                \Delta _{\theta _{d}}^{\left( j,k\right) }\leftarrow \left. \nabla _{\theta
                _{d}}\left( \frac{1}{m}\sum_{i=1}^{m}\log \Discriminator\left( x_1^{\left(
                i\right) };\theta _{d}\right) +\log \left( 1-\Discriminator\left( x_0^{\left(i\right)} \right) ;\theta
                _{d}\right) \right) \right\vert _{\theta _{d}=\theta _{d}^{\left(j,k-1\right) }}
            \end{equation*}
        \STATE Update the discriminator parameters $\theta _{d}$ using a backpropagation learning rule. For instance, in the case
        of the steepest descent method, we obtain:
            \begin{equation*}
                \theta _{d}^{\left( j,k\right) }\leftarrow \theta _{d}^{\left( j,k-1\right)
                }+\eta _{d}\cdot \Delta _{\theta _{d}}^{\left( j,k\right) }
            \end{equation*}
    \ENDFOR
    \STATE $\theta _{d}^{\left( j\right) }\leftarrow \theta _{d}^{\left(j,n_{\max }\right) }$
    \STATE Sample $m$ random noise vector $\left(z^{\left(1\right)}, \ldots ,z^{\left(m\right)}\right)$
                from the prior distribution $\mathbb{P}_{\mathrm{noise}}\left(z\right)$
    \STATE Compute the gradient vector of the minimization problem with respect to the parameter $\theta_{g}$:
        \begin{equation*}
            \Delta _{\theta _{g}}^{\left( j\right) }\leftarrow \left. \nabla _{\theta
            _{g}}\left( \frac{1}{m}\sum_{i=1}^{m}\log \left( 1-\Discriminator\left( %
            \Generator\left( z^{\left( i\right) };\theta _{g}\right) ;\theta
            _{d}^{\left( j\right) }\right) \right) \right) \right\vert _{\theta
            _{g}=\theta _{g}^{\left( j-1\right) }}
\end{equation*}%
    \STATE Update the generator parameters $\theta _{g}$ using a backpropagation learning rule. For instance, in the case
        of the steepest descent method, we obtain:
        \begin{equation*}
            \theta _{g}^{\left( j\right) }\leftarrow \theta _{g}^{\left( j-1\right)
            }-\eta _{g}\cdot \Delta _{\theta _{g}}^{\left( j\right) }
        \end{equation*}
\ENDFOR
\RETURN $\hat{\theta}_g \leftarrow \theta _{g}^{\left(n_{\min}\right) }$ and
$\hat{\theta}_d \leftarrow \theta _{d}^{\left(n_{\min}\right) }$
\end{algorithmic}
\begin{flushright}
Source: \citet[Algorithm 1]{Goodfellow-2014}.
\end{flushright}
\end{algorithm}

\noindent
In Algorithm (\ref{alg:minimax1}), we describe the stochastic
gradient optimization method proposed by \citet{Goodfellow-2014}
when the learning rule corresponds to the steepest descent method.
But other learning rules can be used such as the momentum method
or the adaptive learning method. In practice, these algorithms may not converge%
\footnote{For instance, we may observe a cycle because the parameters oscillate}.
Another big issue known as \textquoteleft \textit{mode collapse} \textquoteright\
concerns the diversity of generated samples. In this case, the simulated data drawn by the generator
exhibit small and limited differences, meaning all generated samples are almost identical.
This is why researchers have proposed alternatives to Algorithm (\ref{alg:minimax1})
in order to solve these two problems \citep{Mao-2017, Karras-2018, Metz-2017}.

\subsubsection{Time series modeling with GANs}

It is possible to generate fake time series from a single random noise vector
$z$ without any labels. However, this implies the use of complex structures for
both generator and discriminator. These structures can cause the models to be
computationally expensive, particularly when using convolutional networks.
Moreover, such structures don't address the issue concerning the lack of data.
Let us remember that we only have one single historical scenario. A less
expensive alternative is to use labels combined with simple multi-layer
perceptrons for the generator and the discriminator. This approach has been
introduced by \citet{Mirza-2014} and is called
conditional generative adversarial networks (cGANs).\smallskip

Before training a GAN, the data need to be labelled. An additional vector that
encodes structural information about real historical data must then be defined.
It will help the GAN to generate specific scenarios depending on labels defined
before. Therefore, the learning process will be supervised. The previous
framework remains valid, but the generative and discriminative models are
written as $\Generator\left( {z\mid v;\theta }_{g}\right) $ and
$\Discriminator\left( x{\mid v};\theta _{d}\right) $, where $v\in \mathcal{V}$
is the label vector. Therefore, the cost function becomes:
\begin{eqnarray*}
\mathcal{C}\left( \theta _{g},\theta _{d}\right)  &=&\mathbb{E}\left[ \log %
\Discriminator\left( x_{1}\mid v;\theta _{d}\right) \mid x_{1}\sim \mathbb{P}%
_{\mathrm{data}}\left( x\mid v\right) \right] + \\
&&\mathbb{E}\left[ \log \left( 1-\Discriminator\left( x_{0}\mid v;\theta
_{d}\right) \right) \mid x_{0}\sim \mathbb{P}_{\mathrm{\func{mod}el}}\left(
x\mid v\right) \right]
\end{eqnarray*}%
where $\mathbb{P}_{\mathrm{\func{mod}el}}\left( x\mid v\right) =\Generator\left(
\mathbb{P}_{\mathrm{noise}}\left( z\right) \mid v;\theta _{g}\right) $.\smallskip

The label vector may encode various types of information, and can be
categorical or continuous. For example, if we consider the S\&P 500 index, we
can specify $\mathcal{V}=\{-1,0,+1\}$ depending on its short trend. Let $y_{t}$
be the value of the S\&P 500 index and $m_{t}$ the corresponding 10-day moving
average. $v_{t}$ is equal to $-1$ if the S\&P 500 index exhibits a negative
trend, $0$ if it has no trend, and $+1$ otherwise\footnote{For instance, we can
define the labels in the following way: $v_{t}=\mathds{1}\left\{ \varepsilon
_{t}>\epsilon \right\} -\mathds{1}\left\{ \varepsilon _{t}<-\epsilon \right\} $
where $\varepsilon _{t}=y_{t}-m_{t}$ and $\epsilon >0$ is a threshold.}. An
example of continuous labels is a vector composed of the last $p$ values
\citep{Koshiyama-2019}. This type of labels acts as a time memory
and helps to reproduce auto-correlation patterns of the stochastic process.

\subsection{Wasserstein GAN models}
\label{section:wasserstein-gan}
In Section \ref{section:financial-applications}, we will see that the basic
GAN model using the cross-entropy as the loss function for the discriminator
suffers from three main problems. First, the visualization of the training
process is not obvious. Traditionally, we focus on the loss error curve to
decide whether or not the network is trained correctly. In the case of
financial applications, looking at binary cross entropy is not relevant.
Computer vision algorithms can trust generated images to evaluate the GAN,
financial time series are extremely noisy. Therefore, a visual evaluation is
much more questionable. Second, the basic GAN model is not suitable for
generating multi-dimensional time series. A possible alternative is to modify
the GAN's structure. For instance, we can replace the binary cross entropy
function by the Wasserstein distance for the training error
\citep{Arjovsky-2017a, Arjovsky-2017b}, or temporal information of
multi-dimensional time series can be encoded using a more complex structure
such as convolutional neural
networks \citep{Radford-2016} or recurrent neural networks %
\citep{Hyland-2017}. Third, the mode collapse phenomenon must be addressed in
order to manage the poor diversity of generated samples, because we would like to have
several simulated time series, as we have when performing Monte Carlo methods.

\subsubsection{Optimization problem}

Let $\mathbb{P}$ and $\mathbb{Q}$ be two univariate probability
distributions. The Wasserstein (or Kantorovich) distance between $\mathbb{P}$
and $\mathbb{Q}$ is defined as:
\begin{equation}
W_{p}\left( \mathbb{P},\mathbb{Q}\right) =\left( \inf_{\mathbb{F}\in
\mathcal{F}\left( \mathbb{P},\mathbb{Q}\right) }\int \left\Vert
x-y\right\Vert ^{p}\,\mathrm{d}\mathbb{F}\left( x,y\right) \right) ^{1/p}
\label{eq:wasserstein1}
\end{equation}%
where $\mathcal{F}\left( \mathbb{P},\mathbb{Q}\right) $ denotes the Fr\'echet
class of all joint distributions $\mathbb{F}\left( x,y\right) $ whose
marginals are equal to $\mathbb{P}$ and $\mathbb{Q}$. In the case $p=1$, the
Wasserstein distance represents the cost of the optimal transport problem%
\footnote{See Appendix \ref{appendix:Monge-Kantorovich} on page
\pageref{appendix:Monge-Kantorovich} for an introduction of Monge-Kantorovich
problems, and the relationship between optimal transport and the Wasserstein
distance.} \citep{Villani-2008}:
\begin{equation}
W\left( \mathbb{P},\mathbb{Q}\right) =\inf_{\mathbb{F}\in \mathcal{F}\left(
\mathbb{P},\mathbb{Q}\right) }\mathbb{E}\left[ \left\Vert X-Y\right\Vert
\mid \left( X,Y\right) \sim \mathbb{F}\right]   \label{eq:wasserstein2}
\end{equation}%
because $\mathbb{F}\left( x,y\right) $ describes how many masses are needed in
order to transport one distribution to another \citep{Arjovsky-2017a,
Arjovsky-2017b}. The Kantorovich-Rubinstein duality introduced by
\citet{Villani-2008} allows us to rewrite Equation (\ref{eq:wasserstein2}) as
follows:
\begin{equation}
W\left( \mathbb{P},\mathbb{Q}\right) =\sup_{\varphi }\mathbb{E}\left[
\varphi \left( X\right) \mid X\sim \mathbb{P}\right] -\mathbb{E}\left[
\varphi \left( Y\right) \mid Y\sim \mathbb{Q}\right]
\label{eq:wasserstein3}
\end{equation}%
where $\varphi $ is a $1$-Lipschitz function, such that $\left\vert \varphi
\left( x\right) -\varphi \left( y\right) \right\vert \leq \left\Vert
x-y\right\Vert $ for all $\left( x,y\right) $.\smallskip

In the case of a generative model, we would like to check whether sample and
generated data follows the same distribution. Therefore, we obtain
$\mathbb{P}=\mathbb{P}_{\mathrm{data}}$ and
$\mathbb{Q}=\mathbb{P}_{\mathrm{model}}$. In the special case of a GAN model,
$\mathbb{P}_{\mathrm{model}}\left( x;\theta _{g}\right) $ is given by
$\Generator\left( \mathbb{P}_{\mathrm{noise}}\left( z\right) ;\theta
_{g}\right) $. \citet{Arjovsky-2017a, Arjovsky-2017b} demonstrated that if
$\Generator\left( z;\theta _{g}\right) $ is continuous with respect to
$\theta $, then $W\left(
\mathbb{P}_{\mathrm{data}},\mathbb{P}_{\mathrm{model}}\right) $ is continuous
everywhere and differentiable almost everywhere. This result implies that GAN
can be trained until it has converged to its optimal solution contrary to
basic GANs, where the training loss were bounded and not
continuous\footnote{There is also no problem in computing the gradient.}.
Moreover, \citet{Gulrajani-2017} adapted Equation (\ref{eq:wasserstein3}) in
order to recover the two player min-max game:
\begin{equation*}
\mathcal{C}\left( \theta _{g},\theta _{d}\right)  =\mathbb{E}\left[ %
\Discriminator\left( x_{1};\theta _{d}\right) \mid x_{1}\sim \mathbb{P}_{%
\mathrm{data}}\left( x\right) \right] -
\mathbb{E}\left[ \Discriminator\left( x_{0};\theta _{d}\right) \mid
x_{0}\sim \Generator\left( z;\theta _{g}\right) ,z\sim \mathbb{P}_{\mathrm{%
noise}}\left( z\right) \right]
\end{equation*}%
The Wasserstein GAN (WGAN) plays the same min-max optimization problem as
previously:
\begin{equation*}
\hat{\theta}_{g}^{\left( \min \right) }=\arg \underset{\theta _{g}\in \Theta
_{\Generator}}{\min }\mathcal{C}^{\left( \min \right) }\left( \theta
_{g}\mid \theta _{d}\right)
\end{equation*}%
This implies that the discriminator needs to be a $1$-Lipschitz function.

\begin{remark}
The discriminator function is not necessarily a classifier, since the output
of the function $\Discriminator$ can take a value in $\left[ 0,1\right] $.
It can then be a general scoring method.
\end{remark}

\begin{remark}
The Kantorovich-Rubinstein duality makes the link between Wasserstein
distance and integral probability metrics presented in Appendix
\ref{appendix:ipm-gan} on page \pageref{appendix:ipm-gan}.
\end{remark}

\subsubsection{Properties of the optimal solution}

In Wasserstein GAN models, the cost function becomes:
\begin{equation*}
\mathcal{C}\left( \theta _{g},\theta _{d}\right) =\mathbb{E}\left[ \varphi
\left( x_{1};\theta _{d}\right) \mid x_{1}\sim \mathbb{P}_{\mathrm{data}}%
\right] -\mathbb{E}\left[ \varphi \left( x_{0};\theta _{d}\right) \mid
x_{0}\sim \Generator\left( z;\theta _{g}\right) ,z\sim \mathbb{P}_{\mathrm{%
noise}}\left( z\right) \right]
\end{equation*}%
and we have:
\begin{equation*}
\nabla _{\theta _{g}}\mathcal{C}\left( \theta _{g},\theta _{d}\right) =-%
\mathbb{E}\left[ \nabla _{\theta _{g}}\varphi \left( \Generator\left(
z;\theta _{g}\right) ;\theta _{d}\right) \right]
\end{equation*}%
We recall that the discriminator $\varphi $ needs to satisfy the Lipschitz
property otherwise the loss gradient will explode\footnote{As said
previously, $\varphi $ is not necessarily a discriminator function, but we
continue to use this term to name $\varphi $.}. A first alternative proposed
by \citet{Arjovsky-2017a, Arjovsky-2017b} is to clip the weights into a
closed space $\left[ -c,c\right] $. However, this method is limited because
the gradient can vanish and weights can saturate. A second alternative
proposed by \citet{Gulrajani-2017} is to focus on the properties of the
optimal discriminator gradient. They showed that the optimal solution
$\varphi ^{\star }$ has a gradient norm 1 almost everywhere under
$\mathbb{P}$ and $\mathbb{Q}$. Therefore, they proposed to add a
regularization term to the cost function in order to constraint the gradient
norm to converge to 1. This gradient penalty leads to define a new cost
function:
\begin{equation*}
\mathcal{C}_{\mathrm{regularized}}\left( \theta _{g},\theta _{d};\lambda
\right) =\mathcal{C}\left( \theta _{g},\theta _{d}\right) +\lambda \mathbb{E}%
\left[ \left( \left\Vert \nabla _{x}\varphi \left( x_{2};\theta _{d}\right)
\right\Vert _{2}-1\right) ^{2}\mid x_{2}\sim \mathbb{P}_{\mathrm{mixture}}%
\right]
\end{equation*}%
where $\mathbb{P}_{\mathrm{mixture}}$ is a mixture distribution of $\mathbb{P}$
and $\mathbb{Q}$. More precisely, \citet{Gulrajani-2017} proposed to sample
$x_{2}$ as follows: $x_{2}=\alpha x_{1}+\left( 1-\alpha \right) x_{0}$ where
$\alpha \sim \mathcal{U}_{\left[ 0,1\right] }$, $x_{0}\sim \mathbb{Q}$ and
$x_{1}\sim \mathbb{P}$.

\begin{remark}
\citet{Gulrajani-2017} found that a good value of the coefficient $\lambda $
is 10.
\end{remark}

\subsection{Convolutional neural networks}
\label{section:cnn}
A convolutional neural network (CNN) is a class of deep neural networks,
where the inputs are transformed using convolution and filtering operators.
For instance, this type of neural networks is extensively used in computer
vision. Recently, \citet{Radford-2016} implemented a new version of GANs
using convolutional networks as generator and discriminator in order to
improved image generation\footnote{This class of CNNs is called deep
convolutional generative adversarial networks (DCGANs).}. In this approach,
the underlying idea is to extract pertinent features. In finance, inputs are
different and correspond to $n_{x} $-dimensional time series representing
asset prices over $n_{t}$ days. Therefore, inputs are represented by a matrix
belonging to $\mathcal{M}_{n_{t},n_{x}}\left( \mathbb{R}\right) $, and are
characterized by the time axis, where order matters and the asset axis, where
order doesn't matter. Therefore, the input transformation is referring to
1-dimensional convolution (1D-CNN).

\begin{figure}[tbph]
\centering
\caption{Input architecture of convolutional neural networks}
\label{fig:cnn1}
\figureskip
\begin{tikzpicture}[x={(1,0)},y={(0,1)},z={({cos(60)},{sin(60)})},font=\sffamily\small,scale=2]

\tikzset{pics/fake box/.style args={#1 with dimensions #2 and #3 and #4 with text #5}{
code={

\draw[gray,ultra thin,fill=#1] (0,0,0) coordinate(-front-bottom-left) to ++ (0,#3,0)
coordinate(-front-top-right) -- ++ (#2,0,0) coordinate(-front-top-right) -- ++ (0,-#3,0)
coordinate(-front-bottom-right) -- cycle;

\draw[gray,ultra thin,fill=#1] (0,#3,0) -- ++ (0,0,#4) coordinate(-back-top-left) -- ++ (#2,0,0)
coordinate(-back-top-right) -- ++ (0,0,-#4) -- cycle;

\draw[gray,ultra thin,fill=#1!80!black] (#2,0,0) --++ (0,0,#4) coordinate(-back-bottom-right) -- ++ (0,#3,0) -- ++ (0,0,-#4) -- cycle;

\path[gray,decorate,decoration={text effects along path,text={#5}}] (#2/2,{2+(#3-2)/2},0) -- (#2/2,0,0);
}}}

\def\x{2.0}
\def\y{1.0}
\draw pic (box1-\y) at (\y,-\x/2,0) {fake box=green!40!white with dimensions {1.0} and {2*\x} and {1*\x} with text {} };

\def\x{1.5}
\def\y{3.0}
\draw pic (box1-\y) at (\y,-\x/2,0) {fake box=red!40!white   with dimensions {1.0} and {2*\x} and {1*\x} with text {} };

\def\x{1.0}
\def\y{5.0}
\draw pic (box1-\y) at (\y,-\x/2,0) {fake box=blue!40!white  with dimensions {1.0} and {2*\x} and {1*\x} with text {} };

\node[draw] at (1.75,2.15) {raw inputs};

\node[draw] at (3.55,1.675) {filtered inputs};

\node[draw] at (5.45,1.225) {filtered inputs};

\end{tikzpicture}

\end{figure}
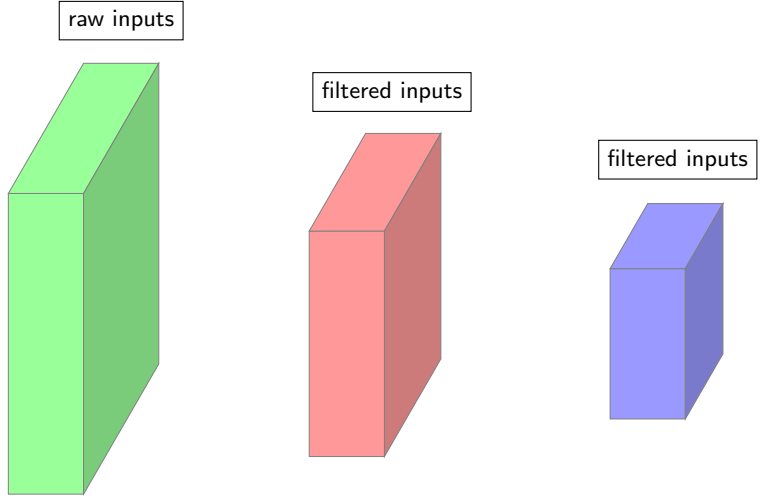

\subsubsection{Extracting features using convolution}

Convolution is no more than a linear transformation of a given input. But, contrary to simple multilayer perceptrons, convolutional operations preserve
the notion of ordering according to a specific axis. To achieve this, a
weight matrix of a given length $n_{k}$ called kernel will slide across the
input data. At each location, a continuous part of the input data is
selected, and the kernel receives this vector as input in order to produce a
single output by matrix product. This process is repeated with $n_{f}$
different filters of similar dimension. Consequently, we obtained $n_{f}$
down-sampled versions of the original input. The way the kernel can slide
across the input data is controlled by the stride $n_{s}$. It is defined as
the distance between two consecutive positions of the kernel
\citep{Dumoulin-2016}. The higher the stride value, the more important the
sub-sampling. If $s$ is equal to one, all the data are considered and there
is no sub-sampling. Finally, the notion of padding allows us to address the
situation when the kernel arrive at the end or the beginning of an axis. So,
padding is defined as the number of zeros concatenated at the beginning or at
the end of a given axis.\smallskip

In finance, we want to down-sample a given multi-dimensional time series
belonging to $\mathcal{M}_{n_{t},n_{x}}\left( \mathbb{R}\right) $ into a
subspace $\mathcal{M}_{n_{t}^{\prime },n_{x}}\left( \mathbb{R}\right) $ with
the condition that $n_{t}^{\prime }<n_{t}$. With a 1-dimensional convolution,
output time axis is automatically defined with respect to the dimension of
the kernel. The first dimension of the kernel is free, whereas the second
dimension is set to the number of time series $n_{x}$. Thus, the kernel is
defined by a weight matrix belonging to $\mathcal{M}_{n_{k},n_{x}}\left(
\mathbb{R}\right) $. In order to sample the input data, the stride is chosen, such that $s\geq 1$. The padding is set, such that $n_{p} $ rows of zeros is
padded at the (bottom and top) limits of the input data. Finally, the output
will belong to $\mathcal{M}_{n_{t}^{\prime },n_{x}}\left( \mathbb{R}\right)
$, such that:
\begin{equation}
n_{t}^{\prime }=\frac{n_{t}-n_{k}+2n_{p}}{n_{s}}+1
\end{equation}%
This type of convolution will be used to build the discriminator that should
output a single scalar. In this case, down-sampling real or fake time series
become essential.

\subsubsection{Up-sampling a feature using transpose convolution}

This transformation also called \textquoteleft
\textit{deconvolution}\textquoteright\ is useful to build the generator when
we would like to up-sample a random noise in order to produce a fake time
series belonging to $\mathcal{M}_{n_{t},n_{x}}\left( \mathbb{R}\right) $.
More generally, it could be used as the decoding layer in all forms of
auto-encoders. Transpose convolution is the exact inverse transformation of
the convolution previously defined. Considering an input data belonging to
$\mathcal{M}_{n_{t},n_{x}}\left( \mathbb{R}\right) $, we obtain an output
data belonging to $\mathcal{M}_{n_{t}^{\prime },n_{x}}\left(
\mathbb{R}\right) $ such that:
\begin{equation}
n_{t}^{\prime }=n_{s}\left( n_{t}-1\right) +n_{k}-2n_{p}
\end{equation}%
The term transpose comes from the fact that convolution is in fact a matrix
operation. When we perform a convolution, input matrix flattens into a
$n_{t}n_{x}$-dimensional vector. All the parameters (stride, kernel settings
or padding) are encoded into the weight convolution matrix belonging to
$\mathcal{M}_{n_{t}n_{x},n_{t}^{\prime }}\left( \mathbb{R}\right) $.
Therefore, the output data is obtained by computing the product between the
input vector and the convolution matrix. Performing a transpose convolution
is equivalent of taking the transpose of this convolution matrix.

\section{Financial applications}
\label{section:financial-applications}

\subsection{Application of RBMs to generate synthetic financial time series}
\label{section:applications-RBM}

A typical financial time series of length $T$ may be described by a real-valued
vector $y=\left(y_1,...,y_T\right)$. As we have introduced in Sections
\ref{section:bernoulli-rbm} and \ref{section:gaussian-rbm} on page
\pageref{section:bernoulli-rbm}, Bernoulli RBMs take binary vectors as input
for training and Gaussian RBMs take the data with unit variance as input in
order to ignore the parameter $\sigma$. Therefore, data preprocessing is
necessary before any training process of RBMs. For a Bernoulli RBM, each value
of $y_t$ needs to be converted to an $N$-digit binary vector using the
algorithm proposed by \citet{Kondratyev-2019}. The underlying idea consists in
discretizing the distribution of the training dataset and representing them
with binary numbers. For instance, $N$ may be set to 16 and a Bernoulli RBM,
which takes a one-dimensional time series as training dataset, will have 16
visible layer units. In the same way, the visible layer will have $16 \times n$
neurons in the case of an $n$-dimensional time series. Moreover, samples
generated by a Bernoulli RBM after Gibbs sampling are also binary vectors and
we need another algorithm to transform binary vectors into real values. These
transformation algorithms between real values and binary vectors are detailed
in Appendix \ref{appendix:binary transformation} on page
\pageref{appendix:binary transformation}. For a Gaussian RBM, we need only to
normalize data to unit variance and scale generated samples after Gibbs
sampling.\smallskip

For the training process of RBMs, we use the contrastive divergence
algorithm $\limfunc{CD}\nolimits^{\left( 1\right) }$ to estimate the
log-likelihood gradient and the learning rate
$\eta^{\left(t\right)}$ is set to $0.01$. All models are trained
using mini-batch gradient descent with batch size $500$ and we apply
$100\,000$ epochs to ensure that models are well-trained. Using a
larger mini-batch size will get a more stable gradient estimate at
each step, but it will also use more memory and take a longer time
to train the model.\smallskip

After having trained the RBMs, we may use these models to generate
synthetic (or simulated) samples to match training samples.
Theoretically, a well-trained RBM should be able to transform random
noise into samples that have the same distribution as the training
dataset after doing Gibbs sampling for a sufficiently long time.
Therefore, the trained RBM is fed by samples drawn from the Gaussian
distribution $\mathcal{N}\left(0, 1\right)$ as input data. We then perform a large number of forward and backward passes through the
network to ensure convergence between the probability distribution
of generated samples and the probability distribution of the
training sample. In the particular case of Bernoulli RBMs, after the
last backward pass from the hidden layer to the visible layer, we
need to transform generated binary vectors into real-valued vectors.

\subsubsection{Simulating multi-dimensional datasets}

\begin{figure}[tbh]
\caption{Histograms of simulated training samples}
\label{fig:HistMultiSimulated}
\begin{subfigure}[b]{0.5\linewidth}
\centering
\includegraphics[width=0.9\linewidth]{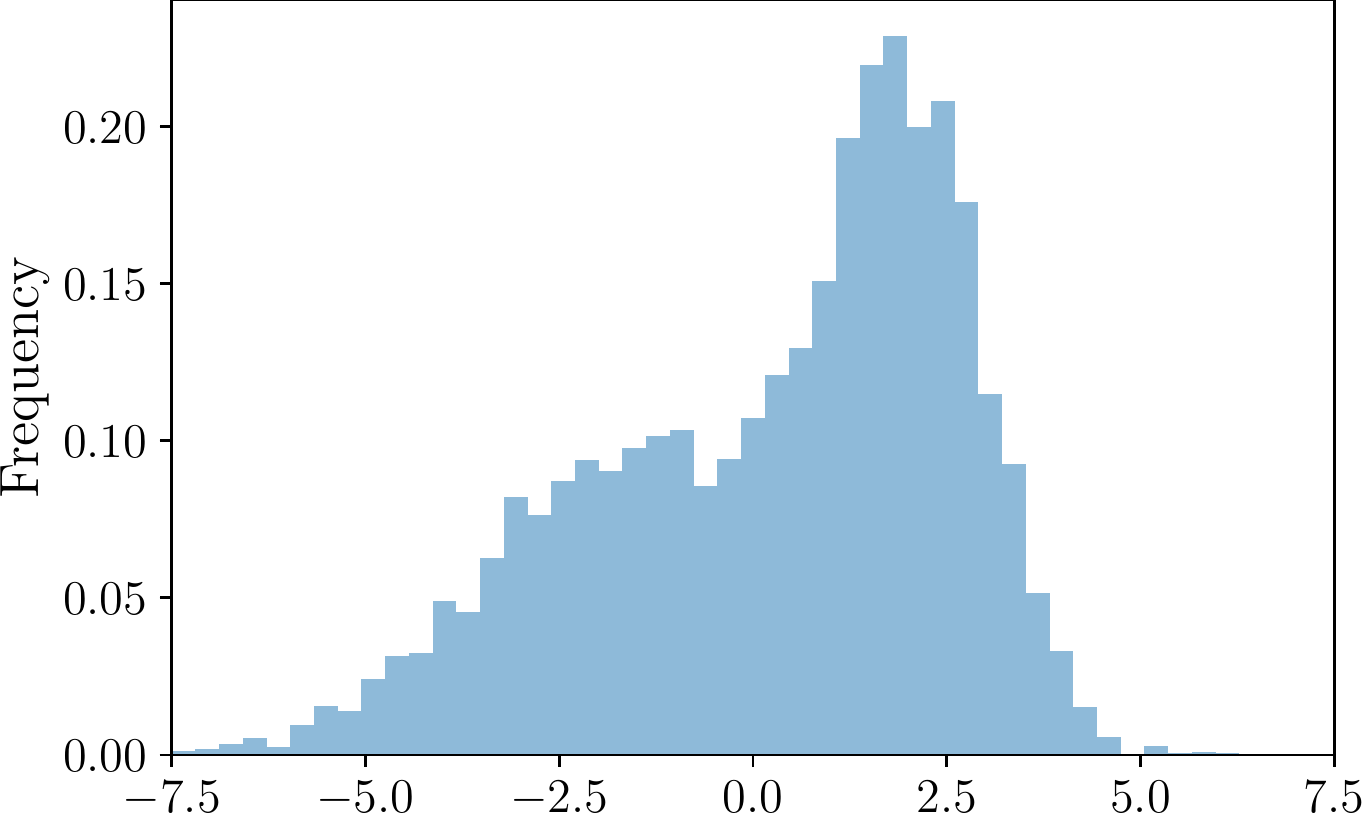}
\caption{Gaussian mixture model}
\vspace*{0.5cm}
\end{subfigure}
\begin{subfigure}[b]{0.5\linewidth}
\centering
\includegraphics[width=0.9\linewidth]{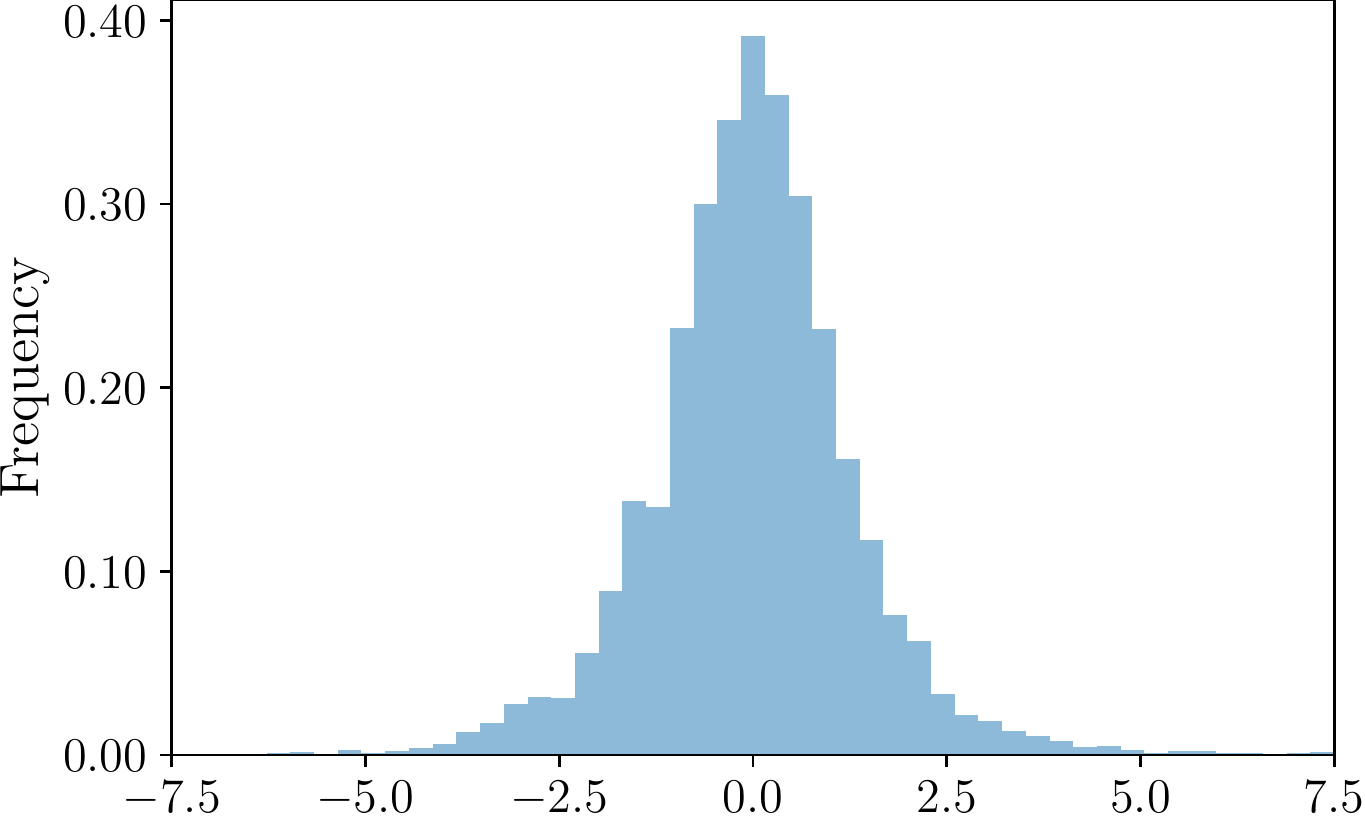}
\caption{Student's \textit{t}-distribution with $\nu = 4$}
\vspace*{0.5cm}
\end{subfigure}
\begin{subfigure}[b]{0.5\linewidth}
\centering
\includegraphics[width=0.9\linewidth]{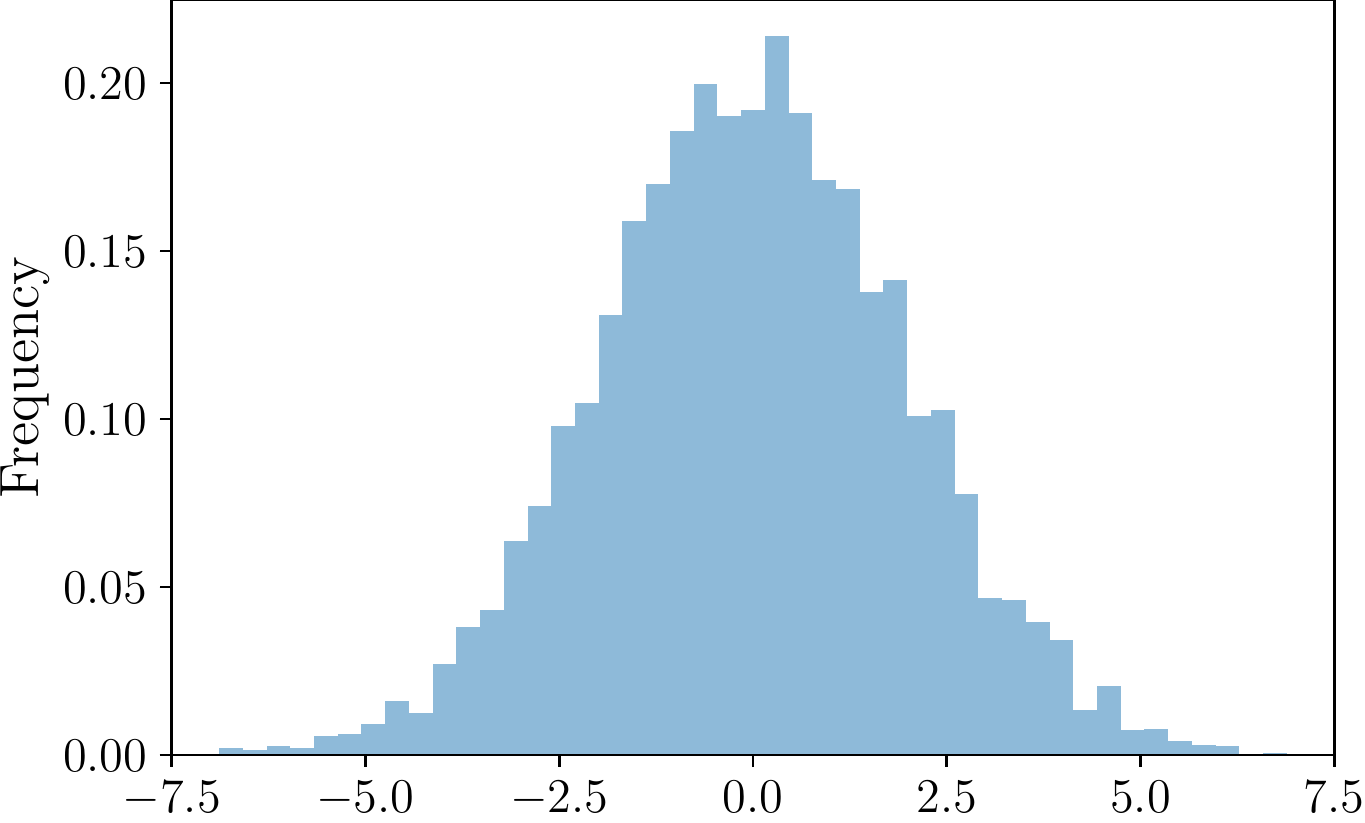}
\caption{Gaussian distribution $\mathcal{N}\left(0, 2\right)$}
\end{subfigure}
\begin{subfigure}[b]{0.5\linewidth}
\centering
\includegraphics[width=0.9\linewidth]{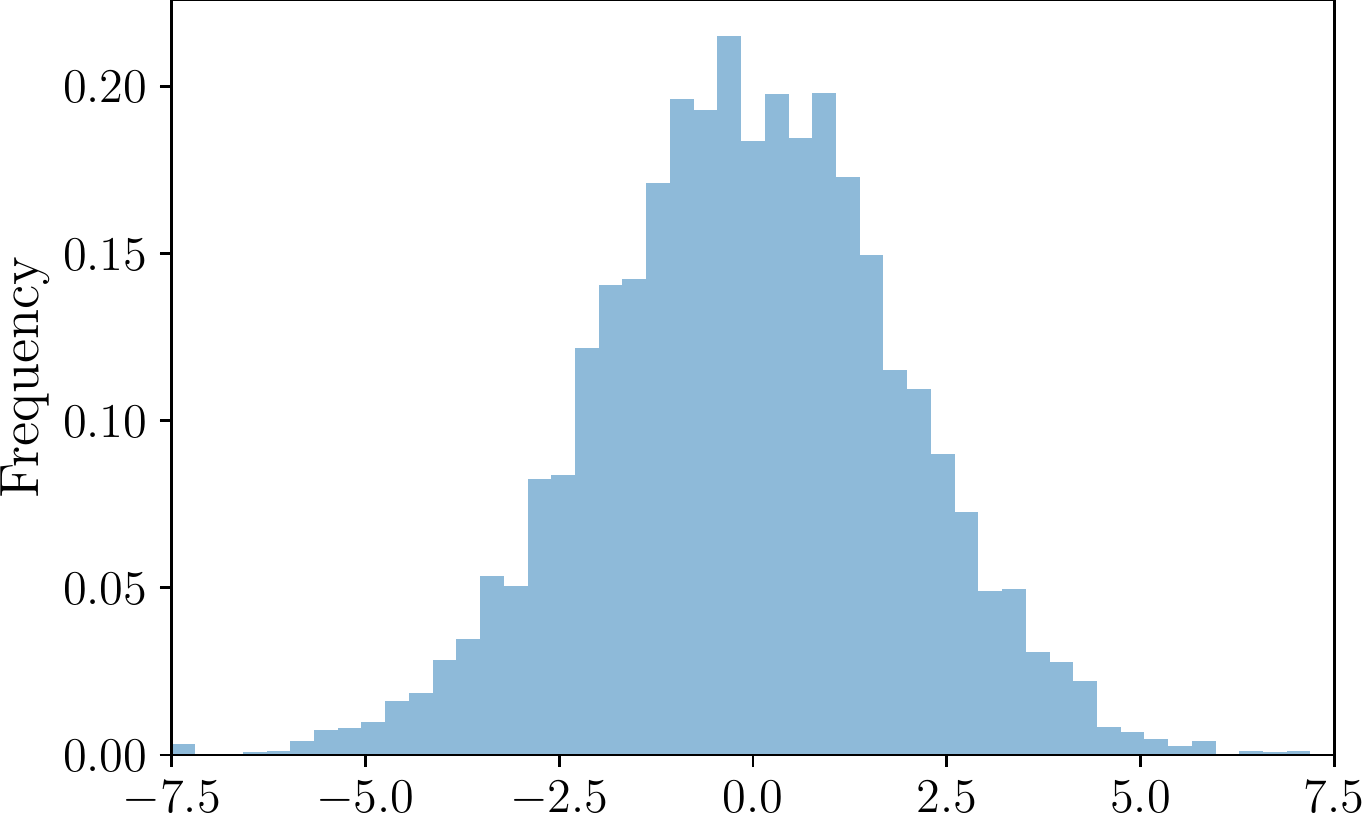}
\caption{Gaussian distribution $\mathcal{N}\left(0, 2\right)$}
\end{subfigure}
\end{figure}

In this first study, we test Bernoulli and Gaussian RBMs on a
simulated multi-dimensional dataset to check whether RBMs can learn
the joint distribution of training samples that involves marginal
distributions and a correlation structure. We simulate $10\,000$
observations of $4$-dimensional data with different marginal
distributions, as shown in Figure \ref{fig:HistMultiSimulated}. The
first dimension consists of samples drawn from a Gaussian mixture
model\footnote{The mixture probability is equal to $50\%$, whereas
the two underlying probability distributions correspond to two
Gaussian random variables $\mathcal{N}\left(-1.5, 2\right)$ and
$\mathcal{N}\left(2, 1\right)$.}. The samples in the second
dimension are drawn from a Student's $t$ distribution with 4 degrees
of freedom. For the third and fourth dimensions, we draw
respectively samples from the Gaussian distribution
$\mathcal{N}\left(0, 2\right)$. The empirical probability
distributions of these $4$-dimensional data are reported in Figure
\ref{fig:HistMultiSimulated}.
\begin{figure}[h!]
\centering
\caption{The theoretical correlation matrix of the Gaussian copula}
\label{fig:CorrMatCopula}
\includegraphics[scale=0.45]{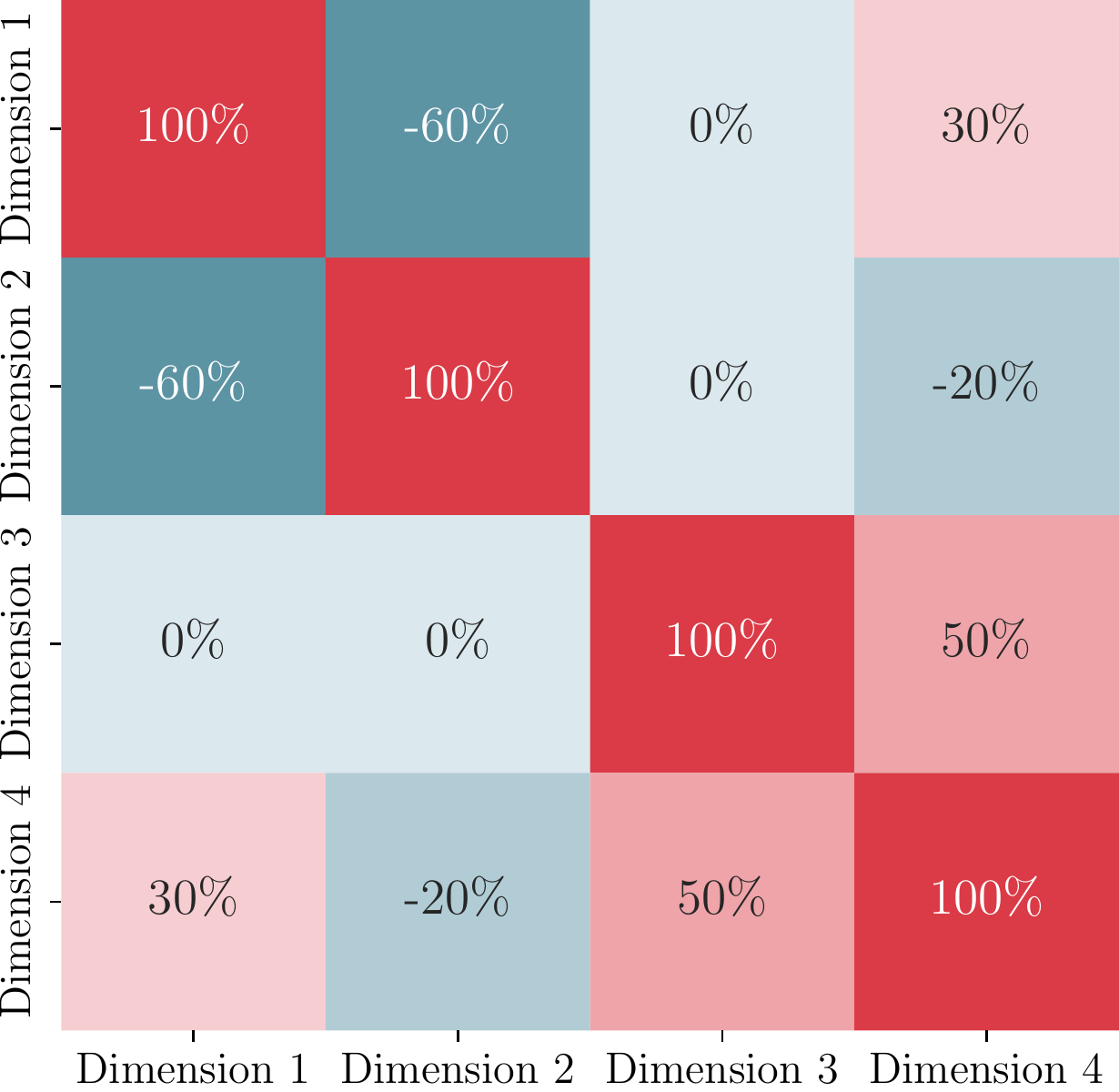}
\end{figure}
In addition, we use a Gaussian copula to construct a simple correlation
structure of the simulated samples with the correlation matrix of the Gaussian
copula given in Figure \ref{fig:CorrMatCopula}. We set a strong negative
correlation $-60\%$ between the Gaussian mixture model and the Student's $t$
distribution and a strong positive correlation $50\%$ between the two Gaussian
distributions. Moreover, the Gaussian distribution in the fourth dimension has
a more complicated correlation structure\footnote{The correlations are equal to
$30\%$ and $-20\%$ with respect to the first and second dimensions.} than the
Gaussian distribution in the third dimension, which is independent from the
Gaussian mixture model and the Student's $t$ distribution or the first and
second dimensions. Here, the challenge is to learn both the marginal
distribution and the copula function.\smallskip

Before implementing the training process, we need to update the
parameters for the RBMs to adjust multi-dimensional input. In the
case of the Bernoulli RBM, each value should be transformed into a
$16$-digit binary vector and we need to concatenate them to form a
$64$-digit binary vector. So, the visible layer of the Bernoulli RBM
has $64$ neurons. For the Gaussian RBM, we have simply $4$ visible
layer units. According to our experience, we need to set a large
number of hidden layer units in order to learn marginal
distributions and the correlation structure at the same time.
Therefore, we choose $256$ hidden layer units for the Bernoulli RBM
and $128$ hidden layer units for the Gaussian RBM. We also recall
that the number of epochs is set to $100\,000$ in order to ensure
that models are well-trained. After the training process, we perform
$1\,000$ steps of the Gibbs sampling on a $4$-dimensional random
noise to generate $10\,000$ samples with the same size as the
training dataset. These simulated samples are expected to have not
only the same marginal distributions but also the same correlation
structure as the training dataset.

\paragraph{Bernoulli RBM}

\begin{figure}[tbph]
\caption{Histogram of training and Bernoulli RBM simulated samples}
\label{fig:multiBRBMfake}
\begin{subfigure}[b]{0.5\linewidth}
\centering
\includegraphics[width=0.9\linewidth]{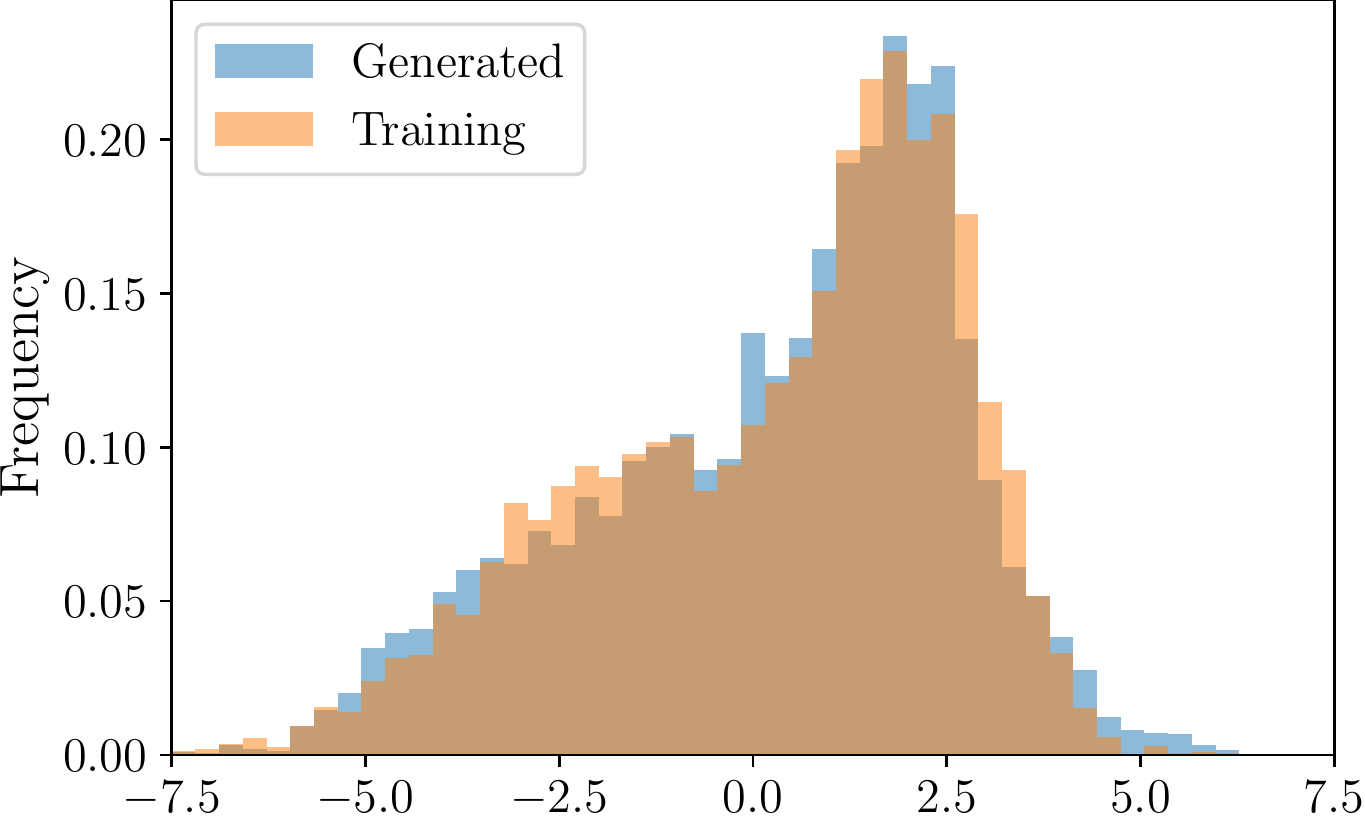}
\caption{Gaussian mixture model}
\vspace*{0.5cm}
\end{subfigure}
\begin{subfigure}[b]{0.5\linewidth}
\centering
\includegraphics[width=0.9\linewidth]{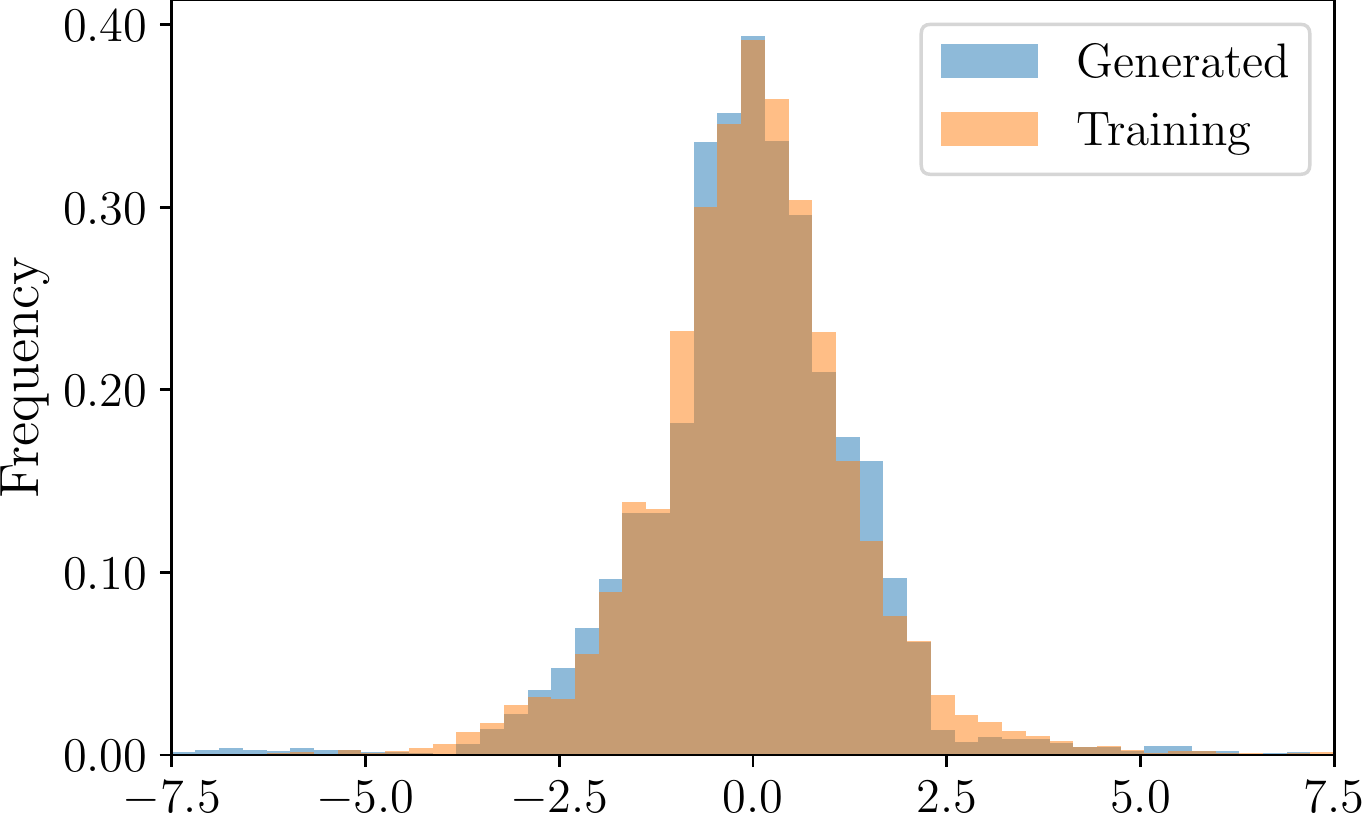}
\caption{Student's \textit{t}-distribution with $\nu = 4$}
\vspace*{0.5cm}
\end{subfigure}
\begin{subfigure}[b]{0.5\linewidth}
\centering
\includegraphics[width=0.9\linewidth]{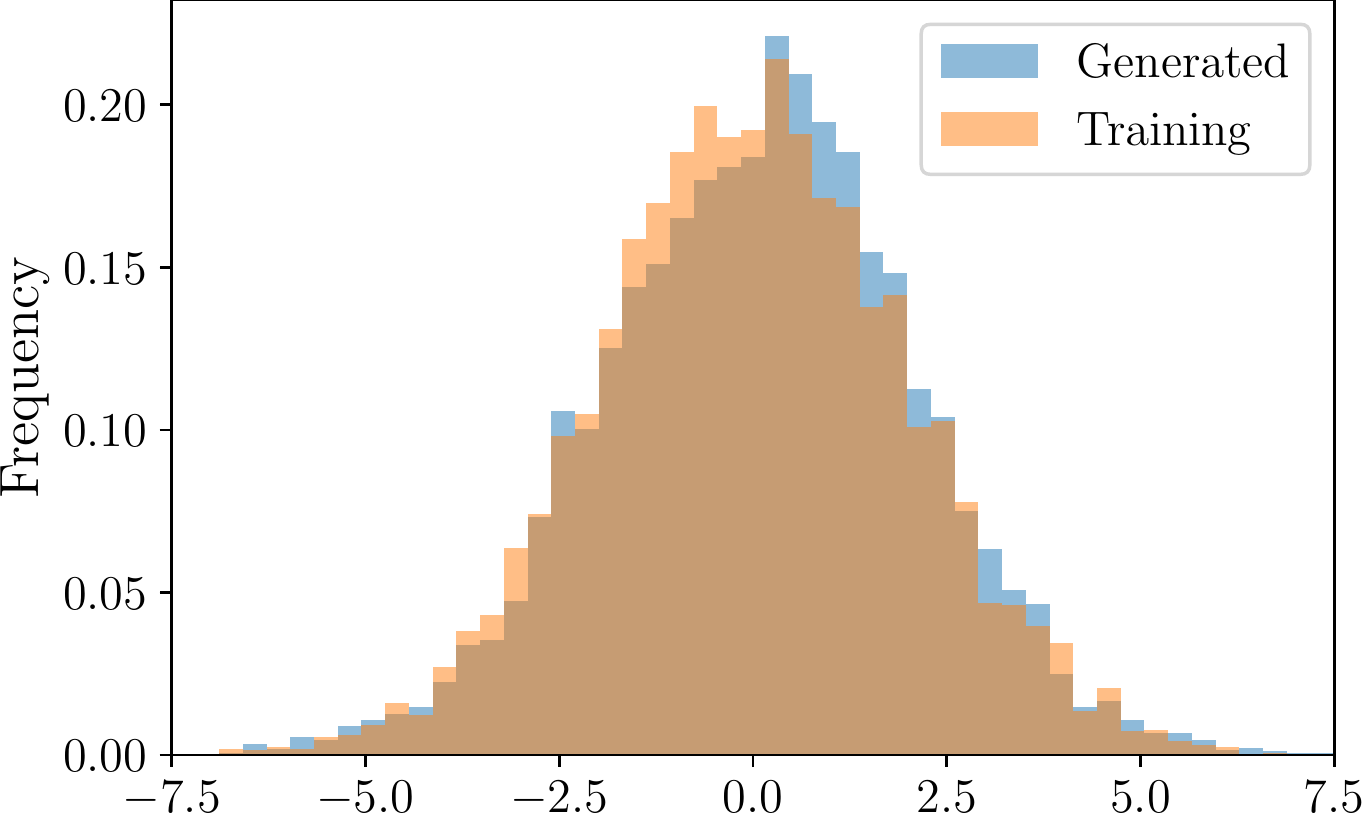}
\caption{Gaussian distribution $\mathcal{N}\left(0, 2\right)$}
\end{subfigure}
\begin{subfigure}[b]{0.5\linewidth}
\centering
\includegraphics[width=0.9\linewidth]{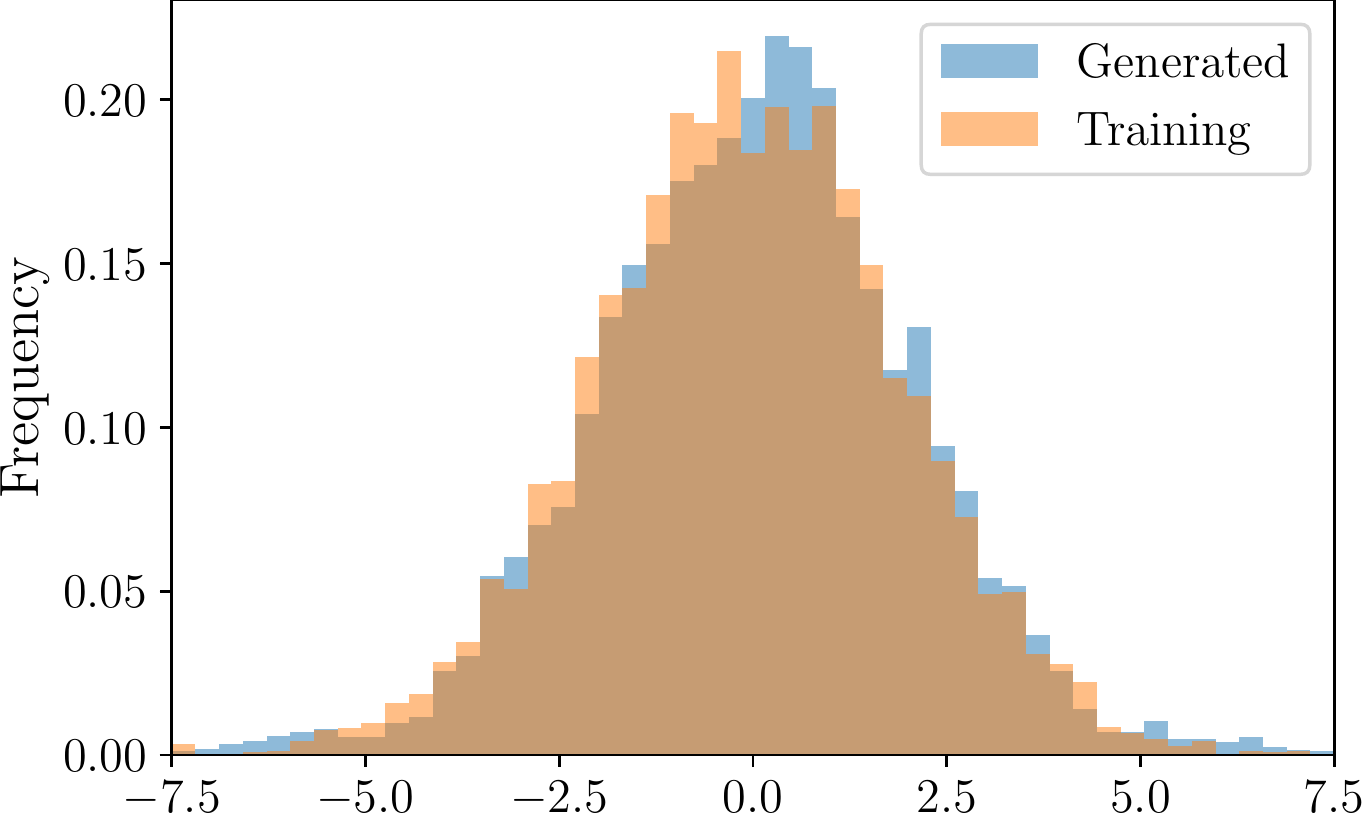}
\caption{Gaussian distribution $\mathcal{N}\left(0, 2\right)$}
\end{subfigure}
\end{figure}

\begin{figure}[tbph]
\caption{QQ-plot of training and Bernoulli RBM simulated samples}
\label{fig:multiBRBMqqplot}
\begin{subfigure}[b]{0.5\linewidth}
\centering
\includegraphics[width=0.9\linewidth]{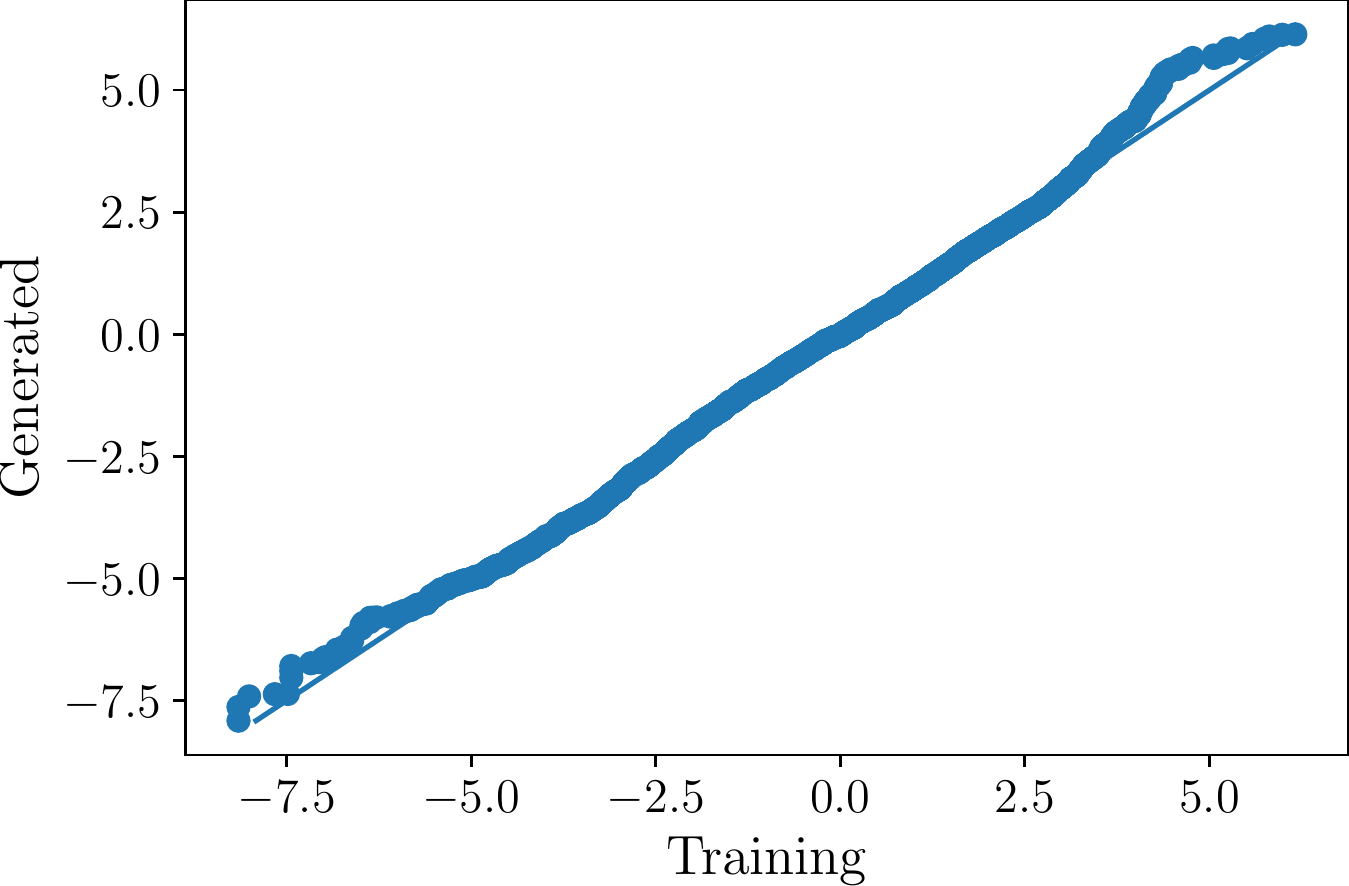}
\caption{Gaussian mixture model}
\vspace*{0.5cm}
\end{subfigure}
\begin{subfigure}[b]{0.5\linewidth}
\centering
\includegraphics[width=0.9\linewidth]{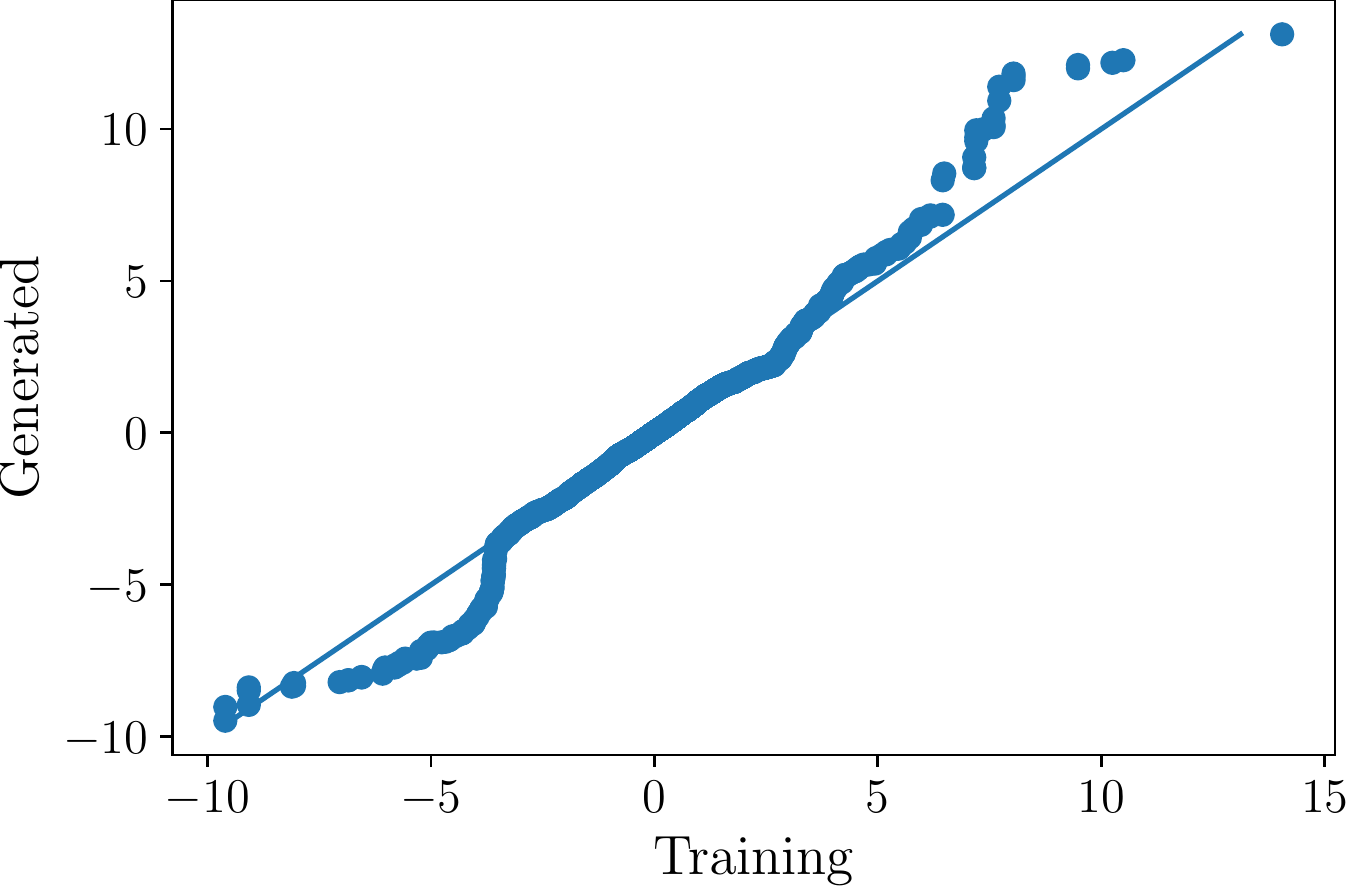}
\caption{Student's \textit{t}-distribution with $\nu = 4$}
\vspace*{0.5cm}
\end{subfigure}
\begin{subfigure}[b]{0.5\linewidth}
\centering
\includegraphics[width=0.9\linewidth]{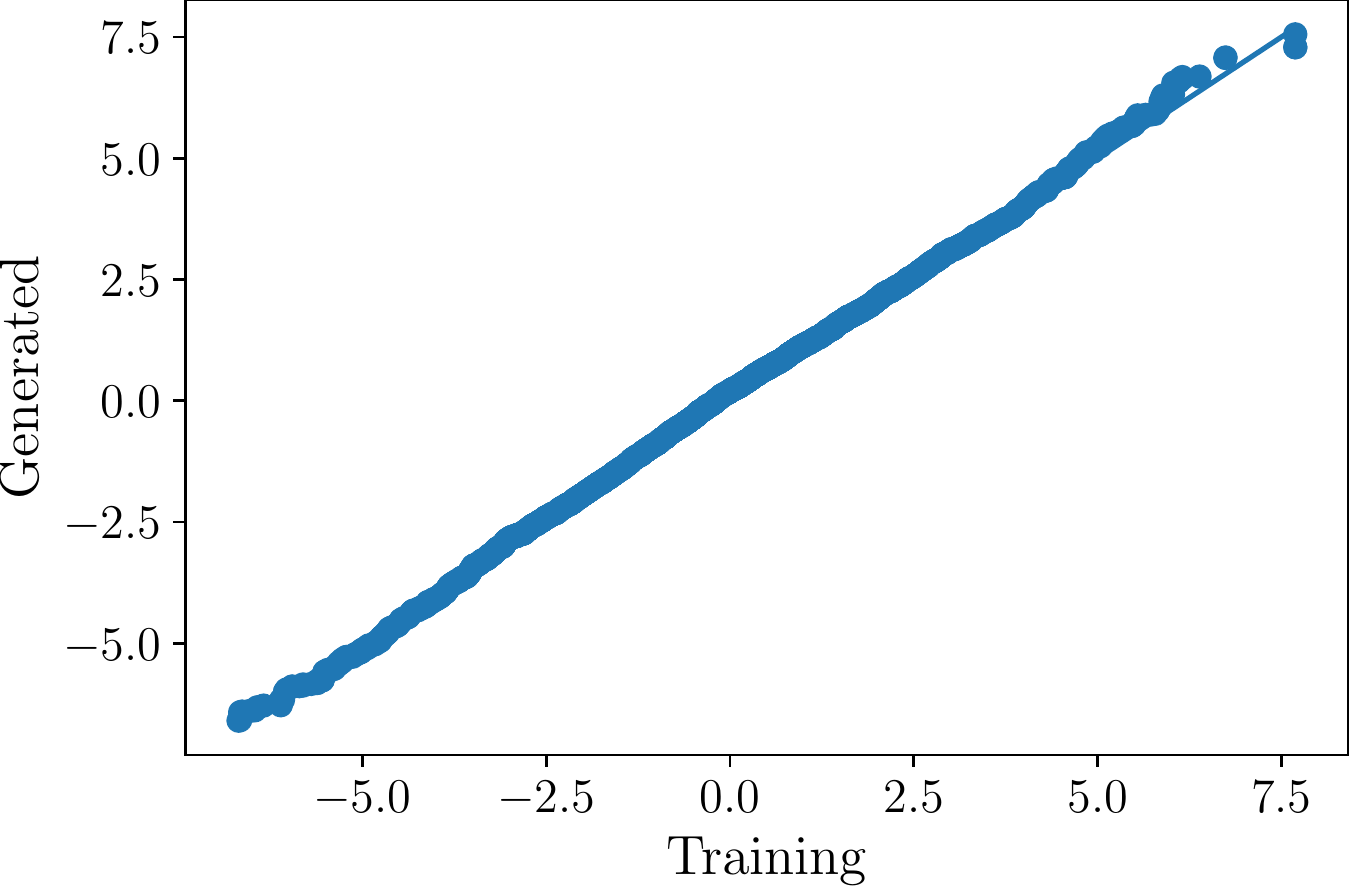}
\caption{Gaussian distribution $\mathcal{N}\left(0, 2\right)$}
\end{subfigure}
\begin{subfigure}[b]{0.5\linewidth}
\centering
\includegraphics[width=0.9\linewidth]{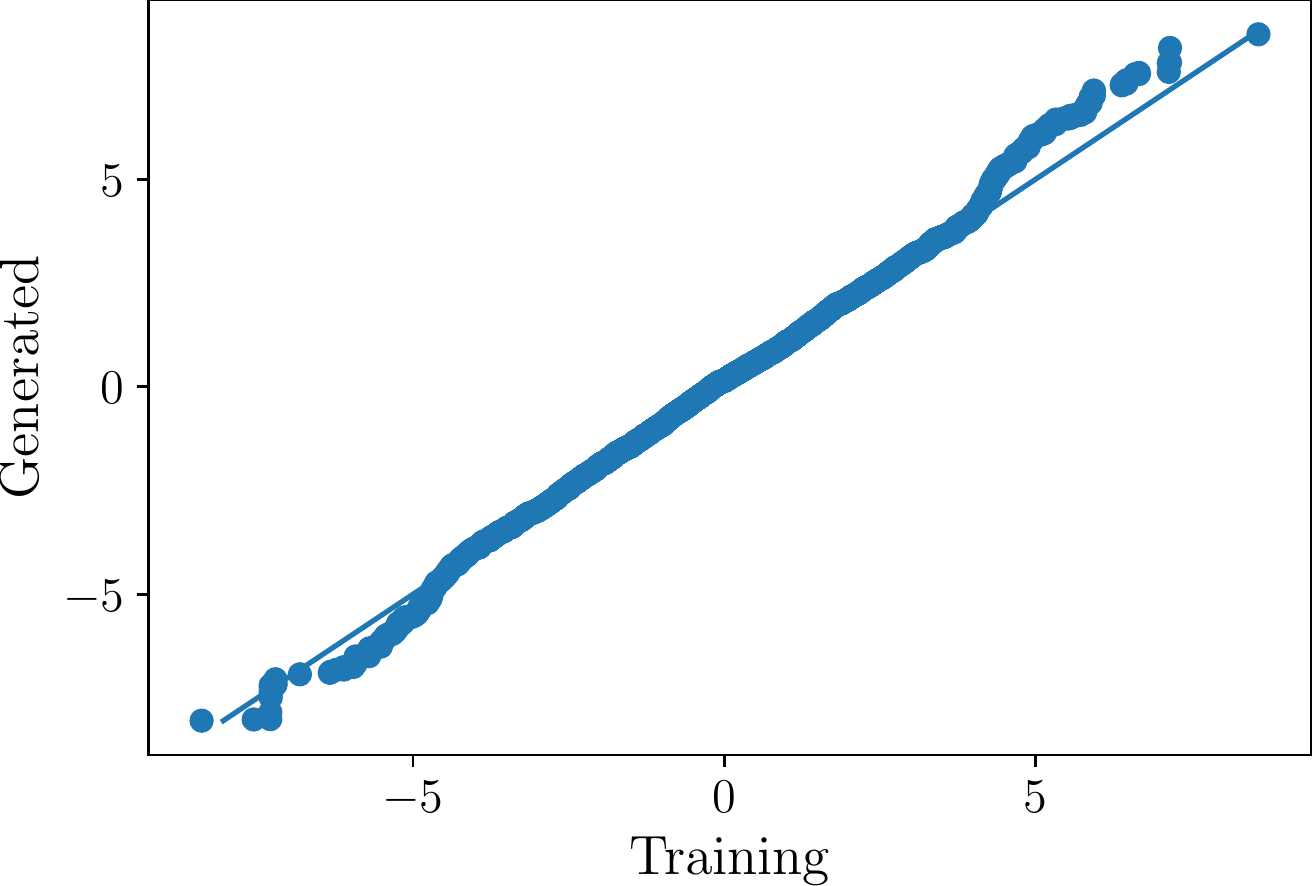}
\caption{Gaussian distribution $\mathcal{N}\left(0, 2\right)$}
\end{subfigure}
\end{figure}

Figures \ref{fig:multiBRBMfake} and \ref{fig:multiBRBMqqplot}
compare the histograms and QQ-plots between training samples and
generated samples after $1\,000$ Gibbs sampling steps in the case of
the Bernoulli RBM. According to these figures, we observe that the
Bernoulli RBM can learn pretty well each marginal distribution of
training samples. However, we also find that the Bernoulli RBM
focuses on extreme values in the training dataset instead of the
whole tail of the distribution and this phenomenon is more evident
for heavy-tailed distributions. For example, in the case of the
Student's $t$ distribution (Panel (b) in Figure
\ref{fig:multiBRBMqqplot}), the Bernoulli RBM tries to learn the
behavior of extreme values, but ignores the other part of the
distribution tails. Comparing the results for the two Gaussian
distributions (Panels (c) and (d) in Figure
\ref{fig:multiBRBMqqplot}), we notice that the learning of the
Gaussian distribution in the fourth dimension, which has a more
complicated correlation structure with the other dimensions, is not
as good as the result for the Gaussian distribution in the third
dimension.\smallskip

We have replicated the previous simulation $50$ times\footnote{Each
replication corresponds to $10\,000$ generated samples of the four
random variables, and differs because they use different random
noise series as input data.} and we compare the average of the mean,
the standard deviation, the $1^{\mathrm{st}}$ percentile and the
$99^{\mathrm{th}}$ percentile between training and simulated samples
in Table \ref{table:multiBRBM}. Moreover, in the case of simulated
samples, we indicate the confidence interval by $\pm$ one standard
deviation. These statistics demonstrate that the Bernoulli RBM
overestimates the value of standard deviation, the $1^{\mathrm{st}}$
percentile and the $99^{\mathrm{th}}$ percentile, especially in the
case of Student's $t$ distribution. This means that the Bernoulli
RBM is sensitive to extreme values and the learned probability
distribution is more heavy-tailed than the empirical probability
distribution of the training dataset.\smallskip

\begin{table}[tbh]
\centering
\caption{Comparison between training and Bernoulli RBM simulated samples}
\label{table:multiBRBM}
\begin{tabular}{l|cc|cc}
\hline
\multirow{2}{*}{Statistic}
& \multicolumn{2}{c|}{Dimension 1} & \multicolumn{2}{c}{Dimension 2}  \\
& Training & Simulated & Training & Simulated \\
Mean                          &  ${\TsVIII}0.303$ & ${\TsVIII}0.319$ ($\pm$ 0.039) &         $-0.006$ &         $-0.054$ ($\pm$ 0.029) \\
Standard deviation            &  ${\TsVIII}2.336$ & ${\TsVIII}2.385$ ($\pm$ 0.124) & ${\TsVIII}1.409$ & ${\TsVIII}1.540$ ($\pm$ 0.044) \\
$1^{\mathrm{st}}$ percentile  &          $-5.512$ &         $-5.380$ ($\pm$ 0.121) &         $-3.590$ &         $-5.008$ ($\pm$ 0.905) \\
$99^{\mathrm{th}}$ percentile &  ${\TsVIII}4.071$ & ${\TsVIII}4.651$ ($\pm$ 0.143) & ${\TsVIII}3.862$ & ${\TsVIII}4.255$ ($\pm$ 0.374) \\
\hline
\multirow{2}{*}{Statistic}
& \multicolumn{2}{c|}{Dimension 3} & \multicolumn{2}{c}{Dimension 4}  \\
& Training & Simulated & Training & Simulated \\
Mean                          &         $-0.002$ & ${\TsVIII}0.071$ ($\pm$ 0.039) &         $-0.063$ & ${\TsVIII}0.033$ ($\pm$ 0.043) \\
Standard deviation            & ${\TsVIII}1.988$ & ${\TsVIII}2.044$ ($\pm$ 0.028) & ${\TsVIII}1.982$ & ${\TsVIII}2.037$ ($\pm$ 0.030) \\
$1^{\mathrm{st}}$ percentile  &         $-4.691$ &         $-4.987$ ($\pm$ 0.167) &         $-4.895$ &         $-5.227$ ($\pm$ 0.244) \\
$99^{\mathrm{th}}$ percentile & ${\TsVIII}4.677$ & ${\TsVIII}4.918$ ($\pm$ 0.190) & ${\TsVIII}4.431$ & ${\TsVIII}4.938$ ($\pm$ 0.322) \\
\hline
\end{tabular}
\end{table}

\begin{figure}[tbh]
\vspace*{10pt}
\caption{Comparison between the empirical correlation matrix and the average correlation matrix of Bernoulli RBM simulated data}
\label{fig:multiBRBMcorr}
\begin{subfigure}[b]{0.5\linewidth}
\centering
\includegraphics[width=0.8\linewidth]{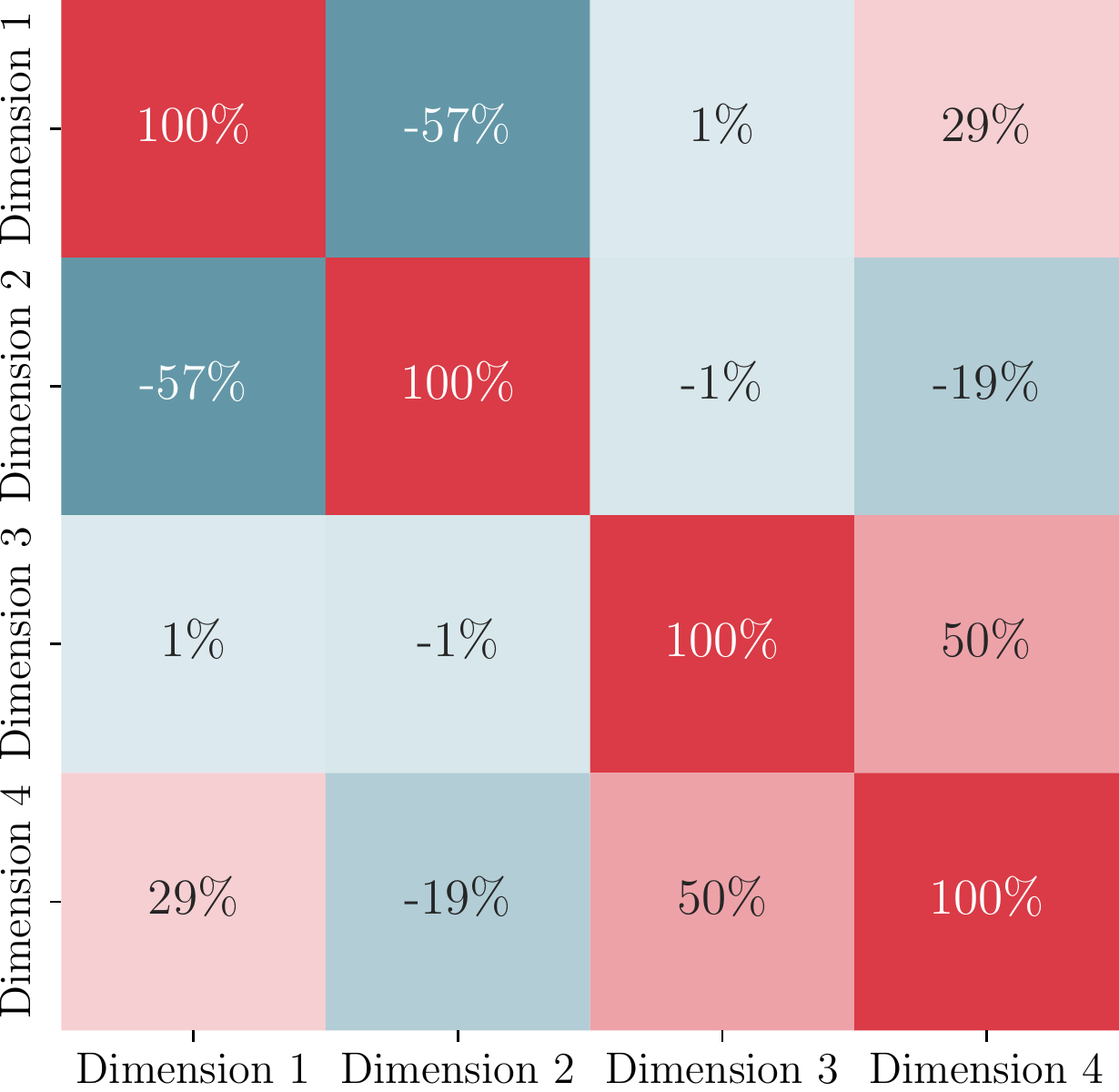}
\caption*{Training samples}
\end{subfigure}
\begin{subfigure}[b]{0.5\linewidth}
\centering
\includegraphics[width=0.8\linewidth]{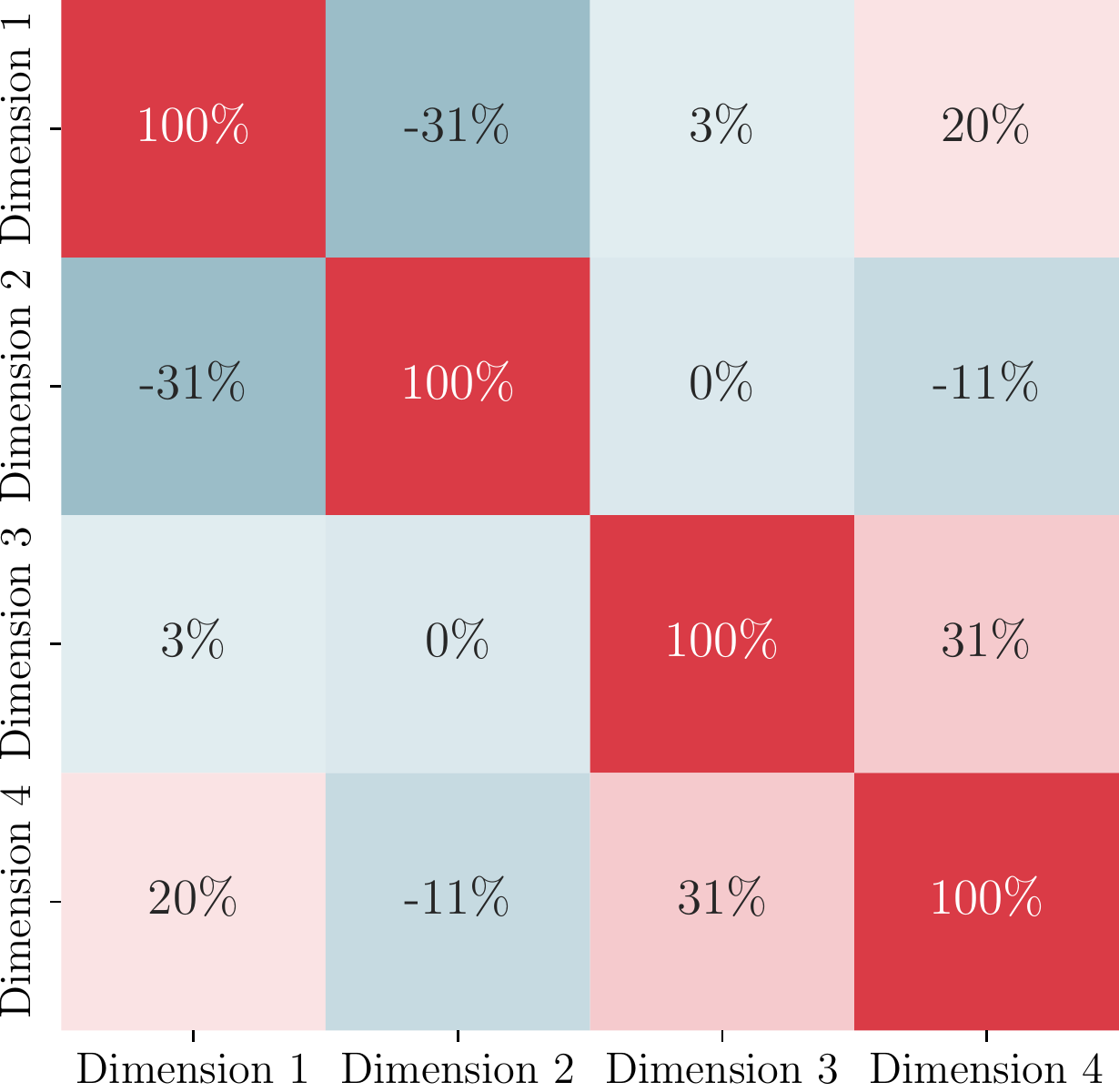}
\caption*{Generated samples}
\end{subfigure}
\end{figure}

In Figure \ref{fig:multiBRBMcorr}, we show the comparison between
the empirical correlation matrix\footnote{This empirical correlation
matrix is not exactly equal to the correlation matrix of the copula
function, because the marginals are not all Gaussian.} of the
training dataset and the average of the correlation matrices
computed by the 50 Monte Carlo replications. We notice that the
Bernoulli RBM does not capture perfectly the correlation structure
since the correlation coefficient values are less significant
comparing with the empirical correlation matrix. For instance, the
correlation between the first and second dimensions is equal to
$-57\%$ for the training data, but only $-31\%$ for the simulated
data on average\footnote{Similarly, the correlation between third
and fourth dimensions is equal to $50\%$ for the training data, but
only $31\%$ for the simulated data on average.}.

\newpage

\paragraph{Gaussian RBM}

\begin{figure}[tbph]
\caption{Histogram of training and Gaussian RBM simulated samples}
\label{fig:multiGRBMfake}
\begin{subfigure}[b]{0.5\linewidth}
\centering
\includegraphics[width=0.9\linewidth]{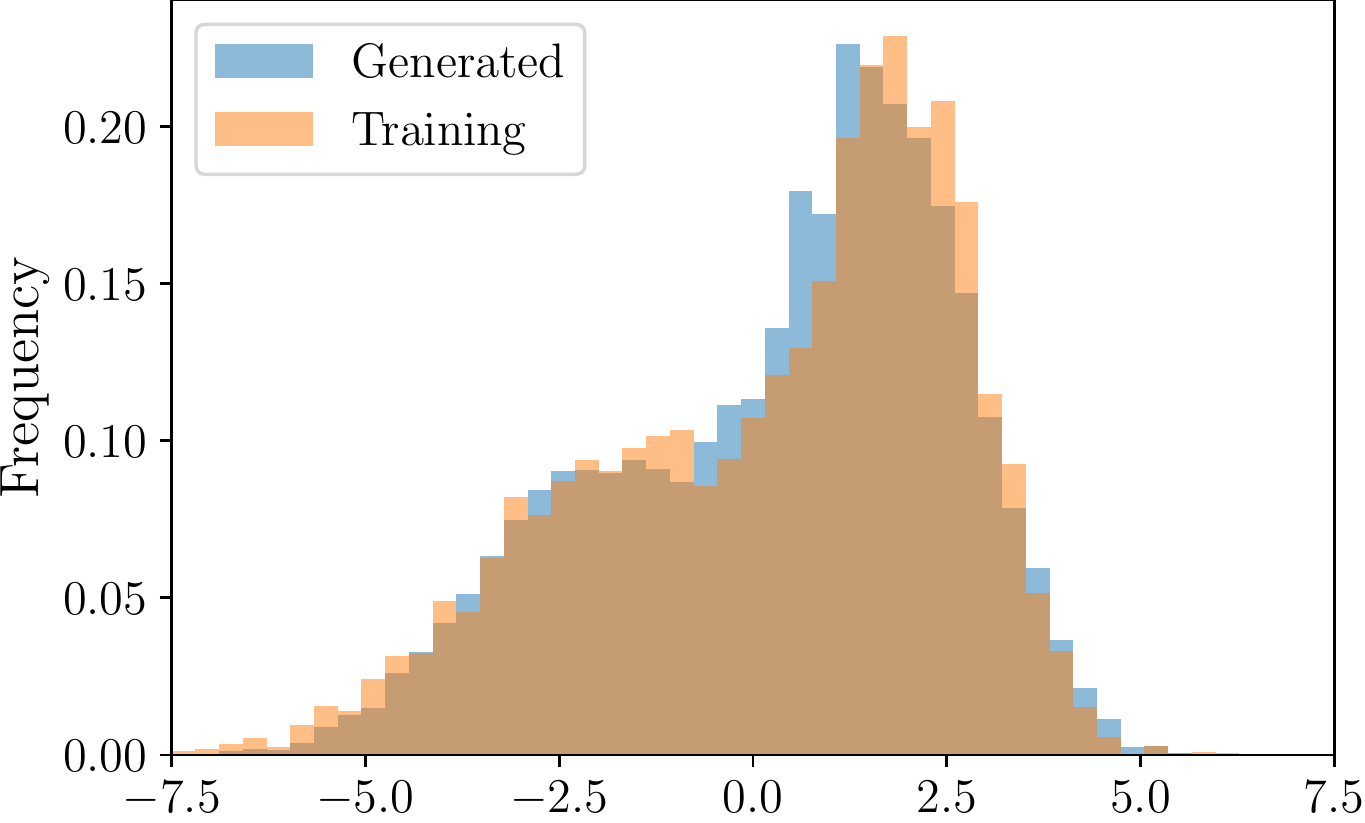}
\caption{Gaussian mixture model}
\vspace*{0.5cm}
\end{subfigure}
\begin{subfigure}[b]{0.5\linewidth}
\centering
\includegraphics[width=0.9\linewidth]{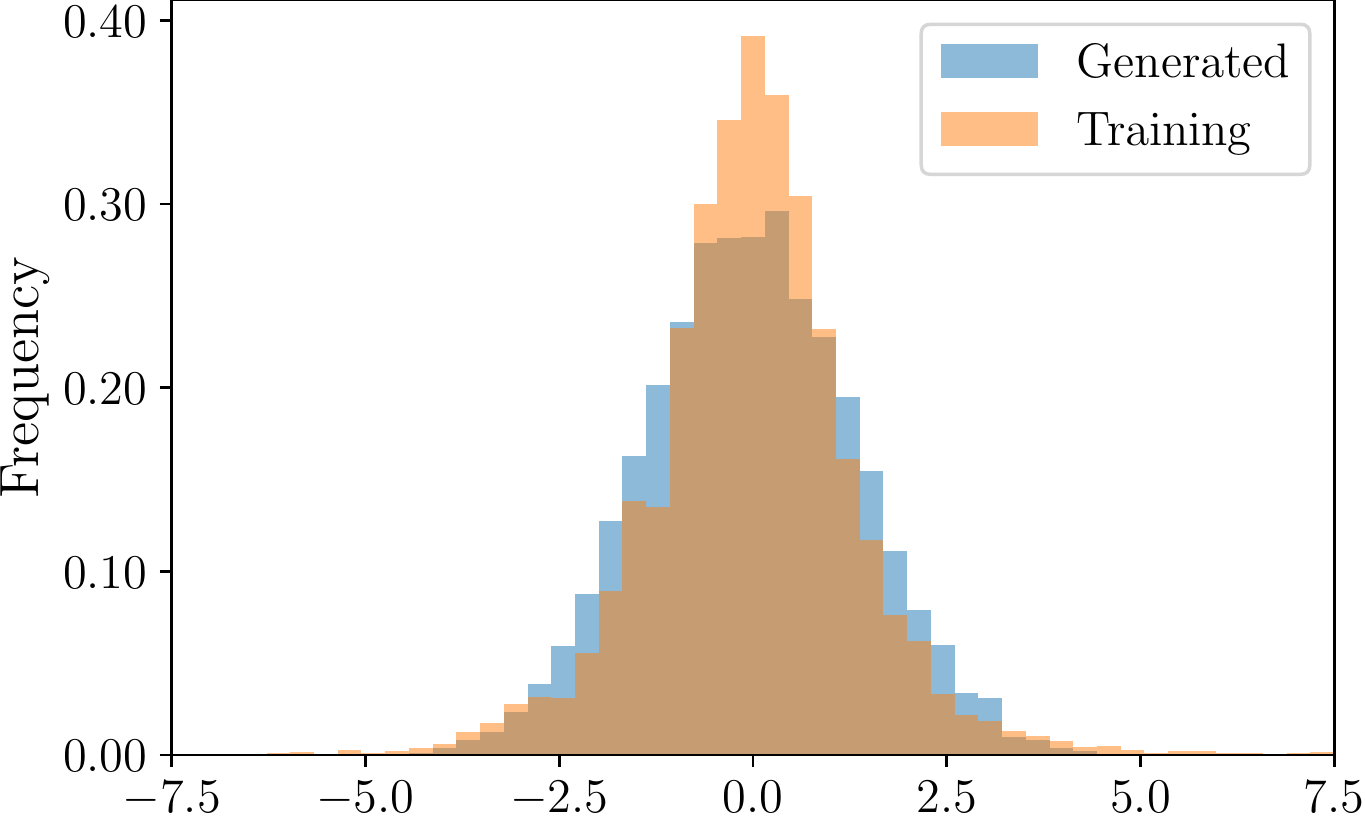}
\caption{Student's $t$ distribution with $\nu = 4$}
\vspace*{0.5cm}
\end{subfigure}
\begin{subfigure}[b]{0.5\linewidth}
\centering
\includegraphics[width=0.9\linewidth]{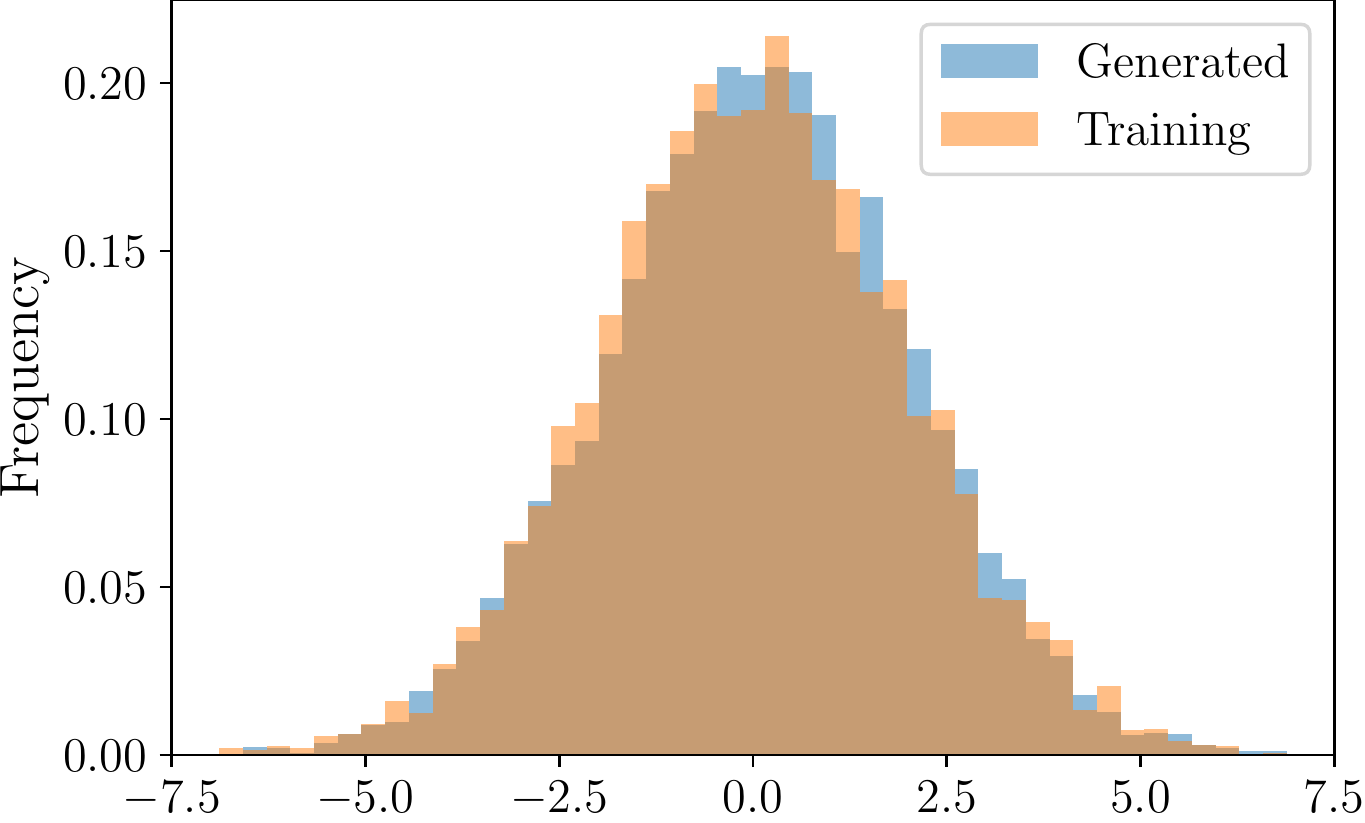}
\caption{Gaussian distribution $\mathcal{N}\left(0, 2\right)$}
\end{subfigure}
\begin{subfigure}[b]{0.5\linewidth}
\centering
\includegraphics[width=0.9\linewidth]{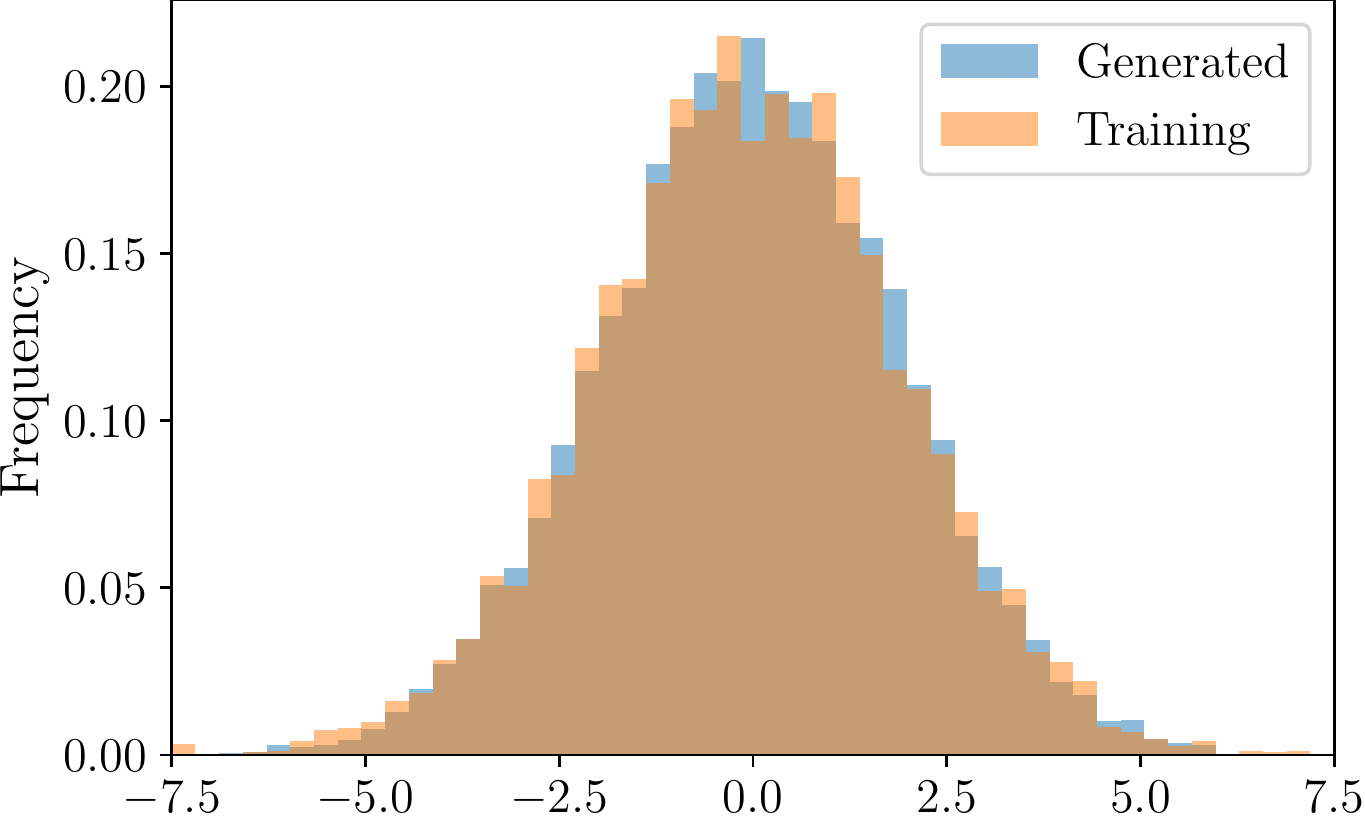}
\caption{Gaussian distribution $\mathcal{N}\left(0, 2\right)$}
\end{subfigure}
\end{figure}

\begin{figure}[tbph]
\caption{QQ-plot of training and Gaussian RBM simulated samples}
\label{fig:multiGRBMqqplot}
\begin{subfigure}[b]{0.5\linewidth}
\centering
\includegraphics[width=0.9\linewidth]{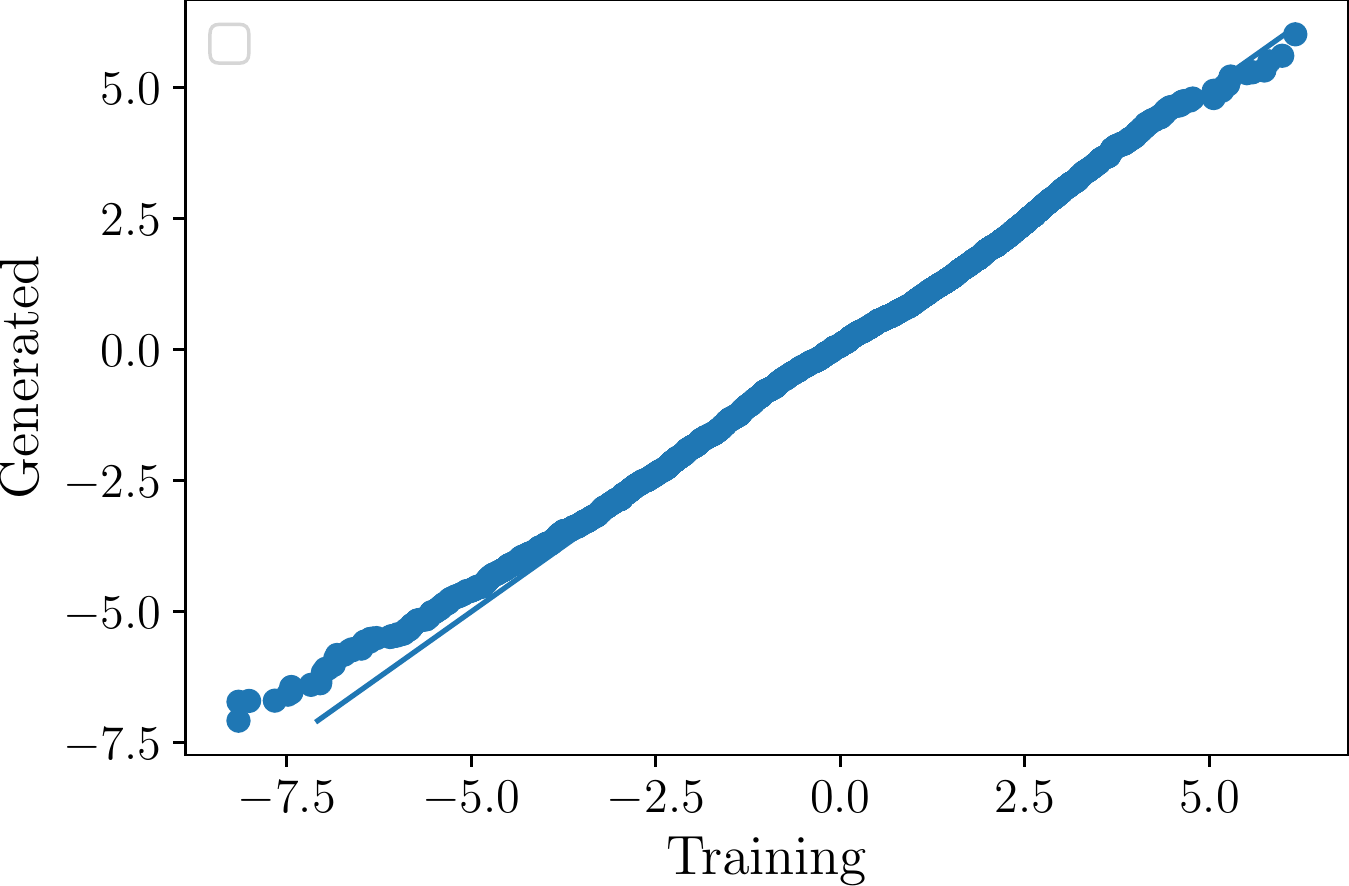}
\caption{Gaussian mixture model}
\vspace*{0.5cm}
\end{subfigure}
\begin{subfigure}[b]{0.5\linewidth}
\centering
\includegraphics[width=0.9\linewidth]{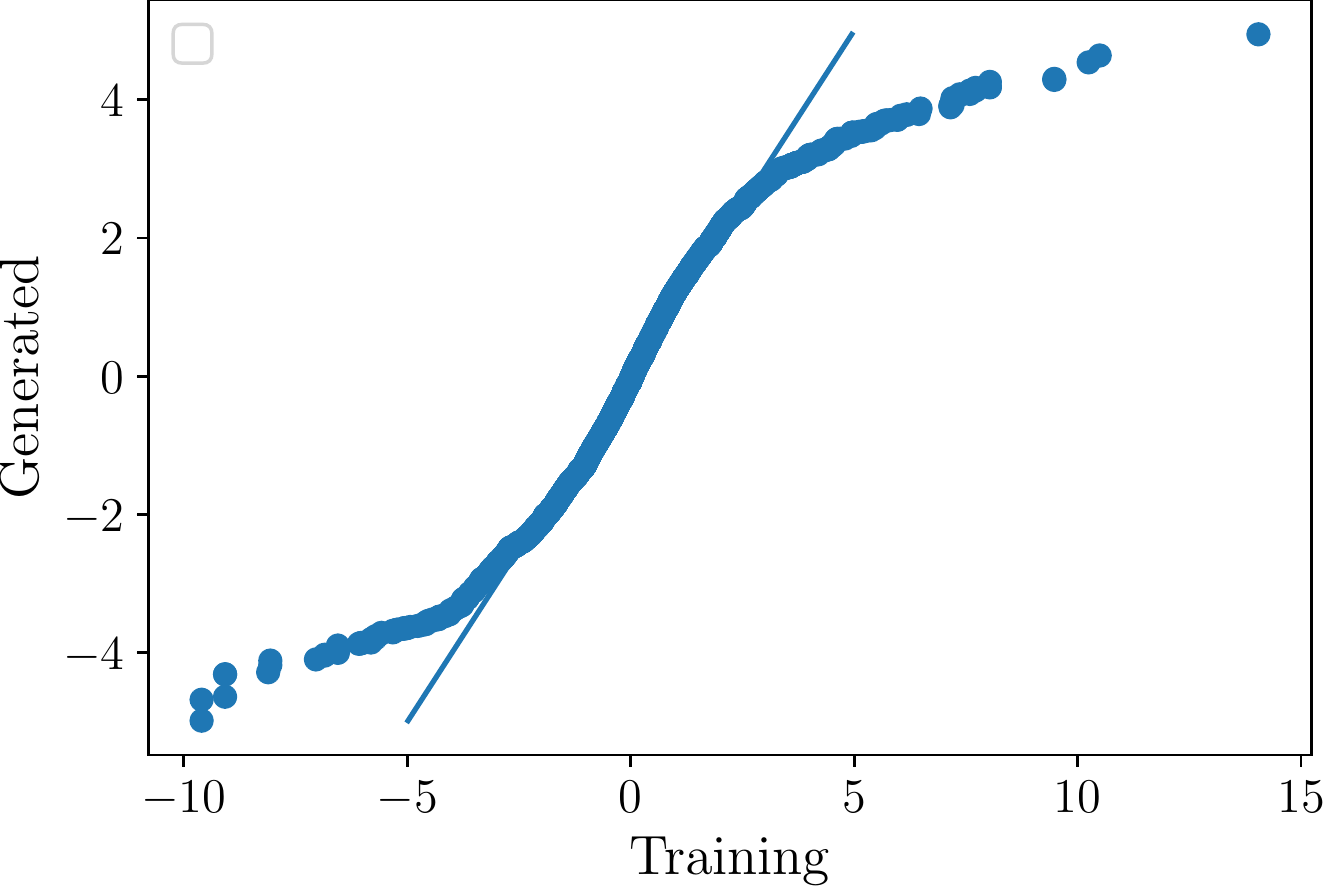}
\caption{Student's \textit{t}-distribution with $\nu = 4$}
\vspace*{0.5cm}
\end{subfigure}
\begin{subfigure}[b]{0.5\linewidth}
\centering
\includegraphics[width=0.9\linewidth]{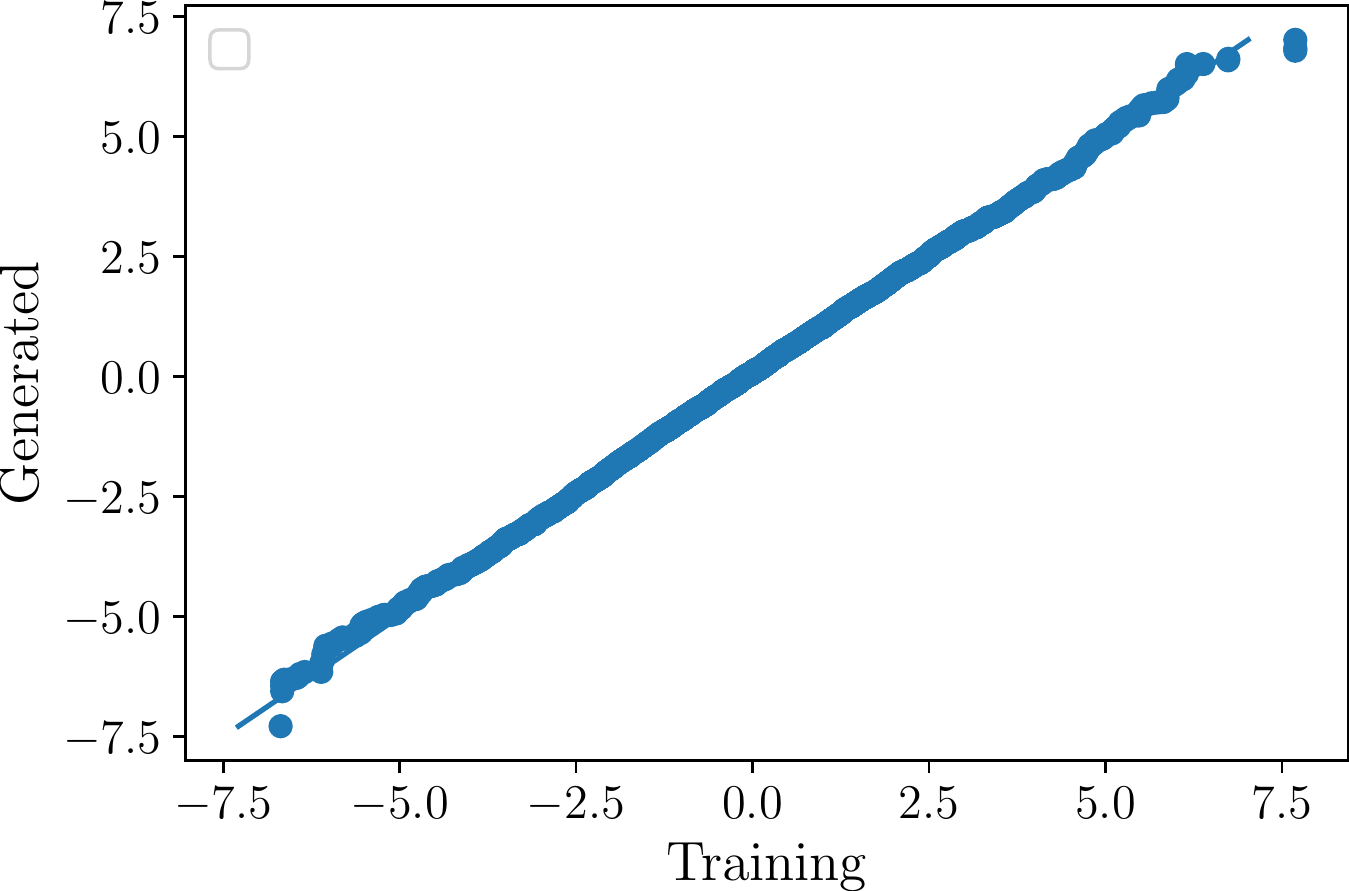}
\caption{Gaussian distribution $\mathcal{N}\left(0, 2\right)$}
\end{subfigure}
\begin{subfigure}[b]{0.5\linewidth}
\centering
\includegraphics[width=0.9\linewidth]{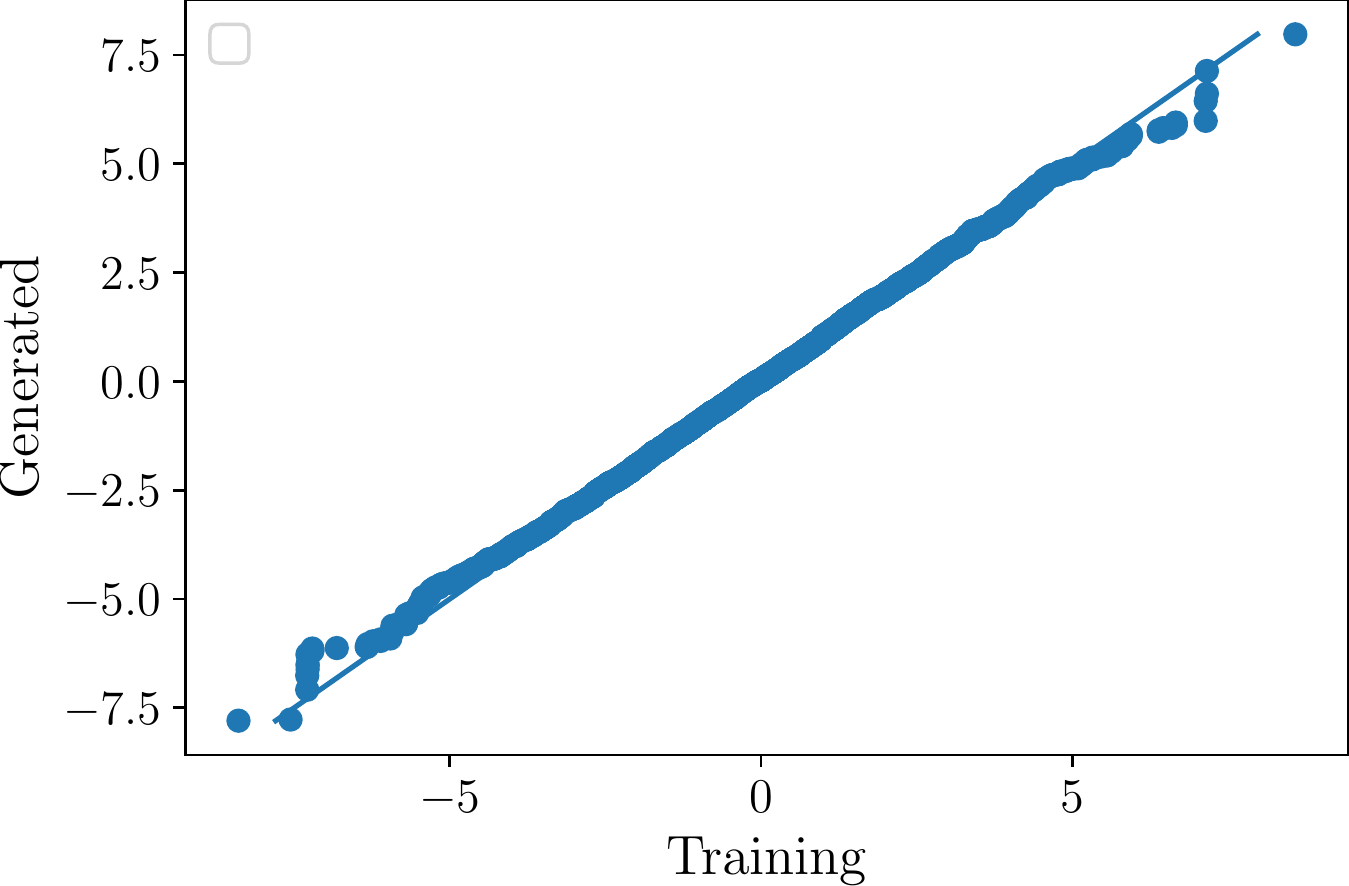}
\caption{Gaussian distribution $\mathcal{N}\left(0, 2\right)$}
\end{subfigure}
\end{figure}

Let us now consider a Gaussian RBM with $4$ visible layer units and
$128$ hidden layer units using the same simulated training dataset.
After the training process, synthetic samples are always generated
using the Gibbs sampling with $1\,000$ steps between visible and
hidden layers. Histograms and QQ-plots of training and generated
samples are reported in Figures \ref{fig:multiGRBMfake} and
\ref{fig:multiGRBMqqplot}. We notice that the Gaussian RBM works
well for the Gaussian mixture model and the two Gaussian
distributions (Panels (a), (c) and (d) in Figure
\ref{fig:multiGRBMqqplot}). Again, we observe that the Gaussian
distribution with the simplest correlation structure is easier to
learn than the Gaussian distribution with a more complicated
correlation structure. However, the Gaussian RBM fails to learn
heavy-tailed distributions such as the Student's $t$ distribution
since the model tends to ignore all values in the distribution
tails.\smallskip

Comparing Tables \ref{table:multiBRBM} on page
\pageref{table:multiBRBM} for the Bernoulli RBM and Tables
\ref{table:multiGRBM} on page \pageref{table:multiGRBM} for the
Gaussian RBM, we notice that the Gaussian RBM generally
underestimates the value of the standard deviation, the
$1^{\mathrm{st}}$ percentile and the $99^{\mathrm{th}}$ percentile.
This means that this is challenging to generate leptokurtic
probability distributions with Gaussian RBMs.\smallskip

\begin{table}[tbh]
\centering
\caption{Comparison between training and Gaussian RBM simulated samples}
\label{table:multiGRBM}
\begin{tabular}{l|cc|cc}
\hline
\multirow{2}{*}{Statistic}
& \multicolumn{2}{c|}{Dimension 1} & \multicolumn{2}{c}{Dimension 2}  \\
& Training & Simulated & Training & Simulated \\
Mean                          &  ${\TsVIII}0.303$ & ${\TsVIII}0.325$ ($\pm$ 0.037) &         $-0.006$ &         $-0.006$ ($\pm$ 0.023) \\
Standard deviation            &  ${\TsVIII}2.336$ & ${\TsVIII}2.232$ ($\pm$ 0.021) & ${\TsVIII}1.409$ & ${\TsVIII}1.355$ ($\pm$ 0.020) \\
$1^{\mathrm{st}}$ percentile  &          $-5.512$ &         $-4.992$ ($\pm$ 0.121) &         $-3.590$ &         $-3.054$ ($\pm$ 0.078) \\
$99^{\mathrm{th}}$ percentile &  ${\TsVIII}4.071$ & ${\TsVIII}4.167$ ($\pm$ 0.078) & ${\TsVIII}3.862$ & ${\TsVIII}3.240$ ($\pm$ 0.105) \\
\hline
\multirow{2}{*}{Statistic}
& \multicolumn{2}{c|}{Dimension 3} & \multicolumn{2}{c}{Dimension 4}  \\
& Training & Simulated & Training & Simulated \\
Mean                          &         $-0.002$ & ${\TsVIII}0.064$ ($\pm$ 0.029) &         $-0.063$ & ${\TsVIII}0.046$ ($\pm$ 0.031) \\
Standard deviation            & ${\TsVIII}1.988$ & ${\TsVIII}1.942$ ($\pm$ 0.023) & ${\TsVIII}1.982$ & ${\TsVIII}1.910$ ($\pm$ 0.026) \\
$1^{\mathrm{st}}$ percentile  &         $-4.691$ &         $-4.382$ ($\pm$ 0.109) &         $-4.895$ &         $-4.512$ ($\pm$ 0.122) \\
$99^{\mathrm{th}}$ percentile & ${\TsVIII}4.677$ & ${\TsVIII}4.572$ ($\pm$ 0.109) & ${\TsVIII}4.431$ & ${\TsVIII}4.303$ ($\pm$ 0.139) \\
\hline
\end{tabular}
\end{table}

\begin{figure}[tbh]
\caption{Comparison between the empirical correlation matrix and the average correlation matrix of Gaussian RBM simulated data}
\label{fig:multiGRBMcorr}
\begin{subfigure}[b]{0.5\linewidth}
\centering
\includegraphics[width=0.8\linewidth]{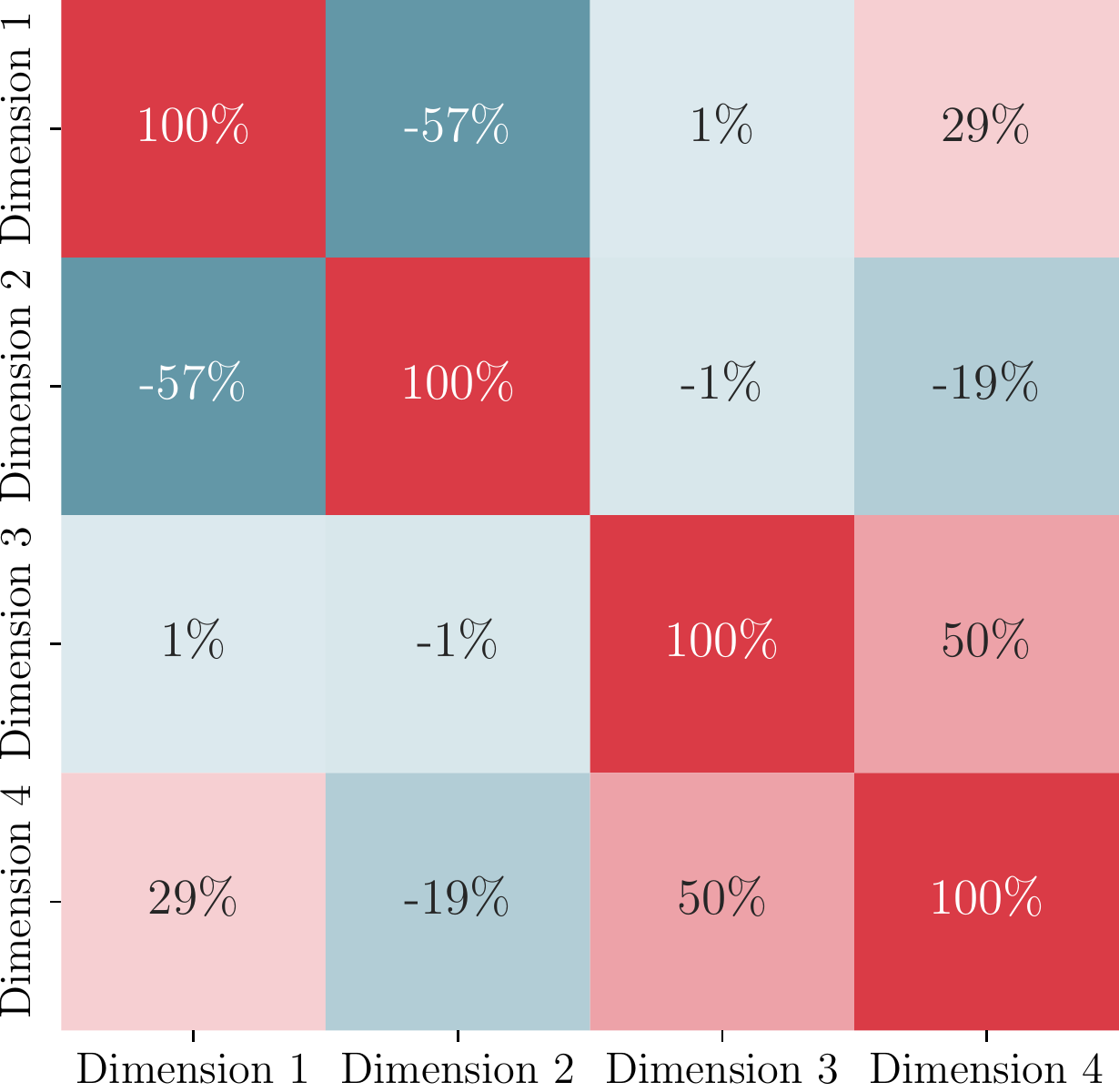}
\caption*{Training samples}
\end{subfigure}
\begin{subfigure}[b]{0.5\linewidth}
\centering
\includegraphics[width=0.8\linewidth]{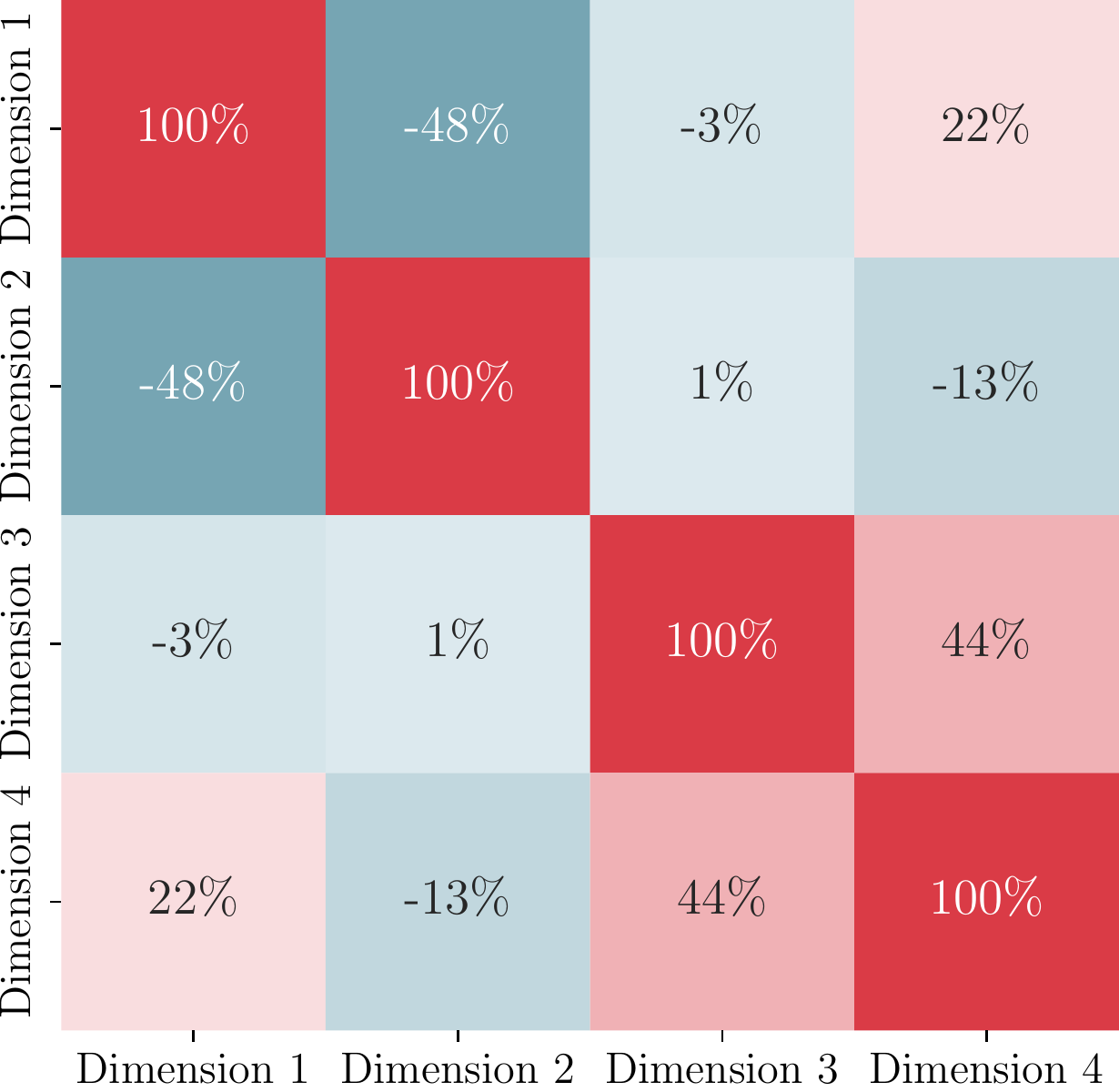}
\caption*{Generated samples}
\end{subfigure}
\end{figure}

In Figure \ref{fig:multiGRBMcorr}, we also compare the empirical
correlation matrix of the training dataset and the average of the
correlation matrices computed with 50 Monte Carlo replications.
Compared to the Bernoulli RBM, the Gaussian RBM captures much better
the correlation structure of the training dataset. This is
particular true for the largest correlation values. For instance,
the correlation between first and second dimensions is equal to
$-57\%$ for the training data, $-31\%$ for the Bernoulli RBM
simulated data and $-48\%$ for the Gaussian RBM simulated data.

\paragraph{Summary of the results}

According to our tests, we consider that both Bernoulli and Gaussian
well-trained RBMs can transform random noise series to the joint distribution
of a training dataset. But they have their own characteristics:
\begin{itemize}
\item A Bernoulli RBM is very sensitive to extreme values of the training
    dataset and has a tendency to overestimate the tail of distribution. On
    the contrary, a Gaussian RBM has a tendency to underestimate the tails of
    probability distribution and faces difficulties in learning leptokurtic probability
    distributions.
\item Gaussian RBMs may capture more accurately the correlation structure of
    the training dataset than Bernoulli RBMs.
\end{itemize}
In practice, we may apply the normal score transformation in order
to transform the training dataset, such that each marginal has a
standard normal distribution. For that, we rank the values of each
dimension from the lowest to the highest, map these ranks to a
uniform distribution and apply the inverse Gaussian cumulative
distribution function. We then train the RBM with these transformed
values. After the training process, the marginals of samples
generated by well-trained RBMs will follow a standard normal
distribution. By using the inverse transformation, we can generate
synthetic samples with the same marginal distributions than those of
the training dataset, and a correlation structure that is closed to
the empirical correlation matrix of the training data. In this case,
we may consider Gaussian RBMs as an alternate method of the
bootstrap sampling approach.

\subsubsection{Application to financial time series}
\label{section:SPXRBM} 
\begin{figure}[tbph]
\caption{Historical prices of the S\&P $500$ and VIX indices}
\label{fig:SpxVixCRBM}
\begin{subfigure}[b]{0.5\linewidth}
\centering
\includegraphics[width=\linewidth]{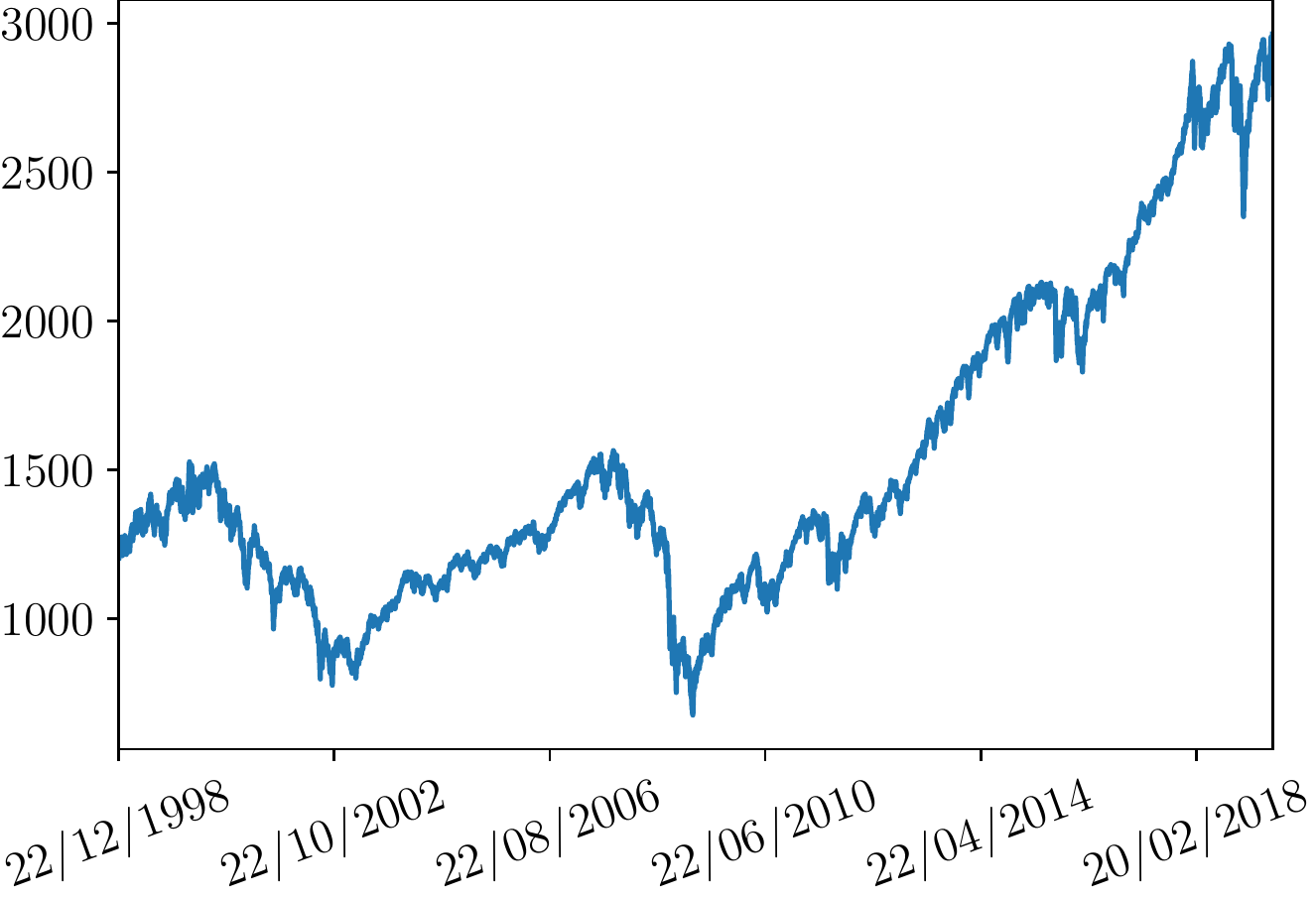}
\caption*{S\&P 500 index}
\end{subfigure}
\begin{subfigure}[b]{0.5\linewidth}
\centering
\includegraphics[width=\linewidth]{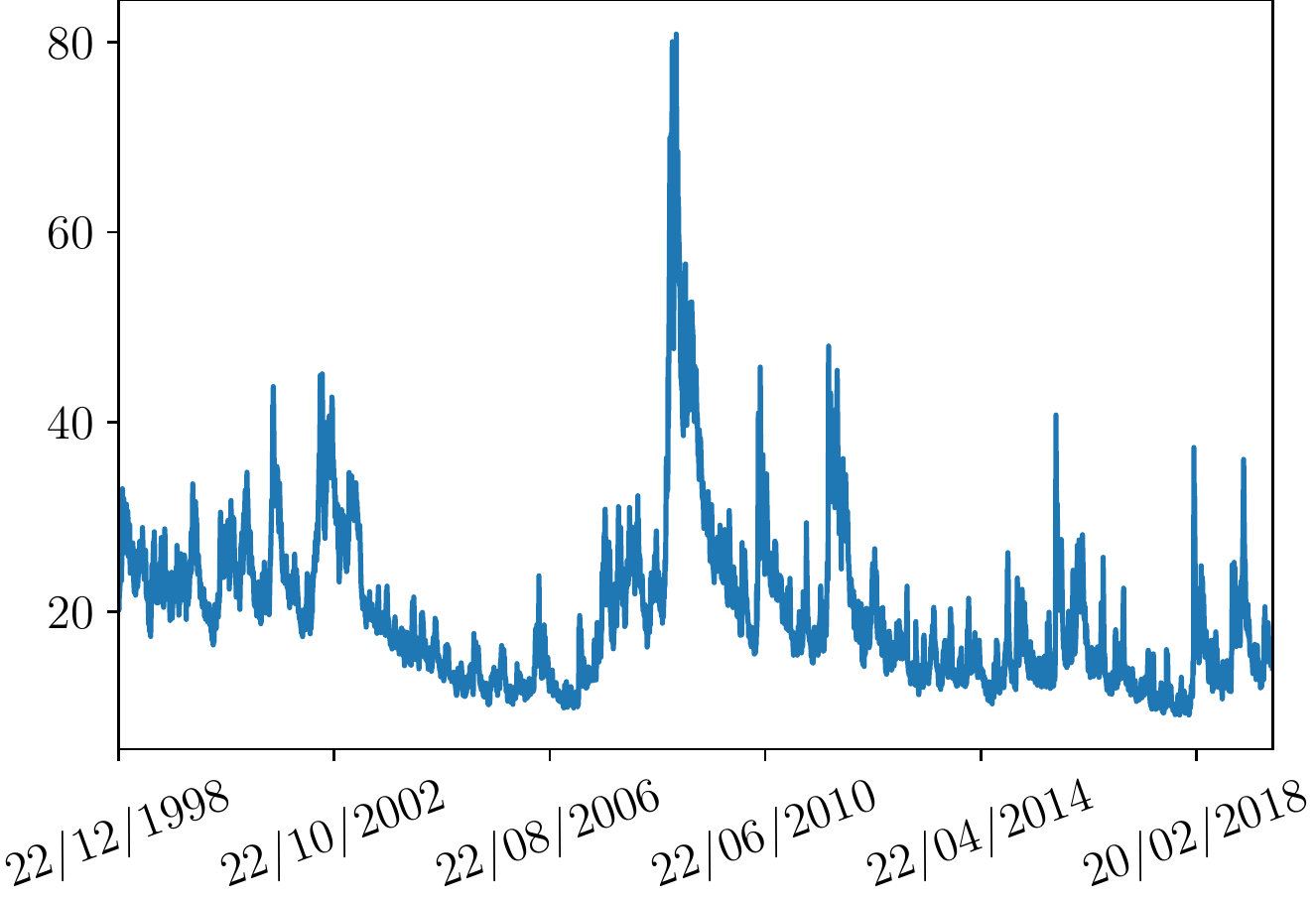}
\caption*{VIX index}
\end{subfigure}
\end{figure}

\begin{figure}[tbph]
\centering
\caption{Autocorrelation function of the training dataset}
\label{fig:AutocorrSpxVix}
\includegraphics[scale=0.5]{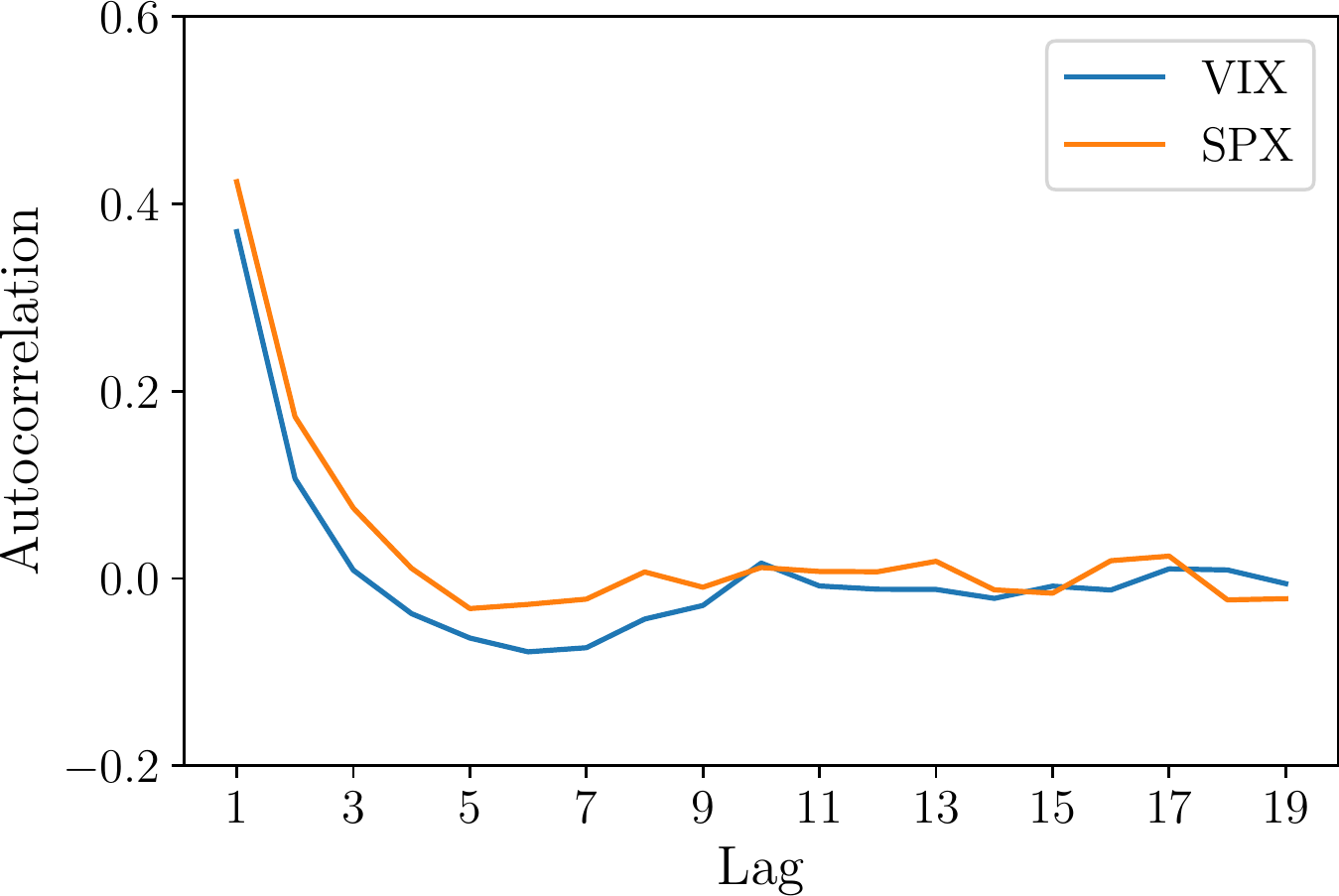}
\end{figure}

According to the results in the previous section,
Gaussian RBMs capture more accurately the correlation structure of the training
dataset and we can avoid its drawbacks by applying the normal score
transformation. In this paragraph, we test and compare a Gaussian RBM and a
conditional RBM on a multi-dimensional autocorrelated time series in order to
check if conditional RBMs may capture at the same time the correlation
structure and the time dependence of the training dataset. The RBM is trained
on a real financial dataset consisting of $5\, 354$ historical daily returns of
the S\&P 500 index and the VIX index from December 1998 to May 2018 (see Figure
\ref{fig:SpxVixCRBM}). During this period, the S\&P 500 index and the VIX index
have a remarkable negative correlation that is equal to $-71\%$. In order to
reinforce the autocorrelation level in the training samples, we apply a $3$-day
exponential weighted moving average approach to historical prices before
calculating daily returns. Figure \ref{fig:AutocorrSpxVix} shows the
autocorrelation of training data. As both daily returns of the S\&P 500 and VIX
indices are leptokurtic and have some extreme values, we need to apply the
normal score transformation on input data before the training
process.\smallskip

\paragraph{Normal score transformation}

In order to verify the learning quality of Gaussian RBMs using the
normal score transformation, we run $50$ Monte Carlo simulations
after the training process and for each simulation, we generate
$5\,354$ simulated observations from different random noise series.
We then calculate the statistics of these samples as we have done in
the previous section. According to Table \ref{table:SpxVixGRBM}, we
find that although the Gaussian RBM still underestimates a little
the standard deviation, the $1^{\mathrm{st}}$ percentile and the
$99^{\mathrm{th}}$ percentile, the learning quality is much more
improved. In addition, the average correlation coefficient between
daily returns of S\&P 500 and VIX indices over 50 Monte Carlo
simulations is equal to $-76\%$, which is very closed to the figure
$71\%$ of the empirical correlation. Therefore, we consider that
training Gaussian RBMs with the normal score transformation may be
used as an alternate method of bootstrap sampling.\smallskip

\begin{table}[tbh]
\centering
\caption{Comparison between the training sample and Gaussian RBM simulated samples using the normal score transformation}
\label{table:SpxVixGRBM}
\begin{tabular}{l|cc|cc}
\hline
\multirow{2}{*}{Statistic}
& \multicolumn{2}{c|}{S\&P 500 index} & \multicolumn{2}{c}{VIX index}  \\
& Training & Simulated & Training & Simulated \\
Mean                          &  ${\TsVIII}0.02\%$ &         $-0.01\%$ ($\pm$ $0.01\%$) & ${\TsVIII}0.06\%$ & ${\TsVIII}0.09\%$ ($\pm$ $0.06\%$) \\
Standard deviation            &  ${\TsVIII}0.64\%$ & ${\TsVIII}0.62\%$ ($\pm$ $0.01\%$) & ${\TsVIII}3.84\%$ & ${\TsVIII}3.65\%$ ($\pm$ $0.12\%$) \\
$1^{\mathrm{st}}$ percentile  &          $-1.93\%$ &         $-1.88\%$ ($\pm$ $0.12\%$) &         $-7.31\%$ &         $-6.97\%$ ($\pm$ $0.30\%$) \\
$99^{\mathrm{th}}$ percentile &  ${\TsVIII}1.61\%$ & ${\TsVIII}1.49\%$ ($\pm$ $0.06\%$) & ${\TsIII}11.51\%$ & ${\TsIII}10.96\%$ ($\pm$ $0.63\%$) \\
\hline
\end{tabular}
\end{table}

\paragraph{Learning time dependence}

As conditional RBM is an extension of Gaussian RBM, applying the
normal score transformation should still work. Moreover, conditional RBM should be
able to learn the time-dependent relationship by its design. So, we train a
conditional RBM on the same training dataset in order to check whether this
model can capture the autocorrelation patterns previously given in Figure
\ref{fig:AutocorrSpxVix}.\smallskip

\begin{figure}[tbph]
\caption{Autocorrelation function of synthetic samples generated by a Gaussian RBM}
\label{fig:SpxVixGRBMAuto}
\begin{subfigure}[b]{0.5\linewidth}
\centering
\includegraphics[width=\linewidth]{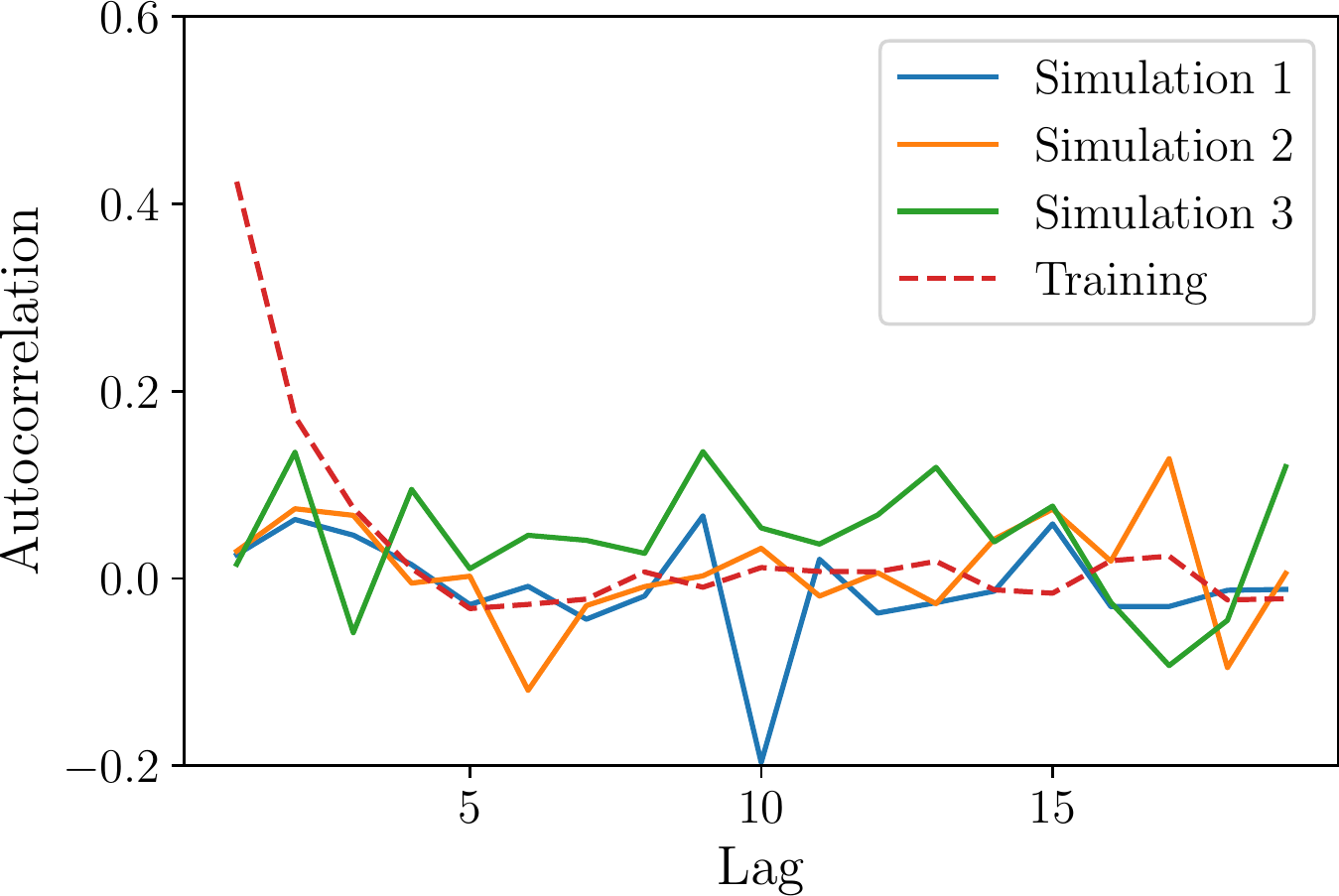}
\caption*{S\&P 500 index}
\end{subfigure}
\begin{subfigure}[b]{0.5\linewidth}
\centering
\includegraphics[width=\linewidth]{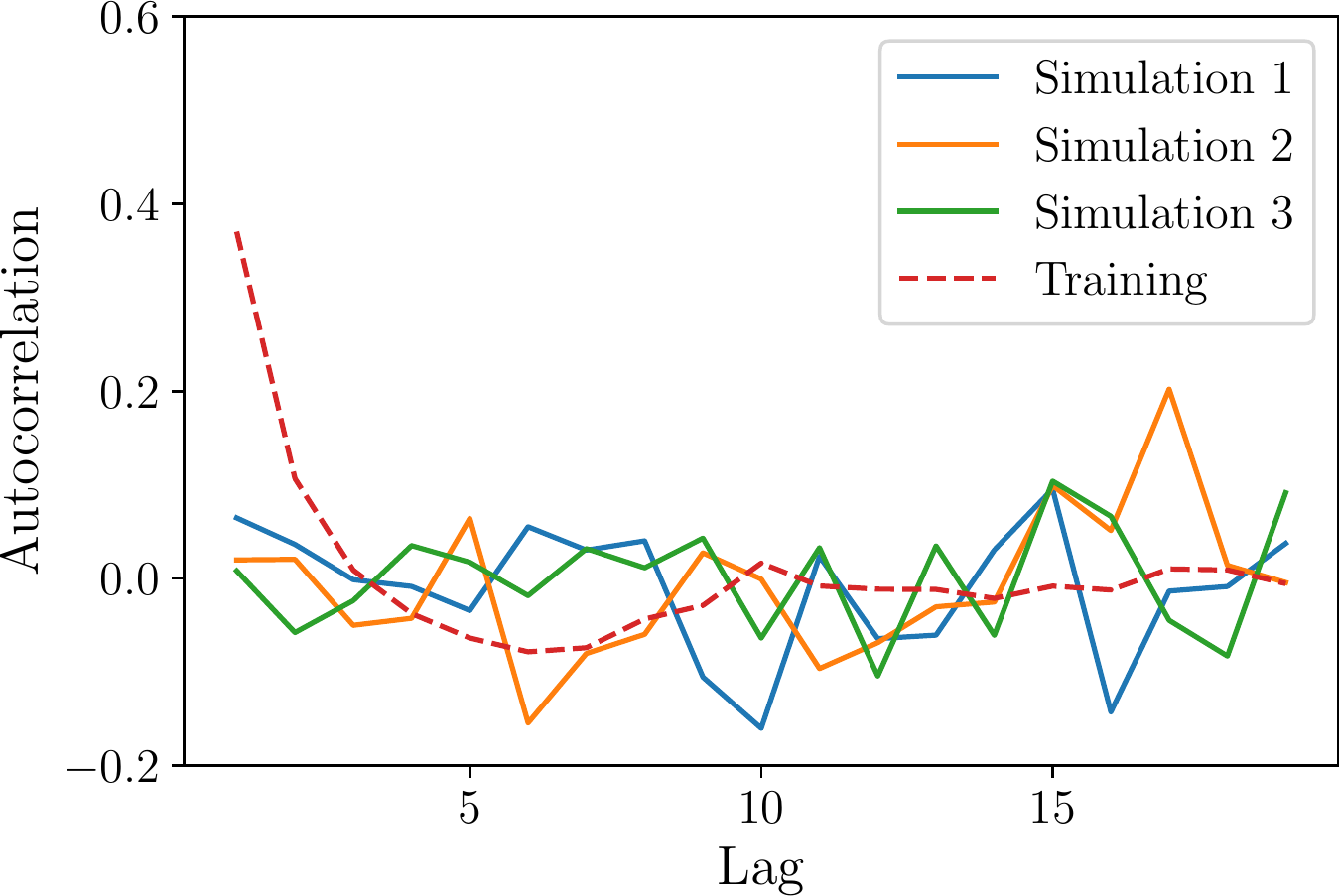}
\caption*{VIX index}
\end{subfigure}
\end{figure}

\begin{figure}[ht]
\caption{Autocorrelation function of synthetic samples generated by a conditional RBM}
\label{fig:SpxVixCRBMAuto}
\begin{subfigure}[b]{0.5\linewidth}
\centering
\includegraphics[width=\linewidth]{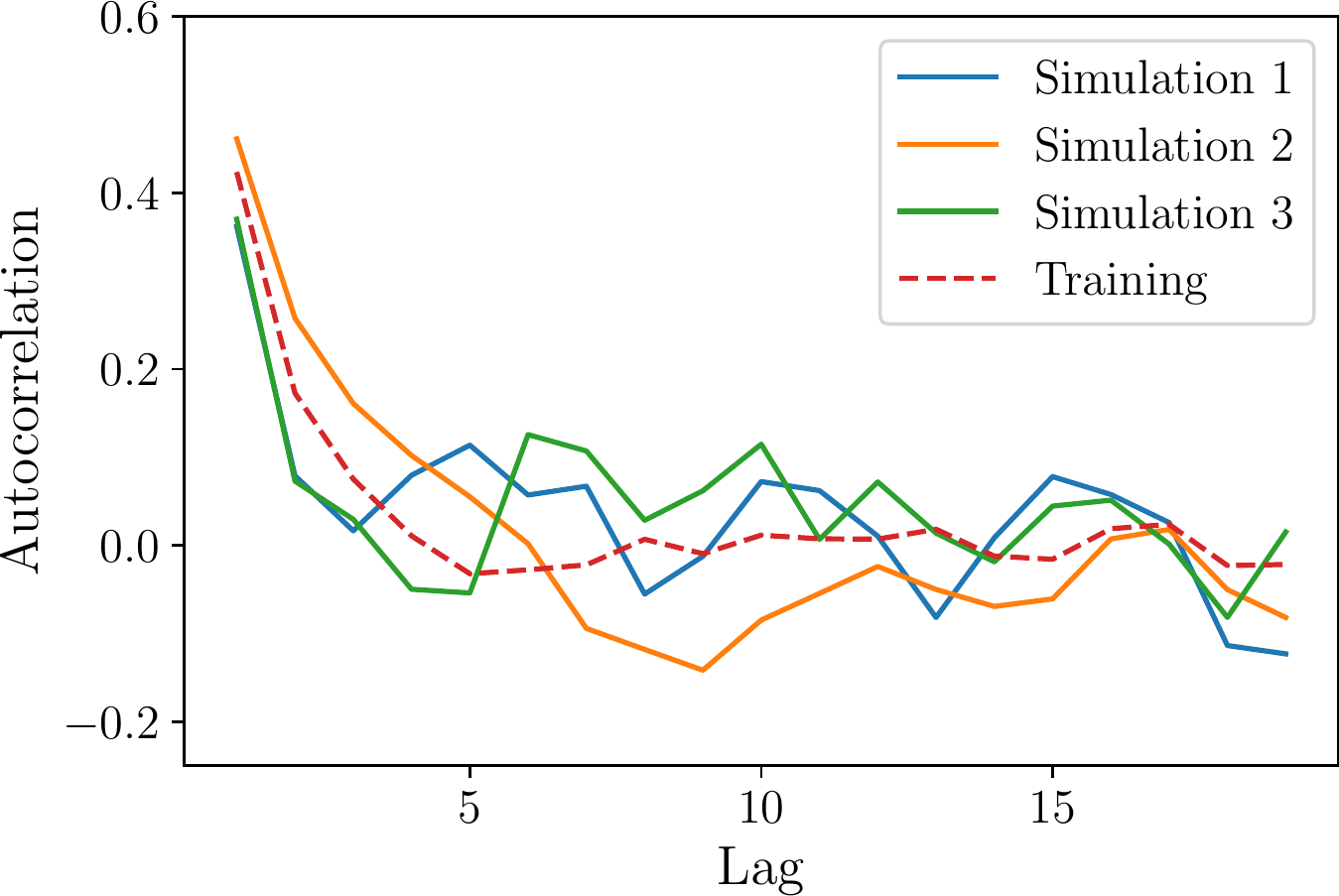}
\caption*{S\&P 500 index}
\end{subfigure}
\begin{subfigure}[b]{0.5\linewidth}
\centering
\includegraphics[width=\linewidth]{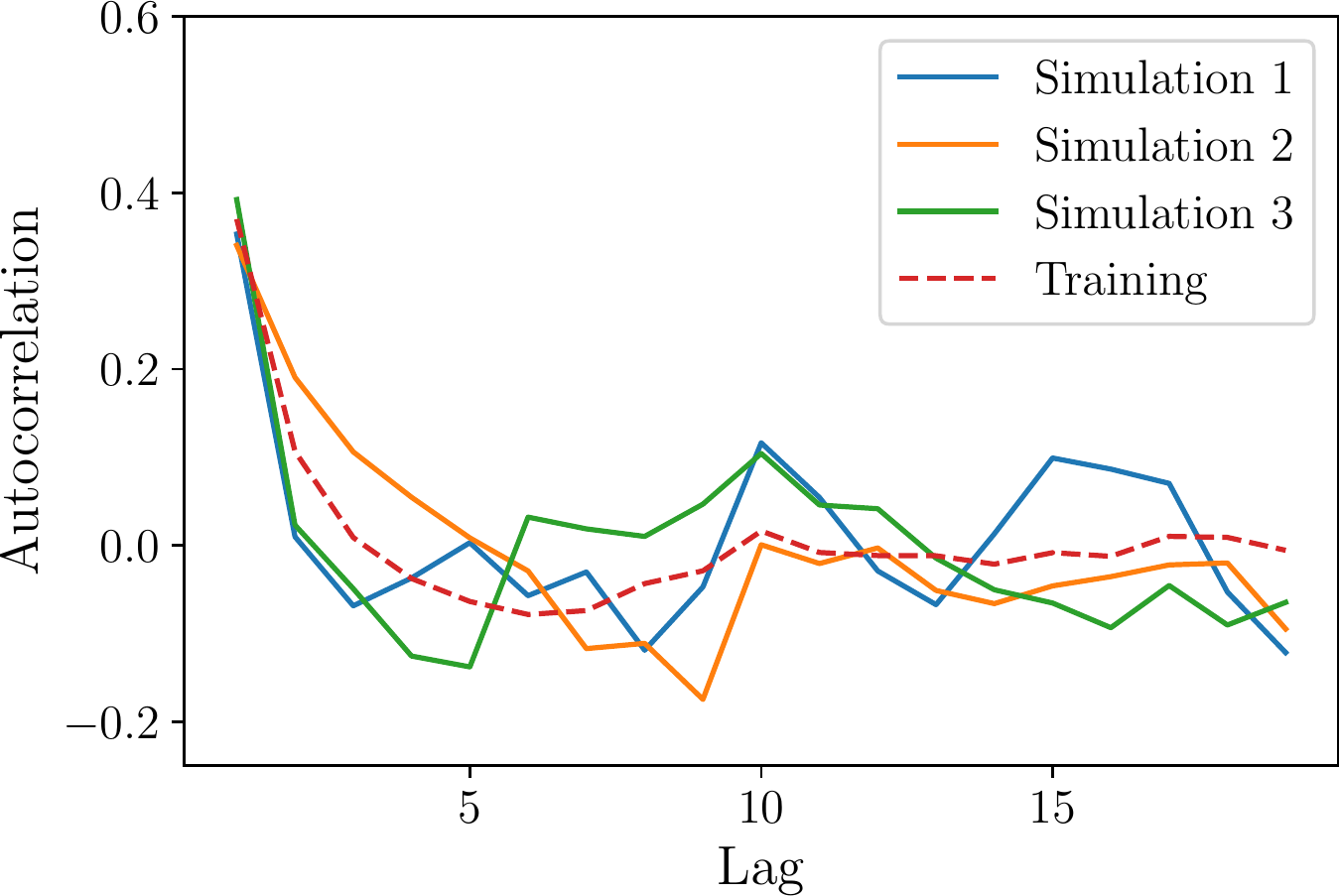}
\caption*{VIX index}
\end{subfigure}
\end{figure}

In practice, we choose to use a long memory for the conditional layer in the
conditional RBM and the values of the last $20$ days will be fed to the model.
Therefore, the conditional RBM has $2$ visible layer units, $128$ hidden layer
units and $40$ conditional layer units\footnote{Corresponding to two
time series of $20$ lags.}. We then use the normal score transformation in
order to transform the training dataset as mentioned before. After having done
a training process of $100\,000$ epochs, we are interested in generating
samples of consecutive time steps to verify if the conditional RBM can learn
the joint distribution but also capture the autocorrelation of the training
dataset. We run $3$ Monte Carlo simulations and for each simulation, we
generate consecutively samples of $250$ observations. In Figures
\ref{fig:SpxVixGRBMAuto} and \ref{fig:SpxVixCRBMAuto}, we compare the
autocorrelation function of generated samples by Gaussian and conditional RBMs.
These results show clearly that a conditional RBM can capture well the
autocorrelation of the training dataset, which is not the case of a traditional
Gaussian RBM.

\paragraph{Market generator}

Based on above results, we consider that conditional RBMs can be used as a
market generator \citep{Kondratyev-2019}. In the example of S\&P 500 and VIX
indices, we have trained the models with all historical observations and for
each date, we have used the values of the last $20$ days as a memory vector for
the conditional layer. In this approach, the normal score transformation
ensures the learning of marginal distributions, whereas the conditional RBM
ensures the learning of the correlation structure and the time dependence.
After having calibrated the training process, we may choose a date as the
starting point. The values of this date and its last 20 days are then passed to
the trained RBM model and, after sufficient steps of Gibbs sampling, we get a
sample for the next day. We then feed this new sample and the updated memory
vector to the model in order to generate a sample for the following day.
Iteratively, the conditional RBM can generate a multi-dimensional time series
that has the same patterns as those of the training data in terms of marginal
distribution, correlation structure and time dependence.\smallskip

\begin{figure}[tbph]
\caption{Comparison between scaling historical prices of S\&P 500 and VIX indices and
synthetic time series generated by the conditional RBM}
\label{fig:SpxVixCRBMFake}
\begin{subfigure}[b]{0.5\linewidth}
\centering
\includegraphics[width=\linewidth]{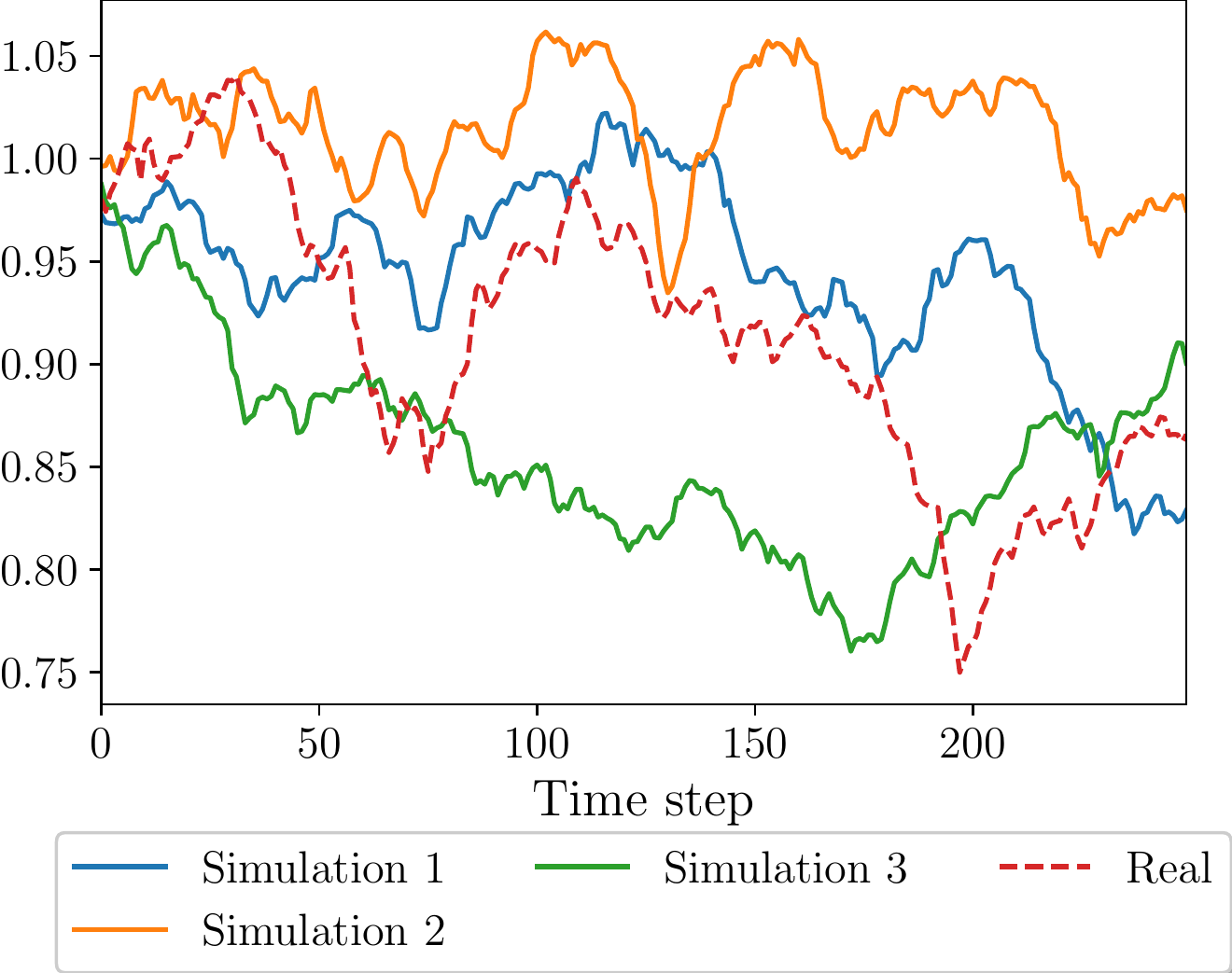}
\caption*{S\&P 500 index}
\end{subfigure}
\begin{subfigure}[b]{0.5\linewidth}
\centering
\includegraphics[width=\linewidth]{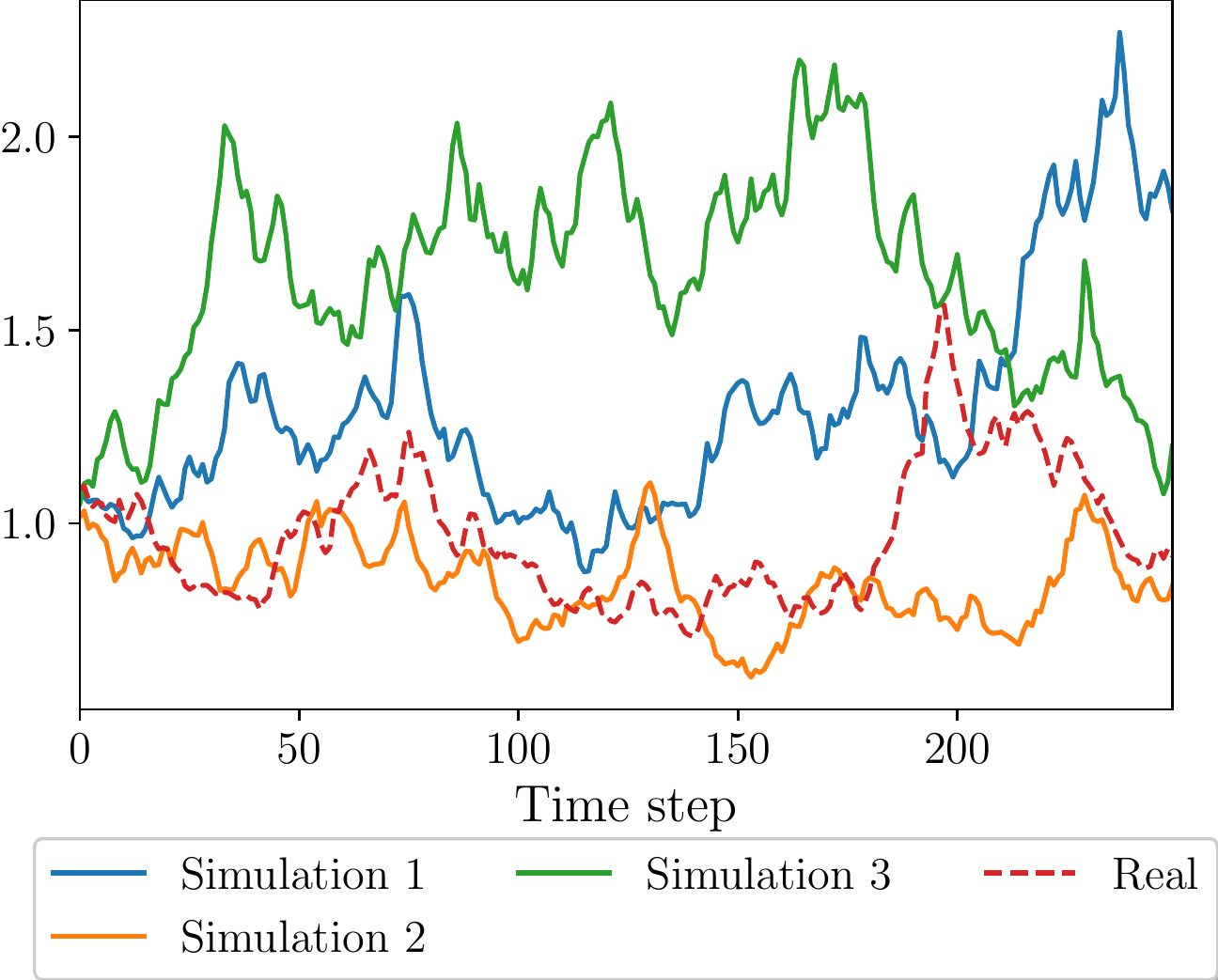}
\caption*{VIX index}
\end{subfigure}
\end{figure}

For instance, Figure \ref{fig:SpxVixCRBMFake} shows three time
series of $250$ trading days starting at the date 22/11/2000, which
are generated by the trained conditional RBM. We notice that these
three simulated time series have the same characteristics as the
historical prices represented by the dashed line. The negative
correlation between S\&P 500 and VIX indices is also learned and
we can see clearly that there exists a positive autocorrelation in
each simulated one-dimensional time series.

\subsection{Application of GANs to generate fake financial time series}

In this section, we perform the same tests for GAN models as those we have done
for RBMs. The objective is to compare the results between these two types of
generative models on the same training dataset. We recall that the first test
consists of learning the joint distribution of a simulated multi-dimensional
dataset with different marginal distributions and a simple correlation
structure driven by a Gaussian copula. In this part, we test a Wasserstein GAN
model with only simple dense layers in the generator and in the discriminator.
In the second test, we train a conditional Wasserstein GAN model on the
historical daily returns of the S\&P 500 index and the VIX index from December
1998 to May 2018. As we have done in the previous tests for RBMs, we also apply
a three-day exponential weighted moving average approach to historical prices
before calculating daily returns. We know that this transformation will
reinforce the autocorrelation level in the training samples. As a result, we
need to choose a more complex structure for the generator and the discriminator
to capture the complex features in the training dataset. For instance, we
construct the generator and the discriminator with convolutional neural network
as introduced in Section \ref{section:cnn} on page \pageref{section:cnn}.
According to the results of these two tests, we check whether the family of GAN
models may also learn the marginal distribution, the correlation structure and
the time dependence of the training dataset as RBMs can do.

\subsubsection{Simulating multi-dimensional datasets}

As we have introduced in Section \ref{section:wasserstein-gan} on
page \pageref{section:wasserstein-gan}, Wasserstein GAN models, which
use the Wasserstein distance as the loss function, have several
advantages comparing the basic traditional GAN models using the
cross-entropy loss measure. Therefore, we test in this first study
Wasserstein GAN models on the simulated multi-dimensional dataset as
we have done for Bernoulli and Gaussian RBMs. We recall that we
simulate $10\,000$ samples of $4$-dimensional data with different
marginal distributions (Gaussian mixture model, Student's $t$
distribution and two Gaussian distributions), as shown in Figure
\ref{fig:HistMultiSimulated} on page
\pageref{fig:HistMultiSimulated} and the training dataset has a
correlation structure simulated by a Gaussian copula with the
correlation matrix given in Figure \ref{fig:CorrMatCopula} on page
\pageref{fig:CorrMatCopula}. Here, the objective of the test is to
check whether a well-trained Wasserstein GAN models can learn the
marginal distribution and the copula function of training samples.

\paragraph{Data preprocessing}

In the practice of GAN models, we need to match the dimension and
the range of the output of the generator and the input of
discriminator. To address this issue, there are two possible ways:

\begin{itemize}
\item We can use a complex activation function for the last layer of the
generator in order to ensure that the generated samples have the
same range as the training samples for the discriminator
\citep{Clevert-2015}.

\item We can modify the range of the training samples by applying a
preprocessing function and choose a usual activation function
\citep{Wiese-2019}.
\end{itemize}

In our study, we consider the second approach by applying the MinMax
scaling function\footnote{We have $f\left( x\right) =\left( x-\min
x\right) /\left(\max x-\min x\right) $.} to the training samples for
the discriminator.  As a result, the input data scaled for the
discriminator will take their values in $\left[0, 1\right]$ and we
may choose the sigmoid activation function for the last layer of the
generator to ensure that the outputs of the generator have theirs
values in $\left[0, 1\right]$. In addition, we need to simulate the
random noise as the inputs for the generator. As the random noise
will play the role of latent variables, we have the freedom to
choose its distribution. For instance, we use the Gaussian
distribution $ z\sim \mathcal{N}\left( 0,1\right) $ in this study.

\paragraph{Structure and training of Wasserstein GAN models}

As we have mentioned, we need to match the output of the generator
and the input of the discriminator. Therefore, the last layer in the
generator should have 4 neurons for our simulated training dataset
and we may use sigmoid activation function for each neuron as we
have applied MinMax scaling function to the training samples. The
discriminator corresponds to a traditional classification problem,
so its last layer should have only one neuron. We may choose the
sigmoid activation function for this neuron if the training samples
are labelled by $\{0, 1\}$ or the tangent hyperbolic activation
function if the training samples are labelled by $\{-1, 1\}$.
\smallskip

In our study, we want to construct a Wasserstein GAN model, which can
convert a $100$-dimensional random noise vector to the data with the
same joint distribution as the training dataset. According to our
tests, a simple structure as multi-layer perceptrons for the
generator and the discriminator is sufficient for learning the
distribution of these simulated training samples. More precisely,
the generator is fed by $100$-dimensional vectors and has 5 layers
with the structure $\mathcal{S}_{\Generator}=\left\{200, 100, 50,
25, 4\right\} $ where $s_{i}\in \mathcal{S}_{\Generator}$ represents
the numbers of neurons of the $i^{\mathrm{th}}$ layer. The
discriminator has 4 layers with
$\mathcal{S}_{\Discriminator}=\left\{100, 50, 10, 1\right\} $ and
takes 4-dimensional vectors as input data. As explained in the
previous paragraph, we may choose the sigmoid activation function
for the last layer of the generator and the tangent hyperbolic
activation function for the last layer of the discriminator. For all
other layers of the discriminator and the generator, the activation
function $f\left( x\right) $ corresponds to a leaky rectified linear
unit (or RELU) function with $\alpha = 0.5$:
\begin{equation*}
f_{\alpha}\left( x\right) =\left\{
\begin{array}{ll}
{\alpha x} & {\text{if }x<0} \\
{x} & {\text{otherwise}}%
\end{array}%
\right.
\end{equation*}
According to \citet{Arjovsky-2017a}, the Wasserstein GAN model
should be trained with the RMSProp optimizer using a small learning
rate. Therefore, we set the learning rate to $\eta = 1.10^{-4}$ and
use mini-batch gradient descent with batch size $500$. Table
\ref{table:WGANconfig} summarizes the setting of the Wasserstein GAN
model implementation using Python and the TensorFlow library. After
the training process, we generate $10\,000$ samples in order to
compare the marginal distributions and  the correlation structure
with the training dataset.

\begin{table}[tbph]
\centering
\caption{The setting of the Wasserstein GAN model}
\label{table:WGANconfig}
\begin{tabular}{p{6cm}p{5cm}}
\hline
Training data dimension & 4 \\
Input feature scaling function       & MinMax                                                        \\
Random noise vector dimension        & 100                                                           \\
Random noise distribution            & $\mathcal{N}\left(0, 1\right)$                                \\ \hdashline
Generator structure                  & $\mathcal{S}_{\Generator}=\left\{200, 100, 50, 25, 4\right\}$ \\
Discriminator structure              & $\mathcal{S}_{\Discriminator}=\left\{100, 50, 10, 1\right\}$  \\ \hdashline
Loss function                        & Wasserstein distance                                          \\
Learning optimizer                   & RMSProp                                                       \\
Learning rate                        & $1.10^{-4}$                                                   \\
Batch size                           & 500                                                           \\
\hline
\end{tabular}
\end{table}

\paragraph{Results}

\begin{figure}[tbph]
\caption{Histogram of training and WGAN simulated samples}
\label{fig:multiWGANfake}
\begin{subfigure}[b]{0.5\linewidth}
\centering
\includegraphics[width=0.9\linewidth]{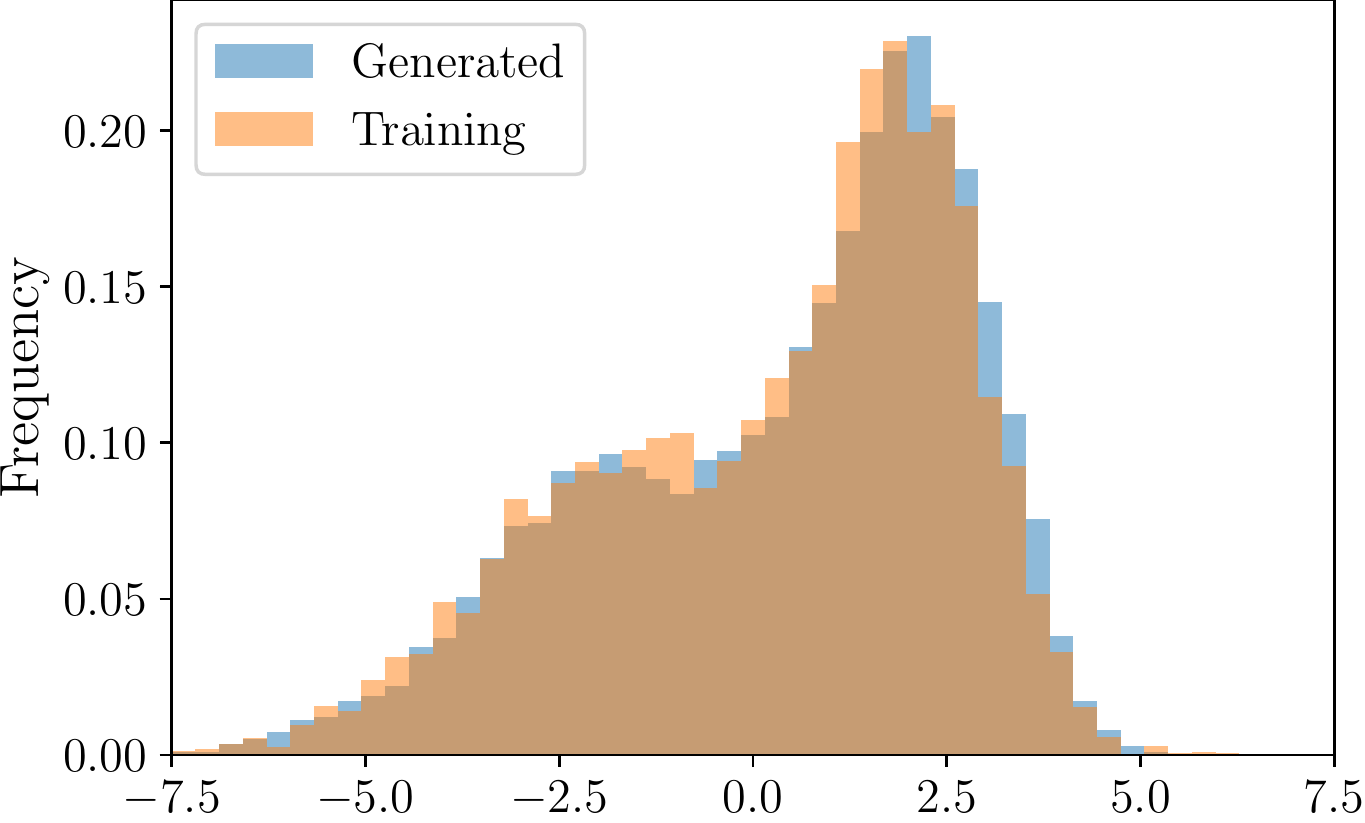}
\caption{Gaussian mixture model}
\vspace*{0.5cm}
\end{subfigure}
\begin{subfigure}[b]{0.5\linewidth}
\centering
\includegraphics[width=0.9\linewidth]{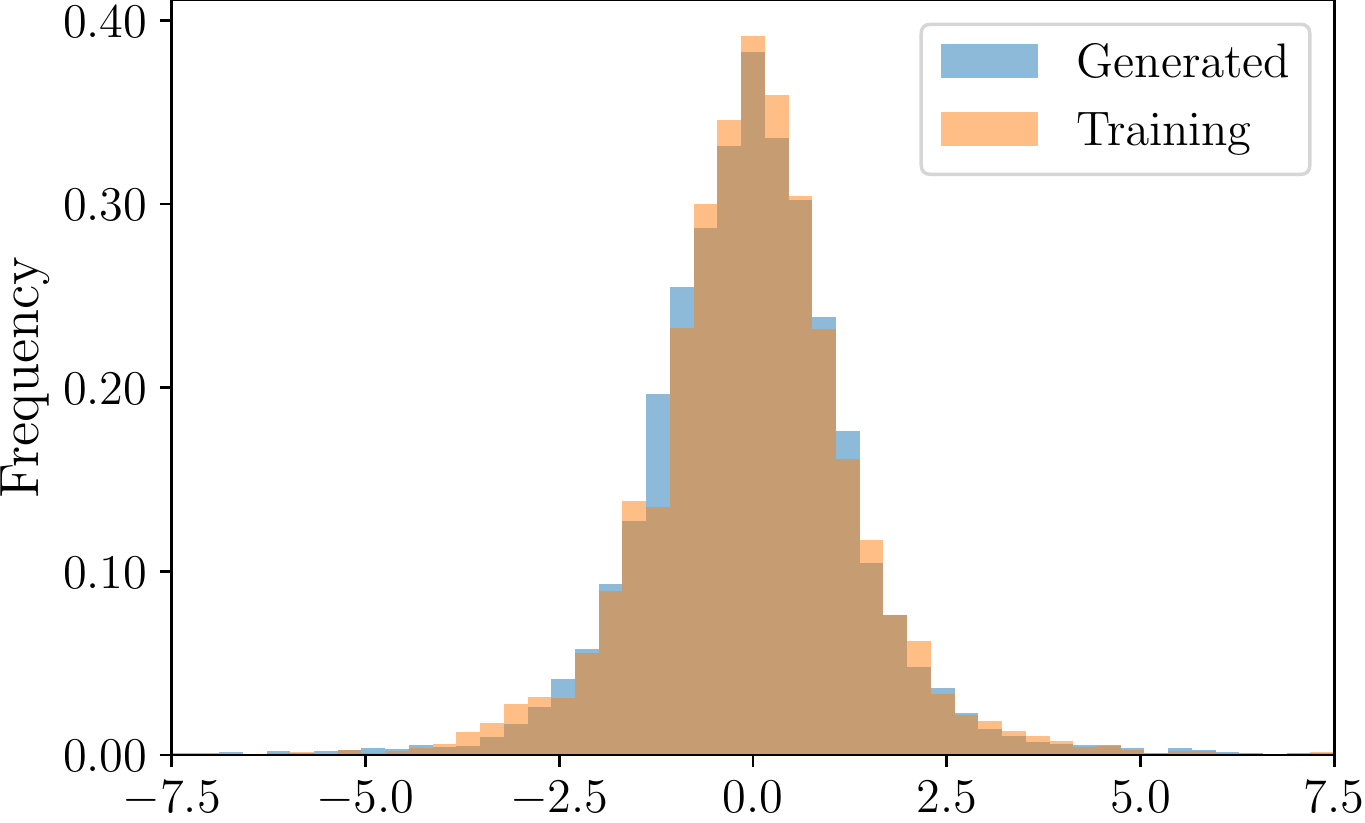}
\caption{Student's \textit{t}-distribution with $\nu = 4$}
\vspace*{0.5cm}
\end{subfigure}
\begin{subfigure}[b]{0.5\linewidth}
\centering
\includegraphics[width=0.9\linewidth]{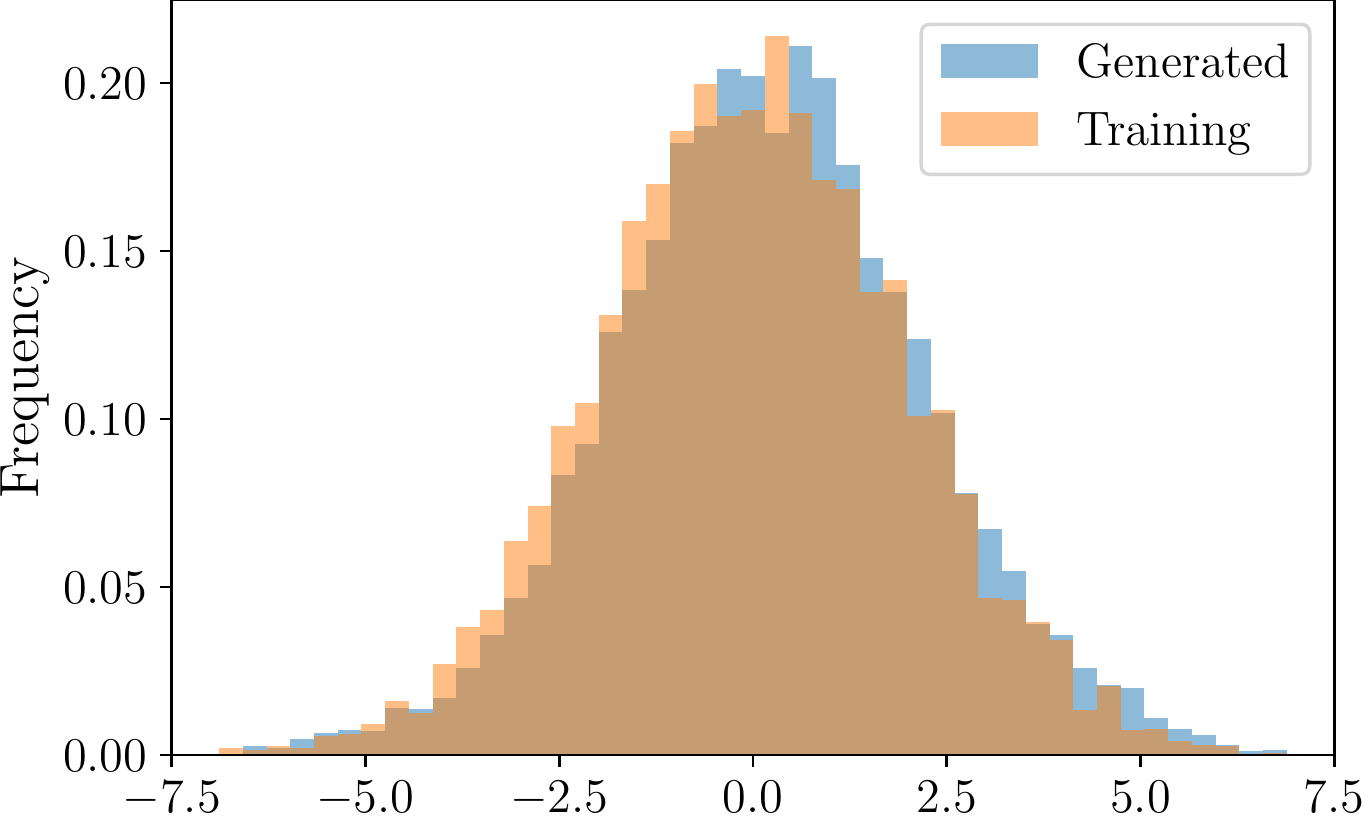}
\caption{Gaussian distribution $\mathcal{N}\left(0, 2\right)$}
\end{subfigure}
\begin{subfigure}[b]{0.5\linewidth}
\centering
\includegraphics[width=0.9\linewidth]{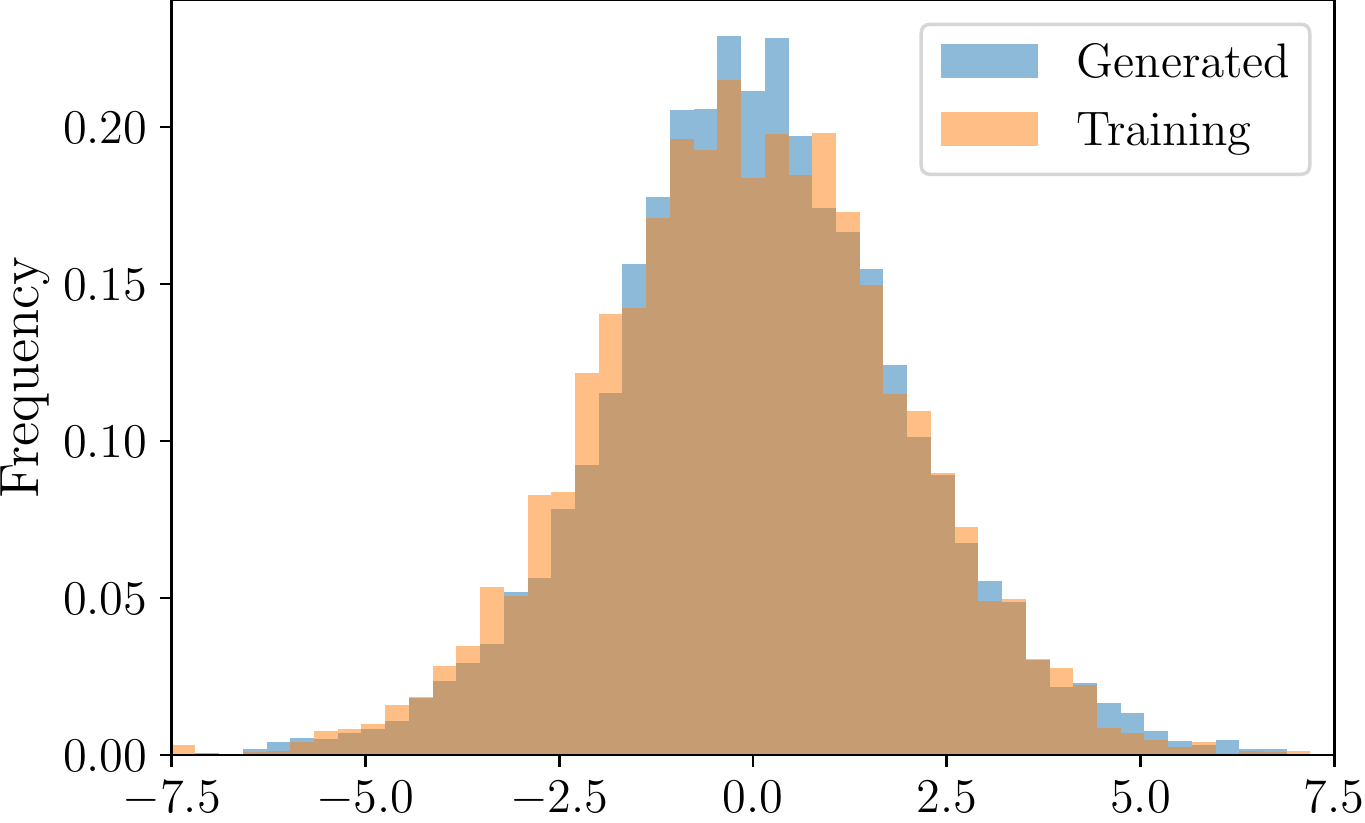}
\caption{Gaussian distribution $\mathcal{N}\left(0, 2\right)$}
\end{subfigure}
\end{figure}

\begin{figure}[tbph]
\caption{QQ-plot of training and WGAN simulated samples}
\label{fig:multiWGANqqplot}
\begin{subfigure}[b]{0.5\linewidth}
\centering
\includegraphics[width=0.9\linewidth]{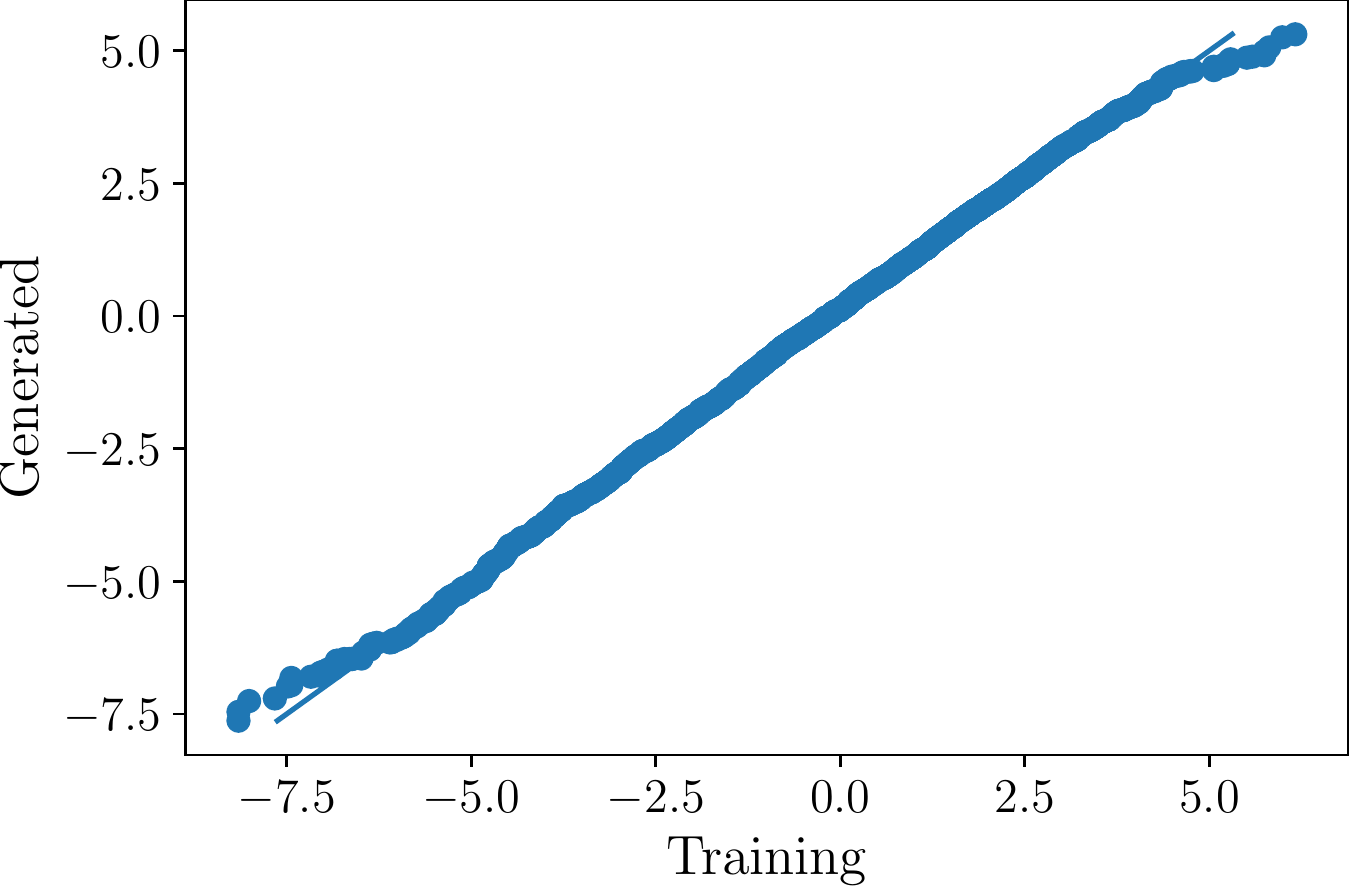}
\caption{Gaussian mixture model}
\vspace*{0.5cm}
\end{subfigure}
\begin{subfigure}[b]{0.5\linewidth}
\centering
\includegraphics[width=0.9\linewidth]{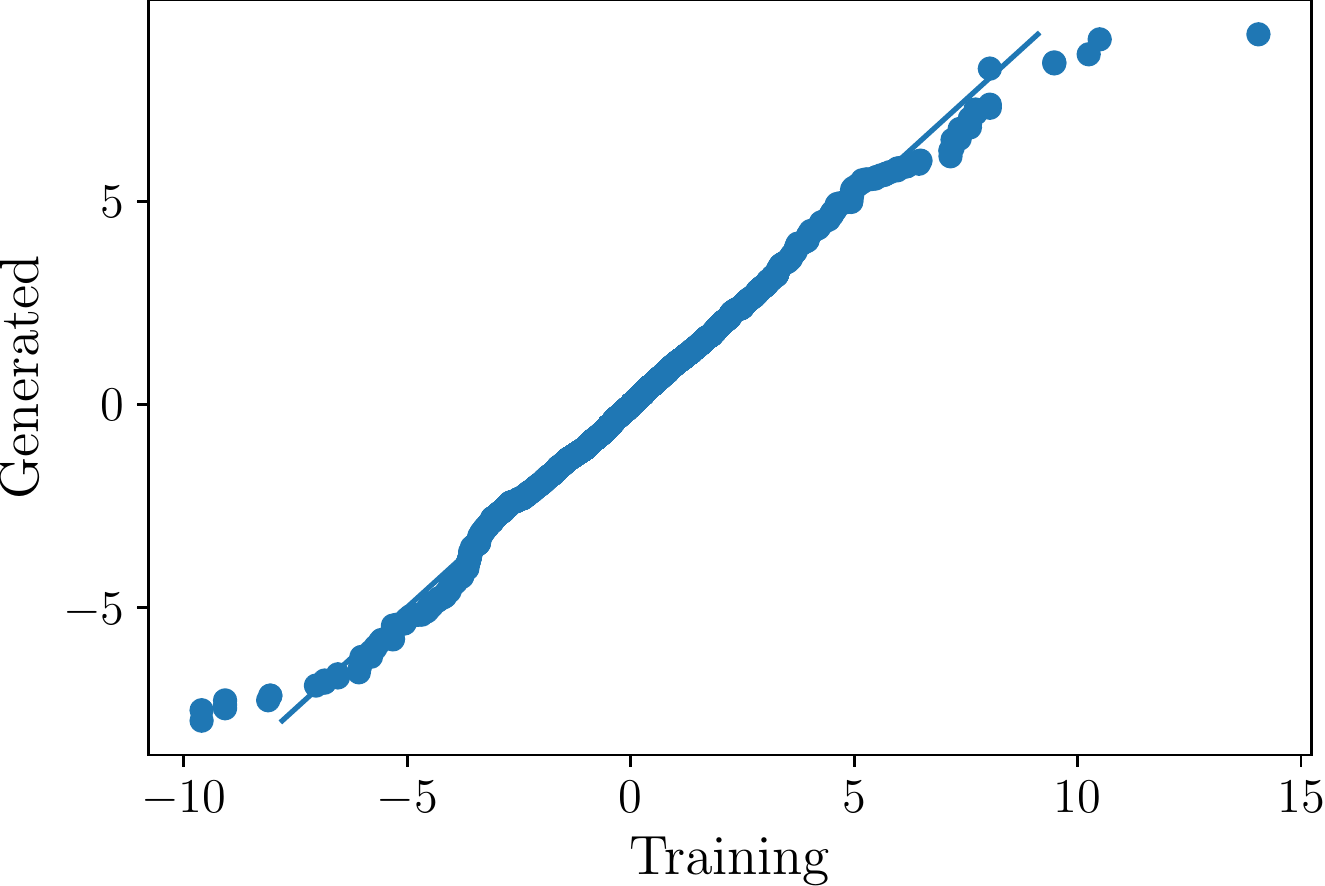}
\caption{Student's \textit{t}-distribution with $\nu = 4$}
\vspace*{0.5cm}
\end{subfigure}
\begin{subfigure}[b]{0.5\linewidth}
\centering
\includegraphics[width=0.9\linewidth]{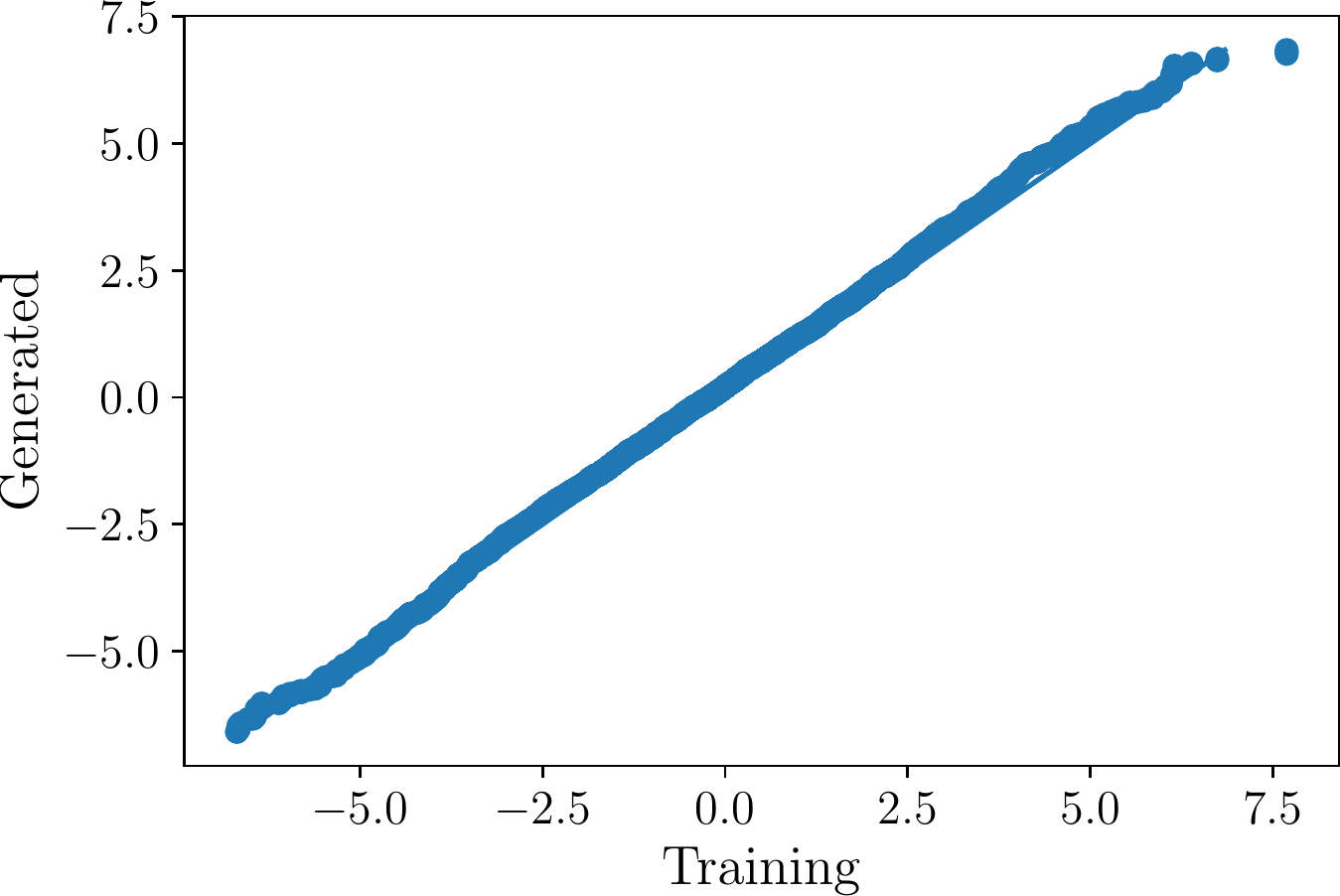}
\caption{Gaussian distribution $\mathcal{N}\left(0, 2\right)$}
\end{subfigure}
\begin{subfigure}[b]{0.5\linewidth}
\centering
\includegraphics[width=0.9\linewidth]{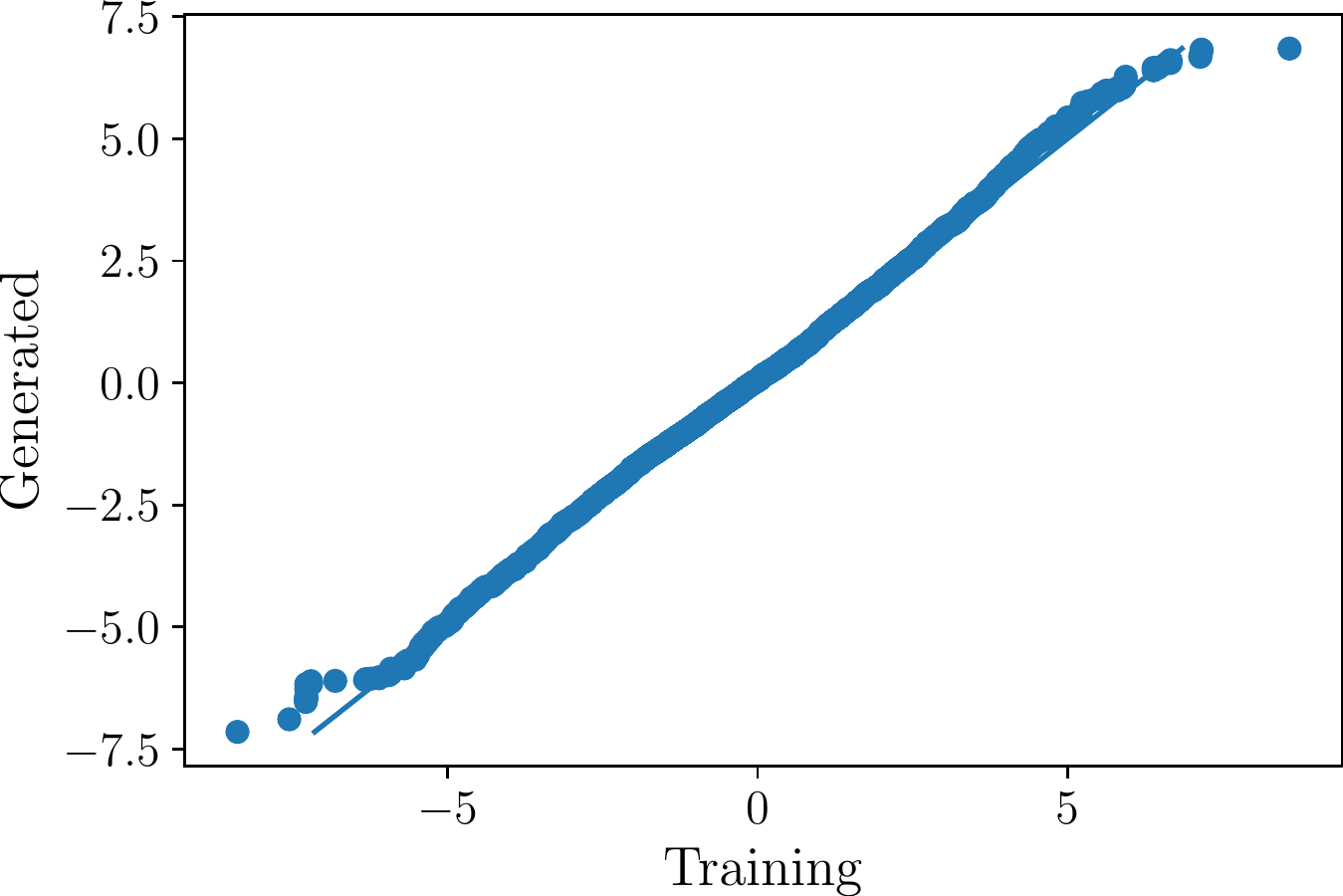}
\caption{Gaussian distribution $\mathcal{N}\left(0, 2\right)$}
\end{subfigure}
\end{figure}

Figures \ref{fig:multiWGANfake} and \ref{fig:multiWGANqqplot}
compare the histograms and QQ-plots between training samples and
generated samples. According to these figures, we observe that the
Wasserstein GAN model can learn very well each marginal distribution
of training samples, even better than Bernoulli and Gaussian RBMs.
In particular, the Wasserstein GAN model fits more accuracy for
heavy-tailed distributions than RBMs as in the case of the Student's
$t$ distribution (Panel (b) in Figure \ref{fig:multiWGANqqplot}).

As we have done for RBMs, we replicate the previous simulation $50$
times and we compare the average of the mean, the standard
deviation, the $1^{\mathrm{st}}$ percentile and the
$99^{\mathrm{th}}$ percentile between training and simulated samples
in Table \ref{table:multiWGAN}. Moreover, in the case of simulated
samples, we indicate the confidence interval by $\pm$ one standard
deviation. These statistics show that the Wasserstein GAN model
estimates very well the value of standard deviation, the
$1^{\mathrm{st}}$ percentile and the $99^{\mathrm{th}}$ percentile
for each marginal distribution, especially for Gaussian mixture
model and Student's $t$ distribution in the first and second
dimensions. One shortcoming is that the Wasserstein GAN model
slightly overestimates the value of the $99^{\mathrm{th}}$
percentile for the two Gaussian distributions. Comparing with the
results shown in Table \ref{table:multiBRBM} on page
\pageref{table:multiBRBM} for the Bernoulli RBM and Table
\ref{table:multiGRBM} on page \pageref{table:multiGRBM} for the
Gaussian RBM, we notice that the Wasserstein GAN model learns better
the marginal distributions of the training dataset and doesn't have
tendency to always underestimate or overestimate the tails of
probability distribution as RBMs. However, there exists a small bias
between the empirical mean computed with 50 Monte Carlo replications
and the mean of the training dataset.\smallskip

\begin{table}[tbh]
\centering
\caption{Comparison between training and WGAN simulated samples}
\label{table:multiWGAN}
\begin{tabular}{l|cc|cc}
\hline
\multirow{2}{*}{Statistic}
& \multicolumn{2}{c|}{Dimension 1} & \multicolumn{2}{c}{Dimension 2}  \\
& Training & Simulated & Training & Simulated \\
Mean                          &  ${\TsVIII}0.303$ & ${\TsVIII}0.427$ ($\pm$ 0.027) &         $-0.006$ &         $-0.026$ ($\pm$ 0.013) \\
Standard deviation            &  ${\TsVIII}2.336$ & ${\TsVIII}2.361$ ($\pm$ 0.014) & ${\TsVIII}1.409$ & ${\TsVIII}1.379$ ($\pm$ 0.019) \\
$1^{\mathrm{st}}$ percentile  &          $-5.512$ &         $-5.539$ ($\pm$ 0.099) &         $-3.590$ &         $-3.638$ ($\pm$ 0.116) \\
$99^{\mathrm{th}}$ percentile &  ${\TsVIII}4.071$ & ${\TsVIII}4.103$ ($\pm$ 0.030) & ${\TsVIII}3.862$ & ${\TsVIII}4.073 $ ($\pm$ 0.181) \\
\hline
\multirow{2}{*}{Statistic}
& \multicolumn{2}{c|}{Dimension 3} & \multicolumn{2}{c}{Dimension 4}  \\
& Training & Simulated & Training & Simulated \\
Mean                          &         $-0.002$ & ${\TsVIII}0.199$ ($\pm$ 0.021) &         $-0.063$ & ${\TsVIII}0.062$ ($\pm$ 0.020) \\
Standard deviation            & ${\TsVIII}1.988$ & ${\TsVIII}2.002$ ($\pm$ 0.013) & ${\TsVIII}1.982$ & ${\TsVIII}1.958$ ($\pm$ 0.016) \\
$1^{\mathrm{st}}$ percentile  &         $-4.691$ &         $-4.717$ ($\pm$ 0.080) &         $-4.895$ &         $-4.821$ ($\pm$ 0.106) \\
$99^{\mathrm{th}}$ percentile & ${\TsVIII}4.677$ & ${\TsVIII}4.954$ ($\pm$ 0.066) & ${\TsVIII}4.431$ & ${\TsVIII}4.865$ ($\pm$ 0.082) \\
\hline
\end{tabular}
\end{table}

\begin{figure}[tbh]
\vspace*{10pt}
\caption{Comparison between the empirical correlation matrix and the average correlation matrix of WGAN simulated data}
\label{fig:multiWGANcorr}
\begin{subfigure}[b]{0.5\linewidth}
\centering
\includegraphics[width=0.8\linewidth]{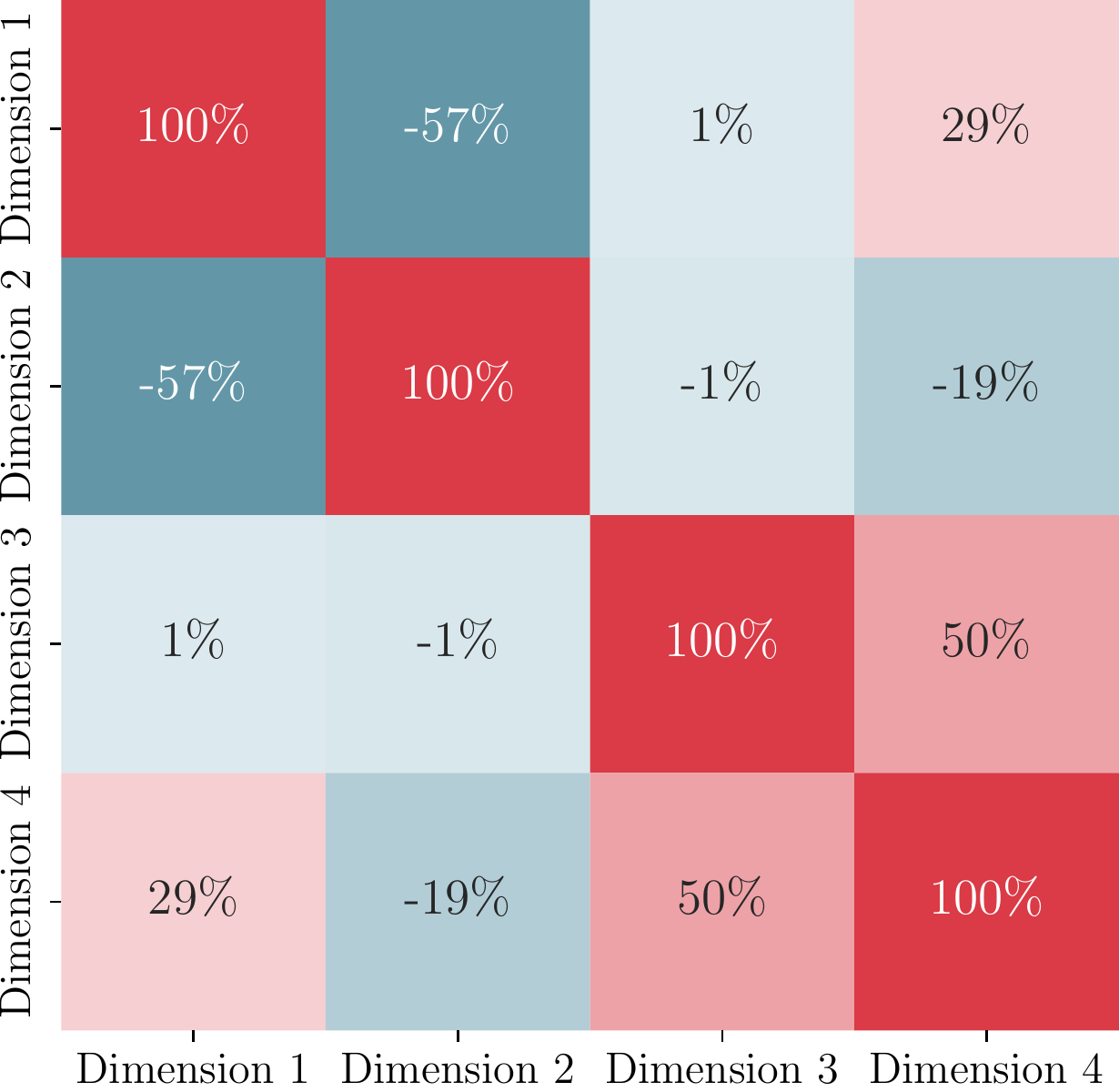}
\caption*{Training sample}
\end{subfigure}
\begin{subfigure}[b]{0.5\linewidth}
\centering
\includegraphics[width=0.8\linewidth]{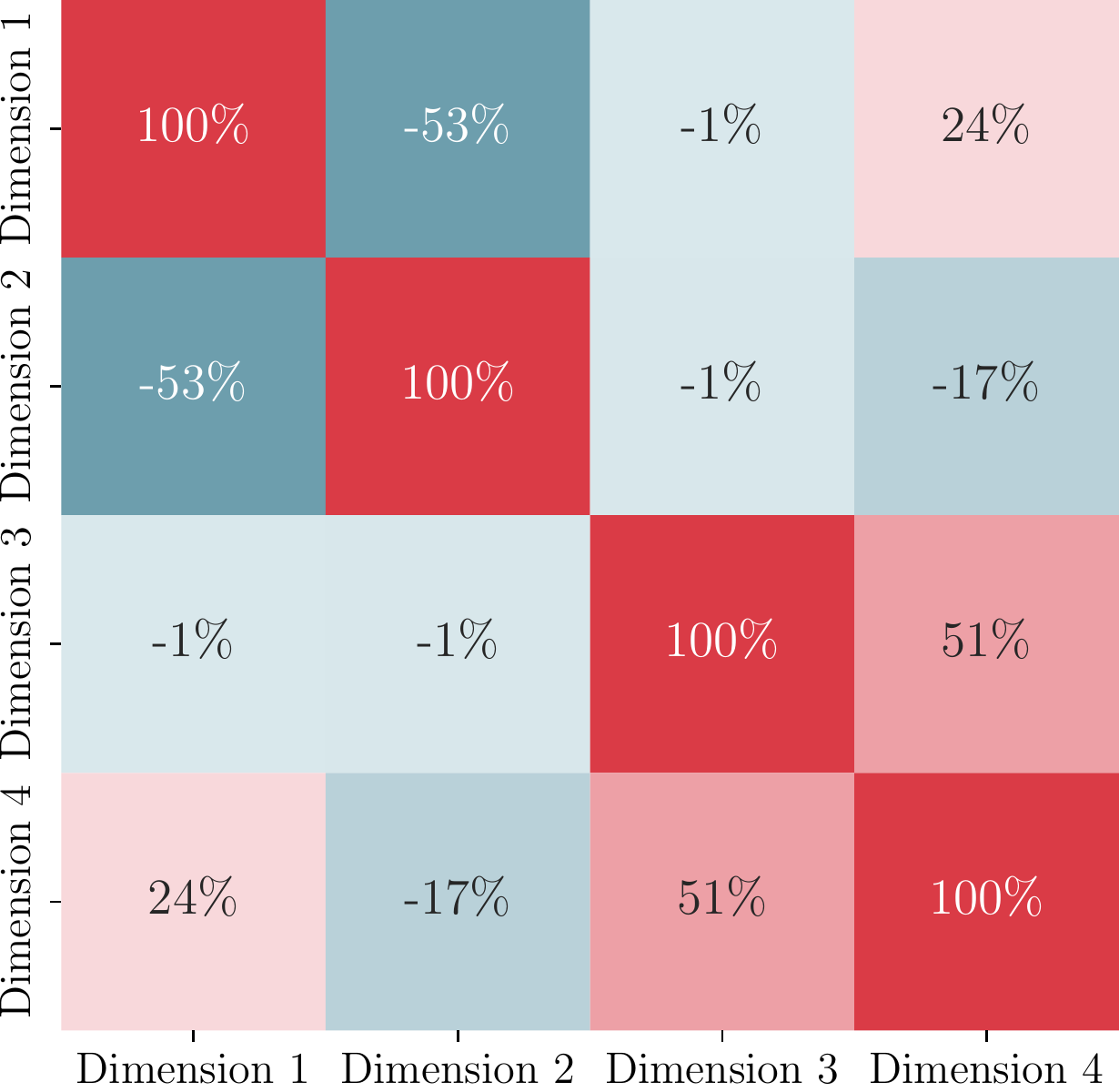}
\caption*{Generated samples}
\end{subfigure}
\end{figure}

In Figure \ref{fig:multiWGANcorr}, we also compare the empirical
correlation matrix of the training dataset and the average of the
correlation matrices computed with 50 Monte Carlo replications. We
notice that the Wasserstein GAN model captures very well the
correlation structure. For each value in the correlation matrix, the
difference is less than 5\%. Compared to the Bernoulli RBM and the
Gaussian RBM,  the Wasserstein GAN model captures much better the
correlation structure of the training dataset. For instance, the
correlation between first and second dimensions is equal to $-57\%$
for the training data, $-53\%$ for the Wasserstein GAN simulated
data, $-48\%$ for the Gaussian RBM simulated data and $-31\%$ for
the Bernoulli RBM simulated data. If we consider the first and
fourth dimensions, these figures become  $29\%$ for the training
data, $24\%$ for the Wasserstein GAN simulated data, $22\%$ for the
Gaussian RBM simulated data and $20\%$ for the Bernoulli RBM
simulated data.

\paragraph{Summary of the results}

According to our tests, the Wasserstein GAN model performs very well
in the task of  learning the joint distribution of our simulated
training dataset. Comparing with the results of Bernoulli and
Gaussian RBMs, the Wasserstein GAN model has several advantages:

\begin{enumerate}
\item The Wasserstein GAN model is less sensitive to
extreme values of the training dataset and doesn't have the tendency
to underestimate or overestimate the tail of probability
distribution.

\item We don't need to apply complex transformation to input data
for Wasserstein GAN model in data preprocessing. In our study, we
just use a MinMax scaling function and we recall that we need to use
the binary transformation for Bernoulli RBM and the normal score
transformation for Gaussian RBM.

\item The Wasserstein GAN model may capture more accurately
the correlation structure of the training dataset than Bernoulli and
Gaussian RBMs.

\end{enumerate}

\subsubsection{Application to financial time series}

\label{section:SPXGAN} According to the results in the previous section, the
Wasserstein GAN model preforms very well in the case of multi-dimensional
simulated dataset without the time dependence relationship. In this paragraph,
we construct a more complex Wasserstein GAN model with convolutional layers in
the generator and the discriminator in order to extract more features in the
real financial dataset consisting of $5\, 354$ historical daily returns of the
S\&P 500 index and the VIX index from December 1998 to May 2018. To capture the
time dependence of the multi-dimensional autocorrelated time series, we also
need to associate the training samples with conditional labels as we have done
in the case of the conditional RBM. \citet{Mariani-2019} proposed a method to
train the model with historical returns over $n_t$ consecutive days conditioned
by the last $n_h$ day historical returns. As a result, the generator of this
model should generate a matrix belonging to $\mathcal{M}_{n_{t},n_{x}}\left(
\mathbb{R}\right)$, which means the generator simulate directly daily returns of
S\&P 500 index and the VIX index for $n_t$ days. For reasons of simplicity,
this model is named in this paper as the conditional deep convolutional
Wasserstein GAN model (or CDCWGAN). In the step of data preprocessing, we need
to apply the MinMax scaling function on input data before the training process
to ensure that historical returns are scaled into the range $ \left [0, 1\right
] $.\smallskip

\paragraph{Structure and training of CDCWGAN models}

In this study on financial time series, the training samples for the
discriminator correspond to a $n_x$-dimensional time series
representing historical returns over $n_t$ days. Therefore, inputs
are represented by a matrix belonging to
$\mathcal{M}_{n_{t},n_{x}}\left( \mathbb{R}\right) $. In addition,
each training sample is conditioned by the past values of returns
over $n_h$ days, which are represented by a matrix belonging to
$\mathcal{M}_{n_{h},n_{x}}\left( \mathbb{R}\right)$. In practice, we
concatenate these two matrices to form a matrix belonging to the
$\mathcal{M}_{n_{h} + n_{t},n_{x}}\left( \mathbb{R}\right)$ and this
matrix will be fed to the discriminator as inputs. In this study,
the inputs of the discriminator will be down sampled using 4
convolutional layers with number of filters $\left\{16, 32, 64,
128\right\}$. We set the kernel length to 3 and the stride to 2 and
we choose to use the leaky RELU function with $\alpha = 0.5$ for
each convolutional layer. For the last layer of the discriminator,
we choose always a 1-neuron dense layer and use the tangent
hyperbolic activation function for this neuron as in the case of the
traditional Wasserstein GAN model.\smallskip

For the generator, we modify the original method of
\citet{Mariani-2019} and we borrow the idea of conditional layer in
the case of conditional RBM. The generator of our CDCWGAN model will
take two inputs: a 100-dimensional random noise vector and $n_h$
past values that are concatenated into a $n_h \times n_x
$-dimensional vector. We then feed these two vectors to the first
layer of the generator, which is a dense layer and we will reshape
the output to a two-dimensional $n_t \times 256$ matrix before
passing them to the second layer. We construct the rest of the
generator using 3 convolutional layers with number of filters
$\left\{256, 64, 2 \right\}$. The kernel length is set to 3 and the
stride is set to 2. We also use a leaky RELU function with $\alpha =
0.5$ for each convolutional layer and since the last convolutional
layer will give directly the simulated scenarios, we choose the
sigmoid activation function in order to get the value in range
$\left[0, 1\right]$.\smallskip

For this conditional deep convolutional Wasserstein GAN model,  we
use always the RMSProp optimizer with a small learning rate. Indeed,
the learning rate is set to $\eta = 1.10^{-4}$ and the batch size is
set to $500$. Table \ref{table:DCWGANconfig} summarizes the setting
of the model implementation using Python and the TensorFlow library.

\begin{table}[tbph]
\centering
\caption{The setting of the CDCWGAN model}
\label{table:DCWGANconfig}
\begin{tabular}{p{6cm}p{6cm}}
\hline
Training data dimension $n_x$        & 2                                                \\
Training data window $n_t$           & 5                                                \\
Training data label length $n_h$     & 20                                               \\
Input feature scaling function       & MinMax                                           \\
Random noise vector dimension        & 100                                              \\
Random noise distribution            & $\mathcal{N}\left(0, 1\right)$                   \\ \hdashline
\multirow{2}{*}{Generator structure}
                                     & $\left\{ \textrm{Dense}(5, 256), \textrm{Conv1D}(256),\right.$                  \\
                                     & $\left. \textrm{Conv1D}(64), \textrm{Conv1D}(2) \right\}$                       \\
\multirow{2}{*}{Discriminator structure}
                                     & $\left\{ \textrm{Conv1D}(16), \textrm{Conv1D}(32),\right.$                      \\
                                     & $\left. \textrm{Conv1D}(64), \textrm{Conv1D}(128), \textrm{Dense}(1) \right \}$ \\ \hdashline
Loss function                        & Wasserstein distance                             \\
Learning optimizer                   & RMSProp                                          \\
Learning rate                        & $1.10^{-4}$                                      \\
Batch size                           & 500                                              \\
\hline
\end{tabular}
\end{table}

\paragraph{Learning joint distribution and time dependence}

\begin{figure}[ht]
\caption{Autocorrelation function of synthetic samples generated by the CDCWGAN model}
\label{fig:SpxVixDCGANAuto}
\begin{subfigure}[b]{0.5\linewidth}
\centering
\includegraphics[width=\linewidth]{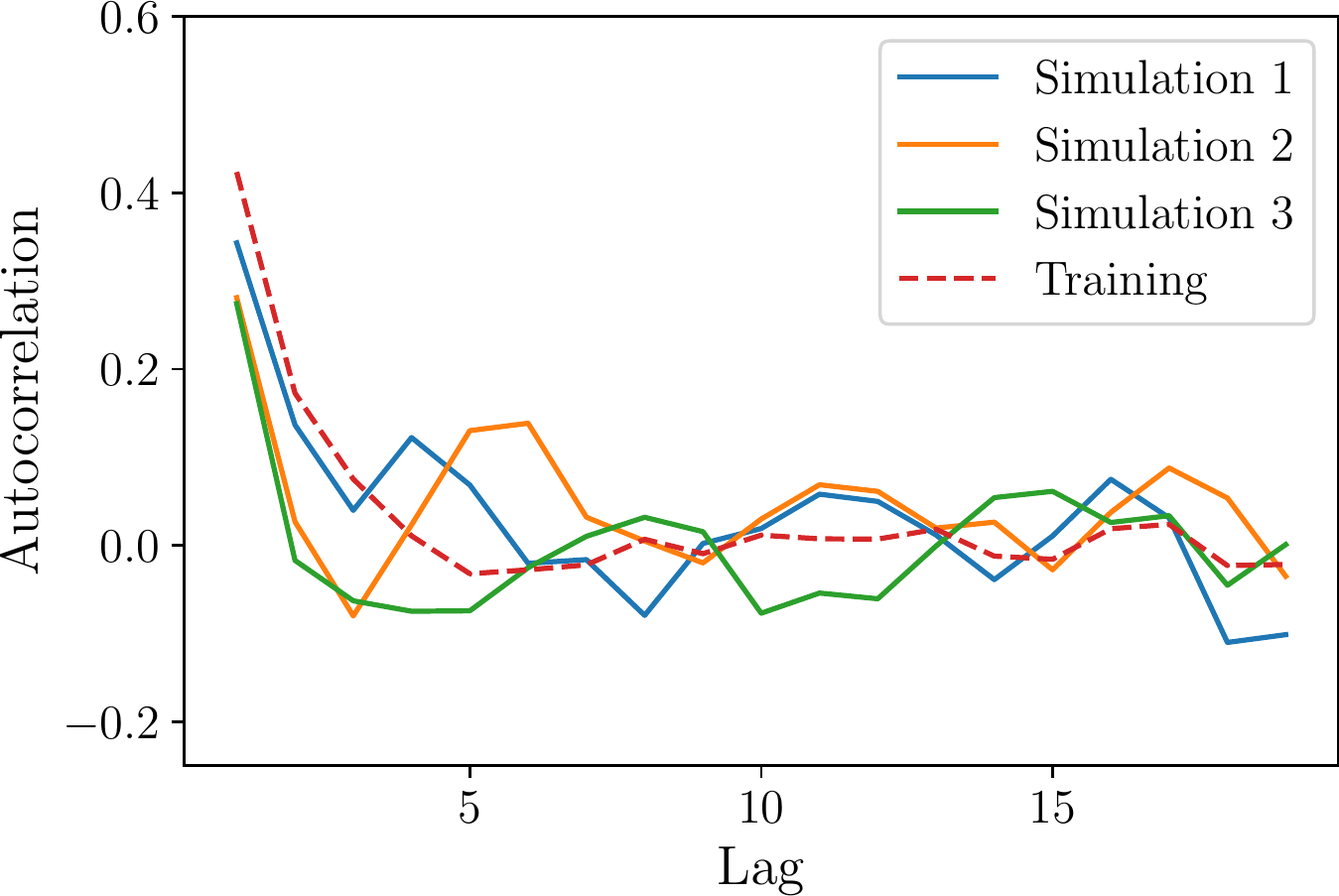}
\caption*{S\&P 500 index}
\end{subfigure}
\begin{subfigure}[b]{0.5\linewidth}
\centering
\includegraphics[width=\linewidth]{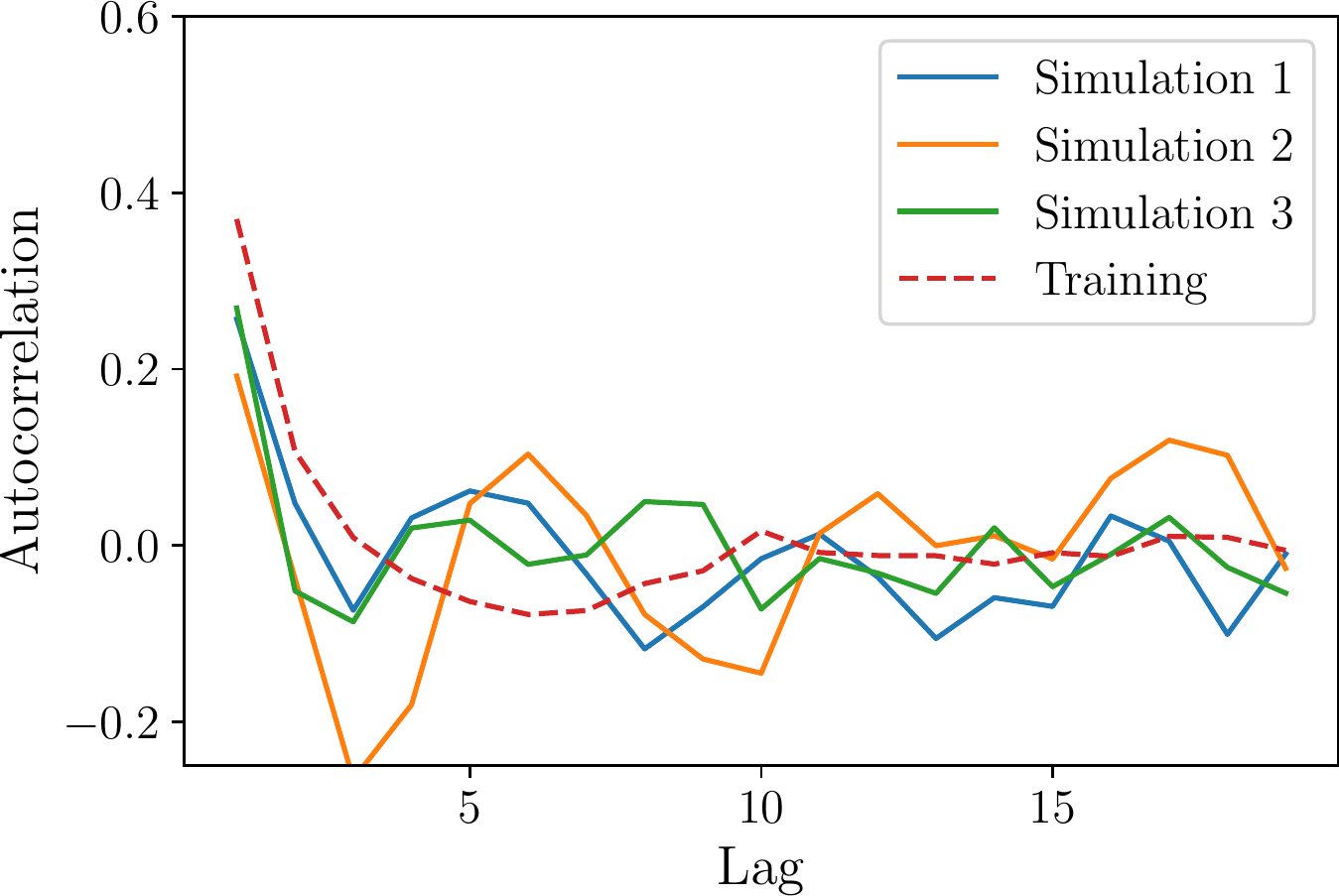}
\caption*{VIX index}
\end{subfigure}
\end{figure}

In this study, we choose to set the length of training data window
$n_t$ to 5, which means that the generator of the model will generate
at each time step the scenarios for the $5$ next days. In addition,
we also use the values of the last $20$ days as a long memory for
input data of the model in order to compare with the results of the
conditional RBM. After the training process, we generate samples of
consecutive time steps to verify the quality of learning the joint
distribution of daily returns of the S\&P index and the VIX index.
We run 50 Monte Carlo simulations and for each simulation, we
generate $5\, 354$ simulated observations from different random
noise series. According to Table \ref{table:SpxVixDCWGAN}, we find
that the CDCWGAN model learns also pretty well the joint
distribution of the S\&P index and the VIX index. In addition, the
average correlation coefficient between daily returns of the two
indices over 50 Monte Carlo replications is equal to $-71\%$, which
is exactly the figure of the empirical correlation. In Figure
\ref{fig:SpxVixDCGANAuto}, we observe clearly that a CDCWGAN model
can capture well the autocorrelation of the training dataset as in
the case of the conditional RBM shown in Figure
\ref{fig:SpxVixCRBMAuto} on page \pageref{fig:SpxVixCRBMAuto}.

\begin{table}[tbh]
\centering
\caption{Comparison between the training sample and simulated samples using the conditional deep convolutional Wasserstein GAN model}
\label{table:SpxVixDCWGAN}
\begin{tabular}{l|cc|cc}
\hline
\multirow{2}{*}{Statistic}
& \multicolumn{2}{c|}{S\&P 500 index} & \multicolumn{2}{c}{VIX index}  \\
& Training & Simulated & Training & Simulated \\
Mean                          &  ${\TsVIII}0.02\%$ &         ${\TsVIII}0.08\%$ ($\pm$ $0.01\%$) & ${\TsVIII}0.06\%$ & $-0.33\%$ ($\pm$ $0.02\%$) \\
Standard deviation            &  ${\TsVIII}0.64\%$ & ${\TsVIII}0.65\%$ ($\pm$ $0.01\%$) & ${\TsVIII}3.84\%$ & ${\TsVIII}3.58\%$ ($\pm$ $0.02\%$) \\
$1^{\mathrm{st}}$ percentile  &          $-1.93\%$ &         $-1.57\%$ ($\pm$ $0.02\%$) &         $-7.31\%$ &         $-7.39\%$ ($\pm$ $0.06\%$) \\
$99^{\mathrm{th}}$ percentile &  ${\TsVIII}1.61\%$ & ${\TsVIII}1.56\%$ ($\pm$ $0.02\%$) & ${\TsIII}11.51\%$ & ${\TsVIII}9.74\%$ ($\pm$ $0.13\%$) \\
\hline
\end{tabular}
\end{table}

\paragraph{Market generator}

Based on above results, we consider that the CDCWGAN model can be also used as
a market generator. In the example of S\&P 500 and VIX indices, we have trained
the model with all historical observations using the method proposed by
\citet{Mariani-2019} and the generator of this method will generate scenarios
over several consecutive days. After having calibrated the training process, we
may choose a date as the starting point and generate new samples for several
days.  We then update the memory vector and generate samples for the following
days. Iteratively, the CDCWGAN model can generate a multi-dimensional time
series that has the same patterns as those of the training data in terms of
marginal distribution, correlation structure and time dependence. Comparing
with the conditional RBM that we have studied, we don't need the normal score
transformation to ensures the learning of marginal distributions and,
convolutional layers in the generator and the discriminator may help us to
extract more features in the training dataset.\smallskip

\begin{figure}[tbph]
\caption{Comparison between scaling historical prices of S\&P 500 and VIX indices and synthetic time series generated by the CDCWGAN model}
\label{fig:SpxVixDCWGANFake}
\begin{subfigure}[b]{0.5\linewidth}
\centering
\includegraphics[width=\linewidth]{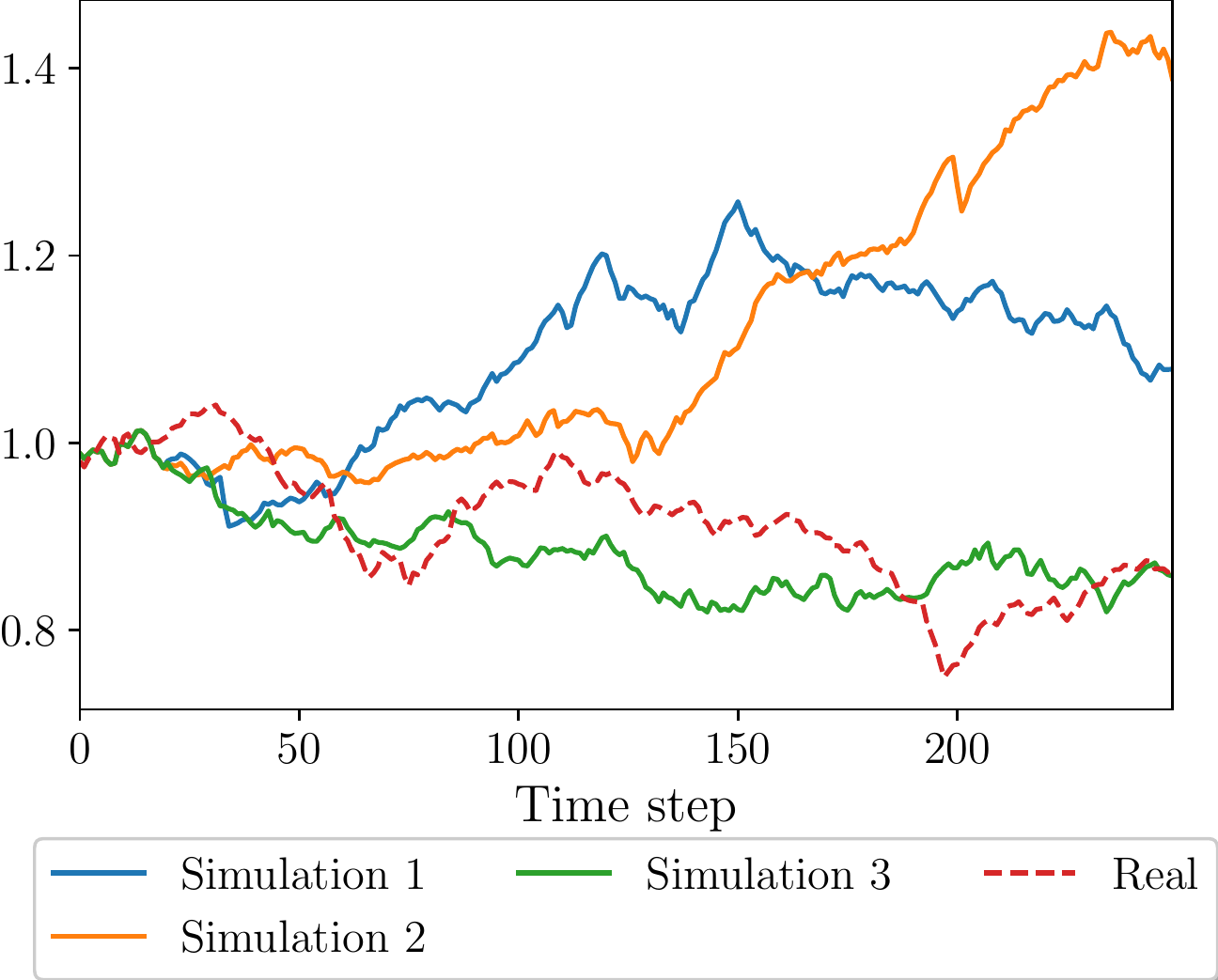}
\caption*{S\&P 500 index}
\end{subfigure}
\begin{subfigure}[b]{0.5\linewidth}
\centering
\includegraphics[width=\linewidth]{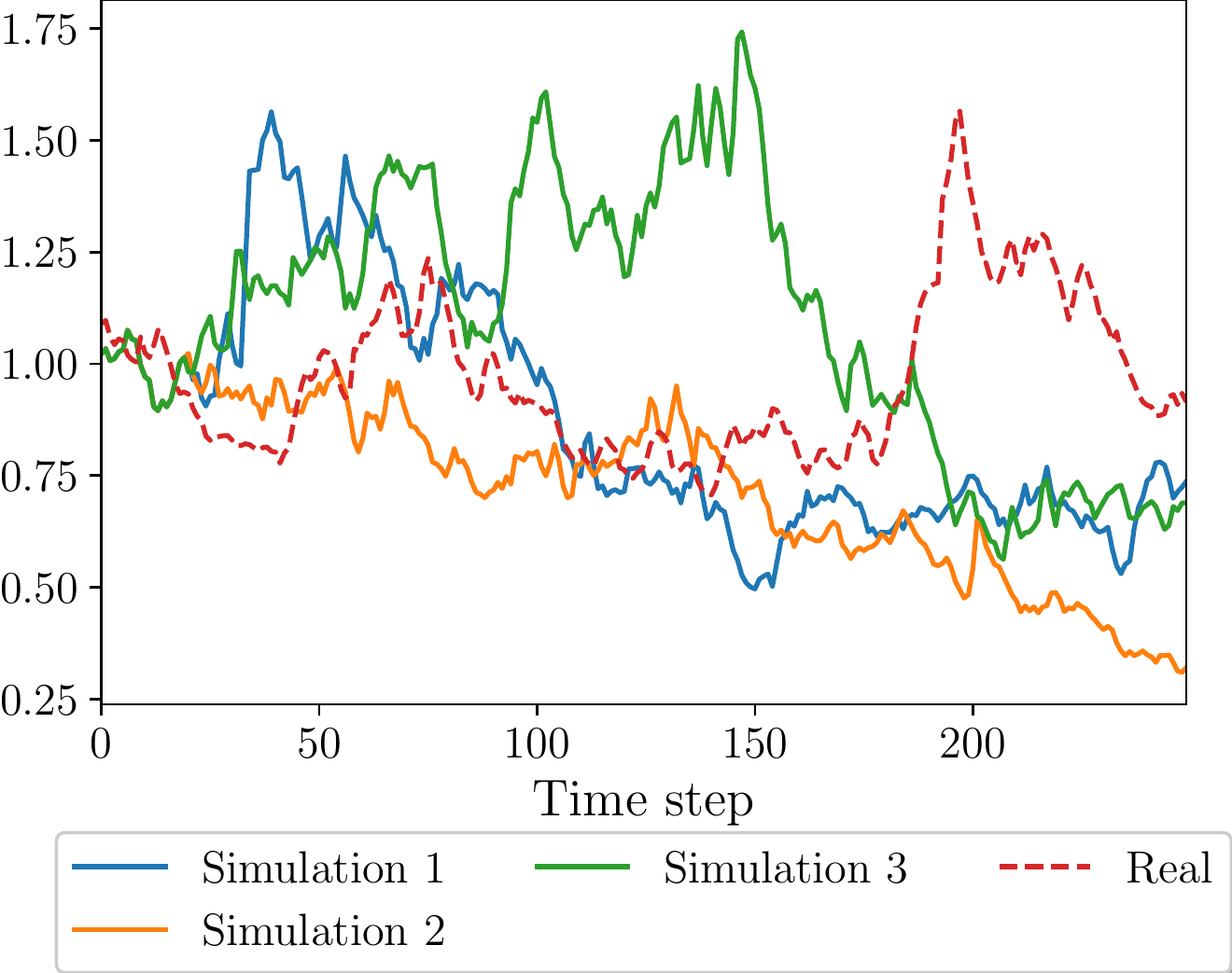}
\caption*{VIX index}
\end{subfigure}
\end{figure}

As we have shown in Figure \ref{fig:SpxVixCRBMFake} on page
\pageref{fig:SpxVixCRBMFake} for the conditional RBM, Figure
\ref{fig:SpxVixDCWGANFake} shows three time series of $250$ trading
days starting at the date 22/11/2000, which are generated by the
trained CDCWGAN model. We notice that these three simulated time
series have clearly the positive autocorrelation in each dimension
and they capture very well the negative correlation between S\&P 500
and VIX indices. Comparing with the historical prices that are
represented by the dashed line, these three simulated time series
have the same characteristics.

\subsection{Managing the out-of-sample robustness of backtesting methods}

We now consider an application of generative models in the context of
quantitative asset management. Traditionally, we use the historical daily
returns of assets to backtest an investment strategy,  meaning that only one
real time series is available. Therefore, we get only one estimated value for
the performance and risk statistics of the strategy, such as the annualized
return, the volatility, the Sharpe ratio, the maximum drawdown, etc. Generative
models can be used here to manage the out-of-sample robustness of backtesting
methods. For instance, we may use a generative model like RBMs and GANs to
learn the joint distribution and the time dependence of historical daily
returns of assets. We may then use these models to generate new time series.
Finally, we may backtest our investment strategy using these new simulated time
series in order to construct the probability distribution of the different
statistics. In this approach, the estimated value of the statistic obtained
with the true real time series becomes one realization of its probability
distribution. Suppose for example that the maximum drawdown of the backtest is
equal to $10\%$, and that the live investment strategy has a maximum drawdown
of $20\%$. Does it mean that the process of the investment strategy has been
overfitted? Certainly \textsl{yes} if there is a zero probability to experience
a maximum drawdown of $20\%$ when the strategy is backtested with generative
models. Definitively \textit{not} if some samples of generative models have
produced a maximum drawdown larger than $20\%$.

\subsubsection{Backtest of the risk parity strategy}

\begin{figure}[tbph]
\caption{Cumulative performance of the risk parity strategy}
\label{fig:RPCumPerf} \centering
\includegraphics[width=0.5\linewidth]{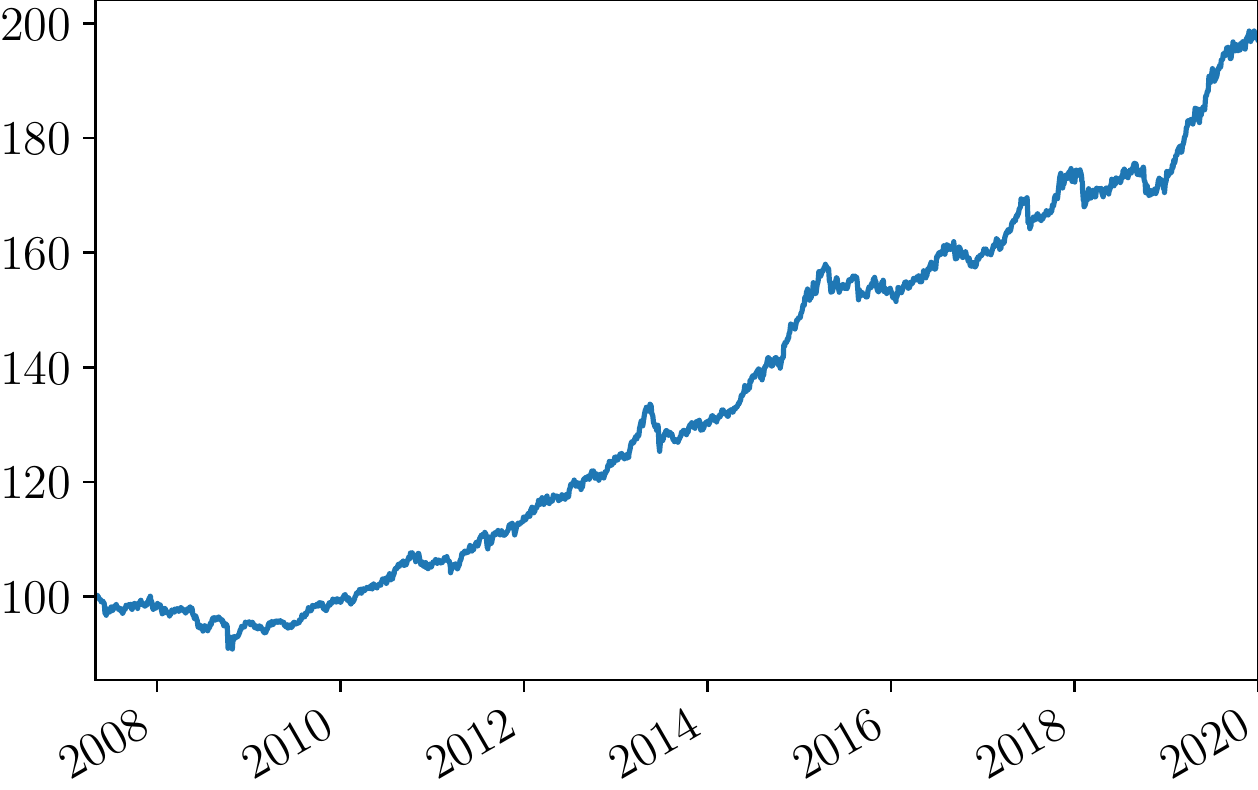}
\end{figure}

Our dataset consists of daily returns of six futures contracts on world-wide
equity indices such as S\&P 500, Eurostoxx 50 and Nikkei 225 indices and 10Y
sovereign bonds such as US, Germany and Australia from January 2007 to December
2019. In this  study, we build a risk parity strategy on this multi-asset
investment universe. Moreover, the strategy is unfunded and we calibrate its
leverage in order to obtain a volatility around $3\%$. In Figure
\ref{fig:RPCumPerf}, we have reported the cumulative performance of the risk
parity strategy. Moreover, the descriptive statistics of performance and risk
are given in Table \ref{table:RPBacktest}, and correspond to the annualized
performance $\mu\left(x\right)$, the volatility $\sigma\left(x\right)$, the
Sharpe ratio $\textrm{SR}\left(x\right)$, the maximum drawdown\footnote{The
maximum drawdown corresponds to a loss. For example, if the maximum drawdown is
equal to $10\%$, the investor may face a maximum loss of $10\%$. This is why
the maximum drawdown is expressed by a positive value.}
$\textrm{MDD}\left(x\right)$ and the skew measure $\xi\left(x\right)$, which is
the ratio between the maximum drawdown and the volatility\footnote{If
$\xi\left(x\right)$ is greater than $3$, this indicates that the strategy has a
high skewness risk.}.

\begin{table}[tbph]
\centering
\caption{Descriptive statistics of the risk parity backtest}
\label{table:RPBacktest}
\begin{tabular}{cccccc}
\hline
Period & $\mu\left(x\right)$ & $\sigma\left(x\right)$ & $\textrm{SR}\left(x\right)$ &
$\textrm{MDD}\left(x\right)$ & $\xi\left(x\right)$ \\\hline
2007 -- 2019 & $5.30\%$ & $3.54\%$ & $1.50$ & $9.40\%$ & $2.65$ \\
2018 -- 2019 & $6.97\%$ & $3.48\%$ & $2.00$ & $3.69\%$ & $1.05$ \\
\hline
\end{tabular}
\end{table}

\subsubsection{Market generator of the risk parity strategy}

We first train the conditional RBM introduced in Section \ref{section:SPXRBM}
on page \pageref{section:SPXRBM} on the daily returns of the six futures
contracts. In this study, the training of the models is done with all
historical observations of futures contracts returns from January $2007$ to
December $2019$. Moreover, for each date, we have used the values of the last
$20$ days as a memory vector for the conditional layer. We also choose a large
number for the number of hidden layers units in order to capture the complex
correlation structure of financial time series. Therefore, the conditional RBM
has $6$ visible layer units, $256$ hidden layer units and $120$ conditional
layer units. In addition, we use the normal score transformation to avoid the
problem of outliers in the training dataset and improve the learning quality.
After having calibrated the training process, we choose the $1^{\mathrm{st}}$
January 2018 as the starting point and generate new futures contracts returns
iteratively for the next two years. We run $500$ Monte Carlo simulations
with different random noise series and for each simulation, we backtest our
risk parity strategy using synthetic time series and calculate the descriptive
statistics of the strategy. Finally, we construct probability distributions for
these descriptive statistics and we compare them with the results obtained by
the traditional method of bootstrap sampling, which consists of using a random
sampling of the historical returns with replacement.\smallskip

\begin{figure}[tbph]
\caption{Comparison between cumulative performance of the risk
parity strategy using synthetic time series generated by the
bootstrap sampling and conditional RBM models}
\label{fig:RPSamples}
\begin{subfigure}[b]{0.5\linewidth}
\centering
\includegraphics[width=\linewidth]{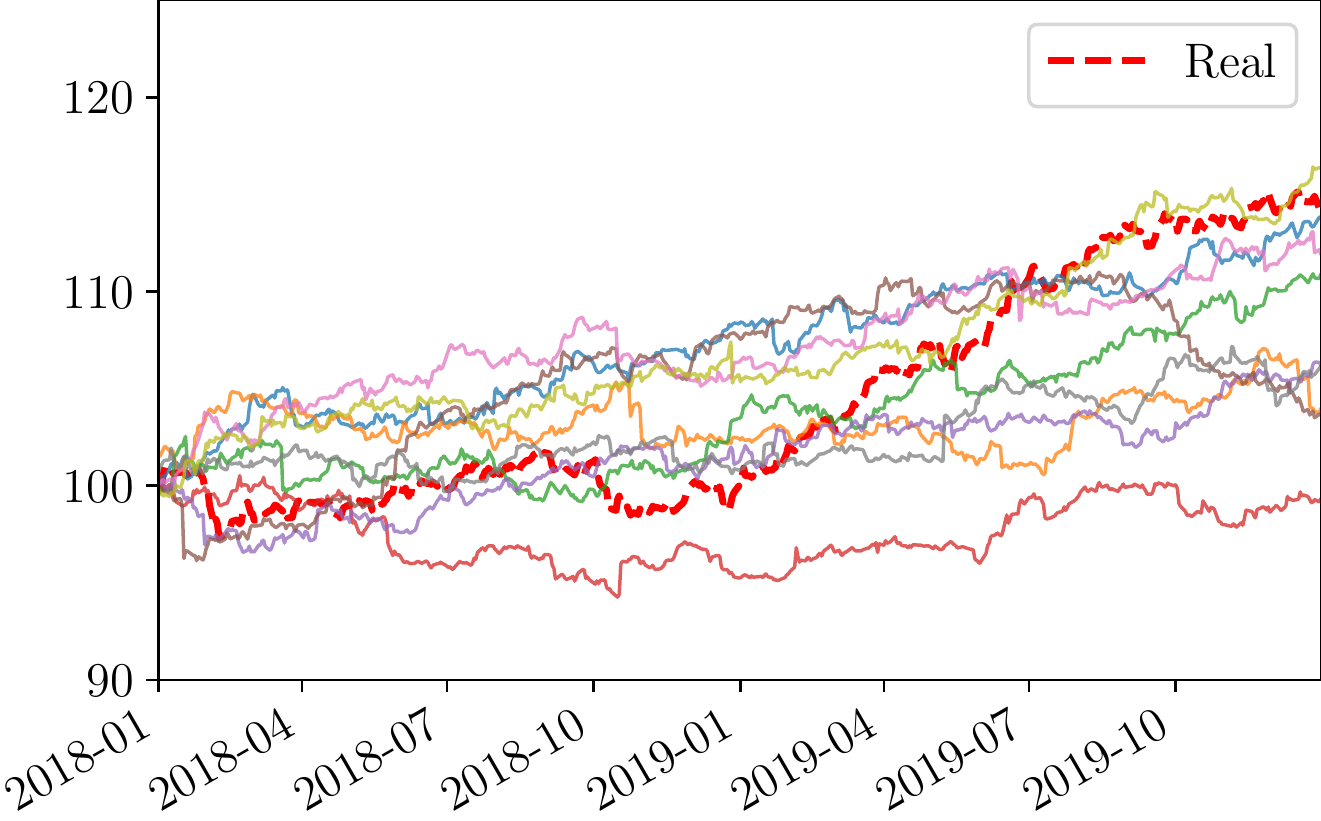}
\caption*{Bootstrap sampling}
\end{subfigure}
\begin{subfigure}[b]{0.5\linewidth}
\centering
\includegraphics[width=\linewidth]{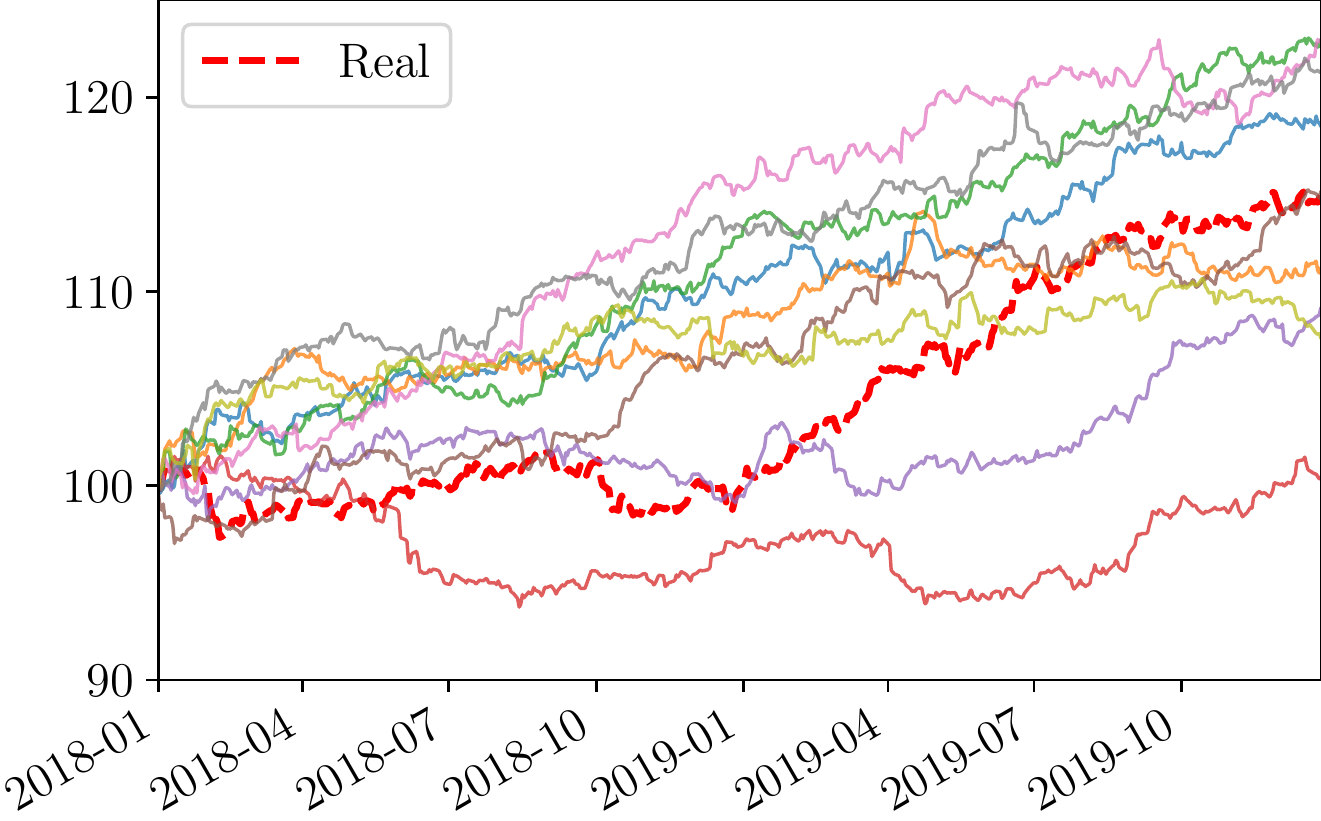}
\caption*{Conditional RBM}
\end{subfigure}
\end{figure}

\begin{figure}[tbph]
\caption{Autocorrelation function of the risk parity strategy using synthetic time series
generated by the bootstrap sampling and conditional RBM methods}
\label{fig:RPAcf}
\begin{subfigure}[b]{0.5\linewidth}
\centering
\includegraphics[width=\linewidth]{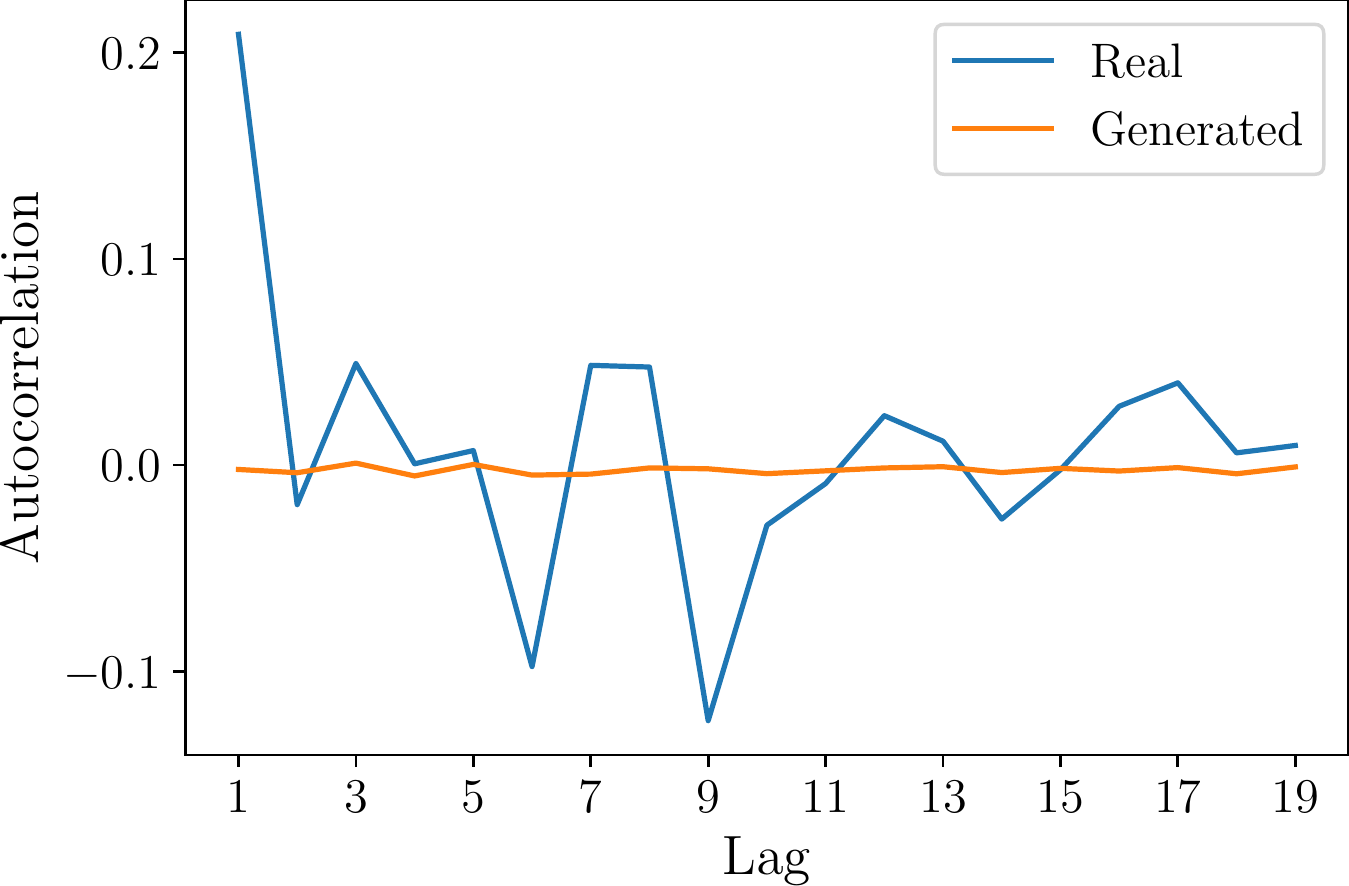}
\caption*{Bootstrap sampling}
\end{subfigure}
\begin{subfigure}[b]{0.5\linewidth}
\centering
\includegraphics[width=\linewidth]{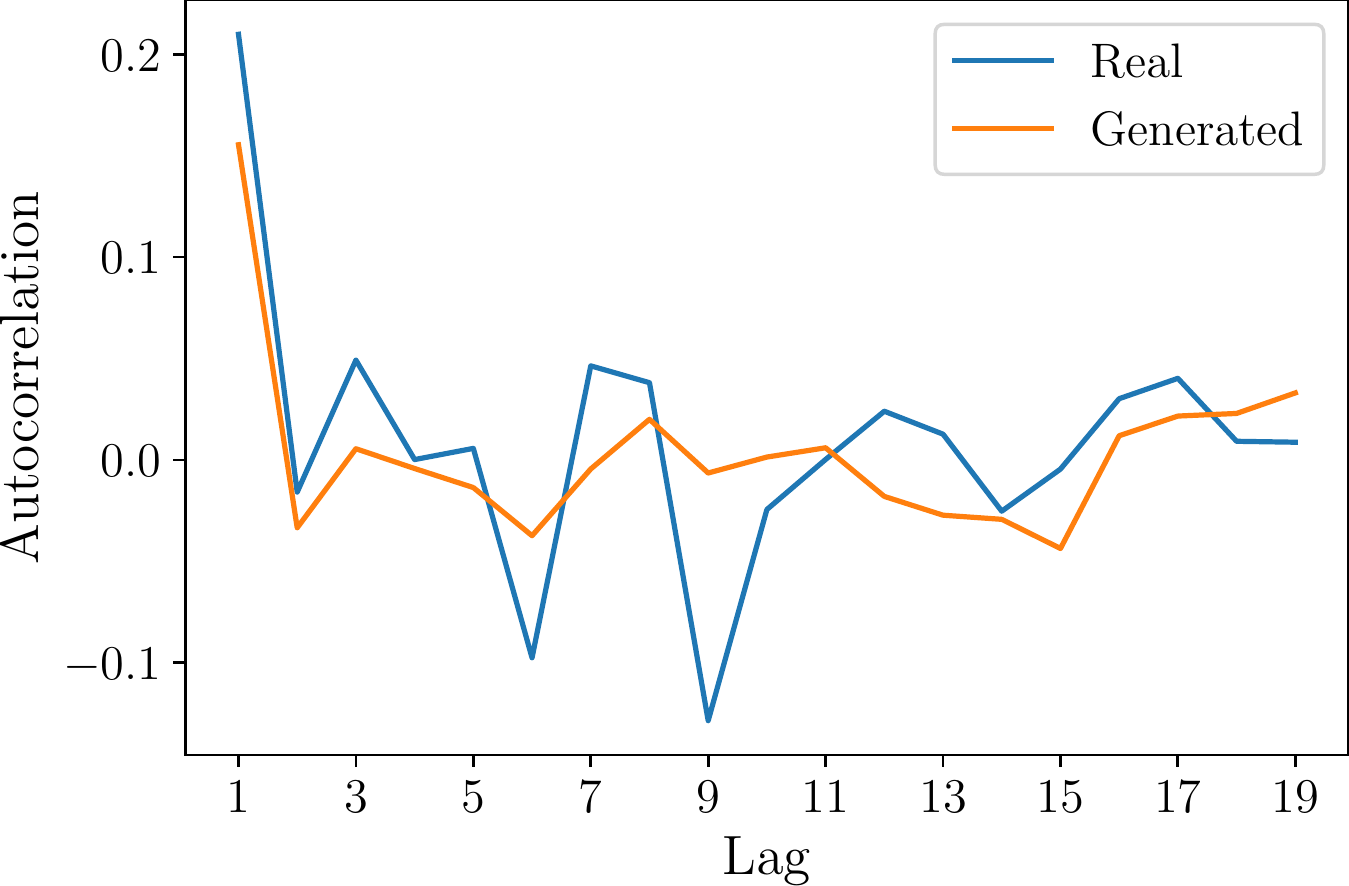}
\caption*{Conditional RBM}
\end{subfigure}
\end{figure}

Figure \ref{fig:RPSamples} shows the cumulative performance of ten risk parity
backtests using synthetic time series generated by the two methods. Among these
simulations, we notice that the cumulative performance of the risk parity
strategy using synthetic time series generated by the bootstrap sampling method
is more centered around the real time series of backtesting. This phenomenon
corresponds to the drawback of traditional bootstrap sampling method, since it
cannot capture the time dependence in risk parity strategy. As shown in Figure
\ref{fig:RPAcf}, our risk parity strategy has a first-lag positive
autocorrelation of $20\%$, and we compare the average autocorrelation function
over 500 Monte Carlo simulations generated by the two approaches. We notice
that the bootstrap sampling method cannot replicate this positive
autocorrelation as the conditional RBM can do. This property is very important
when we backtest meta-strategies, which means a strategy of an existing
strategy. For instance, if we want to design a stop-loss strategy for our risk
parity strategy, we should tune several parameters for implementing this
stop-loss strategy. If we use the real historical time series to calibrate the
stop-loss parameters, we might fall into the trap of overfitting because we
only have one real historical time series. Therefore, if we use the time series
generated by the bootstrap sampling method, we will not find the appropriate
values of the stop-loss parameters, since the autocorrelation plays a key role
in the stop-loss strategy as explained by \citet{Kaminski-2014}. Using the
conditional RBM as a market generator to generate time series is then a better
way to find the appropriate parameters for the meta-strategy and manage the
out-of-sample robustness.

\subsubsection{Building the probability distribution of backtest statistics}

\begin{figure}[tbph]
\caption{Histogram of the maximum drawdown of the risk parity
strategy using synthetic time series generated by the bootstrap
sampling and conditional RBM methods}
\label{fig:RPMDD}
\begin{subfigure}[b]{0.5\linewidth}
\centering
\includegraphics[width=\linewidth]{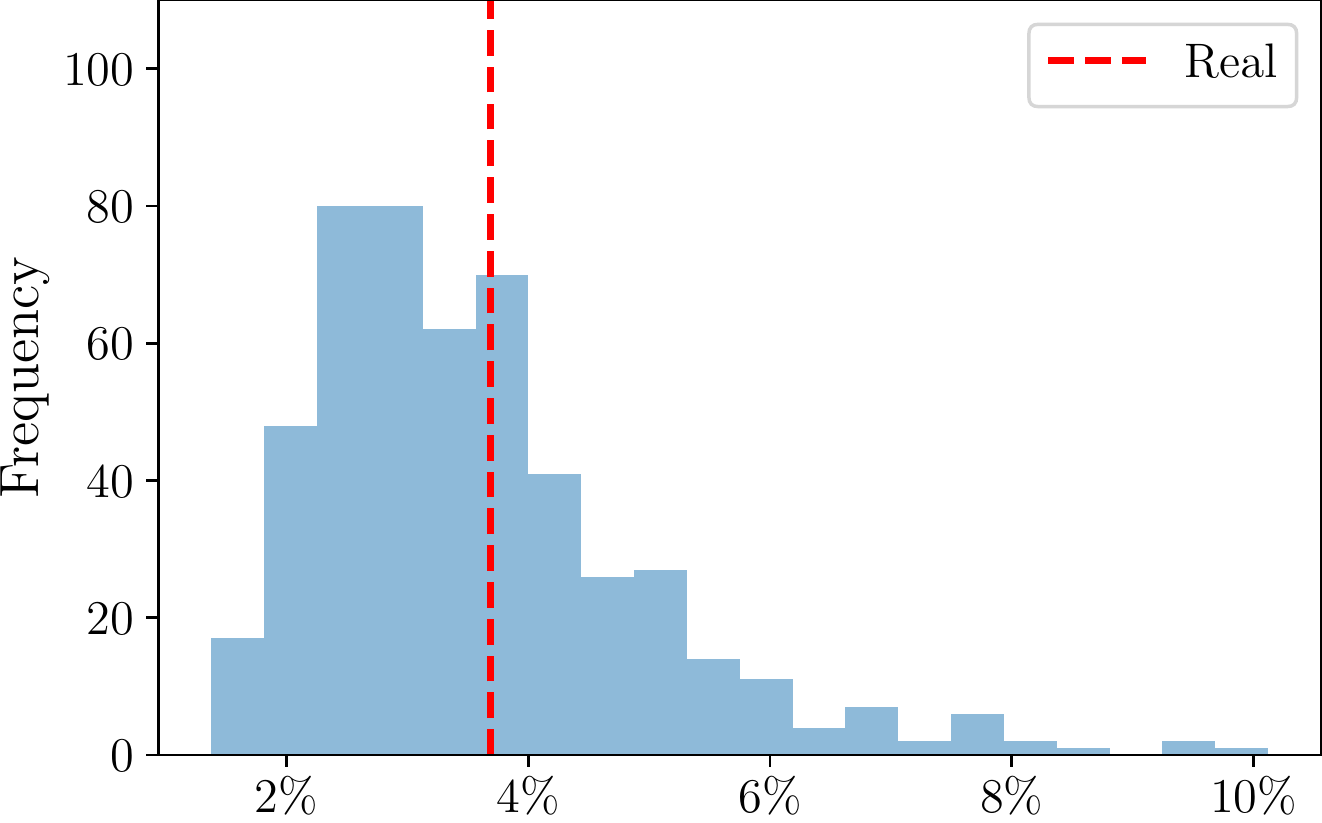}
\caption*{Bootstrap sampling}
\end{subfigure}
\begin{subfigure}[b]{0.5\linewidth}
\centering
\includegraphics[width=\linewidth]{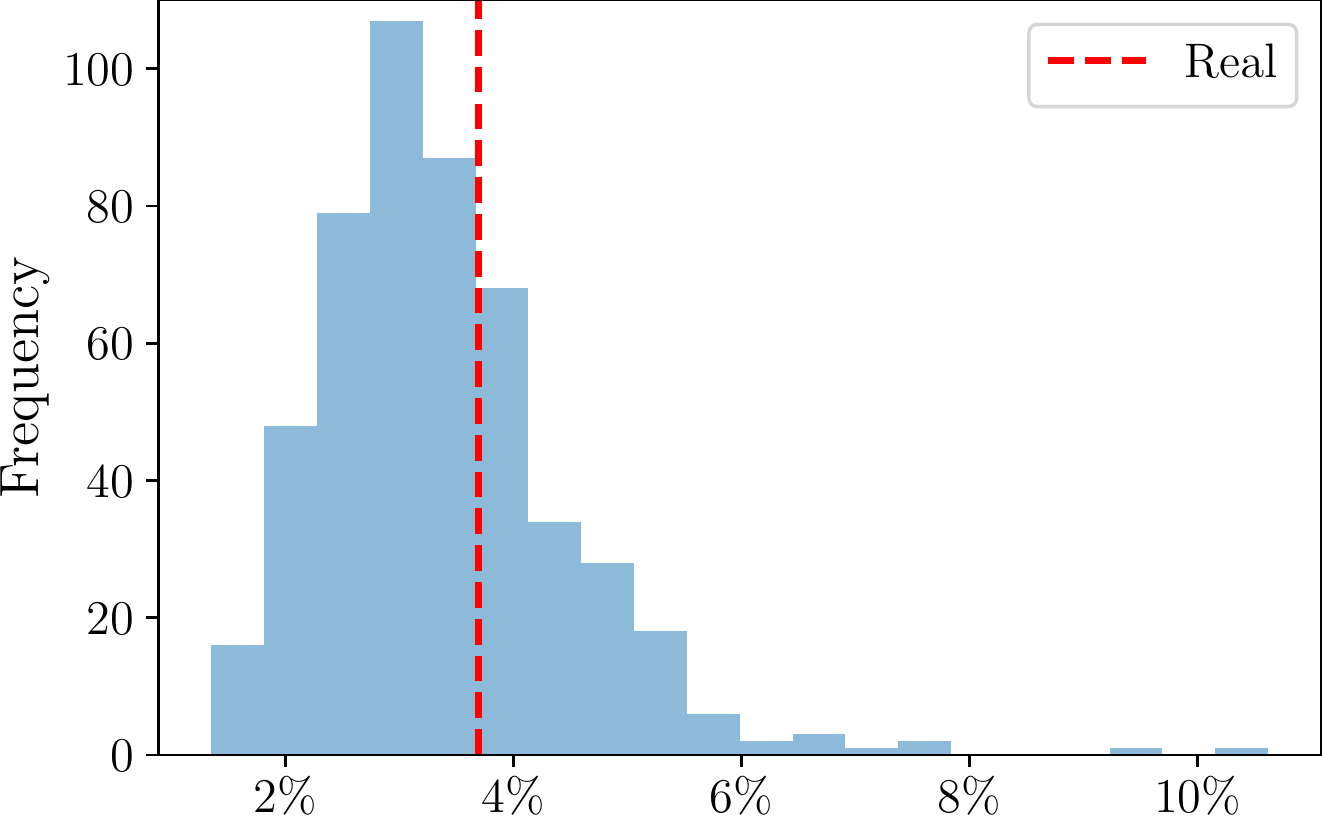}
\caption*{Conditional RBM}
\end{subfigure}
\end{figure}

In our study, we are interested in building the probability distribution of the
maximum drawdown $\textrm{MDD}\left(x\right)$ and the skew measure
$\xi\left(x\right)$ for the risk parity strategy. Figure \ref{fig:RPMDD} shows
the distributions of the maximum drawdown generated by the bootstrap sampling
and conditional RBM methods. We notice that the distribution of the maximum
drawdown generated by the conditional RBM are more centered and have less
severe scenarios. For instance, the value of the maximum drawdown of the real
risk parity strategy from January 2018 to December 2019 is $3.69\%$ and this
value corresponds respectively to the $61.5\%$ quantile in the probability
distribution generated by the bootstrap sampling method and the $69.1\%$
quantile in the probability distribution generated by the conditional RBM. We
consider that the difference comes from the quality of learning about the time
dependence of the dataset.\smallskip

\begin{figure}[tbph]
\caption{Histogram of the skew measure of the risk parity strategy using synthetic time series
generated by the bootstrap sampling and the conditional RBM methods}
\label{fig:RPSKEW}
\begin{subfigure}[b]{0.5\linewidth}
\centering
\includegraphics[width=\linewidth]{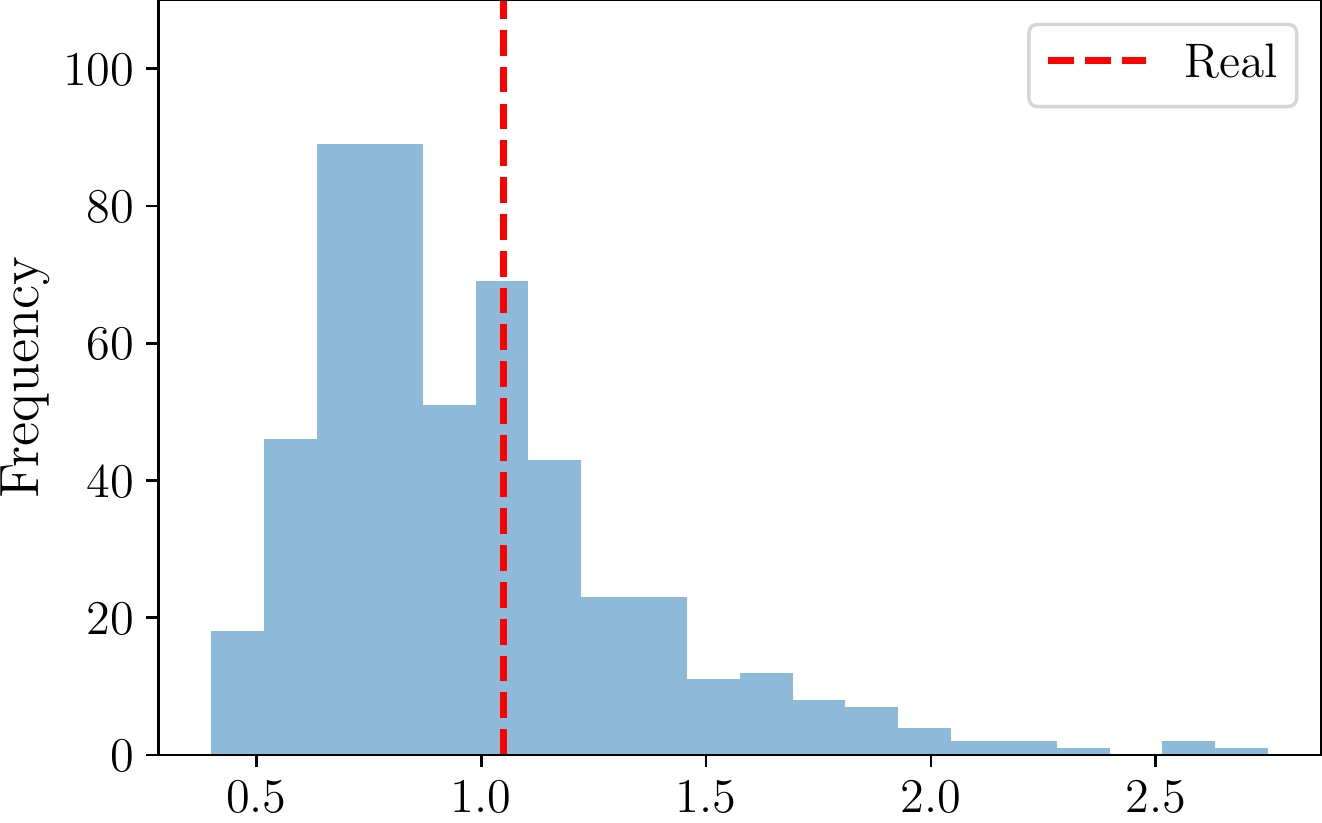}
\caption*{Bootstrap sampling}
\end{subfigure}
\begin{subfigure}[b]{0.5\linewidth}
\centering
\includegraphics[width=\linewidth]{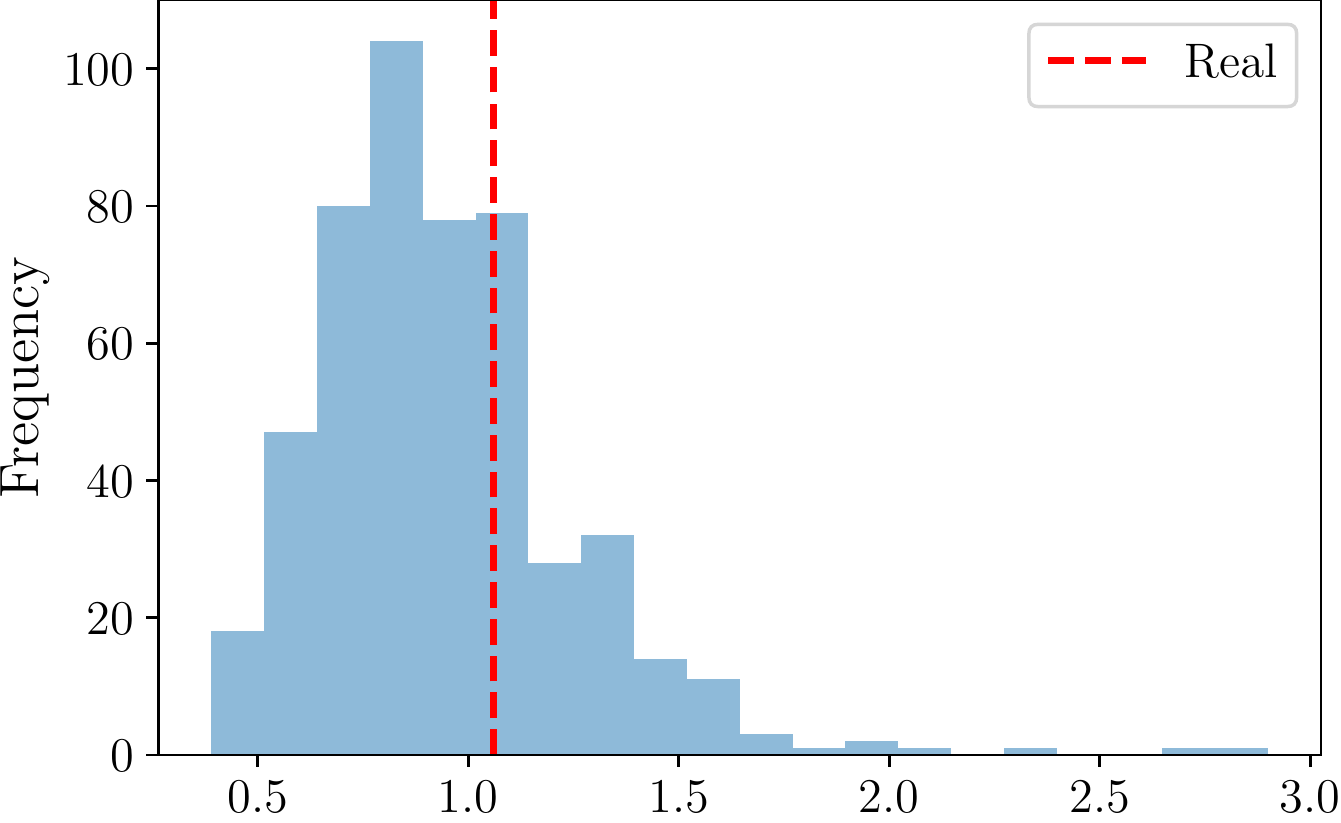}
\caption*{Conditional RBM}
\end{subfigure}
\end{figure}

Although we use a volatility targeting method to control the volatility of risk
parity strategies, we cannot ensure that the strategy volatility is exactly
equal to $3\%$. To avoid the influence of the difference between realized
volatility, we plot in Figure \ref{fig:RPSKEW} the probability distribution of
the skew measure. In this case, we notice that the tail of the probability
distribution generated by the conditional RBM is less fat than that generated
by the bootstrap sampling method but has several extreme severe scenarios. The
skew measure of the real risk parity strategy from January 2018 to December
2019 is equal to $1.05$, and this value corresponds respectively to the
$66.3\%$ quantile in the probability distribution generated by the bootstrap
sampling method and the $70.1\%$ quantile in the probability distribution
generated by the conditional RBM.

\subsubsection{Comparison with Wasserstein GAN models}

\begin{figure}[tbph]
\caption{Histogram of the skew measure of the risk parity strategy using synthetic
time series generated by the conditional RBM and Wasserstein GAN models}
\label{fig:RPRBMvsGAN}
\begin{subfigure}[b]{0.5\linewidth}
\centering
\includegraphics[width=\linewidth]{Part_3_RBM_SKEW}
\caption*{Conditional RBM}
\end{subfigure}
\begin{subfigure}[b]{0.5\linewidth}
\centering
\includegraphics[width=\linewidth]{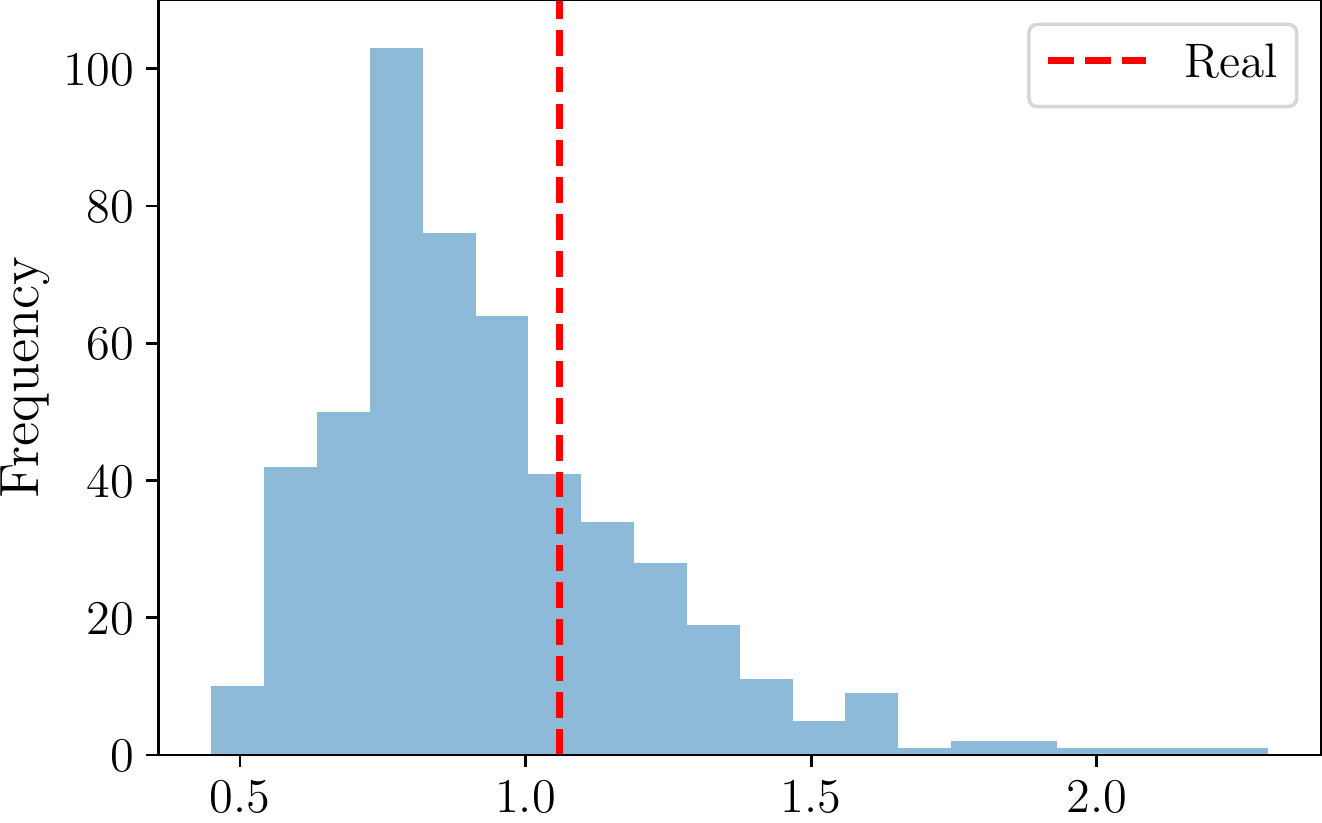}
\caption*{Conditional Wasserstein GAN}
\end{subfigure}
\end{figure}

For the purpose of illustration, we train also a conditional Wasserstein GAN
model with a simple structure as multi-layer perceptrons for the generator and
the discriminator. In order to make a comparison with the results of the
conditional RBM, we also use the values of the last $20$ days as a long memory
for input data of the model. More precisely, the generator will take two
inputs: a $100$-dimensional random noise vector and $20$ past values that are
concatenated into a $20 \times 6 $-dimensional vector. We then construct the
generator using $4$ dense layers with the structure
$\mathcal{S}_{\Generator}=\left\{100, 50, 10, 6\right\} $. We use a leaky RELU
function with $\alpha = 0.5$ for each dense layer and a sigmoid activation
function for the output layer. The discriminator also takes two inputs: a
$6$-dimensional vector of current daily returns of futures contracts and a $20
\times 6 $-dimensional vector of past values. We then construct the
discriminator using $4$ dense layers with the structure
$\mathcal{S}_{\Discriminator}=\left\{100, 50, 10, 1\right\} $. For the first three
dense layers, the activation function corresponds to a leaky RELU
function with $\alpha = 0.5$, whereas the activation function of the output
layer is a tanh activation function. To avoid the problem of outliers, we also
use the normal score transformation as in the case of conditional RBMs before
applying the MinMax scaling function on input data. In Figure
\ref{fig:RPRBMvsGAN}, we compare the distributions of the skew measure of the
risk parity strategy using synthetic time series generated by the conditional
RBM and conditional Wasserstein GAN models. We notice that these two
distributions are similar, except that we have several extreme severe scenarios
in the case of the conditional RBM. We recall that the value of the skew
measure of the real risk parity strategy from January 2018 to December 2019 is
equal to $1.05$ and this value corresponds respectively to the $70.1\%$
quantile in the probability distribution generated by the conditional RBM and
the $72.9\%$ quantile in the probability distribution generated by the
conditional Wasserstein GAN.

\subsubsection{Augmenting the investment universe with market regime indicators}

\begin{figure}[tbph]
\caption{Histogram of the skew measure of the risk parity strategy using
synthetic time series generated by two conditional RBMs}
\label{fig:RPVix}
\begin{subfigure}[b]{0.5\linewidth}
\centering
\includegraphics[width=\linewidth]{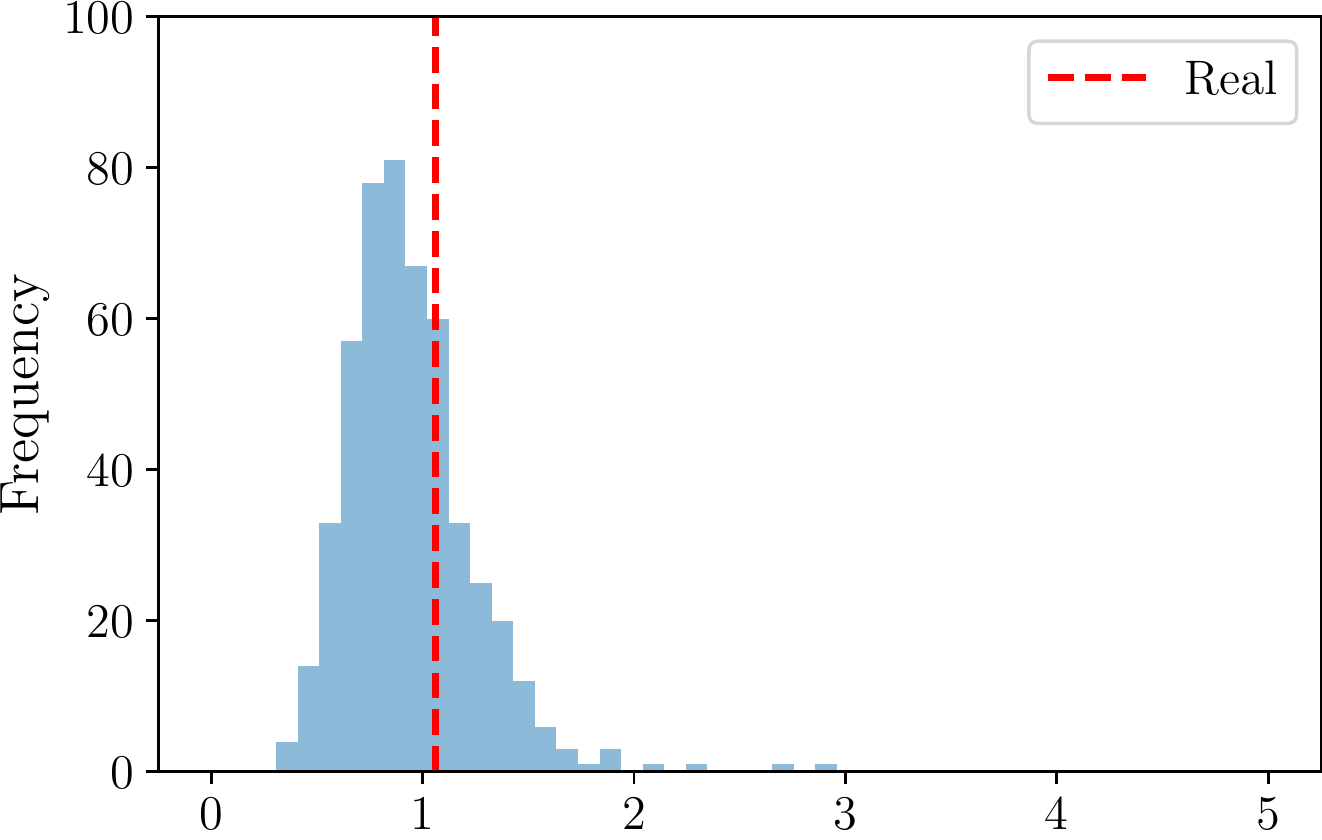}
\caption*{Trained with returns of assets}
\end{subfigure}
\begin{subfigure}[b]{0.5\linewidth}
\centering
\includegraphics[width=\linewidth]{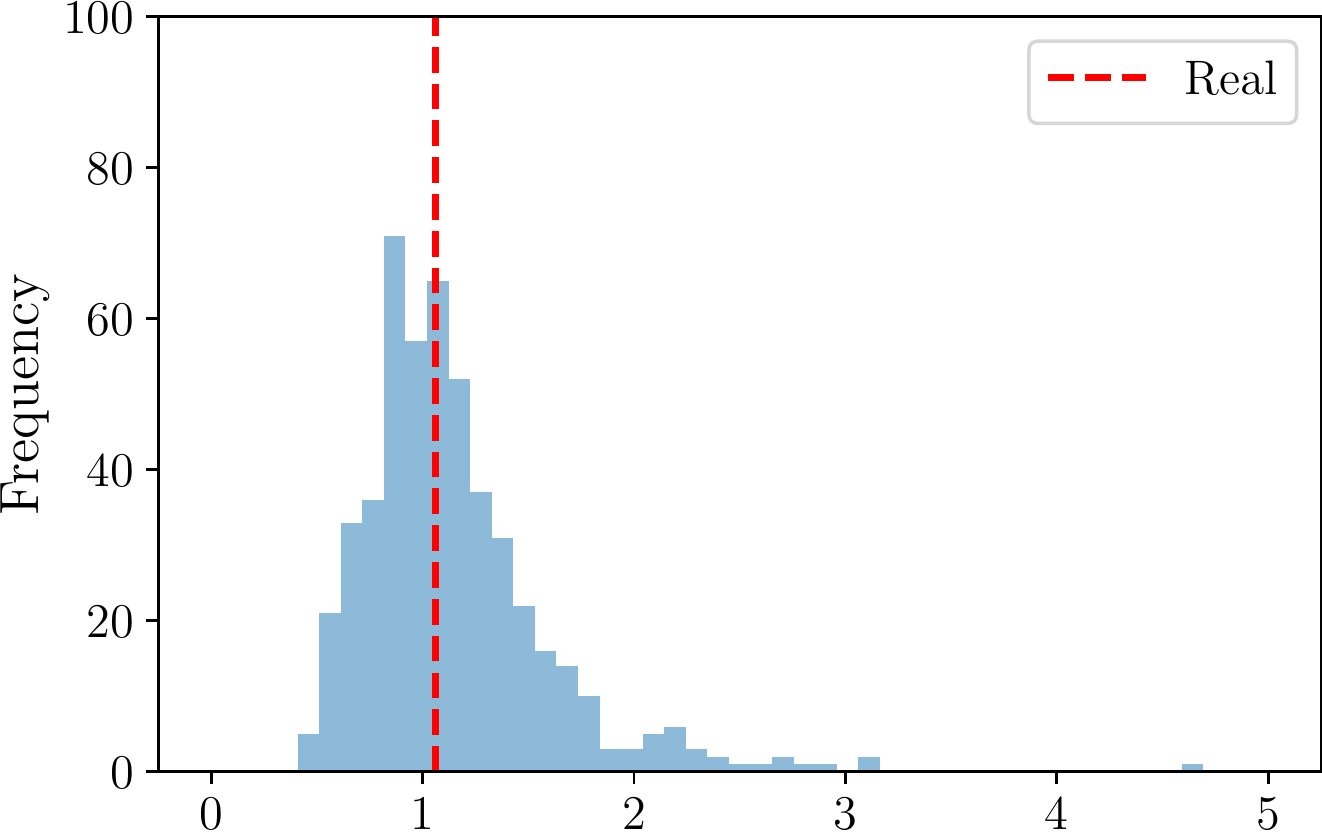}
\caption*{Trained with returns of assets and VIX index}
\end{subfigure}
\end{figure}

In the real world of finance, the correlation structure is very complex and
learning the joint distribution of daily returns of futures contracts using
only time series themselves is not sufficient. For instance, if we train the
generative models by including the time series of the VIX index, it's sure that
we will obtain different results. Indeed, since the daily returns of the VIX
index has a very negative correlation with equity futures contracts and a
positive correlation with bond futures contracts and its distribution is very
leptokurtic, the generative models trained using historical returns of futures
contracts and VIX index will generate more distinct and more severe scenarios.
Figure \ref{fig:RPVix} shows the histograms of the skew measure of the risk
parity strategy using synthetic time series generated by two conditional RBMs.
The first one is trained with only historical daily returns of futures
contracts and the other is trained with historical daily returns of futures
contracts and VIX index. In this figure, we notice that the probability
distribution generated by the model trained with the VIX index has a fatter
tail and more extreme scenarios. Therefore, we consider that this model may
generate more realistic data than the model trained using only the time series
of the future contracts. In order to have a high-quality market generator, we
believe that we should train a conditional RBM or Wasserstein GAN model with
not only the time series of assets that compose the investment portfolio, but
also those of the several market regime indicators such as the VIX index.

\section{Conclusion}

In this article, we explore the use of generative models for improving the
robustness of trading strategies. We consider two approaches for simulating
financial time series. The first one is based on restricted Boltzmann machines,
whereas the second approach corresponds to the family of generative adversarial
networks, including Wasserstein distance models. Given an historical sample of
financial data, we show how to generate new samples of financial data using
these techniques. These new samples of financial data are called synthetic or
fake financial time series, and the objective is to preserve the statistical
properties of the original sample. By statistical properties, we mean the
statistical moments of each univariate time series, the stochastic dependence
between the different variables that compose the multi-dimensional random
vector, and also the time dependence between the observations. If we consider
the financial times series as a multi-dimensional data matrix, the challenge is
then to model both the row and column stochastic structures.\smallskip

There are few satisfactory methods for simulating non-gaussian
multi-dimensional financial time series. For instance, the bootstrap method
does not preserve the cross-correlation between the different variables. The
copula method is better, but it must use the techniques of conditional
augmented data in order to reproduce the autocorrelation functions. Restricted
Boltzmann machines and generative adversarial networks have been successful for
generating complex data with non-linear dependence structures. By applying them
to financial market data, our first results are encouraging and show that these
new alternative techniques may help for simulating non-gaussian
multi-dimensional financial time series. This is particularly true when we
consider the backtesting of trading strategies. In this case, RBMs and GANs may
be used for estimating the probability distribution of performance and risk
statistics of the backtest. This opens the door to a new field of research for
improving the risk management of quantitative investment strategies.

\clearpage

\bibliographystyle{apalike}

\clearpage

\appendix

\section*{Appendix}

\section{Mathematical results}

\subsection{Fundamental concepts of undirected graph model}
\label{appendix:graph-model}

Probabilistic graphical models use graphs to describe interactions between
random variables. Each random variable is represented by a node (or vertice)
and each direct interaction between random variables is represented by an
edge (or link). According to the directionality of the edge in the graph,
probabilistic graphical models can be divided into two categories: directed
graphical model and undirected graphical model. For instance, the Bayesian
networks that are introduced by \citet{Pearl-1985} are a type of directed
graphical models, whereas Markov random fields (or Markov networks) use
undirected graphs \citep{Kindermann-1980}.

\subsubsection{Undirected graph}

An undirected graph $\mathcal{G}=\left( \mathcal{V},\mathcal{E}\right) $ is a
defined by its finite set of nodes $\mathcal{V}$ and its set of undirected
edges $\mathcal{E}$. Each edge $e_{i,j}$ is defined by a pair of two
connected nodes $v_{i}$ and $v_{j}$ from $\mathcal{V}$. We define the
neighborhood $\mathcal{N}\left( v_{i}\right) $ of a given node $v_{i}$ as the
set of all nodes connected to $v_{i}$:
\begin{equation*}
\mathcal{N}\left( v_{i}\right) =\left\{ v_{j}\in \mathcal{V}\mid e_{i,j}\in
\mathcal{E}\right\}
\end{equation*}%
A clique of size $n$ named $\mathcal{C}_{n}$ is a subset of $\mathcal{V}$
containing $n$ nodes $\left( v_{1},v_{2},\ldots ,v_{n}\right) $ defined as:
\begin{equation*}
\forall i,j\in \left\{ 1,2,\ldots ,n\right\} \text{ and }i\neq j:e_{i,j}\in
\mathcal{E}
\end{equation*}%
In other words, each node belonging to $\mathcal{C}_{n}$ is fully connected
with the other nodes of $\mathcal{C}_{n}$. A clique of a graph $\mathcal{G}$
is called maximal if we can't create a bigger clique by adding another node
of $\mathcal{G}$, meaning that no node in $\mathcal{G}$ can be added such
that the resulting set is still a clique.

\subsubsection{Markov random field}

Let $X=\left( X\left( v_{1}\right) ,X\left( v_{2}\right) ,\ldots ,X\left(
v_{n}\right) \right) $ be a set of random variables associated with the
undirected graph $\mathcal{G}=\left( \mathcal{V},\mathcal{E}\right) $ such
that each random variable $X\left( v_{i}\right) $ is linked to the $i^{%
\mathrm{th}}$ node $v_{i}\in \mathcal{V}$. $X$ is said to be a Markov random
field (MRF) if, for all $i\in \left\{ 1,2,\ldots ,n\right\} $, $X\left(
v_{i}\right) $ is conditionally independent from all other variables $%
X\left( v_{j}\right) $, whose nodes $v_{j}$ do not belong to the neighborhood
$\mathcal{N}\left( v_{i}\right) $:
\begin{equation*}
\mathbb{P}\left( x\left( v_{i}\right) \mid x\left( v_{j}\right) ,v_{j}\in
\mathcal{V}\setminus v_{i}\right) =\mathbb{P}\left( x\left( v_{i}\right)
\mid x\left( v_{j}\right) ,v_{j}\in \mathcal{N}\left( v_{i}\right) \right)
\end{equation*}%
In other words, $X$ is said to be a Markov random field if the joint
probability distribution verifies the local Markov property.

\subsubsection{Hammersley-Clifford theorem}
\label{appendix:Hammersley-Clifford}

Let $X=\left( X\left( v_{1}\right) ,X\left( v_{2}\right) ,\ldots ,X\left(
v_{n}\right) \right) $ be a Markov random field and $\mathcal{C}$ the set of
all maximal cliques of the undirected graph $\mathcal{G}=\left( \mathcal{V},%
\mathcal{E}\right) $. According to \citet{Fischer-2014}, a simple version%
\footnote{%
The rigorous formulation of the Hammersley-Clifford theorem can be found in %
\citet{Koller-2009}.} of Hammersley-Clifford theorem states that a strictly
positive distribution $\mathbb{P}$ satisfies the Markov property with respect
to the undirected graph $\mathcal{G}$ if and only if $\mathbb{P}$ factorizes
over $\mathcal{G}$. This means that there exists a set of strictly positive
functions $\left\{ \psi _{C},C\in \mathcal{C}\right\} $, such that the joint
probability distribution is given by a product of factors:
\begin{equation}
\mathbb{P}\left( x\right) =\mathbb{P}\left( x\left( v_{1}\right) ,\ldots
,x\left( v_{n}\right) \right) =\frac{1}{Z}\prod_{C\in \mathcal{C}}\psi
_{C}\left( x\left( v_{C}\right) \right)   \label{eq:HC1}
\end{equation}%
where $\psi _{C}$ is the potential function for the clique $C$, $v_{C}$ are
all the nodes belonging to the clique $C$ and $Z$ is the partition function
given by:
\begin{equation*}
Z=\sum_{x}\prod_{C\in \mathcal{C}}\psi _{C}\left( x\left( v_{C}\right)
\right)
\end{equation*}%
The partition function $Z$ is the normalization constant what ensures the
overall distribution sums to $1$.

\subsubsection{Energy function and Boltzmann distribution}

The Hammersley-Clifford theorem is valid only if each potential function $%
\psi _{C}$ is strictly positive. Thus, we can introduce a new function $E$
in order to rewrite the probability distribution $\mathbb{P}\left( x\right) $%
:
\begin{eqnarray}
\mathbb{P}\left( x\right)  &=&\frac{1}{Z}\exp \left( \sum_{C\in \mathcal{C}%
}\log \psi _{C}\left( x\left( v_{C}\right) \right) \right)   \notag \\
&=&\frac{1}{Z}e^{-E\left( x\right) }  \label{eq:HC2}
\end{eqnarray}%
where $E\left( x\right) =-\sum_{C\in \mathcal{C}}\log \psi _{C}\left( x\left(
v_{C}\right) \right) $ is called the energy function. Because natural
exponential function is always positive, this guarantees that the energy
function will result in a positive probability for any state. In addition, a
large value of energy indicates a low probability of the state. According to
\citet{LeCun-2007}, models of this form are called energy-based
models.\smallskip

Using the energy function described above, the strictly positive probability
distribution of a Markov random field can be expressed in the form $\mathbb{P%
}\left( x\right) =Z^{-1}e^{-E\left( x\right) }$. This form of distribution is
also called Boltzmann (or Gibbs) distribution for a system in statistical
physics. It is defined as:
\begin{equation*}
p_{i}=\frac{1}{Z_{T}}\exp \left(-\frac{E_{i}}{kT}\right)
\end{equation*}%
where $p_{i}$ is the probability of state $i$ of the system, $E_{i}$ is the
energy of state $i$, $k$ is the Boltzmann constant, and $T$ is the
temperature of the system. Since $N$ is the number of states accessible to
the system, the normalization constant is $Z_{T}=\sum_{i=1}^{N}e^{-E_{i}/%
\left( kT\right) }$. If we set $kT$ to $1$, we find that the Boltzmann
distribution and the probability distribution of a Markov random field have
the same formula. For this reason, many energy-based models are called
Boltzmann machines.

\subsubsection{The example of restricted Boltzmann machines}

The restricted Boltzmann machine introduced in Section \ref{section:rbm} on
page \pageref{section:rbm} is a Markov random field associated with a
bipartite undirected graph $\mathcal{G}=\left( \mathcal{V},\mathcal{E}%
\right) $. In other words, all visible layer units and hidden layer units
can be considered as nodes in an undirected graph\footnote{%
For simplicity reasons, we make a little abuse of notation here to use $V_{i}
$ and $H_{j}$ to represent not only nodes in the graph but also random
variables associated with these nodes and we use $v_{i}$ and $h_{j}$ to
denote respectively the possible values of the variables associated with the
$i^{\mathrm{th}}$ visible unit and $j^{\mathrm{th}}$ hidden unit.}:%
\begin{equation*}
\mathcal{V}=\left\{ V_{1},V_{2},\ldots ,V_{m},H_{1},H_{2},\ldots
,H_{n}\right\}
\end{equation*}
and all connections between visible layer and hidden layer are edges of $%
\mathcal{G}$. In the case of RBMs, we know that there are only cliques of
size 1 (one visible unit or one hidden unit) and cliques of size 2 (a pair of
one visible unit and one hidden unit) in the graph $\mathcal{G}$. In
addition, it is easy to show that all these cliques are maximal. Let $%
\mathcal{C}_{1}$ and $\mathcal{C}_{2}$ be respectively the set of all the
cliques of size 1 and the set of all the cliques of size 2. We obtain:
\begin{equation*}
\mathcal{C}_{1}=\left\{ \left\{ V_{1}\right\} ,\left\{ V_{2}\right\} ,\ldots
,\left\{ V_{m}\right\} ,\left\{ H_{1}\right\} ,\left\{ H_{2}\right\} ,\ldots
,\left\{ H_{n}\right\} \right\}
\end{equation*}
and:%
\begin{equation*}
\mathcal{C}_{2}=\left\{ \left\{ V_{1},H_{1}\right\} ,\ldots ,\left\{
V_{i},H_{j}\right\} ,\ldots ,\left\{ V_{m},H_{n}\right\} \right\}
\end{equation*}%
According to the Hammersley-Clifford theorem, the probability distribution of
an RBM is given by:
\begin{eqnarray*}
\mathbb{P}\left( v,h\right)  &=&\frac{1}{Z}\prod_{C\in \left\{ \mathcal{C}%
_{1},\mathcal{C}_{2}\right\} }\psi _{C} \\
&=&\frac{1}{Z}\prod_{i=1}^{m}\psi _{V_{i}}\left( v_{i}\right)
\prod_{j=1}^{n}\psi _{H_{j}}\left( h_{j}\right)
\prod_{i=1}^{m}\prod_{j=1}^{n}\psi _{V_{i},H_{j}}\left( v_{i},h_{j}\right)
\\
&=&\frac{1}{Z}e^{-E\left( v,h\right) }
\end{eqnarray*}%
where:%
\begin{eqnarray*}
E\left( v,h\right)  &=&-\log \left( \prod_{i=1}^{m}\psi _{V_{i}}\left(
v_{i}\right) \prod_{j=1}^{n}\psi _{H_{j}}\left( h_{j}\right)
\prod_{i=1}^{m}\prod_{j=1}^{n}\psi _{V_{i},H_{j}}\left( v_{i},h_{j}\right)
\right)  \\
&=&-\sum_{i=1}^{m}\log \psi _{V_{i}}\left( v_{i}\right) -\sum_{j=1}^{n}\log
\psi _{H_{j}}\left( h_{j}\right) - \\
&&\sum_{i=1}^{m}\sum_{j=1}^{n}\log \psi _{V_{i},H_{j}}\left(
v_{i},h_{j}\right)  \\
&=&\sum_{i=1}^{m}E_{i}\left( v_{i}\right) +\sum_{j=1}^{n}E_{j}\left(
h_{j}\right) +\sum_{i=1}^{m}\sum_{j=1}^{n}E_{i,j}\left( v_{i},h_{j}\right)
\end{eqnarray*}%
In the case of Bernoulli RBMs introduced in Section
\ref{section:bernoulli-rbm} on page \pageref{section:bernoulli-rbm}, we
defined $E_{i}\left( v_{i}\right) =-a_{i}v_{i}$, $E_{j}\left( h_{j}\right)
=-b_{j}h_{j}$ and $E_{i,j}\left( v_{i},h_{j}\right) =-w_{i,j}v_{i}h_{j}$. It
follows that the energy function of a Bernoulli RBM is equal to:
\begin{equation*}
E\left( v,h\right)
=-\sum_{i=1}^{m}a_{i}v_{i}-\sum_{j=1}^{n}b_{j}h_{j}-\sum_{i=1}^{m}%
\sum_{j=1}^{n}w_{i,j}v_{i}h_{j}
\end{equation*}%
where $a_{i}$ and $b_{j}$ are bias terms associated with the visible and
hidden variables $V_{i}$ and $H_{j}$, and $w_{i,j}$ is the weight associated
with the edge between $V_{i}$ and $H_{j}$.

\clearpage

\subsection{Calculus formulas for restricted Boltzmann machines}
\label{appendix:rbm}

\subsubsection{Conditional probability $\mathbb{P}\left( h_{j}=1\mid v\right) $}
\label{appendix:rbm-cp}

We have:%
\begin{eqnarray}
\sum_{h}\mathbb{P}\left( h\mid v\right) h_{j} &=&\sum_{h}\prod_{k=1}^{n}%
\mathbb{P}\left( h_{k}\mid v\right) h_{j} \notag \\
&=&\sum_{h}\mathbb{P}\left( h_{-j}\mid v\right) \mathbb{P}\left( h_{j}\mid
v\right) h_{j} \notag \\
&=&\sum_{h_{j}\in \left\{ 0,1\right\} }\sum_{h_{-j}}\mathbb{P}\left(
h_{-j}\mid v\right) \mathbb{P}\left( h_{j}\mid v\right) h_{j} \notag \\
&=&\sum_{h_{j}\in \left\{ 0,1\right\} }\mathbb{P}\left( h_{j}\mid v\right)
h_{j}\underset{=1}{\cdot \underbrace{\sum_{h_{-j}}\mathbb{P}\left(
h_{-j}\mid v\right) }} \notag \\
&=&\mathbb{P}\left( h_{j}=1\mid v\right) \label{eq:rbm-cp}
\end{eqnarray}
where $h_{-j}=\left( h_{1},h_{2},\ldots ,h_{j-1},h_{j+1},\ldots ,h_{n}\right)
$ denotes the state of all hidden units except the $j^{\mathrm{th}}$ one.

\subsubsection{Bernoulli RBMs and neural networks}
\label{appendix:rbm-nn}

Following \citet{Fischer-2014}, we divide the energy function $E\left(
v,h\right) $ into two parts: one collecting all terms involving $v_{i}$ and
one collecting all the other terms $v_{-i}$:%
\begin{eqnarray*}
E\left( v,h\right)
&=&-\sum_{k=1}^{m}a_{k}v_{k}-\sum_{j=1}^{n}b_{j}h_{j}-\sum_{k=1}^{m}%
\sum_{j=1}^{n}w_{k,j}v_{k}h_{j} \\
&=&-a_{i}v_{i}-\sum_{j=1}^{n}w_{i,j}v_{i}h_{j}-\sum_{k=1,k\neq
i}^{m}a_{k}v_{k}-\sum_{j=1}^{n}b_{j}h_{j}-\sum_{k=1,k\neq
i}^{m}\sum_{j=1}^{n}w_{k,j}v_{k}h_{j} \\
&=&v_{i}\alpha _{i}\left( h\right) +\beta \left( v_{-i},h\right)
\end{eqnarray*}%
where:%
\begin{equation*}
\alpha _{i}\left( h\right) =-a_{i}-\sum_{j=1}^{n}w_{i,j}h_{j}
\end{equation*}%
and:%
\begin{equation*}
\beta \left( v_{-i},h\right) =-\sum_{k=1,k\neq
i}^{m}a_{k}v_{k}-\sum_{j=1}^{n}b_{j}h_{j}-\sum_{k=1,k\neq
i}^{m}\sum_{j=1}^{n}w_{k,j}v_{k}h_{j}
\end{equation*}%
The Bayes theorem gives:%
\begin{eqnarray*}
\mathbb{P}\left( v_{i}=1\mid h\right)  &=&\mathbb{P}\left( v_{i}=1\mid
v_{-i},h\right)  \\
&=&\frac{\mathbb{P}\left( v_{i}=1,v_{-i},h\right) }{\mathbb{P}\left(
v_{-i},h\right) } \\
&=&\frac{\mathbb{P}\left( v_{i}=1,v_{-i},h\right) }{\mathbb{P}\left(
v_{i}=0,v_{-i},h\right) +\mathbb{P}\left( v_{i}=1,v_{-i},h\right) }
\end{eqnarray*}%
We deduce that the conditional probability $\mathbb{P}\left( v_{i}=1\mid
h\right) $ is equal to:%
\begin{eqnarray*}
\mathbb{P}\left( v_{i}=1\mid h\right)  &=&\frac{e^{-E\left(
v_{i}=1,v_{-i},h\right) }}{e^{-E\left( v_{i}=0,v_{-i},h\right) }+e^{-E\left(
v_{i}=1,v_{-i},h\right) }} \\
&=&\frac{e^{-1\cdot \alpha _{i}(h)-\beta \left( v_{-i},h\right) }}{%
e^{-0\cdot \alpha _{i}(h)-\beta \left( v_{-i},h\right) }+e^{-1\cdot \alpha
_{i}(h)-\beta \left( v_{-i},h\right) }} \\
&=&\frac{e^{-\alpha _{i}(h)}}{1+e^{-\alpha _{i}\left( h\right) }} \\
&=&\frac{1}{1+e^{\alpha _{i}\left( h\right) }} \\
&=&\sigma \left( -\alpha _{i}\left( h\right) \right)
\end{eqnarray*}%
where $\sigma \left( x\right) $ is the sigmoid function:%
\begin{equation*}
\sigma \left( x\right) =\frac{1}{1+e^{-x}}
\end{equation*}%
Similarly, we can divide the energy function $E\left( v,h\right) $ into two
parts: one collecting all terms involving $h_{j}$ and one collecting all the
other terms $h_{-j}$:%
\begin{equation*}
E\left( v,h\right) =h_{j}\gamma _{j}\left( v\right) +\delta \left(
v,h_{-j}\right)
\end{equation*}%
where:%
\begin{equation*}
\gamma _{j}\left( v\right) =-b_{j}-\sum_{i=1}^{m}w_{i,j}v_{i}
\end{equation*}%
and:%
\begin{equation*}
\delta \left( v,h_{-j}\right) =-\sum_{i=1}^{m}a_{i}v_{i}-\sum_{k=1,k\neq
j}^{n}b_{k}h_{k}-\sum_{i=1}^{m}\sum_{k=1,k\neq j}^{n}w_{i,k}v_{i}h_{k}
\end{equation*}%
Using the same approach as previously, we can show that:%
\begin{equation*}
\mathbb{P}\left( h_{j}=1\mid v\right) =\sigma \left( -\gamma _{j}\left(
v\right) \right)
\end{equation*}

\subsubsection{Gradient of the Bernoulli RBM log-likelihood function}
\label{appendix:rbm-logl-gradient}

We have:%
\begin{equation*}
\ell \left( \theta \mid v\right) =\log \left( \sum_{h}e^{-E\left( v,h\right)
}\right) -\log \left( \sum_{v^{\prime },h}e^{-E\left( v^{\prime },h\right)
}\right)
\end{equation*}%
\citet{Fischer-2014} computed the log-likelihood gradient $\nabla _{\theta
}\left( v\right) =\partial _{\theta }\ell \left( \theta \mid v\right) $:%
\begin{eqnarray}
\nabla _{\theta }\left( v\right)  &=&\frac{\partial }{\partial \,\theta }%
\left( \log \sum_{h}e^{-E\left( v,h\right) }\right) -\frac{\partial }{%
\partial \,\theta }\left( \sum_{v^{\prime },h}e^{-E\left( v^{\prime
},h\right) }\right)  \notag \\
&=&\nabla _{\theta }^{\left( 1\right) }\left( v\right) +
\nabla _{\theta }^{\left( 2\right) }\left( v^{\prime}\right) \label{eq:rbm-logl-grad1}
\end{eqnarray}%
We have\footnote{Using the Bayes theorem, the conditional probability distribution $\mathbb{P}%
\left( h\mid v\right) $ is equal to:%
\begin{equation*}
\mathbb{P}\left( h\mid v\right) = \frac{\mathbb{P}\left( v,h\right) }{%
\mathbb{P}\left( v\right) } = \frac{e^{-E\left( v,h\right) }}{\sum_{h^{\prime }}e^{-E\left( v,h^{\prime
}\right) }}
\end{equation*}}:%
\begin{eqnarray*}
\nabla _{\theta }^{\left( 1\right) }\left( v\right) &=&-\frac{\sum_{h}e^{-E\left(
v,h\right) }\cdot \partial _{\theta }E\left( v,h\right) }{%
\sum_{h}e^{-E\left( v,h\right) }} \\
&=&-\sum_{h}\frac{e^{-E\left( v,h\right) }}{\sum_{h^{\prime }}e^{-E\left(
v,h^{\prime }\right) }}\partial _{\theta }E\left( v,h\right)  \\
&=&-\sum_{h}\mathbb{P}\left( h\mid v\right) \frac{\partial \,E\left(
v,h\right) }{\partial \,\theta }
\end{eqnarray*}%
and:%
\begin{eqnarray*}
\nabla _{\theta }^{\left( 2\right) }\left( v^{\prime}\right)  &=&\frac{%
\sum_{v^{\prime },h}e^{-E\left( v^{\prime },h\right) }\cdot \partial
_{\theta }E\left( v^{\prime },h\right) }{\sum_{v^{\prime },h}e^{-E\left(
v^{\prime },h\right) }} \\
&=&\sum_{v^{\prime },h}\frac{e^{-E\left( v^{\prime },h\right) }}{%
\sum_{v^{\prime \prime },h^{\prime }}e^{-E\left( v^{\prime \prime
},h^{\prime }\right) }}\frac{\partial \,E\left( v^{\prime },h\right) }{%
\partial \,\theta } \\
&=&\sum_{v^{\prime },h}\mathbb{P}\left( v^{\prime }\mid h\right) \frac{%
\partial \,E\left( v^{\prime },h\right) }{\partial \,\theta } \\
&=&\sum_{v^{\prime }}\sum_{h}\mathbb{P}\left( v^{\prime }\right) \mathbb{P}%
\left( h\mid v^{\prime }\right) \frac{\partial \,E\left( v^{\prime
},h\right) }{\partial \,\theta } \\
&=&\sum_{v^{\prime }}\mathbb{P}\left( v^{\prime }\right) \sum_{h}\mathbb{P}%
\left( h\mid v^{\prime }\right) \frac{\partial \,E\left( v^{\prime
},h\right) }{\partial \,\theta }
\end{eqnarray*}
We recall that $\partial _{a_{i}}E\left( v,h\right) =-v_{i}$ and $\partial
_{b_{j}}E\left( v,h\right) =-h_{j}$. It follows that:%
\begin{eqnarray*}
\frac{\partial \,\ell\left( \theta \mid v\right) }{\partial \,a_{i}} &=&-\sum_{h}%
\mathbb{P}\left( h\mid v\right) \frac{\partial \,E\left( v,h\right) }{%
\partial \,a_{i}}+\sum_{v^{\prime }}\mathbb{P}\left( v^{\prime }\right)
\sum_{h}\mathbb{P}\left( h\mid v^{\prime }\right) \frac{\partial \,E\left(
v^{\prime },h\right) }{\partial \,a_{i}} \\
&=&\sum_{h}\mathbb{P}\left( h\mid v\right) v_{i}-\sum_{v^{\prime }}\mathbb{P}%
\left( v^{\prime }\right) \sum_{h}\mathbb{P}\left( h\mid v^{\prime }\right)
v_{i}^{\prime } \\
&=&v_{i}-\sum_{v^{\prime }}\mathbb{P}\left( v^{\prime }\right) v_{i}^{\prime
}
\end{eqnarray*}%
and\footnote{We use Equation (\ref{eq:rbm-cp}) on page \pageref{eq:rbm-cp}.}:%
\begin{eqnarray*}
\frac{\partial \,\ell\left( \theta \mid v\right) }{\partial \,b_{j}} &=&\sum_{h}%
\mathbb{P}\left( h\mid v\right) h_{j}-\sum_{v^{\prime }}\mathbb{P}\left(
v^{\prime }\right) \sum_{h}\mathbb{P}\left( h\mid v^{\prime }\right) h_{j} \\
&=&\mathbb{P}\left( h_{j}=1\mid v\right) -\sum_{v^{\prime }}\mathbb{P}\left(
v^{\prime }\right) \mathbb{P}\left( h_{j}=1\mid v^{\prime }\right)
\end{eqnarray*}%
For the gradient with respect to $W$, we have $\partial _{w_{i,j}}E\left(
v,h\right) =-v_{i}h_{j}$ and:%
\begin{eqnarray*}
\frac{\partial \,\ell\left( \theta \mid v\right) }{\partial \,w_{i,j}}
&=&\sum_{h}\mathbb{P}\left( h\mid v\right) v_{i}h_{j}-\sum_{v^{\prime }}%
\mathbb{P}\left( v^{\prime }\right) \sum_{h}\mathbb{P}\left( h\mid v^{\prime
}\right) v_{i}^{\prime }h_{j} \\
&=&\mathbb{P}\left( h_{j}=1\mid v\right) v_{i}-\sum_{v^{\prime }}\mathbb{P}%
\left( v^{\prime }\right) \mathbb{P}\left( h_{j}=1\mid v^{\prime }\right)
v_{i}^{\prime }
\end{eqnarray*}
We notice that we have to sum over $2^{m}$ possible combinations of the
visible variables when calculating the second term $\nabla _{\theta
}^{\left( 2\right) }\left( v\right)$. Therefore, we generally approximate the expectation $%
\mathbb{P}\left( v^{\prime }\right) $ by sampling from the model
distribution.

\subsubsection{Gradient of the contrastive divergence}
\label{appendix:rbm-cd-gradient}

The contrastive divergence function is equal to:%
\begin{equation*}
\limfunc{CD}\nolimits^{\left( k\right) }=\func{KL}\left( \mathbb{P}^{\left(
0\right) }\parallel \mathbb{P}^{\left( \infty \right) }\right) -\func{KL}%
\left( \mathbb{P}^{\left( k\right) }\parallel \mathbb{P}^{\left( \infty
\right) }\right)
\end{equation*}%
where:%
\begin{eqnarray*}
\func{KL}\left( \mathbb{P}\parallel \mathbb{Q}\right)  &=&\sum_{v}\mathbb{P}%
\left( v\right) \log \left( \frac{\mathbb{P}\left( v\right) }{\mathbb{Q}%
\left( v\right) }\right)  \\
&=&\sum_{v}\mathbb{P}\left( v\right) \log \mathbb{P}\left( v\right) -\sum_{v}%
\mathbb{P}\left( v\right) \log \mathbb{Q}\left( v\right)
\end{eqnarray*}%
We have:%
\begin{eqnarray}
\frac{\partial \,\func{KL}\left( \mathbb{P}\parallel \mathbb{Q}\right) }{%
\partial \,\theta } &=&\sum_{v}\frac{\partial \,\mathbb{P}\left( v\right) }{%
\partial \,\theta }\log \mathbb{P}\left( v\right) +\sum_{v}\mathbb{P}\left(
v\right) \frac{\partial \,\log \mathbb{P}\left( v\right) }{\partial \,\theta
}-  \notag \\
&&\sum_{v}\frac{\partial \,\mathbb{P}\left( v\right) }{\partial \,\theta }%
\log \mathbb{Q}\left( v\right) -\sum_{v}\mathbb{P}\left( v\right) \frac{%
\partial \,\log \mathbb{Q}\left( v\right) }{\partial \,\theta }
\label{eq:kl-gradient}
\end{eqnarray}%
When $\mathbb{P}$ does not depend on the parameter set $\theta $, we have $%
\partial _{\theta }\mathbb{P}\left( v\right) =0$ and the derivative reduces
to\footnote{%
This is the case when $\mathbb{P}$ is equal to $\mathbb{P}^{\left( 0\right) }
$.}:%
\begin{equation*}
\frac{\partial \,\func{KL}\left( \mathbb{P}\parallel \mathbb{Q}\right) }{%
\partial \,\theta }=-\sum_{v}\mathbb{P}\left( v\right) \frac{\partial \,\log
\mathbb{Q}\left( v\right) }{\partial \,\theta }
\end{equation*}%
Since we have:%
\begin{equation*}
\sum_{v}\mathbb{P}\left( v\right) \frac{\partial \,\log \mathbb{P}\left(
v\right) }{\partial \,\theta }=\sum_{v}\frac{\partial \,\mathbb{P}\left(
v\right) }{\partial \,\theta }
\end{equation*}%
we also notice that another expression of Equation (\ref{eq:kl-gradient}) is:%
\begin{equation*}
\frac{\partial \,\func{KL}\left( \mathbb{P}\parallel \mathbb{Q}\right) }{%
\partial \,\theta }=\sum_{v}\frac{\partial \,\mathbb{P}\left( v\right) }{%
\partial \,\theta }\left( \log \frac{\mathbb{P}\left( v\right) }{\mathbb{Q}%
\left( v\right) }+1\right) -\sum_{v}\mathbb{P}\left( v\right) \frac{\partial
\,\log \mathbb{Q}\left( v\right) }{\partial \,\theta }
\end{equation*}%
Finally, we deduce that:%
\begin{eqnarray}
\frac{\partial \,\limfunc{CD}\nolimits^{\left( k\right) }}{\partial \,\theta
} &=&\frac{\partial }{\partial \,\theta }\left( \sum_{v}\mathbb{P}^{\left(
0\right) }\left( v\right) \log \left( \frac{\mathbb{P}^{\left( 0\right)
}\left( v\right) }{\mathbb{P}^{\left( \infty \right) }\left( v\right) }%
\right) \right) -\frac{\partial }{\partial \,\theta }\left( \sum_{v}\mathbb{P%
}^{\left( k\right) }\left( v\right) \log \left( \frac{\mathbb{P}^{\left(
k\right) }\left( v\right) }{\mathbb{P}^{\left( \infty \right) }\left(
v\right) }\right) \right)   \notag \\
&=&-\sum_{v}\mathbb{P}^{\left( 0\right) }\left( v\right) \frac{\partial
\,\log \mathbb{P}^{\left( \infty \right) }\left( v\right) }{\partial
\,\theta }+\sum_{v}\mathbb{P}^{\left( k\right) }\left( v\right) \frac{%
\partial \,\log \mathbb{P}^{\left( \infty \right) }\left( v\right) }{%
\partial \,\theta }-  \notag \\
&&\sum_{v}\frac{\partial \,\mathbb{P}^{\left( k\right) }\left( v\right) }{%
\partial \,\theta }\left( \log \frac{\mathbb{P}^{\left( k\right) }\left(
v\right) }{\mathbb{P}^{\left( \infty \right) }\left( v\right) }+1\right)
\label{eq:cd-grad1}
\end{eqnarray}%
In Equation (\ref{eq:cd-grad1}), it is often possible to compute the exact
values for the first two terms, but not for the third term. However,
\citet{Hinton-2002} showed experimentally that this last term is so small
compared the other two terms that it can be ignored. Thus, we can consider
the following approximation:%
\begin{eqnarray*}
\frac{\partial \,\limfunc{CD}\nolimits^{\left( k\right) }}{\partial \,\theta
} &\approx &\sum_{v}\mathbb{P}^{\left( k\right) }\left( v\right) \frac{%
\partial \,\log \mathbb{P}^{\left( \infty \right) }\left( v\right) }{%
\partial \,\theta }-\sum_{v}\mathbb{P}^{\left( 0\right) }\left( v\right)
\frac{\partial \,\log \mathbb{P}^{\left( \infty \right) }\left( v\right) }{%
\partial \,\theta } \\
&=&\sum_{v}\mathbb{P}^{\left( k\right) }\left( v\right) \frac{\partial
\,\log \mathbb{P}_{\theta }\left( v\right) }{\partial \,\theta }-\sum_{v}%
\mathbb{P}^{\left( 0\right) }\left( v\right) \frac{\partial \,\log \mathbb{P}%
_{\theta }\left( v\right) }{\partial \,\theta } \\
&=&\frac{1}{N}\sum_{s}\left( \frac{\partial \,\log \mathbb{P}_{\theta
}\left( v_{\left( s\right) }^{\left( k\right) }\right) }{\partial \,\theta }-%
\frac{\partial \,\log \mathbb{P}_{\theta }\left( v_{\left( s\right)
}^{\left( 0\right) }\right) }{\partial \,\theta }\right)
\end{eqnarray*}%
where $v_{\left( s\right) }^{\left( 0\right) }$ is the $s^{\mathrm{th}}$
sample of the training set and $v_{\left( s\right) }^{\left( k\right) }$ is
the associated sample after running $k$ steps of Gibbs sampling\footnote{%
In this case, $v_{\left( s\right) }^{\left( 0\right) }$ is the starting value
of the Gibbs sampler.}. By noticing that $\log \mathbb{P}_{\theta }\left(
v\right) =\ell \left( \theta \mid v\right) $ and using Equation
(\ref{eq:rbm-logl-grad1}),
we obtain:%
\begin{eqnarray*}
\frac{\partial \,\limfunc{CD}\nolimits^{\left( k\right) }}{\partial \,\theta
} &=&\frac{1}{N}\sum_{s=1}^{N}\left( \frac{\partial \,\ell \left( \theta
\mid v_{\left( s\right) }^{\left( k\right) }\right) }{\partial \,\theta }-%
\frac{\partial \,\ell \left( \theta \mid v_{\left( s\right) }^{\left(
0\right) }\right) }{\partial \,\theta }\right)  \\
&=&\frac{1}{N}\sum_{s=1}^{N}\left( \nabla _{\theta }\left( v_{\left(
s\right) }^{\left( k\right) }\right) -\nabla _{\theta }\left( v_{\left(
s\right) }^{\left( 0\right) }\right) \right)  \\
&=&\frac{1}{N}\sum_{s=1}^{N}\left( \nabla _{\theta }^{\left( 1\right)
}\left( v_{\left( s\right) }^{\left( k\right) }\right) -\nabla _{\theta
}^{\left( 1\right) }\left( v_{\left( s\right) }^{\left( 0\right) }\right)
\right)  \\
&=&\frac{1}{N}\sum_{s=1}^{N}\sum_{h}\left( \mathbb{P}\left( h\mid v_{\left(
s\right) }^{\left( 0\right) }\right) \frac{\partial \,E\left( v_{\left(
s\right) }^{\left( 0\right) },h\right) }{\partial \,\theta }-\mathbb{P}%
\left( h\mid v_{\left( s\right) }^{\left( k\right) }\right) \frac{\partial
\,E\left( v_{\left( s\right) }^{\left( k\right) },h\right) }{\partial
\,\theta }\right)
\end{eqnarray*}
Therefore, we can compute the derivatives with respect to the parameters $%
a_{i}$, $b_{j}$ and $w_{i,j}$:%
\begin{eqnarray*}
\frac{\partial \,\limfunc{CD}\nolimits^{\left( k\right) }}{\partial \,a_{i}}
&=&\frac{1}{N}\sum_{s=1}^{N}\left( v_{\left( s\right) ,i}^{\left( k\right)
}-v_{\left( s\right) ,i}^{\left( 0\right) }\right)  \\
\frac{\partial \,\limfunc{CD}\nolimits^{\left( k\right) }}{\partial \,b_{j}}
&=&\frac{1}{N}\sum_{s=1}^{N}\left( \mathbb{P}\left( h_{j}=1\mid v_{\left(
s\right) }^{\left( k\right) }\right) -\mathbb{P}\left( h_{j}=1\mid v_{\left(
s\right) }^{\left( 0\right) }\right) \right)  \\
\frac{\partial \,\limfunc{CD}\nolimits^{\left( k\right) }}{\partial \,w_{i,j}%
} &=&\frac{1}{N}\sum_{s=1}^{N}\left( \mathbb{P}\left( h_{j}=1\mid v_{\left(
s\right) }^{\left( k\right) }\right) \cdot v_{\left( s\right) ,i}^{\left(
k\right) }-\mathbb{P}\left( h_{j}=1\mid v_{\left( s\right) }^{\left(
0\right) }\right) \cdot v_{\left( s\right) ,i}^{\left( 0\right) }\right)
\end{eqnarray*}

\subsubsection{Gradient of the Gaussian-Bernoulli RBM log-likelihood function}
\label{appendix:gaussian-rbm-grad}

By updating Equation (\ref{eq:rbm-logl-grad1}), we can easily show that the
gradient of the log-likelihood function is equal to:
\begin{equation}
\frac{\partial \,\ell \left( \theta \mid v\right) }{\partial \,\theta }%
=-\sum_{h}\mathbb{P}\left( h\mid v\right) \frac{\partial \,E_{g}\left(
v,h\right) }{\partial \,\theta }+\int_{v^{\prime }}p\left( v^{\prime
}\right) \sum_{h}\mathbb{P}\left( h\mid v^{\prime }\right) \frac{\partial
\,E_{g}\left( v^{\prime },h\right) }{\partial \,\theta }\,\mathrm{d}%
v^{\prime }
\end{equation}%
where:%
\begin{equation*}
E_{g}\left( v,h\right) =\sum_{i=1}^{m}\frac{\left( v_{i}-a_{i}\right) ^{2}}{%
2\sigma _{i}^{2}}-\sum_{j=1}^{n}b_{j}h_{j}-\sum_{i=1}^{m}%
\sum_{j=1}^{n}w_{i,j}\frac{v_{i}h_{j}}{\sigma _{i}^{2}}
\end{equation*}%
We have:%
\begin{eqnarray*}
\frac{\partial \,E_{g}\left( v,h\right) }{\partial \,a_{i}} &=&\frac{%
v_{i}-a_{i}}{\sigma _{i}^{2}} \\
\frac{\partial \,E_{g}\left( v,h\right) }{\partial \,b_{j}} &=&-h_{j} \\
\frac{\partial \,E_{g}\left( v,h\right) }{\partial \,w_{i,j}} &=&-\frac{%
v_{i}h_{j}}{\sigma _{i}^{2}} \\
\frac{\partial \,E_{g}\left( v,h\right) }{\partial \,\sigma _{i}} &=&-\frac{%
\left( v_{i}-a_{i}\right) ^{2}}{\sigma _{i}^{3}}+v_{i}\sum_{j}w_{i,j}\frac{%
h_{j}}{2\sigma _{i}^{3}}
\end{eqnarray*}%
We deduce that the derivative of $\ell \left( \theta \mid v\right) $ with
respect to the weight $w_{i,j}$ is given by:%
\begin{eqnarray*}
\frac{\partial \,\ell \left( \theta \mid v\right) }{\partial \,w_{i,j}}
&=&\sum_{h}\mathbb{P}\left( h\mid v\right) \frac{v_{i}h_{j}}{\sigma _{i}^{2}}%
-\int_{v^{\prime }}p\left( v^{\prime }\right) \sum_{h}\mathbb{P}\left( h\mid
v^{\prime }\right) \frac{v_{i}^{\prime }h_{j}}{\sigma _{i}^{2}}\,\mathrm{d}%
v^{\prime } \\
&=&\mathbb{P}\left( h_{j}=1\mid v\right) \frac{v_{i}}{\sigma _{i}^{2}}%
-\int_{v^{\prime }}p\left( v^{\prime }\right) \mathbb{P}\left( h_{j}=1\mid
v^{\prime }\right) \frac{v_{i}^{\prime }}{\sigma _{i}^{2}}\,\mathrm{d}%
v^{\prime }
\end{eqnarray*}
Similarly, we can compute the other derivatives. For $a_{i}$ and $b_{j}$, we
obtain:%
\begin{equation*}
\frac{\partial \,\ell \left( \theta \mid v\right) }{\partial \,a_{i}}=-\frac{%
v_{i}}{\sigma _{i}^{2}}+\int_{v^{\prime }}p\left( v^{\prime }\right) \frac{%
v_{i}^{\prime }}{\sigma _{i}^{2}}\,\mathrm{d}v^{\prime }
\end{equation*}%
and:%
\begin{equation*}
\frac{\partial \,\ell \left( \theta \mid v\right) }{\partial \,b_{j}}=%
\mathbb{P}\left( h_{j}=1\mid v\right) -\int_{v^{\prime }}p\left( v^{\prime
}\right) \mathbb{P}\left( h_{j}=1\mid v^{\prime }\right) \,\mathrm{d}%
v^{\prime }
\end{equation*}%
Finally, we obtain for the parameter $\sigma _{i}$:%
\begin{eqnarray*}
\frac{\partial \,\ell \left( \theta \mid v\right) }{\partial \,\sigma _{i}}
&=&-\frac{\left( v_{i}-a_{i}\right) ^{2}}{\sigma _{i}^{3}}+\frac{v_{i}}{%
2\sigma _{i}^{3}}\sum_{j}w_{i,j}\mathbb{P}\left( h_{j}=1\mid v\right) + \\
&&\int_{v^{\prime }}p\left( v^{\prime }\right) \left( \frac{\left(
v_{i}^{\prime }-a_{i}\right) ^{2}}{\sigma _{i}^{3}}-\frac{v_{i}^{\prime }}{%
2\sigma _{i}^{3}}\sum_{j}w_{i,j}\mathbb{P}\left( h_{j}=1\mid v^{\prime
}\right) \right) \,\mathrm{d}v^{\prime }
\end{eqnarray*}

\subsubsection{Gradient of the conditional RBM log-likelihood function}
\label{appendix:cond-rbm-grad}

The log-likelihood function for the conditional RBM is:
\begin{equation*}
\ell \left( \theta \mid v_{t}\right) =\log p_{\theta }\left( v_{t}\mid
c_{t}\right)
\end{equation*}%
where the set of parameters becomes $\theta =\left( a,b,W,P,Q\right) $. We can then compute the partial derivative of the energy function by using
the chain rule:%
\begin{eqnarray*}
\frac{\partial \,\tilde{E}_{g}\left( v_{t},h_{t},c_{t}\right) }{\partial
\,q_{k,i}} &=&\frac{\partial \,\tilde{E}_{g}\left( v_{t},h_{t},c_{t}\right)
}{\partial \,\tilde{a}_{t}}\cdot \frac{\partial \,\tilde{a}_{t}}{\partial
\,q_{k,i}} \\
&=&\left( \frac{v_{t,i}-\tilde{a}_{t,i}}{\sigma _{i}^{2}}\right) c_{t,k}
\end{eqnarray*}
and we obtain:
\begin{equation*}
\frac{\partial \,\ell \left( \theta \mid v_{t}\right) }{\partial \,q_{k,i}}%
=-\left( \frac{v_{t,i}-\tilde{a}_{t,i}}{\sigma _{i}^{2}}\right)
c_{t,k}+\int_{v^{\prime }}p\left( v^{\prime }\right) \left( \frac{%
v_{t,i}^{\prime }-\tilde{a}_{t,i}}{\sigma _{i}^{2}}\right) c_{t,k}\,\mathrm{d%
}v^{\prime }
\end{equation*}%
Similarly, we have:%
\begin{eqnarray*}
\frac{\partial \,\tilde{E}_{g}\left( v_{t},h_{t},c_{t}\right) }{\partial
\,p_{k,j}} &=&\frac{\partial \,\tilde{E}_{g}\left( v_{t},h_{t},c_{t}\right)
}{\partial \,\tilde{b}_{t}}\cdot \frac{\partial \,\tilde{b}_{t}}{\partial
\,p_{k,j}} \\
&=&-h_{t,j}c_{t,k}
\end{eqnarray*}%
and:%
\begin{eqnarray*}
\frac{\partial \,\ell \left( \theta \mid v_{t}\right) }{\partial \,p_{k,j}}
&=&\mathbb{P}\left( h_{t,j}=1\mid v_{t},c_{t}\right) c_{t,k}- \\
&&\int_{v^{\prime }}p\left( v_{t}^{\prime }\mid c_{t}\right) \mathbb{P}%
\left( h_{j}=1\mid v_{t}^{\prime },c_{t}\right) c_{t,k}\,\mathrm{d}v^{\prime
}
\end{eqnarray*}
The calculation for the other derivatives remains unchanged and are the same
as those obtained for the Gaussian-Bernoulli RBM.

\clearpage

\subsection{A unified approach of GAN models}
\label{appendix:unified}

Generative models are trained to perform a mapping from a latent
space to some specified data manifold, which is generally
represented by the empirical distribution of real data. The problem
consists then in finding a mapping function that best matches the
target data in the sense of a certain discrepancy measure. For
comparing the theoretical distribution with the empirical
distribution, the two important metrics used in the context of
generative modeling are divergence measures ($\phi $-divergence) and
integral probability metrics (IPM), which give two different
families of GANs. In this section, we show that all models share
many significant common points. Let $\mathcal{X}$ be a topological
space that is compact, complete and separable. Let us assume that
there are two probability measures $\mathbb{P}$ and $\mathbb{Q}$
which can be defined on $\mathcal{X}$. GAN optimization relies on
the fact that both $\phi $-divergence and IPM can be written as:
\begin{equation*}
d_{\mathcal{F}}\left( \mathbb{P},\mathbb{Q}\right) =\sup_{\varphi \in
\mathcal{F}}\left\vert \Delta _{\mathbb{P},\mathbb{Q}}\left( \varphi \right)
\right\vert
\end{equation*}%
where $\mathcal{F}$ is the class of functions $\varphi $ defined on
$\mathcal{X}$ and $\Delta :\mathcal{F}\rightarrow \mathcal{X}$ is a discrepancy
operator. In the following, we show that the choice of the class $\mathcal{F}$
and the discrepancy operator $\Delta $ lead to different GAN models.

\subsubsection{$\protect\phi $-GAN models}
\label{appendix:phi-gan}

\paragraph{Variational estimation of $\protect\phi $-divergences}

The very first model of GAN is often presented as a minmax optimization problem
\citep{Goodfellow-2014}. However, it is possible to find a direct
correspondence between Goodfellow's saddle point problem and divergence
minimization. Let us recall the definition of a divergence measure. We assume
that there are two probability measures $\mathbb{P}$ and $\mathbb{Q}$ that can
be defined on $\mathcal{X}$. Moreover, $\mathbb{P}$ must be
absolutely continuous\footnote{%
This means that if we consider a $\sigma $-field $\mathcal{A}\subseteq
\mathcal{X}$ such that $\mathbb{Q}\left( \mathcal{A}\right) =0$, then $%
\mathbb{P}\left( \mathcal{A}\right) =0$.} with respect to $\mathbb{Q}$, which is
denoted $\mathbb{P}\ll \mathbb{Q}$. The $\phi $-divergence $D_{\phi }$ is
defined as:
\begin{equation}
D_{\phi }\left( \mathbb{P}\parallel \mathbb{Q}\right) =\int_{\mathcal{X}%
}\phi \left( \frac{\mathrm{d}\mathbb{P}\left( x\right) }{\mathrm{d}\mathbb{Q}%
\left( x\right) }\right) \,\mathbb{Q}\left( \mathrm{d}x\right)
\end{equation}%
where $\phi :\mathbb{R}^{+}\rightarrow \mathbb{R}\cup \left\{ +\infty
\right\} $ is a convex, lower-semi-continuous function such that\footnote{%
This last condition ensures that $D_{\phi }\left( \mathbb{P}\parallel
\mathbb{Q}\right) =0$ if $\mathbb{P}=\mathbb{Q}$.} $\phi \left( 1\right) =0$%
. Looking at the closed-form solution of $\phi $-divergence, we note that it
can be interpreted as the likelihood ratio between two probability
distributions\footnote{%
The condition that imposes that $\mathbb{P}$ must be absolutely continuous with
respect to $\mathbb{Q}$ is related to the Radon-Nikodym theorem. It
states that if $\mathbb{P}\ll \mathbb{Q}$, then there is a function $f:%
\mathcal{X}\rightarrow \mathbb{R}$ that satisfies $\mathbb{P}\left( A\right)
=\int_{A}f\left( x\right) \,\mathbb{Q}\left( \mathrm{d}x\right) $ for all $%
\sigma $-field $\mathcal{A}$. The function $f$ is often denoted by $\dfrac{%
\mathrm{d}\mathbb{P}}{\mathrm{d}\mathbb{Q}}$.}.

\begin{remark}
Different divergence measures can be used for modeling the function $\phi $:

\begin{itemize}
\item the Kullback-Leibler divergence $D_{KL}$ corresponds to $\phi \left(
    t\right) =t\log \left( t\right) $;

\item the Jensen-Shannon divergence $D_{JS}$ is obtained by setting $\phi
    \left( t\right) =-\left( t+1\right) \log \left( \frac{1+t}{2}\right)
    +t\log \left( t\right) $;

\item the total variation (or energy-based) divergence $D_{TV}$ is defined by
    taking $\phi \left( t\right) =\frac{1}{2}\left\vert t-1\right\vert $.
\end{itemize}

\end{remark}

In order to establish the link with GAN optimization problems, %
\citet{Nguyen-2010} proposed to compute a variational
characterization of these $\phi $-divergence measures by looking at
the convex dual. For that, we need to introduce the Fenchel
conjugate, which is a fundamental
tool in convex analysis \citep{Barbu-2012}. Let us consider a function $f:%
\mathcal{X}\rightarrow \mathbb{R}\cup \left\{ +\infty \right\} $.
According to the Riesz representation theorem, it is possible to
identify the dual space $\mathcal{X^{\ast }}$ of the Banach space
$\mathcal{X}$. Therefore, we can work on the product space
$\mathcal{X^{\ast }}\times \mathcal{X}$
associated with the scalar product $\left\langle x^{\ast },x\right\rangle $%
. The Fenchel transform is then defined on the dual space such that:
\begin{equation*}
f^{\ast }:\left\{
\begin{array}{l}
{\mathcal{X^{\ast }}}\rightarrow \mathbb{R} \\
x^{\ast }\rightarrow \sup_{x\in \limfunc{dom}f}\left\{ \left\langle
x^{\ast },x\right\rangle -f\left( x\right) \right\}
\end{array}%
\right.
\end{equation*}%
where $\limfunc{dom}f=\left\{ x\in \mathcal{X}\mid f\left( x\right)
<+\infty \right\} $. According to the Fenchel-Moreau theorem, if $f$
is convex and continuous, then $f^{\ast \ast }=f^{\ast }\circ
f^{\ast }=f$ and we obtain:
\begin{equation*}
f\left( x\right) =\sup_{x^{\ast }\in \limfunc{dom}f^{\ast }}\left\{
\left\langle x,x^{\ast }\right\rangle -f^{\ast }\left( x^{\ast
}\right) \right\}
\end{equation*}%
for all $x\in \mathcal{X}$. If we consider the function $\phi $ associated with
a given $\phi $-divergence, the space $\mathcal{X}$ is $\mathbb{R}$ and
the dual space is also $\mathbb{R}$. In order to express the $\phi $%
-divergence in terms of loss, \citet{Nguyen-2010} simply expressed
$\phi $ in term of its conjugate:
\begin{equation*}
\phi \left( x\right) =\sup_{x^{\ast }\in \limfunc{dom}\phi ^{\ast
}}\left\{ \left\langle x,x^{\ast }\right\rangle -\phi ^{\ast }\left(
x^{\ast }\right) \right\}
\end{equation*}%
Using Jensen inequality and considering that $\phi ^{\ast }$ is
convex, it follows that:
\begin{eqnarray*}
D_{\phi }\left( \mathbb{P}\parallel \mathbb{Q}\right)  &=&\int_{\mathcal{X}%
}\phi \left( \frac{\mathrm{d}\mathbb{P}\left( x\right) }{\mathrm{d}\mathbb{Q}%
\left( x\right) }\right) \,\mathbb{Q}\left( \mathrm{d}x\right)  \\
&=&\int_{\mathcal{X}}\left( \sup_{x^{\ast }\in \limfunc{dom}\phi
^{\ast
}}\left\{ \left\langle \frac{\mathrm{d}\mathbb{P}\left( x\right) }{\mathrm{d}%
\mathbb{Q}\left( x\right) },x^{\ast }\right\rangle -\phi ^{\ast
}\left(
x^{\ast }\right) \right\} \,\mathrm{d}\mathbb{Q}\left( x\right) \right) \,%
\mathrm{d}x \\
&\geq &\sup_{x^{\ast }\in \limfunc{dom}\phi ^{\ast }}\left\{ \int_{%
\mathcal{X}}\left( x^{\ast }\,\mathrm{d}\mathbb{P}\left( x\right)
-\phi ^{\ast }\left( x^{\ast }\right) \,\mathrm{d}\mathbb{Q}\left(
x\right) \right) \,\mathrm{d}x\right\}
\end{eqnarray*}%
We then introduce a class of functions $\mathcal{F}$ that maps
$\mathcal{X}$ to $\limfunc{dom}\phi ^{\ast }$:
\begin{eqnarray*}
D_{\phi }\left( \mathbb{P}\parallel \mathbb{Q}\right)  &\geq &\sup_{\varphi
\in \mathcal{F}}\left\{ \int_{\mathcal{X}}\varphi \left( x\right) \,\mathbb{P%
}\left( \mathrm{d}x\right) -\int_{\mathcal{X}}\phi ^{\ast }\left(
\varphi
\left( x\right) \right) \,\mathbb{Q}\left( \mathrm{d}x\right) \right\}  \\
&\geq &\sup_{\varphi \in \mathcal{F}}\left\{ \mathbb{E}\left[
\varphi \left( X\right) )\mid X\sim \mathbb{P}\right]
-\mathbb{E}\left[ \phi ^{\ast }\left( \varphi \left( X\right)
\right) \mid X\sim \mathbb{Q}\right] \right\}
\end{eqnarray*}%
where $\left\{ \mathcal{F}=\varphi :\mathcal{X}\rightarrow
\mathbb{R}\mid \phi \left( \mathcal{X}\right) \subseteq
\limfunc{dom}\phi ^{\ast }\right\}
$. We define the discrepancy operator for this specific model as follows:%
\begin{equation}
\Delta _{\mathbb{P},\mathbb{Q}}\left( \varphi \right)
=\mathbb{E}\left[ \varphi \left( X\right) \mid X\sim
\mathbb{P}\right] -\mathbb{E}\left[ \phi ^{\ast }\left( \varphi
\left( X\right) \right) \mid X\sim \mathbb{Q}\right]
\end{equation}%
To find the optimal function such that equality in the supremum is obtained, we
introduce the notation $\tilde{x}=\varphi \left( x\right) $ and the quantity
$\mathcal{C}\left( \tilde{x}\right) $ defined by:
\begin{eqnarray}
\mathcal{C}\left( \tilde{x}\right)  &=&C\left( \varphi \left( x\right)
\right) \notag  \\
&=&\mathbb{E}\left[ \varphi \left( X\right) \mid X\sim \mathbb{P}\right] -%
\mathbb{E}\left[ \phi ^{\ast }\left( \varphi \left( X\right) \right)
\mid X\sim \mathbb{Q}\right] \label{eq-Nowozin1}
\end{eqnarray}%
By computing the derivative $\mathcal{C}\left( \tilde{x}\right)$,
\citet{Nowozin-2016} found that the optimal function $\varphi
^{\star }$ is\footnote{See \citet[Theorem 4.4]{Broniatowski-2006} for a formal proof.}:
\begin{equation}
\varphi ^{\star }\left( x\right) =\phi ^{\prime }\left( \frac{\mathrm{d}%
\mathbb{P}\left( x\right) }{\mathrm{d}\mathbb{Q}\left( x\right) }\right)
\end{equation}%
However, evaluating this quantity is impossible because the
distribution function $\mathbb{P}$ is unknown. Therefore, $\varphi $
should be flexible enough to approximate the derivative $\phi
^{\prime }$ everywhere. This is why GAN models use deep neural
networks to estimate it. This leads us to introduce the parameter
$\theta _{d}$ that will be optimized during the training process. In
this context, we write the parameterized function
$\Discriminator\left( x,\theta _{d}\right) $, which aims to estimate
the function $\varphi ^{\star }\left( x\right) $:
\begin{equation*}
\hat{\varphi}^{\star }\left( x\right) =\Discriminator\left( x,\theta
_{d}\right)
\end{equation*}
Consequently, \citet{Nowozin-2016} proposed to use the resulting
lower bound in order to train GANs. $\Generator\left( z,\theta
_{g}\right) $ represents the generative model that is also a neural
network and allows us to build the probability distribution
$\mathbb{P}_{\mathrm{model}}$, which should estimate the given
probability distribution $\mathbb{P}_{\mathrm{data}}$. Thus, we
obtain the saddle point problem:
\begin{equation}
\min_{\phi }\Delta _{\mathbb{P}_{\mathrm{data}},\mathbb{P}_{\mathrm{model}%
}}\left( \varphi ^{\star }\right) \approx \min_{\theta _{g}}\max_{\theta
_{d}}\mathcal{C}\left( \Discriminator\left( X,\theta _{d}\right) \right)
\end{equation}%
where:%
\begin{equation}
\mathcal{C}\left( \Discriminator\left( X,\theta _{d}\right) \right) =\mathbb{%
E}\left[ \Discriminator\left( X,\theta _{d}\right) \mid X\sim \mathbb{P}_{%
\mathrm{data}}\right] -\mathbb{E}\left[ \phi ^{\ast }\left( \Discriminator%
\left( X,\theta _{d}\right) \right) \mid X\sim \Generator\left( z,\theta
_{g}\right) \right]
\end{equation}%
$\Discriminator\left( x,\theta _{d}\right) \in \mathcal{F}$ and $\mathbb{P}_{%
\mathrm{model}}=\Generator\left( z,\theta _{g}\right) $. Therefore, we have
constrained the generator to be in a smaller class of functions belonging to
$\mathcal{F}$ even if the neural network can approximate any function.

\begin{remark}
In order to obtain the minimax problem of \citet{Goodfellow-2014},
\citet{Nowozin-2016} considered the function:
\begin{equation*}
\phi \left( x\right) =x\log x-\left( x+1\right) \log \left( x+1\right)
\end{equation*}%
We deduce that:%
\begin{equation*}
\phi ^{\prime }\left( x\right) =\log \left( \frac{x}{x+1}\right)
\end{equation*}%
and\footnote{%
We have:%
\begin{equation*}
\phi ^{\ast }\left( t\right) =\sup_{x\in \limfunc{dom}\phi
}\left\langle x,t\right\rangle -\phi \left( x\right)
\end{equation*}%
It follows that $h\left( x\right) =xt-x\log x+\left( x+1\right) \log \left(
x+1\right) $ and:%
\begin{equation*}
h^{\prime }\left( x\right) =t-\log \left( \frac{x}{x+1}\right)
\end{equation*}%
The supremum is reached at the point $h^{\prime }\left( x^{\star }\right) =0$,
implying that:
\begin{equation*}
x^{\star }=\frac{e^{t}}{1-e^{t}}
\end{equation*}%
Finally, we obtain:%
\begin{eqnarray*}
\phi ^{\ast }\left( t\right)  &=&t\left(
\frac{e^{t}}{1-e^{t}}\right)
-\left( \frac{e^{t}}{1-e^{t}}\right) \log \left( \frac{e^{t}}{1-e^{t}}%
\right) +\left( \frac{1}{1-e^{t}}\right) \log \left( \frac{1}{1-e^{t}}%
\right)  \\
&=&t\left( \frac{e^{t}}{1-e^{t}}\right) -\left( \frac{e^{t}}{1-e^{t}}\right)
\left( t-\log \left( 1-e^{t}\right) \right) -\left( \frac{1}{1-e^{t}}\right)
\log \left( 1-e^{t}\right)  \\
&=&\left( \frac{e^{t}}{1-e^{t}}\right) \log \left( 1-e^{t}\right) -\left(
\frac{1}{1-e^{t}}\right) \log \left( 1-e^{t}\right)  \\
&=&-\log \left( 1-e^{t}\right)
\end{eqnarray*}%
}:%
\begin{equation*}
\phi ^{\ast }\left( x\right) =-\log \left( 1-e^{x}\right)
\end{equation*}%
Using Equation (\ref{eq-Nowozin1}), we deduce that:%
\begin{equation*}
\mathcal{C}\left( \tilde{x}\right) =\mathbb{E}\left[ \varphi \left( X\right)
\mid X\sim \mathbb{P}\right] +\mathbb{E}\left[ \log \left( 1-\exp \left(
\varphi \left( X\right) \right) \right) \mid X\sim \mathbb{Q}\right]
\end{equation*}%
If we set $\varphi \left( X\right) = \log \Discriminator\left( X,\theta
_{d}\right) $, we find the minimax function of \citet{Goodfellow-2014}:
\begin{equation*}
\mathcal{C}\left( \Discriminator\left( X,\theta _{d}\right) \right) =\mathbb{%
E}\left[ \log \Discriminator\left( X,\theta _{d}\right) \mid X\sim \mathbb{P}%
_{\mathrm{data}}\right] +\mathbb{E}\left[ \log \left( 1-\Discriminator\left(
X,\theta _{d}\right) \right) \mid X\sim \Generator\left( z,\theta
_{g}\right)\right]
\end{equation*}
\end{remark}

\paragraph{Another representation of $\protect\phi $-divergences}

GAN training can be viewed as a process of successively estimating
the optimal function $\varphi $ and minimizing the $\phi
$-divergence. \citet{Chu-2019} proposed a more general view of this
process. Previously, the study has been done on the space
$\mathbb{R}$ thanks to the dual form of the given function $\phi $
associated with the $\phi $-divergence. Alternatively,
\citet{Chu-2019} proposed to directly work on the probability space
$\mathcal{X}$. This imposes to define a probability functional
$\mathcal{J}_{\mathbb{P}}\left( \mathbb{Q}\right)
:{\mathcal{B}\left( \mathcal{X}\right) }\longrightarrow \mathbb{R}$, where $\mathcal{B}\left( \mathcal{X}\right) $ is the space of Borel
probability measures on $\mathcal{X}$. In the context of GAN
optimization problems, we would like to show that:
\begin{equation}
\min D_{\phi }\left( \mathbb{P}\parallel \mathbb{Q}\right) =\min_{\mathbb{Q}%
\in \mathcal{B}\left( \mathcal{X}\right)
}\mathcal{J}_{\mathbb{P}}\left( \mathbb{Q}\right)
\end{equation}%
However, contrary to the previous case, probability functionals take
value in probability spaces, where functional derivatives need to be
defined. Moreover, the dimension of the space is potentially
infinite. Therefore, the difficulty is to transform functional
optimization into a convex optimization problem that can be solved
using traditional numerical algorithms such as the gradient
descent.\smallskip

In order to define the derivative in $\mathcal{B}\left(
\mathcal{X}\right) $, \citet{Chu-2019} used the G\^{a}teaux
derivative of the functional $\mathcal{J}_{\mathbb{P}}$ at
$\mathbb{Q}$ in the direction of $\delta =\mathbb{P}-\mathbb{Q}$:
\begin{eqnarray}
\mathcal{J}_{\mathbb{P}}^{\prime }\left( \mathbb{P}-\mathbb{Q}\right)  &=&%
\mathrm{d}\mathcal{J}_{\mathbb{P}}\left( \mathbb{Q},\delta \right)
\notag
\\
&=&\left. \frac{\mathrm{d}}{\mathrm{d}\epsilon }\mathcal{J}_{\mathbb{P}%
}\left( \mathbb{Q}+\epsilon \delta \right) \right\vert _{\epsilon
=0}  \notag
\\
&=&\lim_{\epsilon \rightarrow 0}\frac{\mathcal{J}_{\mathbb{P}}\left( \mathbb{%
Q}+\epsilon \delta \right) -\mathcal{J}_{\mathbb{P}}\left(
\mathbb{Q}\right) }{\epsilon }
\end{eqnarray}%
This definition initially comes from Von Mises calculus, and the
G\^{a}teaux derivative is also called the \textquoteleft
\textit{Volterra}\textquoteright\ derivative \citep{Fernholz-2012}.
\citet{Chu-2019} recalled that the G\^{a}teaux derivative has an
integral representation $\mathcal{J}_{\mathbb{P}}^{\prime }\left(
\mathbb{\delta }\right) =\int_{\mathcal{X}}g\left( x\right) \,\delta
\left( \mathrm{d}x\right) $ where the function
$g:\mathcal{X}\rightarrow \mathbb{R}$ is called the \textquoteleft
\textit{influence function}\textquoteright\ or the influence
curve\footnote{The influence function is not unique. Indeed, for any
function $g$ that describes the G\^{a}teaux differential at
$\mathbb{Q}$, $g+c$ also works. Thus, the influence function is
uniquely defined up to an arbitrary additive constant.}. It follows
that:
\begin{eqnarray}
\mathcal{J}_{\mathbb{P}}^{\prime }\left(
\mathbb{P}-\mathbb{Q}\right) &=&\int_{\mathcal{X}}g\left( x\right)
\,\left( \mathbb{P}-\mathbb{Q}\right)
\left( \mathrm{d}x\right)   \notag \\
&=&\mathbb{E}\left[ g\left( X\right) \mid X\sim \mathbb{P}\right] -\mathbb{E}%
\left[ g\left( X\right) \mid X\sim \mathbb{Q}\right]
\end{eqnarray}%
We are now able to compute the derivative in order to recover the
optimal function that satisfy the minimum of the $\phi $-divergence:
\begin{eqnarray}
\mathcal{J}_{\mathbb{P}}^{\prime }\left(
\mathbb{P}-\mathbb{Q}\right) &=&\int_{\mathcal{X}}\left.
\frac{\mathrm{d}}{\mathrm{d}\epsilon }\phi
\left( \frac{\mathrm{d}\mathbb{Q}\left( x\right) +\epsilon \left( \mathrm{d}%
\mathbb{P}\left( x\right) -\mathrm{d}\mathbb{Q}\left( x\right) \right) }{%
\mathrm{d}\mathbb{P}\left( x\right) }\right) \right\vert _{\epsilon =0}\,%
\mathrm{d}\mathbb{P}\left( x\right)   \notag \\
&=&\int_{\mathcal{X}}\left. \phi ^{\prime }\left( \frac{\mathrm{d}\mathbb{Q}%
\left( x\right) +\epsilon \left( \mathrm{d}\mathbb{P}\left( x\right) -%
\mathrm{d}\mathbb{Q}\left( x\right) \right)
}{\mathrm{d}\mathbb{P}\left(
x\right) }\right) \right\vert _{\epsilon =0}\frac{\left( \mathrm{d}\mathbb{P}%
\left( x\right) -\mathrm{d}\mathbb{Q}\left( x\right) \right) }{\mathrm{d}%
\mathbb{P}\left( x\right) }\,\mathrm{d}\mathbb{P}\left( x\right)   \notag \\
&=&\int_{\mathcal{X}}\phi ^{\prime }\left(
\frac{\mathrm{d}\mathbb{Q}\left(
x\right) }{\mathrm{d}\mathbb{P}\left( x\right) }\right) \,\left( \mathbb{P}-%
\mathbb{Q}\right) \left( \mathrm{d}x\right)
\end{eqnarray}%
We deduce that the influence function for the $\phi $-divergence is equal to:%
\begin{equation}
g_{\phi }:\left\{
\begin{array}{l}
{\mathcal{X}}\rightarrow \mathbb{R} \\
x\rightarrow \phi ^{\prime }\left( \dfrac{\mathrm{d}\mathbb{Q}\left(
x\right) }{\mathrm{d}\mathbb{P}\left( x\right) }\right)
\end{array}%
\right.
\end{equation}%
The influence function can then be associated with the optimal
function $\phi $ introduced before because we have $g_{\phi
}=\varphi ^{\star }\in \mathcal{F}$. We conclude that the GAN
discriminator will try to estimate the influence function. While
\citet{Nguyen-2010} focused on the dual form of the function $\phi
$, \citet{Chu-2019} used a more general approach by considering the
dual form of the probability functional in order to recover the
Goodfellow's saddle point problem.\smallskip

In the case of probability functionals which take values in
$\mathcal{B}\left( \mathcal{X}\right) $, the dual space can be
defined as the space $L^{1}\left( \mathcal{X}\right) $ of all
real-valued Lipschitz function that takes values in $\mathcal{X}$
\citep{Laschos-2019}. According to the Riesz representation theorem,
it is possible to identify the space $\mathcal{B}\left(
\mathcal{X}\right) $ to it dual. We consider the product space
$\mathcal{B}\left( \mathcal{X}\right) \times L^{1}\left(
\mathcal{X}\right) $ with the scalar product defined as:
\begin{eqnarray*}
\left\langle \varphi ,\mathbb{Q}\right\rangle
&=&\int_{\mathcal{X}}\varphi
\left( x\right) \,\mathbb{Q}\left( \mathrm{d}x\right)  \\
&=&\int_{\mathcal{X}}\varphi \left( x\right)
\,\mathrm{d}\mathbb{Q}\left( x\right) \,\mathrm{d}x
\end{eqnarray*}%
where $\varphi \in L^{1}\left( \mathcal{X}\right) $ and
$\mathbb{Q}\in \mathcal{B}\left( \mathcal{X}\right) \times
\mathbb{R}$. Thus, we can rewrite $\mathcal{J}_{\mathbb{P}}$ in term
of its convex transform:
\begin{eqnarray}
\mathcal{J}_{\mathbb{P}}(\mathbb{Q}) &=&\sup_{\varphi \in
L^{1}\left(
\mathcal{X}\right) }\left\{ \left\langle \varphi ,\mathbb{Q}\right\rangle -%
\mathcal{J}_{\mathbb{P}}^{\ast }\left( \varphi \right) \right\}  \notag \\
&=&\sup_{\varphi \in L^{1}\left( \mathcal{X}\right) }\left\{ \int_{\mathcal{X%
}}\varphi \left( x\right) \,\mathbb{Q}\left( \mathrm{d}x\right) -\mathcal{J}%
_{\mathbb{P}}^{\ast }\left( \varphi \right) \right\} \notag \\
&=&\sup_{\varphi \in L^{1}\left( \mathcal{X}\right) }\left\{
\mathbb{E}\left[
\varphi \left( X\right) \mid X\sim \mathbb{Q}\right] -\mathcal{J}_{\mathbb{P}%
}^{\ast }\left( \varphi \right) \right\}
\end{eqnarray}%
To bridge the gap between the two approaches, we have to demonstrate
that $\mathcal{J}_{\mathbb{P}}^{\ast }\left( \varphi \right)
=\mathbb{E}\left[ \phi ^{\ast }\left( \varphi \left( X\right)
\right) \mid X\sim \mathbb{P}\right] $ for a given estimate of the
function $\varphi \in \mathcal{F}$. For that, \citet{Chu-2019}
considered the Fenchel conjugate of the Jensen-Shannon divergence%
\footnote{The derivation of this result is given in Appendix
\ref{appendix:jensen-shannon} on page
\pageref{appendix:jensen-shannon}.}:
\begin{eqnarray}
\mathcal{J}_{\mathbb{P}}^{\ast }\left( \varphi \right)  &=&\mathcal{J}_{%
\mathrm{JS}}^{\ast }\left( \varphi \right) \notag \\
&=&-\frac{1}{2}\mathbb{E}\left[ \log \left( 1-e^{2\varphi \left(
X\right) -\log 2}\right) \mid X\sim \mathbb{P}\right]
-\frac{1}{2}\log 2 \label{eq:JS1}
\end{eqnarray}%
Using $\varphi \left( x\right) =\frac{1}{2}\log \left(
1-\Discriminator\left( x,\theta _{d}\right) \right) +\frac{1}{2}\log
2$, we obtain:
\begin{eqnarray}
\mathbb{E}\left[ \varphi \left( X\right) \mid X\sim \mathbb{Q}\right] -%
\mathcal{J}_{\mathbb{P}}^{\ast }\left( \varphi \right)  &=&\frac{1}{2}%
\mathbb{E}\left[ \log \left( 1-\Discriminator\left( x,\theta
_{d}\right)
\right) \mid X\sim \mathbb{Q}\right] + \notag \\
&&\frac{1}{2}\mathbb{E}\left[ \log \Discriminator\left( x,\theta
_{d}\right) \mid X\sim \mathbb{P}\right] +\log 2
\end{eqnarray}%
Therefore, the descent algorithm applied to probability functionals
is equivalent to the Goodfellow's saddle point problem:
\begin{equation*}
\min_{\theta _{g}}\max_{\theta _{d}}\mathbb{E}\left[ \log \left( %
\Discriminator\left( X,\theta _{d}\right) \right) \mid X\sim \mathbb{P}_{%
\mathrm{data}}\right] +\mathbb{E}\left[ \log \left(
1-\Discriminator\left( X,\theta _{d}\right) \right) \mid X\sim
\Generator\left( z,\theta _{g}\right) \right]
\end{equation*}

\subsubsection{IPM-GAN models}
\label{appendix:ipm-gan}

In the case of Wasserstein generative adversarial networks
introduced by \citet{Arjovsky-2017a, Arjovsky-2017b}, $\phi
$-divergences have to be replaced by integral probability metrics
(IPMs) in order to compare two different probability distributions.
Let $\mathcal{F}$ be a class of functions defined on $\mathcal{X}$.
\citet{Muller-1997} defined an IPM $I_{\mathcal{F}}$ between
$\mathbb{P}$ and $\mathbb{Q}$ in the following way:
\begin{eqnarray*}
I_{\mathcal{F}}\left( \mathbb{P},\mathbb{Q}\right)  &=&\sup_{\varphi
\in
\mathcal{F}}\left\{ \left\vert \int_{\mathcal{X}}\varphi \left( x\right) \,%
\mathbb{P}\left( \mathrm{d}x\right) -\int_{\mathcal{X}}\varphi
\left(
x\right) \,\mathbb{Q}\left( \mathrm{d}x\right) \right\vert \right\}  \\
&=&\sup_{\varphi \in \mathcal{F}}\left\{ \left\vert \mathbb{E}\left[
\varphi \left( X\right) \mid X\sim \mathbb{P}\right]
-\mathbb{E}\left[ \varphi \left( X\right) \mid X\sim
\mathbb{Q}\right] \right\vert \right\}
\end{eqnarray*}%
IPMs are looking for a \textit{critic} function that maximizes the
average discrepancy between the two distributions $\mathbb{P}$ and
$\mathbb{Q}$. Contrary to $\phi $-GAN models, where we have to find
the variational form of the divergence $\phi \left( x\right) $, the
definition of an IPM directly gives the discrepancy operator:
\begin{equation*}
\Delta _{\mathbb{P},\mathbb{Q}}\left( \varphi \right)
=\mathbb{E}\left[ \varphi \left( X\right) \mid X\sim
\mathbb{P}\right] -\mathbb{E}\left[ \varphi \left( X\right) \mid
X\sim \mathbb{Q}\right]
\end{equation*}%
for all $\varphi \in \mathcal{F}$. The Wasserstein GAN proposed by
\citet{Arjovsky-2017a, Arjovsky-2017b} considers the function class
$\mathcal{F}$ such that $\varphi \left( x\right) $ is a 1-Lipschitz
function. In order to present different choices for the function
class $\mathcal{F}$, we consider the Lebesgue norm on the measurable
space $\Omega =\left( \mathcal{X},\mathbb{P}\right) $: $\left\Vert
f\right\Vert _{2}^{2}=\int_{\mathcal{X}}f^{2}\left( x\right)
\,\mathbb{P}\left( \mathrm{d}x\right) $. Let us denote the normed
space by $L^{2}\left( \Omega \right) =\left\{ f:
\mathcal{X}\rightarrow \mathbb{R}\mid \left\Vert f\right\Vert
_{2}<+\infty \right\} $ and the unit ball by
$\mathcal{B}_{1}=\left\{ f\in L^{2}\left( \Omega \right) \mid
\left\Vert f\right\Vert _{2}\leq 1\right\} $. Therefore, choosing
the class function such that $\mathcal{F}=\mathcal{B}_{1}$ is called
a Fisher GAN by \citet{Mroueh-2017}. In a similar way,
\citet{Mroueh-2018} proposed to define Sobolev GAN models by
considering the following class of functions $\mathcal{F}=\left\{
f\in L^{2}\left( \Omega \right) \mid \left\Vert \nabla
_{x}f\right\Vert _{2}\leq 1\right\} $.

\clearpage

\subsection{The Jensen-Shannon divergence function}
\label{appendix:jensen-shannon}

To derive the convex conjugate of $\mathcal{J}_{\mathrm{JS}}$, we
follow Appendix A given in \citet{Chu-2019}. Let $\mathbb{P}$ and
$\mathbb{Q}$ be two probability measures. We denote by $p\left(
x\right) $ and $q\left( x\right) $ the associated density functions
$\mathrm{d}\mathbb{P}\left( x\right) $ and
$\mathrm{d}\mathbb{Q}\left( x\right) $. The Jensen-Shannon
divergence function is defined by:
\begin{eqnarray*}
D_{\mathrm{JS}}\left( \mathbb{P}\parallel \mathbb{Q}\right)  &=&\frac{1}{2}%
D_{\mathrm{KL}}\left( \mathbb{P}\left\Vert \frac{1}{2}\mathbb{P+}\frac{1}{2}%
\mathbb{Q}\right. \right) +\frac{1}{2}D_{\mathrm{KL}}\left( \mathbb{Q}%
\left\Vert \frac{1}{2}\mathbb{P+}\frac{1}{2}\mathbb{Q}\right. \right)  \\
&=&\frac{1}{2}\int \left( p\left( x\right) \log \frac{2p\left( x\right) }{%
p\left( x\right) +q\left( x\right) }+q\left( x\right) \log
\frac{2q\left( x\right) }{p\left( x\right) +q\left( x\right)
}\right) \,\mathrm{d}x
\end{eqnarray*}%
where $D_{\mathrm{KL}}\left( \mathbb{P}\parallel \mathbb{Q}\right) $
is the Kullback-Leibler divergence. In the case where
$\mathcal{J}_{\mathrm{JS}}\left( \mathbb{P}\right)
=D_{\mathrm{JS}}\left( \mathbb{P}\parallel \mathbb{Q}\right) $, we
obtain:
\begin{eqnarray*}
\mathcal{J}_{\mathrm{JS}}\left( \mathbb{P}+\epsilon \delta \right)  &=&\frac{%
1}{2}\int \left( p\left( x\right) +\epsilon \delta \left( x\right)
\right) \log \frac{2\left( p\left( x\right) +\epsilon \delta \left(
x\right) \right)
}{p\left( x\right) +q\left( x\right) +\epsilon \delta \left( x\right) }\,%
\mathrm{d}x+ \\
&&\frac{1}{2}\int q\left( x\right) \log \frac{2q\left( x\right)
}{p\left( x\right) +q\left( x\right) +\epsilon \delta \left(
x\right) }\,\mathrm{d}x
\end{eqnarray*}%
Since we have:%
\begin{equation*}
\log \frac{2\left( p\left( x\right) +\epsilon \delta \left( x\right)
\right) }{p\left( x\right) +q\left( x\right) +\epsilon \delta \left(
x\right) }=\log 2+\log \left( p\left( x\right) +\epsilon \delta
\left( x\right) \right) -\log \left( p\left( x\right) +q\left(
x\right) +\epsilon \delta \left( x\right) \right)
\end{equation*}%
and:%
\begin{equation*}
\frac{\mathrm{d}}{\mathrm{d}\epsilon }\log \frac{2\left( p\left(
x\right) +\epsilon \delta \left( x\right) \right) }{p\left( x\right)
+q\left(
x\right) +\epsilon \delta \left( x\right) }=\frac{\delta \left( x\right) }{%
p\left( x\right) +\epsilon \delta \left( x\right) }-\frac{\delta
\left( x\right) }{p\left( x\right) +q\left( x\right) +\epsilon
\delta \left( x\right) }
\end{equation*}%
it follows that:%
\begin{eqnarray*}
\frac{\mathrm{d}}{\mathrm{d}\epsilon
}\mathcal{J}_{\mathrm{JS}}\left(
\mathbb{P}+\epsilon \delta \right)  &=&\frac{1}{2}\int \frac{\mathrm{d}}{%
\mathrm{d}\epsilon }\left( \left( p\left( x\right) +\epsilon \delta
\left( x\right) \right) \log \frac{2\left( p\left( x\right)
+\epsilon \delta \left( x\right) \right) }{p\left( x\right) +q\left(
x\right) +\epsilon \delta
\left( x\right) }\right) \,\mathrm{d}x+ \\
&&\frac{1}{2}\int \frac{\mathrm{d}}{\mathrm{d}\epsilon }\left(
q\left( x\right) \log \frac{2q\left( x\right) }{p\left( x\right)
+q\left( x\right)
+\epsilon \delta \left( x\right) }\right) \,\mathrm{d}x \\
&=&\frac{1}{2}\int \delta \left( x\right) \log \frac{2\left( p\left(
x\right) +\epsilon \delta \left( x\right) \right) }{p\left( x\right)
+q\left( x\right) +\epsilon \delta \left( x\right) }\,\mathrm{d}x+ \\
&&\frac{1}{2}\int \left( p\left( x\right) +\epsilon \delta \left(
x\right) \right) \frac{\delta \left( x\right) }{p\left( x\right)
+\epsilon \delta
\left( x\right) }\,\mathrm{d}x- \\
&&\frac{1}{2}\int \left( p\left( x\right) +\epsilon \delta \left(
x\right) \right) \frac{\delta \left( x\right) }{p\left( x\right)
+q\left( x\right)
+\epsilon \delta \left( x\right) }\,\mathrm{d}x- \\
&&\frac{1}{2}\int q\left( x\right) \frac{\delta \left( x\right)
}{p\left( x\right) +q\left( x\right) +\epsilon \delta \left(
x\right) }\,\mathrm{d}x
\end{eqnarray*}%
We deduce that:%
\begin{eqnarray*}
\left. \frac{\mathrm{d}}{\mathrm{d}\epsilon
}\mathcal{J}_{\mathrm{JS}}\left(
\mathbb{P}+\epsilon \delta \right) \right\vert _{\epsilon =0} &=&\frac{1}{2}%
\int \left( \delta \left( x\right) \log \frac{2p\left( x\right)
}{p\left(
x\right) +q\left( x\right) }\right) \,\mathrm{d}x \\
&=&\frac{1}{2}\int \left( \log \frac{p\left( x\right) }{p\left(
x\right) +q\left( x\right) }+\log 2\right) \,\delta \left( x\right)
\mathrm{d}x
\end{eqnarray*}%
\citet{Chu-2019} concluded that the influence function of $\mathcal{J}_{%
\mathrm{JS}}\left( \mathbb{P}\right) $ is equal to:%
\begin{equation*}
g_{\mathrm{JS}}\left( x\right) =\frac{1}{2}\log \frac{\mathrm{d}\mathbb{P}%
\left( x\right) }{\mathrm{d}\mathbb{P}\left( x\right) +\mathrm{d}\mathbb{Q}%
\left( x\right) }+\frac{1}{2}\log 2
\end{equation*}%
In order to find the convex conjugate of
$\mathcal{J}_{\mathrm{JS}}\left( \mathbb{P}\right) $, we consider
the Fenchel-Moreau theorem:
\begin{eqnarray*}
\mathcal{J}_{\mathrm{JS}}^{\ast }\left( \varphi \right)  &=&\sup_{\mathbb{P}%
\in {\mathcal{B}\left( \mathcal{X}\right) }}\left\{ \left\langle \varphi ,%
\mathbb{P}\right\rangle -\mathcal{J}_{\mathrm{JS}}\left(
\mathbb{P}\right)
\right\}  \\
&=&\sup_{\mathbb{P}\in {\mathcal{B}\left( \mathcal{X}\right) }}\left\{ \int_{%
\mathcal{X}}\varphi \left( x\right) \,\mathbb{P}\left( \mathrm{d}x\right) -%
\mathcal{J}_{\mathrm{JS}}\left( \mathbb{P}\right) \right\}
\end{eqnarray*}%
We have:%
\begin{eqnarray*}
f\left( \mathbb{P}\right)  &=&\int_{\mathcal{X}}\varphi \left( x\right) \,%
\mathbb{P}\left( \mathrm{d}x\right) -\mathcal{J}_{\mathrm{JS}}\left( \mathbb{%
P}\right)  \\
&=&\int_{\mathcal{X}}\varphi \left( x\right) p\left( x\right)
\,\mathrm{d}x-
\\
&&\frac{1}{2}\int_{\mathcal{X}}\left( p\left( x\right) \log
\frac{2p\left(
x\right) }{p\left( x\right) +q\left( x\right) }+q\left( x\right) \log \frac{%
2q\left( x\right) }{p\left( x\right) +q\left( x\right) }\,\right) \,\mathrm{d%
}x \\
&=&\int_{\mathcal{X}}h\left( x\right) \,\mathrm{d}x
\end{eqnarray*}%
where:%
\begin{eqnarray*}
h\left( x\right)  &=&\varphi \left( x\right) p\left( x\right) -\frac{1}{2}%
p\left( x\right) \log 2-\frac{1}{2}p\left( x\right) \log p\left( x\right) +%
\frac{1}{2}p\left( x\right) \log \left( p\left( x\right) +q\left(
x\right)
\right) - \\
&&\frac{1}{2}q\left( x\right) \log 2-\frac{1}{2}q\left( x\right)
\log q\left( x\right) +\frac{1}{2}q\left( x\right) \log \left(
p\left( x\right) +q\left( x\right) \right)
\end{eqnarray*}%
and:%
\begin{eqnarray*}
\frac{\partial h\left( x\right) }{\partial p\left( x\right) }
&=&\varphi
\left( x\right) -\frac{1}{2}\log 2-\frac{1}{2}\log p\left( x\right) -\frac{1%
}{2}+\frac{1}{2}\log \left( p\left( x\right) +q\left( x\right) \right) + \\
&&\frac{1}{2}\frac{p\left( x\right) }{p\left( x\right) +q\left( x\right) }+%
\frac{1}{2}\frac{q\left( x\right) }{p\left( x\right) +q\left( x\right) } \\
&=&\varphi \left( x\right) -\frac{1}{2}\log 2-\frac{1}{2}\log
p\left( x\right) +\frac{1}{2}\log \left( p\left( x\right) +q\left(
x\right) \right)
\end{eqnarray*}%
Since the first-order condition $\partial _{p\left( x\right)
}h\left( x\right) =0$, we deduce that the optimal solution is:
\begin{eqnarray*}
\varphi \left( x\right)  &=&\frac{1}{2}\log 2+\frac{1}{2}\log
\frac{p\left(
x\right) }{p\left( x\right) +q\left( x\right) } \\
&=&\frac{1}{2}\log \frac{2p\left( x\right) }{p\left( x\right)
+q\left( x\right) }
\end{eqnarray*}%
\citet{Chu-2019} noticed that:%
\begin{eqnarray*}
\frac{q\left( x\right) }{p\left( x\right) +q\left( x\right) } &=&1-\frac{%
p\left( x\right) }{p\left( x\right) +q\left( x\right) } \\
&=&1-\frac{1}{2}e^{2\varphi \left( x\right) }
\end{eqnarray*}%
In this case, we obtain:%
\begin{eqnarray*}
h\left( x\right)  &=&\varphi \left( x\right) p\left( x\right) -\frac{1}{2}%
p\left( x\right) \log \frac{2p\left( x\right) }{p\left( x\right)
+q\left(
x\right) }-\frac{1}{2}q\left( x\right) \log \frac{2q\left( x\right) }{%
p\left( x\right) +q\left( x\right) } \\
&=&\varphi \left( x\right) p\left( x\right) -\varphi \left( x\right)
p\left( x\right) -\frac{1}{2}q\left( x\right) \log \left(
2-e^{2\varphi \left( x\right) }\right)
\end{eqnarray*}%
The convex conjugate of $\mathcal{J}_{\mathrm{JS}}$ is then:%
\begin{eqnarray*}
\mathcal{J}_{\mathrm{JS}}^{\ast }\left( \varphi \right)  &=&\int_{\mathcal{X}%
}-\frac{1}{2}q\left( x\right) \log \left( 2-e^{2\varphi \left(
x\right)
}\right) \,\mathrm{d}x \\
&=&-\frac{1}{2}\int_{\mathcal{X}}q\left( x\right) \log \left(
2-e^{2\varphi
\left( x\right) }\right) \,\mathrm{d}x \\
&=&-\frac{1}{2}\int_{\mathcal{X}}\log \left(
1-\frac{1}{2}e^{2\varphi \left(
x\right) }\right) q\left( x\right) \,\mathrm{d}x-\frac{1}{2}\log 2 \\
&=&-\frac{1}{2}\mathbb{E}\left[ \log \left( 1-e^{2\varphi \left(
X\right) -\log 2}\right) \mid X\sim \mathbb{Q}\right]
-\frac{1}{2}\log 2
\end{eqnarray*}

\begin{remark}
In order to retrieve Equation (\ref{eq:JS1}), we have to interchange $\mathbb{P}$ and $%
\mathbb{Q}$, because we need to compute
$\mathcal{J}_{\mathbb{P}}\left(
\mathbb{Q}\right) $ and not $\mathcal{J}_{\mathbb{Q}}\left( \mathbb{P}%
\right) $.
\end{remark}

\clearpage

\subsection{Derivation of the minimax cost function}
\label{appendix:minimax}

The cost function can be viewed as a binary cross-entropy measure. Let $Y$ and
$\hat{Y} $ be two random variables with probability mass function $p$ and $q$.
The cross-entropy function is equal to:
\begin{equation*}
H\left( p,q\right) =\mathbb{E}\left[ -\log q\left( x\right)
\mid x\sim p\left( x\right) \right]
\end{equation*}
For discrete probability distributions, we obtain:
\begin{equation*}
H\left( p,q\right) =-\sum_{x}p\left( x\right) \log q\left( x\right)
\end{equation*}
In a binary classification problem, for a given observation $x_i$, we have $p =
y_i \in \{0, 1\}$, which is the true label and $q = \hat {y}_i \in [0, 1]$ which
is the predicted probability of the current model. We can use binary cross
entropy to get a measure of dissimilarity between ${y_i}$ and $\hat{y_i}$:
\begin{equation*}
H\left( p,q\right) =-y_i \log \hat{y_i}-\left( 1-y_i\right) \log \left( 1-\hat{y_i}\right)
\end{equation*}
In the case of $m$ samples, the loss function is then given by:
\begin{equation*}
\mathcal{L} = -\frac{1}{m} \sum_{i=1}^{m} y_i \log \hat{y_i}+
\left( 1-y_i\right) \log \left( 1-\hat{y_i} \right)
\end{equation*}%
\smallskip

Under the GAN framework, we have $m$ samples of $x_0$ and $m$ samples of $x_1$,
which serve as the input data of the discriminator model. We note $x = \{x_0,
x_1\}$ the set of the two samples. Since $\hat{y_i} = \Discriminator\left(
x_{i};\theta _{d}\right)$, the formula above can be written as:
\begin{equation*}
\mathcal{L}\left( \theta _{g},\theta _{d}\right) = -\frac{1}{2m} \sum_{i=1}^{2m}
y_i \log \Discriminator\left( x_{i};\theta _{d}\right) +\left( 1-y_i\right)
\log \left( 1- \Discriminator\left( x_{i};\theta _{d}\right) \right)
\end{equation*}
If $x_i$ comes from the sample $x_0$, $y_i$ takes the value $0$, otherwise it
takes the value $1$. It follows that:
\begin{eqnarray*}
\mathcal{L}\left( \theta _{g},\theta _{d}\right) &=&
 -\frac{1}{2m} \left( \sum_{i=1}^{m}\log \Discriminator\left( x_{1,i};\theta _{d}\right) +
 \sum_{i=1}^{m} \log \left( 1- \Discriminator\left( x_{0, i};\theta _{d}\right) \right) \right) \\
&=& -\frac{1}{2} \left( \frac{1}{m}\sum_{i=1}^{m}\log \Discriminator\left( x_{1, i};\theta _{d}\right) + \frac{1}{m}
\sum_{i=1}^{m} \log \left( 1- \Discriminator\left( \Generator\left( z_i;\theta _{g}\right);\theta _{d}\right) \right) \right)
\end{eqnarray*}
Therefore, the loss function $\mathcal{L}\left( \theta _{g},\theta _{d}\right)$
corresponds to the average of the cross-entropy when considering several
observations:
\begin{equation*}
\mathcal{L}\left( \theta _{g},\theta _{d}\right) = - \frac{1}{2}
\mathbb{E}\left[ \log \Discriminator\left( x_{1};\theta _{d}\right) \right] - \frac{1}{2}
\mathbb{E}\left[ \log \left( 1- \Discriminator\left( \Generator\left( z;\theta _{g}\right);\theta _{d}\right) \right)\right]
\end{equation*}
We notice that $\mathcal{C}\left( \theta _{g},\theta _{d}\right)$ is equal to
$-2 \cdot \mathcal{L}\left( \theta _{g},\theta _{d}\right)$. Minimizing the
loss function is then equivalent to maximize $\mathcal{C}\left( \theta
_{g},\theta _{d}\right)$ with respect to $\theta _{d}$.

\clearpage

\subsection{An introduction to Monge-Kantorovich problems}
\label{appendix:Monge-Kantorovich}

\subsubsection{Primal formulation of optimal transport}

Optimal transport can be very powerful when comparing two probability
distributions. The geometric approach that has been proposed will allow us to
resolve complex optimization problems. OT problem relies on two probability
spaces $\left( \mathcal{X},\mathbb{P}\right) $ and $\left(
\mathcal{Y},\mathbb{Q}\right) $, and a cost function $c:\mathcal{X}\times
\mathcal{Y}\rightarrow \mathbb{R}^{+}$. For instance, we would like to
transform a pile of sand, where particles are distributed according to
$\mathbb{P}$ to a structured sand castle, where particles are distributed
according to $\mathbb{Q}$. In this case, $c$ denotes the amount of such
effort. Thus, we can find an optimal path that minimizes the cost of this
transformation. Let us introduce a map function $T:\mathcal{X}\rightarrow
\mathcal{Y}$ that allow us to describe how $x\in \mathcal{X}$ is transported
to the target space $\mathcal{Y}$. The optimal transport map function is the
solution of the so-called Monge problem:
\begin{equation}
\inf_{T\in \mathcal{A}}\left\{ \left. \int_{\mathcal{X}}c\left( x,T\left(
x\right) \right) \,\mathrm{d}\mathbb{P}\left( x\right) \right\vert T_{\ast
}\left( \mathbb{P}\right) =\mathbb{Q}\right\}
\end{equation}%
where $\mathcal{A}=\left\{ T:\mathcal{X}\rightarrow \mathcal{Y}\mid \mathbb{Q%
}\left( \mathcal{C}\right) =\mathbb{P}\left( T^{-1}\left( \mathcal{C}\right)
\right) ,\mathcal{C}\subseteq \mathcal{Y}\right\} $. In other words, $T$ must
push-forward the probability measure $\mathbb{P}$ toward $\mathbb{Q}$,
meaning that  $\mathbb{Q=}T_{\ast }\left( \mathbb{P}\right) $.\smallskip

However, Monge problem may be difficult to solve in some cases because the
mapping function may not necessarily exist. For instance, let us consider
that the source distribution $\mathbb{P}$ is a Dirac measure such as $%
\mathrm{d}\mathbb{P}\left( x\right) =\delta \left( x\right) \,\mathrm{d}x$
and the target distribution $\mathbb{Q}$ is a continuous measure such as a
normal distribution. In this particular case, there is no map function $T$
such that the condition $\mathbb{Q=}T_{\ast }\left( \mathbb{P}\right) $ is
satisfied. Moreover, this condition is non-convex. To illustrate this, let
us take $\mathbb{P}$ and $\mathbb{Q}$ two continuous Lebesgue measures of $%
\mathcal{X}$ such that $\mathrm{d}\mathbb{P}\left( x\right) =p\left(
x\right) \,\mathrm{d}x$ and $\mathrm{d}\mathbb{Q}\left( x\right) =q\left(
x\right) \,\mathrm{d}x$ for all $x\in \mathcal{X}$. It can be shown that
satisfying the condition $\mathbb{Q=}T_{\ast }\left( \mathbb{P}\right) $
leads to the constraint $q\left( T\left( x\right) \right) \left\vert \func{%
det}\left( \nabla _{x}T\left( x\right) \right) \right\vert =p\left( x\right)
$ for all $x\in \mathcal{X}$. Since this constraint is non convex, the
uniqueness of the minimization problem is not necessarily guaranteed. This is
why Kantorovich reformulated the problem as follows:
\begin{equation}
\inf_{\mathbb{F}\in \mathcal{F}\left( \mathbb{P},\mathbb{Q}\right) }\left\{
\int_{\mathcal{X}\times \mathcal{Y}}c\left( x,y\right) \,\mathrm{d}\mathbb{F}%
\left( x,y\right) \right\}
\end{equation}%
where $\mathcal{F}\left( \mathbb{P},\mathbb{Q}\right) $ is the Fr\'echet
class\footnote{This means that $\mathcal{F}\left(
\mathbb{P},\mathbb{Q}\right) $ collects
all multivariate joint distributions, whose marginals are exactly equal to $%
\mathbb{P}$ and $\mathbb{Q}$.}. The infimum is then obtained by considering
all joint probability
measures $\mathbb{F}$ on $\mathcal{X}\times \mathcal{Y}$ such that $\mathbb{P%
}$ and $\mathbb{Q}$ are the marginals. Such joint probability measures are
called transportation plans. The Kantorovich transportation problem is now
convex and becomes a linear programming problem easier to solve than the
Monge transportation problem. However, searching among all join probability
measures $\mathbb{F}\in \mathcal{F}\left( \mathbb{P},\mathbb{Q}\right) $ can
also be computationally intractable. For instance, \citet{Seguy-2017} recall
that solving the linear program takes $O\left( n^{3}\log n\right) $ when $n$
is the size of the support in the case of discrete probability distributions.

\begin{remark}
Let us consider the particular case where $\mathbb{P}$ and $\mathbb{Q}$ are
continuous Lebesgue measures and the cost function $c$ correspond to a
$p$-Euclidian distance $d\left( x,y\right) $. The solution to the
Monge-Kantorovich problem is then defined as the $p$-Wasserstein
distance\footnote{It is also known as the earth mover's distance (EMD), which
has been used by \citet{Rubner-2000} for content-based image retrieval in
computer vision.}:
\begin{equation*}
W_{p}\left( \mathbb{P},\mathbb{Q}\right) =\left( \inf_{\mathbb{F}\in
\mathcal{F}\left( \mathbb{P},\mathbb{Q}\right) }\left\{ \int_{\mathcal{X}%
\times \mathcal{Y}}d\left( x,y\right) ^{p}\,\mathrm{d}\mathbb{F}\left(
x,y\right) \right\} \right) ^{\nicefrac{1}{p}}
\end{equation*}
\citet{Brenier-1991} showed that there is a unique solution when $p>1$.
\end{remark}

\begin{remark}
In some particular cases, optimal transport problems can be easily
solved. Let us consider the case where $\mathcal{X}=\mathcal{Y}$ is
a one-dimensional space, and $c\left( x,y\right) $ is a convex
function that satisfies the following condition: if $x_{1}<x_{2}$
and $y_{1}<y_{2}$, then $c\left( x_{2},y_{2}\right) -c\left(
x_{1},y_{2}\right) -c\left( x_{2},y_{1}\right) +c\left(
x_{1},y_{1}\right) <0$, then the optimal transport plan respects the
ordering of the elements. Consequently, the solution corresponds to
a monotone rearrangement of $\mathbb{P}$ into $\mathbb{Q}$. Solving
this problem is no more than sorting elements in a list
\citep{Brenier-1991}.
\end{remark}

\subsubsection{Dual formulation of optimal transport}

Another face of the optimal transport problem is the Kantorovich duality that
can be easily understood from an economic point of view. Following closely
the example given by \citet{Villani-2008}, we consider a manufacturer that
produces goods on a production site located at $x$ and sells them on a store
located at $y$. $c\left( x,y\right) $ is the cost to transport goods from $x$
to $y$. Minimizing his transportation cost for the entire production is
equivalent to solve the Monge-Kantorovich problem. Now, let us assume that he
does not care about transportation. This is why he wants to hire a company
specialized in goods transportation. They offer him to buy each product at
the price $\varphi \left( x\right) $. They will then transport at $y$ and
sell back the product to him at the price $\psi \left( y\right) $. The
manufacturer will accept the deal only if there is a financial interest such
that $\psi \left( y\right) -\varphi \left( x\right) \leq c\left( x,y\right)
$. Despite this constraint, the transportation company will try to maximize
its profits. The new problem corresponds to the dual formulation of the
Monge-Kantorovich problem and can be written as follows:
\begin{equation}
\sup \left\{ \int_{\mathcal{Y}}\psi \left( y\right) \,\mathrm{d}\mathbb{Q}%
\left( y\right) -\int_{\mathcal{X}}\varphi \left( x\right) \,\mathrm{d}%
\mathbb{P}\left( y\right) :\psi \left( y\right) -\varphi \left( x\right)
\leq c\left( x,y\right) \right\}
\end{equation}%
According to \citet{Villani-2008}, $\varphi $ and $\psi $ are integrable such
that $\left( \varphi ,\psi \right) \in L^{1}\left(
\mathcal{X},\mathbb{P}\right) \times L^{1}\left(
\mathcal{Y},\mathbb{Q}\right) $. The functions $\varphi $ and $\psi $ are
often called Kantorovich potentials\footnote{\citet[Theorem
1.3]{Brenier-1991} showed the link between Kantorovich potentials and mapping
functions in the particular case where the cost function is a $2$-Euclidean
distance:
\begin{equation*}
T\left( x\right) =x-\nabla _{x}\varphi (x)=\nabla _{x}\left( \frac{1}{2}%
\left\Vert x\right\Vert ^{2}-\varphi (x)\right)
\end{equation*}%
}.\smallskip

\begin{remark}
Considering the viewpoint of the manufacturer who cares about the cost is
equivalent to look at the solution of the Monge-Kantorovich primal problem.
Considering the viewpoint of the transportation company that cares about
optimizing the profit is equivalent to looking at the solution of the
Monge-Kantorovich dual problem.
\end{remark}

A rigorous proof of dual formulation\footnote{See also \citet{Xia-2008,
Xia-2009} for a geometric interpretation.} is given by \citet{Villani-2008}.
We recall that the condition $\mathbb{F}\in \mathcal{F}\left(
\mathbb{P},\mathbb{Q}\right) $ implies:
\begin{eqnarray}
\int_{\mathcal{X}\times \mathcal{Y}}\left( \psi \left( y\right) -\varphi
\left( x\right) \right) \,\mathrm{d}\mathbb{F}\left( x,y\right) &=&\int_{%
\mathcal{X}\times \mathcal{Y}}\psi \left( y\right) \,\mathrm{d}\mathbb{F}%
\left( x,y\right) -\int_{\mathcal{X}\times \mathcal{Y}}\varphi \left(
x\right) \,\mathrm{d}\mathbb{F}\left( x,y\right)  \notag \\
&=&\int_{\mathcal{Y}}\psi \left( y\right) \int_{\mathcal{X}}\mathrm{d}%
\mathbb{F}\left( x,y\right) -\int_{\mathcal{X}}\varphi \left( x\right) \int_{%
\mathcal{Y}}\mathrm{d}\mathbb{F}\left( x,y\right)  \notag \\
&=&\int_{\mathcal{Y}}\psi \left( y\right) \,\mathrm{d}\mathbb{Q}\left(
y\right) -\int_{\mathcal{X}}\varphi \left( x\right) \,\mathrm{d}\mathbb{P}%
\left( x\right)
\end{eqnarray}
If $\mathbb{F}\notin \mathcal{F}\left( \mathbb{P},\mathbb{Q}\right) $, we
assume by convention that the difference between the two members is infinite
and note $\mathds{1}_{\mathcal{F}\left( \mathbb{P},\mathbb{Q}\right) }\left(
\mathbb{F}\right) $ the convex indicator function of $\mathcal{F}\left(
\mathbb{P},\mathbb{Q}\right) $:
\begin{eqnarray*}
(\ast ) &=&\inf_{\mathbb{F}\in \mathcal{F}\left( \mathbb{P},\mathbb{Q}%
\right) }\left\{ \int_{\mathcal{X}\times \mathcal{Y}}c\left( x,y\right) \,%
\mathrm{d}\mathbb{F}\left( x,y\right) \right\} \\
&=&\inf_{\mathbb{F}\in \mathcal{F}\left( \mathbb{P},\mathbb{Q}\right)
}\left\{ \int_{\mathcal{X}\times \mathcal{Y}}c\left( x,y\right) \,\mathrm{d}%
\mathbb{F}\left( x,y\right) \right\} +\mathds{1}_{\mathcal{F}\left( \mathbb{P%
},\mathbb{Q}\right) }\left( \mathbb{F}\right) \\
&=&\inf_{\mathbb{F}\in \mathcal{F}\left( \mathbb{P},\mathbb{Q}\right)
}\left\{ \int_{\mathcal{X}\times \mathcal{Y}}c\left( x,y\right) \,\mathrm{d}%
\mathbb{F}\left( x,y\right) \right\} + \\
&&\sup_{\left( \varphi ,\psi \right) }\left\{ \int_{\mathcal{Y}}\psi \left(
y\right) \,\mathrm{d}\mathbb{Q}\left( y\right) -\int_{\mathcal{X}}\varphi
\left( x\right) \,\mathrm{d}\mathbb{P}\left( x\right) -\int_{\mathcal{X}%
\times \mathcal{Y}}\left( \psi \left( y\right) -\varphi \left( x\right)
\right) \,\mathrm{d}\mathbb{F}\left( x,y\right) \right\} \\
&=&\inf_{\mathbb{F}\in \mathcal{F}\left( \mathbb{P},\mathbb{Q}\right)
}\sup_{\left( \varphi ,\psi \right) }\Gamma \left( \mathbb{P},\mathbb{Q},%
\mathbb{F},\varphi ,\psi \right)
\end{eqnarray*}%
where:%
\begin{eqnarray*}
\Gamma \left( \mathbb{P},\mathbb{Q},\mathbb{F},\varphi ,\psi \right)
&=&\int_{\mathcal{X}\times \mathcal{Y}}c\left( x,y\right) \,\mathrm{d}%
\mathbb{F}\left( x,y\right) +\int_{\mathcal{Y}}\psi \left( y\right) \,%
\mathrm{d}\mathbb{Q}\left( y\right) - \\
&&\int_{\mathcal{X}}\varphi \left( x\right) \,\mathrm{d}\mathbb{P}\left(
x\right) -\int_{\mathcal{X}\times \mathcal{Y}}\left( \psi \left( y\right)
-\varphi \left( x\right) \right) \,\mathrm{d}\mathbb{F}\left( x,y\right)
\end{eqnarray*}%
Since we have:%
\begin{equation*}
\inf_{\mathbb{F}\in \mathcal{F}\left( \mathbb{P},\mathbb{Q}\right)
}\sup_{\left( \varphi ,\psi \right) }\Gamma \left( \mathbb{P},\mathbb{Q},%
\mathbb{F},\varphi ,\psi \right) =\sup_{\left( \varphi ,\psi \right) }\inf_{%
\mathbb{F}\in \mathcal{F}\left( \mathbb{P},\mathbb{Q}\right) }\Gamma \left(
\mathbb{P},\mathbb{Q},\mathbb{F},\varphi ,\psi \right)
\end{equation*}%
it follows that:%
\begin{eqnarray*}
\inf_{\mathbb{F}\in \mathcal{F}\left( \mathbb{P},\mathbb{Q}\right) }\Gamma
\left( \mathbb{P},\mathbb{Q},\mathbb{F},\varphi ,\psi \right) &=&\int_{%
\mathcal{Y}}\psi \left( y\right) \,\mathrm{d}\mathbb{Q}\left( y\right)
-\int_{\mathcal{X}}\varphi \left( x\right) \,\mathrm{d}\mathbb{P}\left(
x\right) + \\
&&\inf_{\mathbb{F}\in \mathcal{F}\left( \mathbb{P},\mathbb{Q}\right)
}\left\{ \int_{\mathcal{X}\times \mathcal{Y}}\left( c\left( x,y\right)
-\left( \psi \left( y\right) -\varphi \left( x\right) \right) \right) \,%
\mathrm{d}\mathbb{F}\left( x,y\right) \right\}
\end{eqnarray*}%
Finally, we conclude that\footnote{Of course the constraint $\psi \left(
y\right) -\varphi \left( x\right) \leq c\left( x,y\right) $ must be
satisfied.}:
\begin{equation*}
\inf_{\mathbb{F}\in \mathcal{F}\left( \mathbb{P},\mathbb{Q}\right) }\left\{
\int_{\mathcal{X}\times \mathcal{Y}}c\left( x,y\right) \,\mathrm{d}\mathbb{F}%
\left( x,y\right) \right\} =\sup_{\left( \varphi ,\psi \right) }\left\{
\int_{\mathcal{Y}}\psi \left( y\right) \,\mathrm{d}\mathbb{Q}\left( y\right)
-\int_{\mathcal{X}}\varphi \left( x\right) \,\mathrm{d}\mathbb{P}\left(
x\right) \right\}
\end{equation*}%
because:%
\begin{equation*}
\inf_{\mathbb{F}\in \mathcal{F}\left( \mathbb{P},\mathbb{Q}\right) }\left\{
\int_{\mathcal{X}\times \mathcal{Y}}\left( c\left( x,y\right) -\left( \psi
\left( y\right) -\varphi \left( x\right) \right) \right) \,\mathrm{d}\mathbb{%
F}\left( x,y\right) \right\} =0
\end{equation*}

\subsubsection{Semi-dual formulation of optimal transport}

It is possible to go deeper in the proof by introducing the notion of
$c$-convexity \citep{Villani-2008}. The function $\varphi
:\mathcal{X}\rightarrow \mathbb{R}\cup \{+\infty \}$ is said to be $c$-convex
if there exists a function $\zeta :\mathcal{Y}\rightarrow \mathbb{R}\cup
\{\pm \infty \}$ such that:
\begin{equation*}
\varphi \left( x\right) =\sup_{y\in \mathcal{Y}}\left\{ \zeta \left(
y\right) -c\left( x,y\right) \right\}
\end{equation*}%
for all $x\in \mathcal{X}$. With this definition, it is possible to define
its $c$-transform $\varphi ^{c}$:
\begin{equation*}
\varphi ^{c}\left( y\right) =\inf_{x\in \mathcal{X}}\left\{ \varphi \left(
x\right) +c\left( x,y\right) \right\}
\end{equation*}%
for all $y\in \mathcal{Y}$. In the particular case where the cost function is
a distance, a $c$-convex function is simply a $1$-Lipschitz function, and is
equal to its $c$-transform. Indeed, let us consider that $\varphi $ is
$1$-Lipschitz such that $\varphi (x)-\varphi (y)\leq c(x,y)$. We have
$\varphi (x)\leq \varphi (y)+c(x,y)$ and:
\begin{eqnarray*}
\varphi \left( x\right)  &=&\inf_{y}\left\{ \varphi \left( y\right) +c\left(
x,y\right) \right\}  \\
&=&\varphi ^{c}\left( x\right)
\end{eqnarray*}%
Again, it is possible to understand the $c$-transform through the economic
point of view. Let us recall that the transportation company needs to
satisfy the condition $\psi \left( y\right) -\varphi \left( x\right) \leq
c\left( x,y\right) $ in order to remain competitive. It follows that $\psi
\left( y\right) \leq \varphi \left( x\right) +c\left( x,y\right) $ and
$\varphi \left( x\right) \geq \psi \left( y\right) -c(x,y)$. To maximize its
profits, the company will choose the pair $\left( \varphi ,\psi \right) $
such that:
\begin{equation*}
\left\{
\begin{array}{l}
\psi \left( y\right) =\inf_{x}\left\{ \varphi \left( x\right) +c\left(
x,y\right) \right\}  \\
\varphi \left( x\right) =\sup_{y}\left\{ \psi \left( y\right)
-c(x,y)\right\}
\end{array}%
\right.
\end{equation*}%
Therefore, it becomes useful to write $\psi $ in term of $\varphi $. If we
consider that the cost function is a distance and $\mathcal{X}=\mathcal{Y}$,
\citet{Villani-2008} showed that:
\begin{eqnarray*}
(\ast ) &=&\inf_{\mathbb{F}\in \mathcal{F}\left( \mathbb{P},\mathbb{Q}%
\right) }\left\{ \int_{\mathcal{X}\times \mathcal{Y}}c\left( x,y\right) \,%
\mathrm{d}\mathbb{F}\left( x,y\right) \right\}  \\
&=&\sup_{\left( \varphi ,\psi \right) }\left\{ \int_{\mathcal{Y}}\psi \left(
y\right) \,\mathrm{d}\mathbb{Q}\left( y\right) -\int_{\mathcal{X}}\varphi
\left( x\right) \,\mathrm{d}\mathbb{P}\left( x\right) :\psi \left( y\right)
-\varphi \left( x\right) \leq c\left( x,y\right) \right\}  \\
&=&\sup_{\varphi }\left\{ \int_{\mathcal{Y}}\varphi ^{c}\left( y\right) \,%
\mathrm{d}\mathbb{Q}\left( y\right) -\int_{\mathcal{X}}\varphi \left(
x\right) \,\mathrm{d}\mathbb{P}\left( x\right) \right\}  \\
&=&\sup_{\varphi }\left\{ \int_{\mathcal{Y}}\varphi \left( y\right) \,%
\mathrm{d}\mathbb{Q}\left( y\right) -\int_{\mathcal{X}}\varphi \left(
x\right) \,\mathrm{d}\mathbb{P}\left( x\right) \right\}  \\
&=&\sup_{\varphi }\left\{ \mathbb{E}\left[ \varphi \left( Y\right) \mid
Y\sim \mathbb{Q}\right] -\mathbb{E}\left[ \varphi \left( X\right) \mid X\sim
\mathbb{P}\right] \right\}
\end{eqnarray*}%
This is the semi-dual formulation of the problem also called the
Kantorovich-Rubinstein duality. This formulation is used to train the
Wasserstein GAN that estimates the optimal function $\varphi $.

\subsubsection{An example}

If $\mathbb{P}\sim \mathcal{N}\left( \mu _{1},\Sigma _{1}\right) $ and
$\mathbb{Q}\sim \mathcal{N}\left( \mu _{2},\Sigma _{2}\right) $,
\citet{Givens-1984} showed that the $2$-Wasserstein distance is equal to:
\begin{equation}
W_{2}\left( \mathbb{P},\mathbb{Q}\right) =\sqrt{\left\Vert \mu _{1}-\mu
_{2}\right\Vert ^{2}+\func{tr}\left( \Sigma _{1}+\Sigma _{2}-2\left( \Sigma
_{1}^{\nicefrac{1}{2}}\Sigma _{2}\Sigma _{1}^{\nicefrac{1}{2}}\right) ^{\nicefrac{1}{2}%
}\right) }
\end{equation}%
where $A^{\nicefrac{1}{2}}$ is the square root of $A$.

\clearpage

\subsection{Converting real-valued samples into binary features}
\label{appendix:binary transformation}

These transformation methods have been introduced by \citet{Kondratyev-2019}.
Algorithm (\ref{alg:binarize1}) describes how to transform real-valued data
into binary features. Each one-dimensional data sample is represented by a
16-digit binary number and in the case of $n$-dimensional data, we transform
receptively each single value into 16-digit binary vector and concatenate them
to form a $16 \times n$-digit binary vector.

\begin{algorithm}[tbh]
\caption{Real-valued to integer to binary transformation}
\label{alg:binarize1}
\begin{algorithmic}
    \STATE \textbf{Result}: Conversion of real-valued dataset into binary vector
    \STATE \textbf{Input}: A real-valued dataset $X_{\text{real}}$ with $N$ samples

    \STATE $\epsilon \geq 0$
    \STATE $X_{\min} \leftarrow \min \left(X_{\textrm{real}}\right)-\epsilon$
    \STATE $X_{\max} \leftarrow \max \left(X_{\textrm{real}}\right)+\epsilon$
        \FOR{$l = 1, \cdots, N $}
            \STATE $X_{\textrm{integer}}^{(l)} \leftarrow \operatorname{int}\left(65535 \times
                    \left(X_{\textrm{real}}^{(l)}-X_{\min}\right) /\left(X_{\max }-X_{\min }\right)\right)$
            \STATE $X_{\textrm{binary}}^{(l)} \leftarrow$ binarize $\left(X_{\textrm{integer}}^{(l)}\right)$
        \ENDFOR
\end{algorithmic}
\end{algorithm}

Algorithm (\ref{alg:binarize2}) performs the inverse transformation. Similarly,
in the case of $n$-dimensional data, we transform receptively each 16 binary
numbers into a real value and concatenate them to form a $n$-dimensional
real-valued vector.

\begin{algorithm}[tbh]
\caption{Binary to integer to real-valued transformation}
\label{alg:binarize2}
\begin{algorithmic}
    \STATE \textbf{Result}: Conversion of binary vector into a real-valued sample
    \STATE \textbf{Input}: A 16-digit binary vector $X = \left(X_{1},\cdots, X_{16} \right)$

    \STATE $X_{\textrm{integer}} \leftarrow 0$
    \FOR{$i = 1, \cdots, 16$}
        \STATE $$\hat{X}_{\text {integer}} \leftarrow \hat{X}_{\text {integer}}+2^{i-1} \times \hat{X}_{16-i}$$
    \ENDFOR
    \STATE $\hat{X}_{\text {real}} \leftarrow X_{\text {min}} + \hat{X}_{\text {integer}} \times\left(X_{\text {max}}-X_{\text {min}}\right) / 65535$
\end{algorithmic}
\end{algorithm}


\end{document}